\documentclass[dsingle]{dissertation}

\usepackage[italian,english]{babel}
\usepackage{hhline}
\usepackage{algorithm}
\usepackage[noend]{algorithmic}
\usepackage{eqparbox}
\usepackage{sidecap}
\usepackage[labelfont=bf,font=footnotesize]{caption}
\usepackage[labelfont=it,textfont={it},justification=centering]{subcaption}
\usepackage{comment}
\usepackage{booktabs}
\usepackage{arydshln}
\usepackage{wrapfig}
\usepackage{bm}
\usepackage{bbm}
\usepackage{textcomp,gensymb}
\usepackage{dsfont}
\usepackage{xstring}
\usepackage{lscape}
\usepackage{makecell}
\usepackage{enumitem}

\usepackage{xcolor}

\usepackage[export]{adjustbox}
\usepackage{lipsum}
\usepackage{pgf}
\usepackage{rotating,multirow}

\usepackage{tikz}
\usepackage{cases}
\usepackage{tablefootnote}

\usepackage{glossaries}
\makeglossaries

\makeatletter
\newcommand\footnoteref[1]{\protected@xdef\@thefnmark{\ref{#1}}\@footnotemark}
\makeatother

\def\-{\raisebox{.75pt}{-}}

\newcommand{\tit}[1]{\smallbreak\noindent\textmd{\textsc{#1.}}}

\begin{document}

%!TEX root = ../dissertation.tex
% Some details about the dissertation.
\title{You can have multiple titles}
\title{and have them listed here for considering}
\title{but the only one counted is the last and }
\title{Autonomous Embodied Agents: \\ \Large When Robotics Meets Deep Learning
Reasoning}
\titleIt{Agenti Incorporati Autonomi: \\ \Large Quando la Robotica Incontra il Ragionamento con Apprendimento Profondo}
\author{Roberto Bigazzi}
\phdfrom{Modena} % only city, not country, from the Beadle Office
\defendVenue{[Aula der Universiteit/Agnietenkapel*]}

\defendDOW{d.o.w}               % dinsdag day of week, in Dutch
\defendDay{XX}         % 18
\defendMonth{XX}     % mei 
\defendYear{2023}           % 2021 
\defendTime{[...time...]}       % 12:00

%If you have one advisor
\advisor{Prof. Rita Cucchiara}
\advisorSchool{University of Modena and Reggio Emilia}

\schoolCoordinator{Prof. Sonia Bergamaschi}
\schoolCoordinatorSchool{University of Modena and Reggio Emilia}

\committeeOne{Prof. Shreyas Kousik}
\committeeOneSchool{Georgia Institute of Technology}
\committeeTwo{Prof. Matteo Matteucci}
\committeeTwoSchool{Polytechnic University of Milan}
\committeeThree{prof. dr. committee member 3}
\committeeThreeSchool{Universiteit van Amsterdam}
\committeeFour{dr. committee member 4}
\committeeFourSchool{Universiteit van Amsterdam}
\committeeFive{dr. committee member 5}
\committeeFiveSchool{Universiteit van Amsterdam}

\definecolor{SchoolColor}{HTML}{8C1515}  % cardinal red 
\definecolor{chaptergrey}{HTML}{8C1515} % dialed back a little   
\definecolor{midgrey}{rgb}{0.4, 0.4, 0.4}

    \pagestyle{plain}
    \frontmatter
    \setstretch{\dnormalspacing}
    \graphicspath{{figures/}}

\fancyhead[LO,RE]{}
\fancyhead[LE]{\slshape \small \leftmark}
\fancyhead[RO]{\slshape \small \rightmark}

\fancypagestyle{nosection}{%
    \fancyhead{}\fancyfoot[C]{\thepage}
}

\renewcommand\chaptermark[1]{\markboth{\MakeUppercase{\thechapter. \footnotesize #1}}{\MakeUppercase{\thesection. \footnotesize #1}}}
\renewcommand\sectionmark[1]{\markright{\MakeUppercase{\thesection. \footnotesize #1}}}
\newacronym[description={branch of computer science that develops machines and software with human-like intelligence}]{ai}{AI}{Artificial Intelligence}

\newacronym[description={research field that strives for the creation of embodied agents which learn, through interaction and exploration, to solve challenging tasks within their environments}]{eai}{Embodied AI}{Embodied Artificial Intelligence}

\newglossaryentry{robotics}{
    name=Robotics,
    description={branch of technology that deals with the design, construction, operation, and application of robots}}
    
\newglossaryentry{cv}{
    name=Computer Vision,
    description={scientific field that deals with how computers can gain high-level understanding from digital images or videos}}
    
\newglossaryentry{ml}{
    name=Machine Learning,
    description=scientific study of algorithms and statistical models that computer systems use to perform tasks without explicit instructions}

\newglossaryentry{dl}{
    name=Deep Learning,
    description={a subfield of machine learning concerned with algorithms inspired by the structure and function of the brain called artificial neural networks}
}

\newacronym[description={specialized electronic circuit designed to manipulate and alter memory to accelerate the creation of images}]{gpu}{GPU}{Graphics Processing Unit}

\newacronym[description={computational model that mimics the way nerve cells work in the human brain}]{ann}{ANN}{artificial neural network}

\newacronym[description={interdisciplinary subfield of linguistics, computer science, and artificial intelligence concerned with the interactions between computers and human language, in particular how to program computers to process and analyze large amounts of natural language data}]{nlp}{NLP}{Natural Language Processing}

\newacronym[description={the process of transferring knowledge or skills learned in a simulated environment to a real-world setting}]{sim2real}{Sim2Real}{simulation-to-reality}

\newacronym[description={area of machine learning concerned with how intelligent agents ought to take actions in an environment in order to maximize the notion of cumulative reward. Reinforcement learning is one of three basic machine learning paradigms, alongside supervised learning and unsupervised learning}]{rl}{RL}{Reinforcement Learning}

\newacronym[description={Deep \acrlong{rl}}]{drl}{DRL}{Deep Reinforcement Learning}

\newacronym[description={Vision-and-Language Navigation}]{vln}{VLN}{Vision-and-Language Navigation}

\newacronym[description={convolutional neural network}]{cnn}{CNN}{convolutional neural network}

\newacronym[description={simultaneous localization and mapping}]{slam}{SLAM}{simultaneous localization and mapping}

\newacronym[description={Point Goal navigation}]{pointnav}{PointNav}{Point Goal navigation}

\newacronym[description={Object Goal navigation}]{objectnav}{ObjectNav}{Object Goal navigation}

\newglossaryentry{ic}{
    name=Image Captioning,
    description={a task at the intersection between \gls{cv} and \gls{nlp} whose goal is to generate a natural language description of a given image}
}

\newacronym[description={semantic occupancy map}]{som}{SOM}{semantic occupancy map}
\newacronym[description={Matterport 3D}]{mp3d}{MP3D}{Matterport 3D}
\newacronym[description={multilayer perceptron}]{mlp}{MLP}{multilayer perceptron}
\newacronym[description={prediction gain}]{pg}{PG}{prediction gain}
\newacronym[description={Density Model Estimation}]{dme}{DME}{Density Model Estimation}
\newacronym[description={intersection-over-union}]{iou}{IoU}{intersection-over-union}
\newacronym[description={masked intersection-over-union}]{miou}{mIoU}{masked intersection-over-union}
\newacronym[description={accuracy}]{acc}{Acc}{accuracy}
\newacronym[description={masked accuracy}]{macc}{mAcc}{masked accuracy}
\newacronym[description={area seen}]{as}{AS}{area seen}
\newacronym[description={\acrlong{iou} of the free space}]{fiou}{FIoU}{\acrlong{iou} of the free space}
\newacronym[description={\acrlong{iou} of the occupied space}]{oiou}{OIoU}{\acrlong{iou} of the occupied space}
\newacronym[description={\acrlong{as} of the free space}]{fas}{FAS}{\acrlong{as} of the free space}
\newacronym[description={\acrlong{as} of the occupied space}]{oas}{OAS}{\acrlong{as} of the occupied space}
\newacronym[description={translation error}]{te}{TE}{translation error}
\newacronym[description={angular error}]{ae}{AE}{angular error}
\newacronym[description={orientation error}]{oe}{OE}{orientation error}
\newacronym[description={Occupancy Anticipation}]{occant}{OccAnt}{Occupancy Anticipation}
\newacronym[description={Active Neural SLAM}]{ans}{ANS}{Active Neural SLAM}
\newacronym[description={distance to the goal}]{d2g}{D2G}{distance to the goal}
\newacronym[description={success rate}]{sr}{SR}{success rate}
\newacronym[description={\acrshort{pointnav} success rate}]{pgsr}{PNSR}{\acrshort{pointnav} success rate}
\newacronym[description={angular success rate}]{asr}{ASR}{angular success rate}
\newacronym[description={success weighted by path length}]{spl}{SPL}{success weighted by path length}
\newacronym[description={Soft\acrshort{spl}}]{sspl}{sSPL}{Soft\acrshort{spl}}
\newacronym[description={loquacity}]{loq}{Loq}{loquacity}
\newacronym[description={recall oriented understudy of gisting evaluation}]{rouge}{ROUGE}{recall oriented understudy of gisting evaluation}
\newacronym[description={bilingual evaluation understudy}]{bleu}{BLEU}{bilingual evaluation understudy}
\newacronym[description={consensus-based image description evaluation}]{cider}{CIDEr}{consensus-based image description evaluation}
\newacronym[description={diversity}]{div}{Div}{diversity}
\newacronym[description={coverage}]{cov}{Cov}{coverage}
\newacronym[description={\textit{Explore and Explain} method}]{ex2}{eX\textsuperscript{2}}{Explore and Explain}
\newacronym[description={episode description score}]{eds}{ED\text{-}S}{episode description score}
\newacronym[description={multi-head self- and cross-attention }]{msca}{MSCA}{multi-head self- and cross-attention}
\newacronym[description={multi-head self-attention}]{msa}{MSA}{multi-head self-attention}
\newacronym[description={semantic propositional image caption evaluation}]{spice}{SPICE}{semantic propositional image caption evaluation}
\newacronym[description={metric for evaluation of translation with explicit ordering}]{meteor}{METEOR}{metric for evaluation of translation with explicit ordering}
\newacronym[description={CLIP score}]{clips}{CLIP\text{-}S}{CLIP score}
\newacronym[description={recurrent neural network}]{rnn}{RNN}{recurrent neural network}
\newacronym[description={long short-term memory}]{lstm}{LSTM}{long short-term memory}
\newacronym[description={anticipation reward}]{ar}{AR}{anticipation reward}
\newacronym[description={coverage reward}]{cr}{CR}{coverage reward}
\newacronym[description={difference reward}]{dr}{DR}{difference reward}
\newacronym[description={hard failure rate}]{hfr}{HFR}{hard failure rate}
\newacronym[description={bump rate}]{br}{BR}{bump rate}
\newacronym[description={field of view}]{fov}{FoV}{field of view}
\newacronym[description={Art Gallery 3D}]{ag3d}{AG3D}{Art Gallery 3D}
\newacronym[description={large language model}]{llm}{LLM}{large language model}

\newcommand{\todo}[1]{\textcolor{red}{[#1]}}
\newcommand{\tocheck}[1]{\textcolor{blue}{[#1]}}

\def\eg{\emph{e.g}\onedot} \def\Eg{\emph{E.g}\onedot}
\def\ie{\emph{i.e}\onedot} \def\Ie{\emph{I.e}\onedot}
\def\cf{\emph{c.f}\onedot} \def\Cf{\emph{C.f}\onedot}
\def\etc{\emph{etc}\onedot} \def\vs{\emph{vs}\onedot}
\def\wrt{w.r.t\onedot} \def\dof{d.o.f\onedot}
\def\etal{\emph{et al}\onedot}

\def \ours {\acrshort{ex2}}
\newcommand{\DATASET}{Art Gallery 3D}
\newcommand{\DATASETS}{AG3D}
\newcommand{\POINTNAV}{PointNav$\mathcal{++}$}
\newcommand{\mathbbmm}[1]{\text{\usefont{U}{bbm}{m}{n}#1}}
\newcommand\blfootnote[1]{%
  \begingroup
  \renewcommand\thefootnote{}\footnote{#1}%
  \addtocounter{footnote}{-1}%
  \endgroup
}

\pagestyle{fancy}
\pagenumbering{arabic}

% \begin{savequote}[65mm]
% If there were a little good in the world and everyone considered himself his brother, there would be less thought and less sorrow and the world would be much more beautiful.
% \qauthor{P.P.}
% \end{savequote}

\hyphenation{Ro-bot-ics}

\chapter{Introduction}
\label{chap:intro}

\label{sec:introduction}
\lettrine[lines=1]{\textcolor{SchoolColor}{W}}{hen} we think about life in $100$ years from now, how do we imagine it?
Most optimist viewers envision flying cars, autonomous transportation, and hi-tech cities, others instead, see a general improvement in life quality and automation in all aspects of everyday life.
In any of the ``good'' scenarios, one detail that cannot be missed in the vision is the inclusion of ubiquitous robotic helpers interacting seamlessly with humans and the surrounding environment. 

Recently, we have already seen robots gradually enter industry and society; for example, medical robots can assist surgeons during procedures, help patients with physical therapy, and perform diagnostic tests. Another example is shown in manufacturing settings, where robots perform tasks such as welding, painting, and assembly.
At the same time, we have seen an increasing interest in service robots designed for interaction with humans in a variety of settings, such as hospitals, museums, and homes. These robots can assist with tasks such as delivering items, providing information, and performing basic maintenance.

However, robots that are able to really understand and learn, like the ones we see in some movies, are still a long way off. In fact, such robots can perform the task they are programmed for, but they are not capable of reasoning about their surroundings, neither they learn new concepts or knowledge.

\AddLabels 

As a response to this lack, a new field called \gls{eai} gained attention in the research community. Such a field has the objective of fostering the development of the intelligent autonomous agents of the future by combining knowledge in various research areas, including \gls{robotics}, \gls{cv}, and \gls{nlp}.

\Gls{eai} considers robots equipped with sensors and actuators that allow them to perceive, manipulate the environment, move around, and navigate; however, differently from pure \gls{robotics}, \gls{eai} research mainly focuses on high-level interactions between the agents and the surrounding environment.

The reasoning behind \gls{eai} is to create \acrlong{ai} systems that can interact with and operate within the physical world in a natural and intuitive way. Embodied agents can also be designed to be more efficient and effective at performing certain tasks, such as manufacturing or search and rescue, by taking advantage of their physical bodies and ability to move through space. In a nutshell, the goal of \gls{eai} is to create intelligent systems that can operate and interact with the world in a manner similar to humans.

Regarding the factors that lead to the emergence of \gls{eai}, we can identify some main causes:
the advances brought by the so-called \gls{dl} revolution, the release of new data devoted to robotic simulation, and the increasing availability of computational power.

By \gls{dl} revolution, we refer to the recent breakthroughs in \gls{ai} research that have been made possible using \gls{dl} techniques. %, such as the adoption of \gls{ann}. that are one the fundamental bases of \gls{dl}, 
In fact, \gls{dl} lead to many of the major advances in \gls{ai} over the past decade, for example in image and speech recognition, language translation, or self-driving cars. \gls{ai} models can now effectively recognize objects, generate natural language sentences, and make decisions aimed at fulfilling a task.
Speaking about robotic simulation, in the last years we have seen the introduction of large-scale datasets of real-life apartments, houses, and offices \cite{chang2017matterport3d,xia2018gibson,savva2019habitat,ramakrishnan2021hm3d}, allowing researchers to use simulation to train agents for navigation and other tasks.
Besides that, the growth in available computing power, boosted also by the usage of dedicated general purpose \glspl{gpu}, has allowed the development of models that can process and analyze large amounts of data in real-time, and be deployed on physical robots.

\section{Problem Statement}
In order to obtain systems and robots that behave and reason like humans, probably decades of research are still needed, requiring advances in all areas affecting \gls{eai} before being able to reach such milestones. With the work presented in this dissertation, we aim to help future research on this topic and take a step toward real intelligent autonomous agents. 

One of the problems presented in the previous section is the ineffective ability of embodied agents to reason about their surroundings. 
To start coping with this issue, an effective way is to design efficient representations to extract useful information from the agent’s sensing. Finding the best way to process visual inputs or other auxiliary representations to improve agents' performance is an object of research, and finding efficient representations would allow faster learning phases using fewer data to train agents for a particular task.

Once the agent is capable of performing a task efficiently, it should be able to transfer its knowledge to different tasks without reinitialization. In light of this, a desired goal of our work is finding a task that can be used as a proxy for downstream.

At the same time, a smart agent should also be able to update its knowledge if something changes in the surrounding environment. We want to tackle this issue by finding effective ways to train the agent to search for fresh available information in the environment.

A totally perpendicular direction of research considers embodied agents' inability to provide feedback about their behavior and their perception. We study a way to overcome this problem by providing the robot with the capability to communicate with humans to describe its observations. As a part of this capability, the agent should consider the right moment for producing descriptions and interacting with humans without constantly outputting useless information. How to choose when the agent should speak is also another interesting aspect we consider in this thesis.

A final problem that we want to address in this work is the correct deployment of agents trained in simulation, where the training is safer, faster, and cheaper, to real robotic platforms without major degradation in their performance. In this respect, we have to consider what discrepancies exist between simulation and reality that we need to address in order to transfer knowledge between simulation and the real world smoothly.

Addressing the aforementioned problems and weaknesses in current embodied agents is a small step towards robots that behave and operate in the real world seamlessly like humans. In the following pages of this dissertation, we will tackle these open questions trying to improve available agents.

\section{Organization}
With this dissertation, we present a complete pipeline for the creation of a smart autonomous agent starting from the design of its knowledge representation to the real-world deployment and test.
In the following, a brief description of the organization of the work is given.

Starting with Chapter \ref{chap:survey}, we present a detailed description of the current literature on \gls{eai} and complementary research fields that are going to be relevant throughout the work proposed in the thesis.

In Chapter \ref{chap:focus} and \ref{chap:ex2}, we present two new implicit rewards for embodied agents for exploration that present interesting benefits also to downstream tasks. 
Specifically, in Chapter \ref{chap:focus}, the agent is enriched by a neural mapper that produces occupancy grid maps and we propose a more efficient intrinsic reward named Impact, which encourages the agent to perform actions that produce high variation in its internal representation of the environment. We evaluate agent's performance in navigation towards coordinates showing that an agent able to efficiently explore the environment can be used without major modifications for downstream tasks.

In Chapter \ref{chap:ex2} we present a new setting where the objective is to produce natural language descriptions while exploring the environment. We name this setting \emph{Explore and Explain}. To tackle this task, we propose an intrinsic reward based on artificial curiosity. This reward pushes the agent towards states where a model that generates future observations, produces wrong predictions.

Chapter \ref{chap:eds} improves the work presented in the previous chapter testing different map-based exploration agents available in the literature on the new \emph{Explore and Explain} setting, as well as adopting state-of-the-art captioning methods. To evaluate the performance of the models we present a novel metric considering both exploration and language generation measures. This metric highlights the agents that are able to explore effectively the environment and generate informative descriptions without repeating the same information multiple times.

In Chapter \ref{chap:sd}, we move away from the task of exploration by proposing a new task where the agent has already collected the map of the environment, but the layout of the map is changed from the state stored in the memory of the agent. Some objects are added, moved, or removed. The goal of the agent is to spot the outdated parts in the map and repair it in a given time window. We call this new task \emph{Spot the Difference}.

In Chapter \ref{chap:out}, we show that the models trained in simulation can be deployed on a Low-Cost Robot (LoCoBot) \cite{locobot} without major redesigns. The performance degradation of the models trained in simulation and tested in the real world shows that \gls{sim2real} transfer still poses a problem and further research is needed to fill the performance gap.

A final contribution is presented in Chapter \ref{chap:gallerie}, where we introduce a new dataset for photo-realistic robotic simulation that, instead of containing houses, flats, or offices as the available benchmark datasets, has been collected in an art museum. We show that the peculiar topology of such a building increases the difficulty of embodied tasks such as coordinate-driven navigation, enabling further research on the task in large environments.

Chapter \ref{chap:conclusions} presents the conclusions of the dissertation with some personal considerations and some possible future work and directions of research.

\hyphenation{par-a-digms au-tonomous}

\chapter[Literature Survey]{Literature Survey}
\label{chap:survey}
\RemoveLabels 

\lettrine[lines=1]{\textcolor{SchoolColor}{I}}{n} this chapter, we present a literature overview of the relevant work that is related to the tasks and settings proposed in this thesis. 
We first review the literature on visual exploration, listing related work in both traditional \gls{robotics} and \gls{eai} fields (Section \ref{sec:exploration_survey}). 
Following, Section \ref{sec:implicit_rewards_survey} presents implicit rewards for exploration agents with the work done in the \gls{ml} field and the recent work in \gls{eai}. 
Section \ref{sec:navigation_survey} reviews the recent advances in embodied navigation, listing both map-based and map-less methods.
In Section \ref{sec:simulators_survey}, we show some of the advances in simulating platforms for research, from the simulators used in traditional \gls{rl} to the photo-realistic simulators for \gls{eai} agents, including also complementary datasets of 3D models of environments and task-specific datasets. 
Section \ref{sec:sim2real_survey} presents the research on \gls{sim2real} transfer of smart autonomous agents or work that includes real-world deployment. 
The last part of this chapter contains three sections to briefly present literature on the knowledge of the environment exploited by embodied agents (Section \ref{sec:dynamic_survey}), \gls{ic} (Section \ref{sec:captioning_survey}), and deep generative models (Section \ref{sec:generative_survey}) that is related to the work presented in this dissertation. 

\section{Embodied Agents for Exploration}
\label{sec:exploration_survey}
Classical heuristic and geometric-based exploration methods rely on two main strategies: frontier-based exploration \cite{yamauchi1997frontier} and next-best-view planning \cite{gonzalez2002navigation}.

The former entails iteratively navigating towards the closest point of the closest frontier, which is defined as the boundary between the explored free space and the unexplored space. The latter entails sequentially reaching cost-effective unexplored points, \ie points from where the gain in the explored area is maximum, weighed by the cost to reach them. 

These methods have been largely used and improved \cite{holz2010evaluating,bircher2016receding,niroui2019deep} or combined in a hierarchical exploration algorithm \cite{zhu2018deep,selin2019efficient}. However, when applied with noisy odometry and localization sensors or in highly complex environments, geometric approaches tend to fail \cite{chen2019learning,niroui2019deep,ramakrishnan2020exploration}. In light of this, increasing research effort has been dedicated to the development of learning-based approaches, which usually exploit \gls{drl} to learn robust and efficient exploration policies.

\AddLabels

In fact, even if the final goal of current research on \gls{eai} mainly focuses on tasks that require navigating to indoor locations or coordinates, such as \gls{vln} \cite{anderson2018vision,landi2019embodied,landi2021multimodal,chen2021history,guhur2021airbert,chen2022think,chen2022learning}, \gls{pointnav}, and \gls{objectnav} \cite{wijmans2019dd,anderson2018evaluation,zhu2017target}, Ramakrishnan \etal \cite{ramakrishnan2020exploration} highlighted the importance of visual exploration in order to pretrain a generic embodied agent and identified four paradigms for learning-based visual exploration: novelty-based, curiosity-based, reconstruction-based, and coverage-based.
Each paradigm is characterized by a different reward function used as a self-supervision signal for optimizing the exploration policy.
A coverage-based reward, considering the area seen, is also used in the modular approach to \acrfull{ans} presented in \cite{chaplot2019learning}, which combines a neural mapper module with a hierarchical navigation policy. 
To enhance exploration efficiency in complex environments, Ramakrishnan \etal \cite{ramakrishnan2020occupancy} resorted to an extrinsic reward by introducing the occupancy anticipation reward, which aims to maximize the agent accuracy in predicting occluded unseen areas, and a method combining frontier-based exploration with a learning-based approach \cite{ramakrishnan2022poni}.
A different approach is followed by Georgakis \etal \cite{georgakis2022uncertainty} that implies an uncertainty-based exploration agent.

We contribute to current research on exploration methods for embodied agents in Chapter \ref{chap:focus} and Chapter \ref{chap:ex2}.

\section{Implicit Rewards for Exploration}
\label{sec:implicit_rewards_survey}

The lack of ground truth in the exploration task forces the adoption of \acrfull{rl} for training exploration methods. 
Unfortunately, \gls{rl} methods have low sample efficiency, even when applied to tasks different from robot exploration. Thus, they require designing intrinsic reward functions that encourage visiting novel states or learning the environment dynamics. Furthermore, the use of intrinsic motivation is beneficial also in case the external task-specific rewards are sparse or absent.

In this context, Oudeyer and Kaplan \cite{oudeyer2009intrinsic} provide a summary of early work on intrinsic motivation. Among them, Schmidhuber \cite{schmidhuber2010formal} and Sun \etal \cite{sun2011planning} proposed to use information gain and compression as intrinsic rewards, while Klyubin \etal \cite{klyubin2005empowerment}, and Mohamed and Rezende \cite{mohamed2015variational} adopted the concept of empowerment as reward during training.

Among the intrinsic rewards that motivate the exploration of novel states, Bellemare \etal \cite{bellemare2016unifying} introduced the notion of pseudo visitation count by using a Context-Tree Switching (CTS) density model to extract a pseudo-count from raw pixels and applied count-based algorithms. Similarly, Ostrovski \etal \cite{ostrovski2017count} applied the autoregressive deep generative model PixelCNN\footnote{Related work on deep generative models is presented in Section \ref{sec:generative_survey}.} \cite{oord2016conditional} to estimate the pseudo-count of the visited state.
Recently, Zhang \etal \cite{zhang2020bebold} proposed a criterion to mitigate common issues in count-based methods.

Rewards that promote the learning of the environment dynamics comprehend Random Network Distillation (RND) \cite{burda2018exploration}, Disagreement \cite{pathak2019self}, and Curiosity.
Among curiosity-driven exploration methods, the strategy of jointly training forward and backward dynamics models for learning a feature space has been demonstrated to be effective in Atari games \cite{bellemare2013arcade} and other exploration games \cite{agrawal2015learning,pathak2017curiosity,burda2018large}.
Differently, Houthooft \etal \cite{houthooft2016vime} presented an exploration strategy based on the maximization of information gain about the agent’s belief of environment dynamics.
Another common approach for exploration is that of using state visitation counts as intrinsic rewards \cite{bellemare2016unifying,tang2017exploration}.
Recently, Raileanu \etal \cite{raileanu2020ride} proposed to jointly encourage both the visitation of novel states and the learning of the environment dynamics, by devising a paradigm that rewards the agent proportionally to the change in the state representation caused by its actions. However, their approach is developed for grid-like environments with a finite number of states, where the visitation count can be easily employed as a discount factor.

In Chapter \ref{chap:focus} and Chapter \ref{chap:ex2}, we present two intrinsic reward paradigms that are suited for robotic exploration in continuous photo-realistic environments.

\section{Embodied Agents for Navigation}
\label{sec:navigation_survey}
After having presented the literature on embodied exploration in Section \ref{sec:exploration_survey}, in this section we discuss relevant research work on navigation. 
The tasks of embodied exploration and navigation are very closely related in \gls{eai} literature, in fact, several research papers on exploration methods also validate their approaches on the downstream task of \acrfull{pointnav}. 

In the last few years, performance on \acrshort{pointnav} has improved significantly saturating very quickly the task of noise-free navigation. For example, Wijmans \etal \cite{wijmans2019dd} achieved $95\%$ of \acrfull{sr} on Habitat Challenge 2019 test dataset.
However, moving on to noisy navigation the task becomes problematic. 

Literature on \gls{eai} methods for navigation can be divided in two main categories: map-based and map-less approaches.

Regarding the map-based approaches, \acrfull{ans} \cite{chaplot2019learning} was among the first methods to adopt a hierarchical planning module coupled with a learning-based neural mapper. Ramakrishnan \etal \cite{ramakrishnan2020occupancy} improved \acrshort{ans} results by inferring unseen or occluded regions in the scene. 

Among the mapless methods, Sax \etal \cite{sax2018mid} transferred visual features from Taskonomy \cite{zamir2018taskonomy}, showing that using diverse representation sets for downstream tasks improves performance. Ye \etal \cite{ye2021auxiliaryPN, ye2021auxiliaryON} takes inspiration from this approach, devising a method that involves learning auxiliary tasks to tackle both \acrlong{pointnav} and \acrlong{objectnav}. Instead, Karkus \etal \cite{karkus2021differentiable} took inspiration from traditional \gls{robotics} proposing a learning-based particle filter to improve agents' pose estimation during navigation. Zhao \etal \cite{zhao2021surprising} is the first to apply visual odometry combined with a learning-based approach effectively. Recently, Partsey \etal \cite{partsey2022mapping} developed data-augmentation techniques that do not require human annotations to train models for visual odometry, achieving significant results on the task of \gls{pointnav}.

We tackle the task of embodied navigation in Chapter \ref{chap:eds}, \ref{chap:out}, and \ref{chap:gallerie}.

\section{Interactive Environments and Datasets}
\label{sec:simulators_survey}
When it comes to training intelligent agents, the underlying environment plays an important role. A first testbed for research in \gls{rl} has been provided by the Atari games of the Arcade Learning Environment (ALE) \cite{bellemare2013arcade,brockman2016openai,machado2018revisiting}.
Using these environments researchers were able to make remarkable progress on learning-based agents surpassing human performance on most games \cite{horgan2018distributed}. 
However, these kinds of settings are not suitable for navigation and exploration in general. To solve this problem, many maze-like environments have been proposed \cite{kempka2016vizdoom,beattie2016deepmind}, but nevertheless, agents trained in synthetic environments hardly adapt to real-world scenarios, because of the drastic change in terms of visual appearance.

Simulating platforms like Habitat \cite{savva2019habitat} and Matterport3D simulator \cite{anderson2018vision} provide a photo-realistic environment to train navigation agents. Some of these simulators only provide RGB equirectangular images as visual input \cite{anderson2018vision}, while others employ the full 3D model and implement physic interactions with the environment \cite{xia2018gibson,savva2019habitat,straub2019replica,xia2020interactive,li2021igibson}. In these simulating platforms, algorithms for intelligent exploration and navigation can be developed safely and more quickly than in the real world, before being easily deployed on real robotic platforms \cite{kadian2020sim2real,anderson2021sim}.

Alongside the introduction of these simulators, a large number of task-specific datasets for \gls{eai} has been released fostering the research on multiple aspects of \gls{eai}. The main tasks covered by these datasets are \gls{pointnav} and \gls{objectnav}, \gls{vln} \cite{anderson2018vision,jain2019stay,ku2020room,krantz2020beyond,qi2020reverie,zhu2021soon}, and household tasks \cite{ALFRED20,deitke2022procthor}.

Among the other datasets, a relevant contribution is given by the release of photo-realistic indoor environments that contain semantic annotations of the objects or regions in the scene \cite{chang2017matterport3d,armeni20193d,yadav2022habitat}. Such datasets enabled multiple tasks that involve learning semantic relations and characteristics of the environment.

We aim to enlarge the possibilities enabled by the presence of specific data for \gls{eai} research by introducing two new datasets in Chapter \ref{chap:sd} and Chapter \ref{chap:gallerie}.

\section{Simulation-to-Reality Transfer}
\label{sec:sim2real_survey}
As mentioned in the previous sections, simulation allowed an impressive boost in training efficiency and final performance of embodied agents on a multitude of tasks. Some of these embodied tasks were solved, \ie Wijmans \etal \cite{wijmans2019dd} achieved nearly perfect results on noise-free \acrlong{pointnav}. Nevertheless, their model is trained using $2.5$ billion frames and requires experience acquired over more than half a year of \gls{gpu} time, and unfortunately, models tend to learn simulator-specific tricks to circumvent navigation difficulties \cite{kadian2020sim2real}. Since such shortcuts do not work in the real world, there is a significant \acrlong{sim2real} performance gap.

Recent work has studied how to deploy models trained on simulation to the real world \cite{deitke2020robothor,kadian2020sim2real,rosano2020embodied}. In their work, Kadian \etal \cite{kadian2020sim2real} make a 3D acquisition of a real-world scene and study the \acrshort{sim2real} gap for various setups and metrics. However, their environment is very simple as the obstacles are large boxes, the floor has an even and regular surface in order to facilitate the actuation system, and there are no doors or other navigation bottlenecks. 
Another recent study on \gls{sim2real} transfer is done by Truong \etal \cite{truong2022rethinking} employing complex legged robots.
Gervet \etal \cite{gervet2022navigating}, instead, extensively study the \acrlong{sim2real} transfer of agents for visual semantic navigation in a real-world apartment.

In Chapter \ref{chap:out}, we present some guidelines for the deployment of models that are trained in simulation to real robotic platforms and we test our method in a realistic environment with multiple rooms and typical office furniture.

\section{Environment Knowledge in Agents}
\label{sec:dynamic_survey}
Current research on Embodied AI for navigation agents can be categorized according to the quantity of knowledge about the environment provided to the agent prior to performing the task \cite{anderson2018evaluation}. The most common approaches focus on the scenario in which the agent is deployed in a completely new environment for which it has no prior knowledge \cite{gupta2017cognitive,chaplot2019learning,karkus2021differentiable,wijmans2019dd}. 

In a different but similar approach, the agent has no knowledge of the environment, but the exploration is run in parallel with a target-driven navigation task resulting in an effective approach to solving the latter (\eg, \gls{objectnav} \cite{chaplot2020object} and \gls{pointnav} \cite{ramakrishnan2020occupancy}). 

Other approaches consider the case in which the agent is able to exploit preacquired information about the environment \cite{da2018autonomously,chaplot2019embodied} when performing a navigation task. Such preacquired information can be either partial \cite{savinov2018semi,sridharan2020commonsense,zhang2020diagnosing} or complete \cite{chen2019learning,cartillier2020semantic,ramakrishnan2020exploration}. A major limitation of such approaches is that the obtained map representation is assumed to conform perfectly with the environment where the downstream task will be performed.

A last option is the case in which the preacquired map provided to the agent is incomplete or incorrect due to changes that occurred in the environment over time. Common strategies to deal with changing environments entail disregarding dynamic objects as landmarks when performing \gls{slam} \cite{saputra2018visual,biswas2019quest} and applying local policies to avoid them when navigating \cite{mac2016heuristic}. 
An alternative strategy is learning to predict geometric changes based on experience, as done by Nardi and Stachniss \cite{nardi2020long}, where the environment is represented as a traversability graph. The main limitation of this strategy is its computational intractability when considering dense metric maps of wide areas. 

In Chapter \ref{chap:sd} we present a learning-based method to efficiently update an obsolete map of the environment when using occupancy grids.

\section{Image Captioning}
\label{sec:captioning_survey}
In order to enable the capability of autonomous agents to provide user-under-standable representations of the perceived environment \cite{poppi2023towards}, in the form of natural language descriptions, \glspl{ic} methods need to be implemented on the robot.
\gls{ic} is a task at the intersection between vision and language whose goal is to generate a natural language description of a given image. To represent images, \gls{cnn} can be used to extract global features \cite{karpathy2015deep,rennie2017self,vinyals2017show}, grids of features \cite{xu2015show,lu2017knowing}, or features for image regions containing visual entities \cite{anderson2018bottom,lu2018neural}.
In most cases, attention mechanisms are applied to enhance the visual input representation. More recent approaches employ fully-attentive Transformer-like architectures \cite{vaswani2017attention} as visual encoders, which can also be applied directly to image patches \cite{liu2021cptr,cornia2021universal,barraco2022camel}.
The image representation is used to condition a language model that generates the caption. The language model can be implemented as a recurrent neural network \cite{karpathy2015deep,rennie2017self,huang2019attention,landi2021working} or Transformer-based fully-attentive models \cite{cornia2020meshed,zhang2021rstnet}. 

We adopt captioning models in Chapter \ref{chap:ex2} and Chapter \ref{chap:eds} to generate descriptions from the point of view of the embodied agent.

\section{Deep Generative Models}
\label{sec:generative_survey}
Deep generative models are trained to approximate high-dimensional probability distributions by means of a large set of training samples. In recent years, literature on deep generative models followed three main approaches: latent variable models like VAE \cite{kingma2013auto}, implicit generative models like GANs or the more recent Style-GAN and VQ-GAN \cite{goodfellow2014generative,karras2019style,esser2021taming}, and exact likelihood models. 
Exact likelihood models can be classified in non-autoregressive flow-based models, like RealNVP \cite{dinh2016density} and Flow++ \cite{ho2019flow++}, and autoregressive models, like PixelCNN \cite{oord2016conditional} and Image Transformer \cite{parmar2018image}. Non-autoregressive flow-based models consist of a sequence of invertible transformation functions to compose a complex distribution modeling the training data. Autoregressive models decompose the joint distribution of images as a product of conditional probabilities of the single pixels. Usually, each pixel is computed using as input only the previously predicted ones, following a raster scan order.

Regarding the work presented in this dissertation, we use a generative model in the architecture presented in Chapter \ref{chap:focus}.

\chapter[Exploration with Intrinsic Motivation]{Focus on Impact: \\ \Large Embodied Exploration with Intrinsic Motivation}
\label{chap:focus}
\blfootnote{This Chapter is related to the publication ``R. Bigazzi \etal, Focus on Impact: Indoor Exploration with Intrinsic Motivation, RA-L 2022'' \cite{bigazzi2022impact}. See the list of Publications on page \pageref{publications} for more details.}
\RemoveLabels

% \section{Introduction}
% \label{sec:introduction_impact}
\lettrine[lines=1]{\textcolor{SchoolColor}{R}}{obotic} exploration can be defined as the task of autonomously navigating an unknown environment with the goal of gathering sufficient information to represent it, often via a spatial map \cite{stachniss2009robotic}. This ability is key to enabling many downstream tasks such as planning \cite{selin2019efficient} and goal-driven navigation \cite{savva2019habitat,wijmans2019dd,morad2021embodied}. 
Although a vast portion of existing literature tackles this problem \cite{osswald2016speeding,chaplot2019learning,chen2019learning,ramakrishnan2020occupancy}, it is not yet completely solved, especially in complex indoor environments.
The introduction of large datasets of photo-realistic indoor environments \cite{chang2017matterport3d,xia2018gibson} has eased the development of robust exploration strategies\cite{rawal2023aigen}, which can be validated safely and quickly thanks to powerful simulating platforms \cite{deitke2020robothor,savva2019habitat}. Moreover, exploration algorithms developed on simulated environments can be deployed in the real world with little hyperparameter tuning \cite{kadian2020sim2real,truong2021bi}, if the simulation is sufficiently realistic. 

Most of the recently devised exploration algorithms exploit \acrfull{drl} \cite{zhu2018deep}, as learning-based exploration and navigation algorithms are more flexible and robust to noise than geometric methods \cite{chen2019learning,niroui2019deep,ramakrishnan2020exploration}. 
Despite these advantages, one of the main challenges in training DRL-based exploration algorithms is designing appropriate rewards. 
\begin{figure}[t!]
\centering
\includegraphics[width=0.82\linewidth]{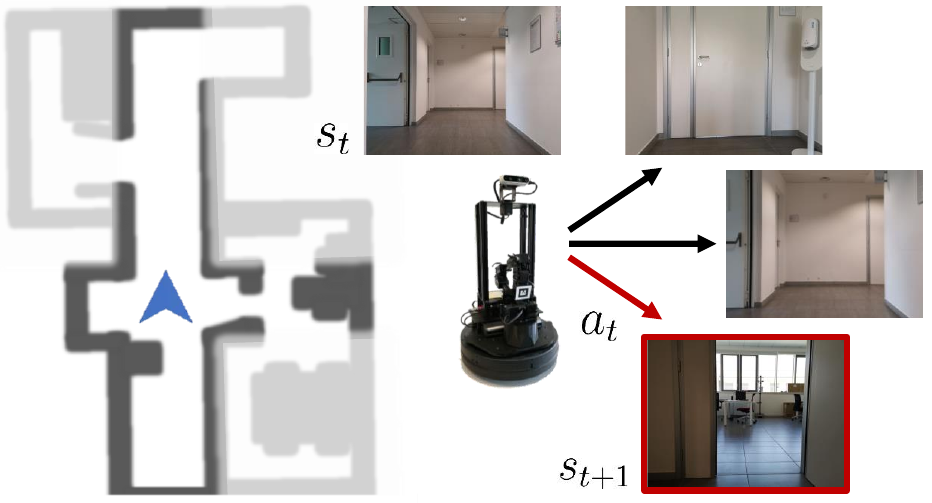}
\caption{We propose an impact-based reward for robot exploration of continuous indoor spaces. The robot is encouraged to take actions that maximize the difference between two consecutive observations.}
\label{fig:fig1_impact}
\end{figure}

In this work, we propose a new reward function that employs the impact of the agent actions on the environment, measured as the difference between two consecutive observations \cite{raileanu2020ride}, discounted with a pseudo-count \cite{bellemare2016unifying} for previously-visited states (see Fig \ref{fig:fig1_impact}).
So far, impact-based rewards \cite{raileanu2020ride} have been used only as an additional intrinsic reward in procedurally-generated (\eg~grid-like mazes) or singleton (\ie~the test environment is the same employed for training) synthetic environments.
Instead, our reward can deal with photo-realistic non-singleton environments.

\AddLabels

Recent research on robot exploration proposes the use of an extrinsic reward based on the prediction of the occupancy of the surrounding environment \cite{ramakrishnan2020occupancy}.
This type of reward encourages the agent to navigate toward areas that can be easily mapped without errors. Unfortunately, this approach presents a major drawback, as this reward heavily depends on the mapping phase, rather than focusing on what has been already seen. In fact, moving towards new places that are difficult to map would produce a low occupancy-based reward. Moreover, the precise layout of the training environments is not always available, especially in real-world applications.
To overcome these issues, a possible solution is given by the use of intrinsic reward functions, so that the agent does not need to rely on annotated data and can compute its reward by means of its observations. 
Some examples of recently proposed intrinsic rewards for robot exploration are based on curiosity \cite{bigazzi2020explore}, novelty \cite{ramakrishnan2020exploration}, and coverage \cite{chaplot2019learning}. All these rewards, however, tend to vanish with the length of the episode because the agent quickly learns to model the environment dynamics and appearance (for curiosity and novelty-based rewards) or tends to stay in previously-explored areas (for the coverage reward). Impact, instead, provides a stable reward signal throughout the episode \cite{raileanu2020ride}. 

\thispagestyle{nosection}

Since robot exploration takes place in complex and realistic environments that can present an infinite number of states, it is impossible to store a visitation count for every state. Furthermore, the vector of visitation counts would consist of a very sparse vector, and that would cause the agent to give the same impact score to nearly identical states. To overcome this issue, we introduce an additional module in our design to keep track of a pseudo-count for visited states. The pseudo-count is estimated by a density model that is trained end-to-end and together with the policy.
We integrate our newly-proposed reward in a modular embodied exploration and navigation system inspired by that proposed by Chaplot \etal \cite{chaplot2019learning} and consider two commonly adopted collections of photo-realistic simulated indoor environments, namely Gibson \cite{xia2018gibson} and \gls{mp3d} \cite{chang2017matterport3d}. Furthermore, we also deploy the devised algorithm in the real world. The results in both simulated and real environments are promising: we outperform state-of-the-art baselines in simulated experiments and demonstrate the effectiveness of our approach in real-world experiments. We make the source code of our approach and pretrained models publicly available\footnote{\url{https://github.com/aimagelab/focus-on-impact}}.

\begin{landscape}
\begin{figure}[!t]
\centering
\includegraphics[width=0.95\linewidth]{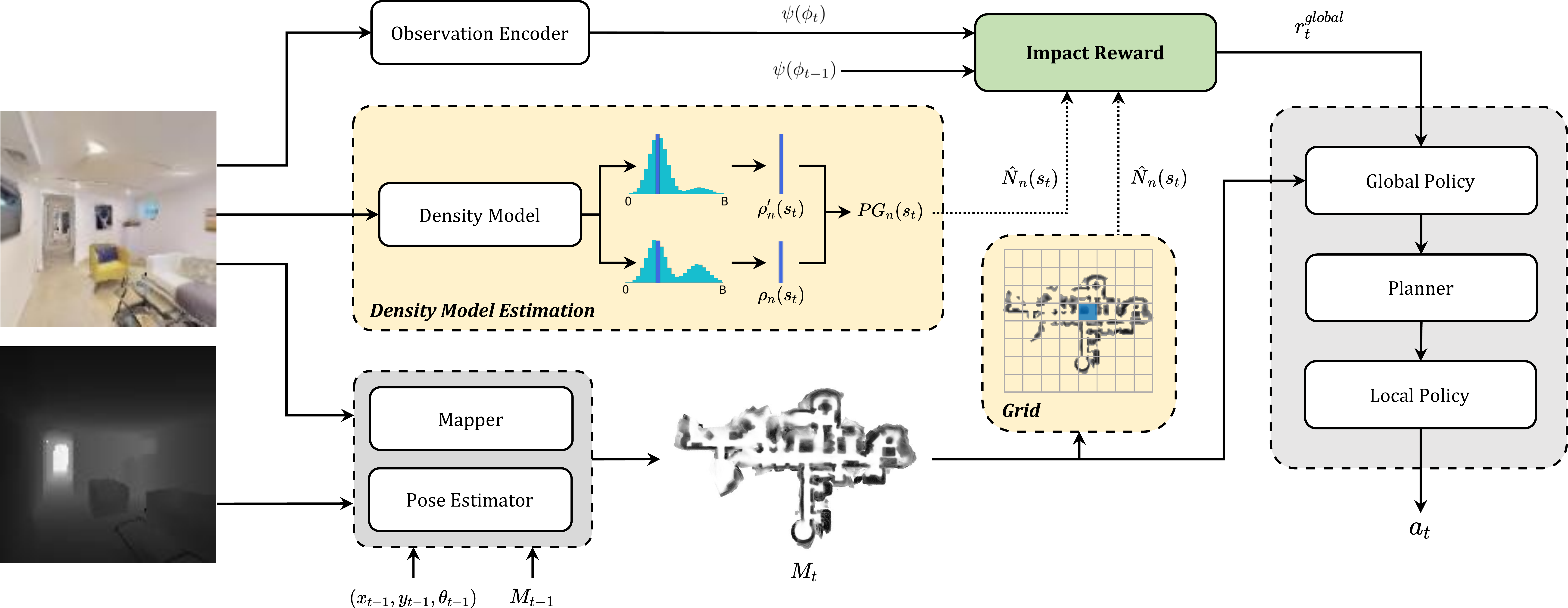}
\caption{Our modular exploration architecture consists of a Mapper that iteratively builds a top-down occupancy map of the environment, a Pose Estimator that predicts the pose of the robot at every step, and a hierarchical self-supervised Navigation Module in charge of sequentially setting exploration goals and predicting actions to navigate towards it. We exploit the impact-based reward to guide the exploration and adapt it for continuous environments, using an Observation Encoder to extract observation features and depending on the method, a Density Model or a Grid to compute the pseudo-count.}
\label{fig:fig2_impact}
\end{figure}
\end{landscape}

\section{Proposed Method}
\label{sec:method_impact}

Following the current state-of-the-art architectures for navigation for embodied agents \cite{chaplot2019learning,ramakrishnan2020occupancy}, the proposed method comprises three major components: a \gls{cnn}-based mapper, a pose estimator, and a hierarchical navigation policy. The navigation policy defines the actions of the agent, the mapper builds a top-down map of the environment to be used for navigation, and the pose estimator locates the position of the agent on the map. Our architecture is depicted in Fig. \ref{fig:fig2_impact} and described below.

\subsection{Mapper}
\label{ssec:mapper_impact}
The mapper generates a map of the free and occupied regions of the environment discovered during the exploration. 
At each timestep, the RGB observation $\phi^{rgb}_t$ and the depth observation $\phi^{d}_t$ are processed to output a two-channel $V\times V$ local map $m_t$ depicting the area in front of the agent, where each cell describes the state of a $5\times 5$ cm area of the environment, the channels measure the probability of a cell being occupied and being explored, as in \cite{chaplot2019learning}.

Specifically, the RGB observation $\phi^{rgb}_t$ is encoded using the first two blocks of ResNet-18 \cite{he2016deep} pretrained on ImageNet, followed by a three-layered CNN. We project the depth image $\phi^d_t$ using the camera intrinsics \cite{chen2019learning} and obtain a preliminary map for the visible occupancy. We name the obtained feature representations $\hat{\phi}^{rgb}_t$ and $\hat{\phi}^d_t$, respectively. We then encode the two feature maps using a U-Net \cite{ronneberger2015u}:
\begin{equation}
    f_\mu(\hat{\phi}^{rgb}_t, \hat{\phi}^d_t) = \text{U-Net}_\text{enc}(\hat{\phi}^{rgb}_t, \hat{\phi}^d_t, \mu_\text{enc}) , 
\end{equation}
and decode the $2 \times V \times V$ matrix of probabilities as:
\begin{equation}
    v_t = \sigma(\text{U-Net}_\text{dec}(f_\mu(\hat{\phi}^{rgb}_t, \hat{\phi}^d_t), \mu_\text{dec})) ,
\end{equation}
where $\mu_\text{enc}$ and $\mu_\text{dec}$ represent the learnable parameters in the U-Net encoder and decoder, respectively, and $\sigma$ is the sigmoid activation function.

Please note that this module performs anticipation-based mapping, as defined in \cite{ramakrishnan2020occupancy}, where the predicted local map $m_t$ includes also unseen/occluded portions of space.
The local maps are aggregated and registered to the $W\times W\times 2$ global map $M_t$ of the environment.
To that end, we use a geometric transformation to project $v_t$ in the global coordinate system, for which we need a triple $(x,y,\theta)$ corresponding to the agent position and heading in the environment. This triple is estimated by a specific component that tracks the agent displacements across the environment, as discussed in the following paragraph.
The resulting global map is used by the navigation policy for action planning and in this setting, it is initially empty and is built incrementally with the exploration of the environment.

\subsection{Pose Estimator} 
\label{ssec:poseest_impact}
The pose estimator is used to predict the displacement of the agent as a consequence of an action. The considered atomic actions $a_t$ of the agent are: \textit{go forward 0.25m, turn left 10°, turn right 10°}. However, the noise in the actuation system and the possible physical interactions between the agent and the environment could produce unexpected outcomes causing positioning errors. The pose estimator reduces the effect of such errors by predicting the real displacement $(\Delta x_t, \Delta y_t, \Delta\theta_t)$. 
According to \cite{ramakrishnan2020occupancy}, the input of this module consists of the RGB-D observations $(\phi^{rgb}_{t-1}, \phi^{rgb}_{t})$ and $(\phi^d_{t-1}, \phi^d_t)$ and the local maps $(m_{t-1}, m_t)$. Each modality $i = \{0, 1, 2\}$ is encoded singularly to obtain three different estimates of the displacement:
\begin{equation}
    p_i(e_{t-1}, e_t) = W_1 \text{max}(W_2(e_{t-1}, e_t) + b_2,0) + b_1,
    \label{eq:g_i}
\end{equation}
where $e_t \in \{\phi^{rgb}_t, \phi^d_t, m_t\}$ and $W_{1,2}$ and $b_2$ are weights matrices and bias. Eventually, the displacement estimates are aggregated with a weighted sum:
\begin{align}
    \alpha_i = \text{softmax}(\text{\acrshort{mlp}}_i([\bar{\phi}^{rgb}_t, \bar{\phi}^d_t, \bar{m}_t])), \\
    (\Delta x_t,\Delta y_t,\Delta \theta_t) = \sum_{i=0}^{2}{\alpha_i \cdot p_i},
\label{eq:displacement}
\end{align}
where \gls{mlp} is a three-layered fully-connected network, ($\bar{\phi}^{rgb}_t$, $\bar{\phi}^d_t$, $\bar{l}_t$) are the inputs encoded by a \gls{cnn}, and $[\cdot, \cdot, \cdot]$ denotes tensor concatenation. The estimated pose of the agent at time $t$ is given by:
\begin{equation}
(x_t, y_t, \theta_t) = (x_{t-1}, y_{t-1}, \theta_{t-1}) + (\Delta x_t,\Delta y_t,\Delta \theta_t).
\label{eq:pose}
\end{equation}
Note that, at the beginning of each exploration episode, the agent sets its position to the center of its environment representation, \ie $(x_0, y_0, \theta_0) = (0, 0, 0)$.

\subsection{Navigation Policy}
\label{ssec:navpolicy_impact}
The sampling of the atomic actions of the agent relies on the hierarchical navigation policy that is composed of the following modules: the global policy, the planner, and the local policy. This architecture is in line with current literature on embodied exploration \cite{chaplot2019learning,ramakrishnan2020occupancy,ramakrishnan2022poni}. 
The hierarchical policy is adopted to decouple high-level and low-level concepts like moving across rooms and avoiding obstacles. It samples a goal coordinate on the map, while the deterministic planner uses the global goal to compute a local goal in close proximity of the agent. The local policy then predicts actions to reach the local goal. 

The global policy samples a point on an augmented global map of the environment, $M_t^+$, that represents the current global goal of the agent. The augmented global map $M_t^+$ is a $W\times W\times 4$ map obtained by stacking the two-channel global map $M_t$ from the Mapper with the one-hot representation of the agent position $(x_t, y_t)$ and the map of the visited positions, which collects the one-hot representations of all the positions assumed by the agent from the beginning of the exploration. 
Moreover, $M_t^+$ is in parallel cropped with respect to the position of the agent and max-pooled to a spatial dimension $H\times H$ where $H<W$. These two versions of the augmented global map are concatenated to form the $H\times H\times 8$ input of the global policy that is used to sample a goal in the global action space $H\times H$. The global policy is trained with reinforcement learning with our proposed impact-based reward $r^{global}_t$, defined below, that encourages exploration. 

The deterministic planner adopts the A* algorithm to compute a feasible trajectory from the agent's current position to the global goal using the current state of the map $m_t$. A point on the trajectory within $1.25$m from the agent is extracted to form the local goal $l_t$.

The local policy outputs the atomic actions needed to reach the local goal and is trained to minimize the euclidean distance to the local goal, which is expressed via the following reward:

\begin{equation}
r^{local}_t = d(\omega_{t-1}, l_{t}, M_{t-1}) - d(\omega_{t}, l_{t}, M_{t}),
\label{eq:r_local}
\end{equation} 
where $d(\cdot,\cdot,\cdot)$ is the function computing the euclidean distance between two positions using the map $M_t$ to take into account possible obstacles on the way, while $\omega=(x_t, y_t, \theta_t)$ and $l_t$ are respectively, the pose of the agent and the local goal at timestep $t$.
Note that the output actions in our setup are discrete. These platform-agnostic actions can be translated into signals for specific robot actuators, as we do in this work. Alternatively, based on the high-level predicted commands, continuous actions can be predicted, \eg~in the form of linear and angular velocity commands to the robot, by using an additional, lower-level policy, as done in \cite{irshad2021hierarchical}. The implementation of such a policy is beyond the scope of our work.

Following the hierarchical structure, the global goal is reset every $\eta$ steps, and the local goal is reset if at least one of the following conditions verifies: a new global goal is sampled, the agent reaches the local goal, the local goal location is discovered to be in an occupied area.

\subsection{Impact-Driven Exploration}
The exploration ability of the agent relies on the design of an appropriate reward for the global policy. In this setting, the lack of external rewards from the environment requires the design of a dense intrinsic reward. To the best of our knowledge, our proposed method presents the first implementation of impact-driven exploration in photo-realistic environments.
The key idea of this concept is encouraging the agent to perform actions that have impact on the environment and the observations retrieved from it, where the impact at timestep $t$ is measured as the $l_2$-norm of the encodings of two consecutive states $\psi(s_{t})$ and $\psi(s_{t+1})$, considering the RGB observation $\phi_t^{rgb}$ to compute the state $s_{t}$. Following the formulation proposed in \cite{raileanu2020ride}, the reward of the global policy for the proposed method is calculated as:
\begin{equation}
    r^{global}_t(s_{t}, s_{t+1}) = \frac{\left\| \psi(s_{t+1}) - \psi(s_{t}) \right\|_2}{\sqrt{N(s_{t+1})}},
\label{eq:global_reward}
\end{equation}
where $N(s_{t})$ is the visitation count of the state at timestep $t$, \ie~how many times the agent has observed $s_{t}$. The visitation count is used to drive the agent out of regions already seen in order to avoid trajectory cycles. Note that the visitation count is episodic, \ie~$N_{ep}(s_t) \equiv N(s_t)$. For simplicity, in the following, we denote the episodic visitation count as $N(s_t)$.

\tit{Visitation Counts} The concept of normalizing the reward using visitation count, as in \cite{raileanu2020ride}, fails when the environment is continuous since during exploration is unlikely to visit exactly the same state more than once. In fact, even microscopic changes in terms of translation or orientation of the agent cause shifts in the values of the RGB observation, thus resulting in new states.
Therefore, using a photo-realistic continuous environment nullifies the scaling property of the denominator of the global reward in Eq. \ref{eq:global_reward} because every state $s_t$ during the exploration episode is, most of the times, only encountered for the first time. To overcome this limitation, we implement two types of pseudo-visitation counts $\hat{N}(s_t)$ to be used in place of $N(s_t)$, which extend the properties of visitation counts to continuous environments: \textit{Grid} and \textit{Density Model Estimation}.

\tit{Grid} With this approach, we consider a virtual discretized grid of cells with fixed size in the environment. We then assign a visitation count to each cell of the grid. 
Note that, different from approaches working on procedurally-generated environments like \cite{raileanu2020ride}, the state space of the environment we consider is continuous also in this formulation, and depends on the pose of the agent $(x,y,\theta)$. The grid approach operates a quantization of the agent's positions, and that allows to cluster observation made from similar positions. 
To this end, we take the global map of the environment and divide it into cells of size $G\times G$. The estimated pose of the agent, regardless of its orientation $\theta_t$, is used to select the cell that the agent occupies at time $t$. 
In the \textit{Grid} formulation, the visitation count of the selected cell is used as $N(s_t)$ in Eq. \ref{eq:global_reward} and is formalized as:
\begin{equation}
    \hat{N}(s_t) = \hat{N}(\text{grid}(x_t, y_t)),
\label{eq:pseudo_grid}
\end{equation}
where $\text{grid}(\cdot)$ returns the block corresponding to the estimated position of the agent. 

\tit{Density Model Estimation (\gls{dme})}
Let $\rho$ be an autoregressive density model defined over the states $s \in S$, where $S$ is the set of all possible states.
We call $\rho_n(s)$ the probability assigned by $\rho$ to the state $s$ after being trained on a sequence of states $s_1, ..., s_n$, and $\rho'_n(s)$, or recoding probability \cite{bellemare2016unifying,ostrovski2017count}, the probability assigned by $\rho$ to $s$ after being trained on $s_1, ..., s_n, s$.
The prediction gain $\acrshort{pg}$ of $\rho$ describes how much the model has improved in the prediction of $s$ after being trained on $s$ itself, and is defined as
\begin{equation}
    \acrshort{pg}_n(s) = \log \rho'_n(s) - \log \rho_n(s).
\label{eq:prediction_gain}
\end{equation}
In this work, we employ a lightweight version of Gated PixelCNN \cite{oord2016conditional} as density model. This model is trained from scratch along with the exploration policy using the states visited during the exploration, which are fed to PixelCNN one at a time, as they are encountered.
The weights of PixelCNN are optimized continually over all the environments. As a consequence, the knowledge of the density model is not specific for a particular environment or episode.
To compute the input of the PixelCNN model, we transform the RGB observation $\phi^{rgb}_t$ to grayscale and we crop and resize it to a lower size $P\times P$. The transformed observation is quantized to $B$ bins to form the final input to the model, $s_t$. The model is trained to predict the conditional probabilities of the pixels in the transformed input image, with each pixel depending only on the previous ones following a raster scan order. The output has shape $P\times P\times B$ and each of its elements represents the probability of a pixel belonging to each of the $B$ bins. The joint distribution of the input modeled by PixelCNN is:
\begin{equation}
    p(s_t) = \prod_1^{P^2}p(\chi_i|\chi_1,...,\chi_{i-1}),
    \label{eq:pixel_cnn}
\end{equation}
where $\chi_i$ is the $i^{\textit{th}}$ pixel of the image $s_t$. $\rho$ is trained to fit $p(s_t)$ by using the negative log-likelihood loss. 

Let $\hat{n}$ be the pseudo-count total, \ie~the sum of all the visitation counts of all states during the episode. The probability and the recoding probability of $s$ can be defined as: 
\begin{align}
    \rho_n(s) &= \frac{\hat{N}_n(s)}{\hat{n}}, & \rho'_n(s) &= \frac{\hat{N}_n(s) + 1}{\hat{n} + 1}.
\label{eq:probabilities}
\end{align}
Note that, if $\rho$ is learning-positive, \ie~if $\acrshort{pg}_n(s) > 0$ for all possible sequences $s_1,...,s_n$ and all $s \in S$, we can approximate $\hat{N}_n(s)$ as: 
\begin{equation}
    \hat{N}_n(s) = \frac{\rho_n(s)(1-\rho'_n(s))}{\rho'_n(s)-\rho_n(s)} \approx (e^{\acrshort{pg}_n(s)} - 1)^{-1}.
\label{eq:pseudo_count}
\end{equation}

To use this approximation in Eq. \ref{eq:global_reward}, we still need to address three problems: it does not scale with the length of the episode, the density model could be not learning-positive, and $\hat{N}_n(s)$ should be large enough to avoid the reward becoming too large regardless the goal selection. In this respect, to take into account the length of the episode, we introduce a normalizing factor $n^{-1/2}$, where $n$ is the number of steps done by the agent since the start of the episode. Moreover, to force $\rho$ to be learning-positive, we clip $\acrshort{pg}_n(s)$ to 0 when it becomes negative.
Finally, to avoid small values at the denominator of $r^{global}_t$ (Eq. \ref{eq:global_reward}), we introduce a lower bound of 1 to the pseudo visitation count.
The resulting definition of $\hat{N}_n(s)$ in the \textit{\acrlong{dme}} formulation is:
\begin{align}
    \widetilde{\acrshort{pg}}_n &= c \cdot n^{-1/2} \cdot (\acrshort{pg}_n(s))_+, \\
    \hat{N}_n(s) &= \max \Big\{\Big(e^{\widetilde{\acrshort{pg}}_n(s)}-1\Big)^{-1}, 1\Big\},
\label{eq:pseudo_count_final}
\end{align}
where $c$ is a term used to scale the prediction gain.
It is worth noting that, unlike the Grid approach that can be applied only when $s_t$ is representable as the robot location, the \gls{dme} can be adapted to a wider range of tasks, including settings where the agent alters the environment.

\section{Experimental Setup}
\label{sec:experiments_impact}

\subsection{Datasets} For comparison with state-of-the-art \gls{drl}-based methods for embodied exploration, we employ the photo-realistic simulated 3D environments contained in the Gibson dataset \cite{xia2018gibson} and the \acrshort{mp3d} dataset \cite{chang2017matterport3d}. Both these datasets consist of indoor environments where different exploration episodes take place. In each episode, the robot starts exploring from a different point in the environment.
Environments used during training do not appear in the validation/test split of these datasets.
Gibson contains $106$ scans of different indoor locations, for a total of around $5$M exploration episodes ($14$ locations are used in $994$ episodes for test in the so-called Gibson Val split). 
\acrshort{mp3d} consists of $90$ scans of large indoor environments ($11$ of those are used in $495$ episodes for the validation split and $18$ in $1008$ episodes for the test split).

\subsection{Evaluation Protocol}
We train our models on the Gibson train split. Then, we perform model selection basing on the results obtained on Gibson Val. We then employ the \acrshort{mp3d} validation and test splits to benchmark the generalization abilities of the agents.
To evaluate exploration agents, we employ the following metrics. \Acrlong{iou} (\textbf{$\mathsf{\acrshort{iou}}$}) between the reconstructed map and the ground-truth map of the environment: here we consider two different classes for every pixel in the map (free or occupied). Similarly, the map \acrlong{acc} (\textbf{$\mathsf{\acrshort{acc}}$}, expressed in $m^2$) is the portion of the map that has been correctly mapped by the agent. The \acrlong{as} (\textbf{$\mathsf{\acrshort{as}}$}, in $m^2$) is the total area of the environment observed by the agent. For both the \textbf{$\mathsf{\acrshort{iou}}$} and the \acrlong{as}, we also present the results relative to the two different classes: free space and occupied space respectively (\textbf{$\mathsf{\acrshort{fiou}}$}, \textbf{$\mathsf{\acrshort{oiou}}$}, \textbf{$\mathsf{\acrshort{fas}}$}, \textbf{$\mathsf{\acrshort{oas}}$}). Finally, we report the mean positioning error achieved by the agent at the end of the episode. A larger translation error (\textbf{$\mathsf{\acrshort{te}}$}, expressed in $m$) or angular error (\textbf{$\mathsf{\acrshort{ae}}$}, in degrees) indicates that the agent struggles to keep a correct estimate of its position throughout the episode. For all the metrics, we consider episodes of length $T=500$ and $T=1000$ steps.

For our comparisons, we consider five baselines trained with different rewards. \textit{Curiosity} employs a surprisal-based intrinsic reward as defined in \cite{pathak2017curiosity}. \textit{Coverage} and \textit{Anticipation} are trained with the corresponding coverage-based and accuracy-based rewards defined in \cite{ramakrishnan2020occupancy}. For completeness, we include two count-based baselines, obtained using the reward defined in Eq. \ref{eq:global_reward}, but ignoring the contribution of impact (\ie setting the numerator to a constant value of 1). These are \textit{Count (Grid)} and \textit{Count (\gls{dme})}. All the baselines share the same overall architecture and training setup of our main models.

\begin{table}[t]
\footnotesize
\centering
\setlength{\tabcolsep}{.35em}
\resizebox{.9\linewidth}{!}{
\begin{tabular}{l c ccccccccc}
\toprule
\textbf{Model} & & \textbf{$\mathsf{\acrshort{iou}}$} $\uparrow$ & \textbf{$\mathsf{\acrshort{fiou}}$} $\uparrow$ & \textbf{$\mathsf{\acrshort{oiou}}$} $\uparrow$ & \textbf{$\mathsf{\acrshort{acc}}$} $\uparrow$ & \textbf{$\mathsf{\acrshort{as}}$} $\uparrow$ & \textbf{$\mathsf{\acrshort{fas}}$} $\uparrow$ & \textbf{$\mathsf{\acrshort{oas}}$} $\uparrow$ & \textbf{$\mathsf{\acrshort{te}}$} $\downarrow$ & \textbf{$\mathsf{\acrshort{ae}}$} $\downarrow$ \\
\midrule
\textbf{Grid} \\
\textit{  G = 2} & & 0.726 & 0.721 & 0.730 & 51.41 & 61.88 & 34.17 & 27.71 & 0.240 & 4.450 \\
\textit{  G = 4} & & 0.796 & 0.792 & 0.801 & 54.34 & 61.17 & 33.74 & 27.42 & 0.079 & 1.055 \\
\textit{  G = 5} & & \textbf{0.806} & \textbf{0.801} & \textbf{0.813} & \textbf{55.21} & \textbf{62.17} & \textbf{34.31} & \textbf{27.87} & \textbf{0.077} & \textbf{0.881} \\
\textit{  G = 10} & & 0.789 & 0.784 & 0.794 & 54.26 & 61.67 & 34.06 & 27.61 & 0.111 & 1.434 \\
\midrule
\textbf{\acrshort{dme}} \\
\textit{  B = 64} & & 0.773 & 0.768 & 0.778 & 53.58 & 61.00 & 33.79 & 27.21 & 0.131 & 2.501 \\
\textit{  B = 128} & & \textbf{0.796} & \textbf{0.794} & \textbf{0.799} & \textbf{54.73} & \textbf{62.07} & \textbf{34.27} & \textbf{27.79} & \textbf{0.095} & \textbf{1.184} \\
\textit{  B = 256} & & 0.685 & 0.676 & 0.695 & 49.27 & 61.40 & 33.95 & 27.45 & 0.311 & 6.817 \\
\bottomrule
\end{tabular}
}
% \captionsetup{justification=centering}
\caption{Results for our model selection on Gibson validation split for $T=500$. $G=5$ and $B=128$ are the best models for the Grid-based and the DME-based models, respectively.}
\label{tab:tab1_impact}
\end{table}

\begin{table*}[!t]
\footnotesize
\centering
\setlength{\tabcolsep}{.32em}
\resizebox{\linewidth}{!}{
\begin{tabular}{>{\color{black}}l>{\color{black}}c >{\color{black}}c>{\color{black}}c>{\color{black}}c>{\color{black}}c>{\color{black}}c>{\color{black}}c>{\color{black}}c>{\color{black}}c>{\color{black}}c>{\color{black}}c >{\color{black}}c}
\toprule
& & \multicolumn{9}{>{\color{black}}c}{\textbf{Gibson Val (T = 500)}} \\
\cmidrule{3-11}
\textbf{Model} & & \textbf{$\mathsf{\acrshort{iou}}$} $\uparrow$ & \textbf{$\mathsf{\acrshort{fiou}}$} $\uparrow$ & \textbf{$\mathsf{\acrshort{oiou}}$} $\uparrow$ & \textbf{$\mathsf{\acrshort{acc}}$} $\uparrow$ & \textbf{$\mathsf{\acrshort{as}}$} $\uparrow$ & \textbf{$\mathsf{\acrshort{fas}}$} $\uparrow$ & \textbf{$\mathsf{\acrshort{oas}}$} $\uparrow$ & \textbf{$\mathsf{\acrshort{te}}$} $\downarrow$ & \textbf{$\mathsf{\acrshort{ae}}$} $\downarrow$ \\
\midrule
Curiosity
& & 0.678 & 0.669 & 0.688 & 49.35 & 61.67 & 34.16 & 27.51 & 0.330 & 7.430 \\
Coverage 
& & 0.721 & 0.715 & 0.726 & 51.47 & 61.13 & 34.07 & 27.06 & 0.272 & 5.508 \\
Anticipation 
& & 0.783 & 0.778 & 0.789 & 54.68 & 60.96 & 34.15 & 26.81 & 0.100 & 1.112 \\
\midrule
Count (Grid)
& & 0.714 & 0.706 & 0.721 & 50.85 & 61.61 & 34.17 & 27.44 & 0.258 & 5.476 \\
Count (\acrshort{dme})
& & 0.764 & 0.757 & 0.772 & 52.81 & 60.69 & 33.68 & 27.01 & 0.148 & 2.888 \\
\midrule
\textbf{Impact (Grid)}
& & \textbf{0.803} & \textbf{0.797} & \textbf{0.809} &          54.94 & 61.90 & 34.07 & 27.83 & \textbf{0.079} & \textbf{0.878} \\
\textbf{Impact (\acrshort{dme})}
& & 0.800 & 0.796 & 0.803 & \textbf{55.10} & \textbf{62.59} & \textbf{34.45} & \textbf{28.14} & 0.095 & 1.166 \\
\midrule
& & \multicolumn{9}{>{\color{black}}c}{\textbf{\acrshort{mp3d} Val (T = 500)}} \\
\cmidrule{3-11}
\textbf{Model} & & \textbf{$\mathsf{\acrshort{iou}}$} $\uparrow$ & \textbf{$\mathsf{\acrshort{fiou}}$} $\uparrow$ & \textbf{$\mathsf{\acrshort{oiou}}$} $\uparrow$ & \textbf{$\mathsf{\acrshort{acc}}$} $\uparrow$ & \textbf{$\mathsf{\acrshort{as}}$} $\uparrow$ & \textbf{$\mathsf{\acrshort{fas}}$} $\uparrow$ & \textbf{$\mathsf{\acrshort{oas}}$} $\uparrow$ & \textbf{$\mathsf{\acrshort{te}}$} $\downarrow$ & \textbf{$\mathsf{\acrshort{ae}}$} $\downarrow$ \\
\midrule
Curiosity
& & 0.339 & 0.473 & 0.205 & 97.82 & 118.13 & 75.73 & 42.40 & 0.566 & 7.290 \\
Coverage 
& & 0.352 & 0.494 & 0.210 & 102.05 & 120.00 & 76.78 & 43.21 & 0.504 & 5.822 \\
Anticipation 
& & 0.381 & 0.530 & 0.231 & 106.02 & 114.06 & 72.94 & 41.13 & 0.151 & 1.280 \\
\midrule
Count (Grid)
& & 0.347 & 0.488 & 0.206 & 99.00 & 116.77 & 75.00 & 41.76 & 0.466 & 5.828 \\
Count (\acrshort{dme})
& & 0.359 & 0.493 & 0.225 & 101.73 & 112.65 & 72.22 & 40.43 & 0.268 & 3.318 \\
\midrule
\textbf{Impact (Grid)} 
& &          0.383 &          0.531 & \textbf{0.234} &          107.41 & 116.60 & 74.44 & 42.17 & \textbf{0.120} & \textbf{0.860} \\
\textbf{Impact (\acrshort{dme})}
& & \textbf{0.396} & \textbf{0.560} & 0.233 & \textbf{111.61} & \textbf{124.06} &  \textbf{79.47} & \textbf{44.59} & 0.232 & 1.988 \\
\midrule
& & \multicolumn{9}{>{\color{black}}c}{\textbf{\acrshort{mp3d} Test (T = 500)}} \\
\cmidrule{3-11}
\textbf{Model} & & \textbf{$\mathsf{\acrshort{iou}}$} $\uparrow$ & \textbf{$\mathsf{\acrshort{fiou}}$} $\uparrow$ & \textbf{$\mathsf{\acrshort{oiou}}$} $\uparrow$ & \textbf{$\mathsf{\acrshort{acc}}$} $\uparrow$ & \textbf{$\mathsf{\acrshort{as}}$} $\uparrow$ & \textbf{$\mathsf{\acrshort{fas}}$} $\uparrow$ & \textbf{$\mathsf{\acrshort{oas}}$} $\uparrow$ & \textbf{$\mathsf{\acrshort{te}}$} $\downarrow$ & \textbf{$\mathsf{\acrshort{ae}}$} $\downarrow$ \\
\midrule
Curiosity
& & 0.362 & 0.372 & 0.352 & 109.66 & 130.48 & 85.98 & 44.50 & 0.620 & 7.482 \\
Coverage 
& & 0.390 & 0.401 & 0.379 & 116.71 & 134.89 & 88.15 & 46.75 & 0.564 & 5.938 \\
Anticipation 
& & 0.424 & 0.433 & \textbf{0.415} & 117.87 & 124.24 & 81.31 & 42.93 & 0.151 & 1.306 \\
\midrule
Count (Grid)
& & 0.364 & 0.381 & 0.348 & 117.50 & 134.85 & 89.81 & 45.05 & 0.525 & 5.790 \\
Count (\acrshort{dme})
& & 0.391 & 0.397 & 0.385 & 114.02 & 123.86 & 81.86 & 42.00 & 0.287 & 3.322 \\
\midrule
\textbf{Impact (Grid)}
& & 0.420 & 0.430 & 0.409 & 124.44 & 130.98 & 86.08 & 44.90 & \textbf{0.124} & \textbf{0.834} \\
\textbf{Impact (\acrshort{dme})}
& & \textbf{0.426} & \textbf{0.444} & 0.409 & \textbf{133.51} & \textbf{144.64} & \textbf{95.70}  & \textbf{48.94} & 0.288 & 2.312 \\
\bottomrule
\end{tabular}
}
% \captionsetup{justification=centering}
\caption{Exploration results on Gibson and \acrshort{mp3d} datasets, at $T=500$ timesteps. On the small environments of Gibson Val, Impact (Grid) is the best-performing model, while on larger environments like those of MP3D, Impact (DME) achieves the best results.}
\label{tab:tab2_impact}
\end{table*}

\begin{table*}[!t]
\footnotesize
\centering
\setlength{\tabcolsep}{.32em}
\resizebox{\linewidth}{!}{
\begin{tabular}{>{\color{black}}l>{\color{black}}c >{\color{black}}c>{\color{black}}c>{\color{black}}c>{\color{black}}c>{\color{black}}c>{\color{black}}c>{\color{black}}c>{\color{black}}c>{\color{black}}c>{\color{black}}c >{\color{black}}c}
\toprule
& & \multicolumn{9}{>{\color{black}}c}{\textbf{Gibson Val (T = 1000)}} \\
\cmidrule{3-11}
\textbf{Model} & & \textbf{$\mathsf{\acrshort{iou}}$} $\uparrow$ & \textbf{$\mathsf{\acrshort{fiou}}$} $\uparrow$ & \textbf{$\mathsf{\acrshort{oiou}}$} $\uparrow$ & \textbf{$\mathsf{\acrshort{acc}}$} $\uparrow$ & \textbf{$\mathsf{\acrshort{as}}$} $\uparrow$ & \textbf{$\mathsf{\acrshort{fas}}$} $\uparrow$ & \textbf{$\mathsf{\acrshort{oas}}$} $\uparrow$ & \textbf{$\mathsf{\acrshort{te}}$} $\downarrow$ & \textbf{$\mathsf{\acrshort{ae}}$} $\downarrow$ \\
\midrule
Curiosity
& & 0.560 & 0.539 & 0.581 & 45.71 & 67.64 & 37.19 & 30.45 & 0.682 & 14.862 \\
Coverage 
& & 0.653 & 0.641 & 0.664 & 50.10 & 66.15 & 36.77 & 29.38 & 0.492 & 10.796 \\
Anticipation 
& & 0.773 & 0.763 & 0.782 & 56.37 & 66.61 & 37.17 & 29.44 & 0.155 & 1.876 \\
\midrule
Count (Grid)
& & 0.608 & 0.592 & 0.624 & 48.22 & 67.80 & 37.31 & 30.50 & 0.520 & 10.996 \\
Count (\acrshort{dme})
& & 0.708 & 0.694 & 0.722 & 52.67 & 66.91 & 36.81 & 30.12 & 0.282 & 5.802 \\
\midrule
\textbf{Impact (Grid)}
& & \textbf{0.802} & \textbf{0.793} & \textbf{0.811} & \textbf{57.21} & 67.74 & 37.04 & 30.69 & \textbf{0.119} & \textbf{1.358} \\
\textbf{Impact (\acrshort{dme})}
& & 0.789 & 0.783 & 0.796 &          56.77 & \textbf{68.34} & \textbf{37.42} & \textbf{30.92} & 0.154 & 1.958 \\
\midrule
& & \multicolumn{9}{>{\color{black}}c}{\textbf{\acrshort{mp3d} Val (T = 1000)}} \\
\cmidrule{3-11}
\textbf{Model} & & \textbf{$\mathsf{\acrshort{iou}}$} $\uparrow$ & \textbf{$\mathsf{\acrshort{fiou}}$} $\uparrow$ & \textbf{$\mathsf{\acrshort{oiou}}$} $\uparrow$ & \textbf{$\mathsf{\acrshort{acc}}$} $\uparrow$ & \textbf{$\mathsf{\acrshort{as}}$} $\uparrow$ & \textbf{$\mathsf{\acrshort{fas}}$} $\uparrow$ & \textbf{$\mathsf{\acrshort{oas}}$} $\uparrow$ & \textbf{$\mathsf{\acrshort{te}}$} $\downarrow$ & \textbf{$\mathsf{\acrshort{ae}}$} $\downarrow$ \\
\midrule
Curiosity
& & 0.336 & 0.449 & 0.223 & 109.79 & 157.27 & 100.07 & 57.20 & 1.322 & 14.540 \\
Coverage 
& & 0.362 & 0.492 & 0.232 & 116.58 & 158.83 & 100.76 & 58.07 & 1.072 & 11.624 \\
Anticipation 
& & 0.420 & 0.568 & 0.272 & 126.86 & 147.33 & 93.56 & 53.78 & 0.267 & 2.436 \\
\midrule
Count (Grid)
& & 0.350 & 0.474 & 0.226 & 112.75 & 157.13 & 100.03 & 57.10 & 1.074 & 11.686 \\
Count (\acrshort{dme})
& & 0.379 & 0.505 & 0.254 & 119.07 & 149.62 & 95.16 & 54.46 & 0.590 & 6.544 \\
\midrule
\textbf{Impact (Grid)} 
& & \textbf{0.440} & \textbf{0.595} & \textbf{0.285} & \textbf{133.97} & 157.19 & 99.61 & 57.58 & \textbf{0.202} & \textbf{1.294} \\
\textbf{Impact (\acrshort{dme})}
& &          0.427 &          0.587 & 0.268 &          133.27 & \textbf{166.20} & \textbf{105.69} & \textbf{60.50} & 0.461 & 3.654 \\
\midrule
& & \multicolumn{9}{>{\color{black}}c}{\textbf{\acrshort{mp3d} Test (T = 1000)}} \\
\cmidrule{3-11}
\textbf{Model} & & \textbf{$\mathsf{\acrshort{iou}}$} $\uparrow$ & \textbf{$\mathsf{\acrshort{fiou}}$} $\uparrow$ & \textbf{$\mathsf{\acrshort{oiou}}$} $\uparrow$ & \textbf{$\mathsf{\acrshort{acc}}$} $\uparrow$ & \textbf{$\mathsf{\acrshort{as}}$} $\uparrow$ & \textbf{$\mathsf{\acrshort{fas}}$} $\uparrow$ & \textbf{$\mathsf{\acrshort{oas}}$} $\uparrow$ & \textbf{$\mathsf{\acrshort{te}}$} $\downarrow$ & \textbf{$\mathsf{\acrshort{ae}}$} $\downarrow$ \\
\midrule
Curiosity
& & 0.361 & 0.365 & 0.357 & 130.10 & 185.36 & 121.65 & 63.71 & 1.520 & 14.992 \\
Coverage 
& & 0.409 & 0.418 & 0.399 & 142.86 & 193.20 & 126.21 & 66.99 & 1.240 & 11.814 \\
Anticipation 
& & 0.484 & 0.491 & 0.478 & 153.83 & 174.76 & 114.29 & 60.47 & 0.289 & 2.356 \\
\midrule
Count (Grid)
& & 0.377 & 0.391 & 0.363 & 144.26 & 194.76 & 129.22 & 65.53 & 1.246 & 11.608 \\
Count (\acrshort{dme})
& & 0.418 & 0.419 & 0.418 & 140.21 & 172.44 & 113.25 & 59.19 & 0.657 & 6.572 \\
\midrule
\textbf{Impact (Grid)}
& & \textbf{0.502} & \textbf{0.510} & \textbf{0.494} & 168.55 & 190.03 & 124.44 & 65.60 & \textbf{0.218} & \textbf{1.270} \\
\textbf{Impact (\acrshort{dme})}
& &          0.481 &          0.498 &          0.464 & \textbf{174.18} & \textbf{212.00} & \textbf{140.10} & \textbf{71.90} & 0.637 & 4.390 \\
\bottomrule
\end{tabular}
}
% \captionsetup{justification=centering}
\caption{Exploration results on Gibson and \acrshort{mp3d} datasets, at $T=1000$ timesteps. Impact (Grid) is the best-performing model with respect to \textbf{$\mathsf{\acrshort{iou}}$}, while Impact (DME) achieves the best results on \textbf{$\mathsf{\acrshort{as}}$}.}
\label{tab:tab3_impact}
\end{table*}

\subsection{Implementation Details} The experiments are performed using the Habitat Simulator \cite{savva2019habitat} with observations of the agent set to be $128\times 128$ RGB-D images and episode length during training set to $T=500$. Each model is trained with the training split of the Gibson dataset \cite{xia2018gibson} with $40$ environments in parallel for $\approx5$M frames.\\
\tit{Navigation Module} The reinforcement learning algorithm used to train the global and local policies is PPO \cite{schulman2017proximal} with Adam optimizer and a learning rate of $2.5\times 10^{-4}$. The global goal is reset every $\eta=25$ timesteps and the global action space hyperparameter $H$ is $240$. The local policy is updated every $\eta$ steps and the global policy is updated every $20\eta$ steps.\\
\tit{Mapper and Pose Estimator} These models are trained with a learning rate of $10^{-3}$ with Adam optimizer, the local map size is set with $V=101$ while the global map size is $W=961$ for episodes in the Gibson dataset and $W=2001$ in the \acrshort{mp3d} dataset. Both models are updated every $4\eta$ timesteps, where $\eta$ is the reset interval of the global policy.\\
\tit{Density Model} The model used for density estimation is a lightweight version of Gated PixelCNN \cite{oord2016conditional} consisting of a $7\times 7$ masked convolution followed by two residual blocks with $1\times 1$ masked convolutions with $16$ output channels, a $1\times 1$ masked convolutional layer with $16$ output channels, and a final $1\times 1$ masked convolution that returns the output logits with shape $P\times P\times B$, where $B$ is the number of bins used to quantize the model input. We set $P=42$ for the resolution of the input and the output of the density model, and $c=0.1$ for the prediction gain scale factor.

\section{Experimental Results}

\subsection{Exploration Results}
As a first step, we perform model selection using the results on the Gibson Val split (Table \ref{tab:tab1_impact}). Our agents have different hyperparameters that depend on the implementation for the pseudo-counts. When our model employs grid-based pseudo-counts, it is important to determine the dimension of a single cell in this grid-based structure. In our experiments, we test the effects of using $G \times G$ squared cells, with $G \in \{ 2, 4, 5, 10 \}$. The best results are obtained with $G=5$, with small differences among the various setups.
When using pseudo-counts based on a density model, the most relevant hyperparameters depend on the particular model employed as density estimator. In our case, we need to determine the number of bins $B$ for PixelCNN, with $B\in\{64,128,256\}$. We find out that the best results are achieved with $B = 128$.
\begin{figure*}[t!]
\centering
\scriptsize
\setlength{\tabcolsep}{.3em}
\begin{tabular}{cc ccc}
& & \textbf{Gibson Val} &\textbf{\acrshort{mp3d} Val} & \textbf{\acrshort{mp3d} Test} \\
\addlinespace[0.08cm]
\rotatebox{90}{\parbox[t]{0.8in}{\hspace*{\fill}\textbf{Curiosity}\hspace*{\fill}}} & &
\includegraphics[height=0.17\linewidth]{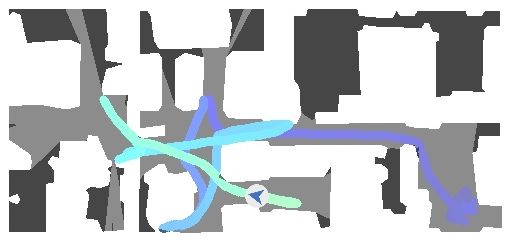} &
\includegraphics[height=0.17\linewidth]{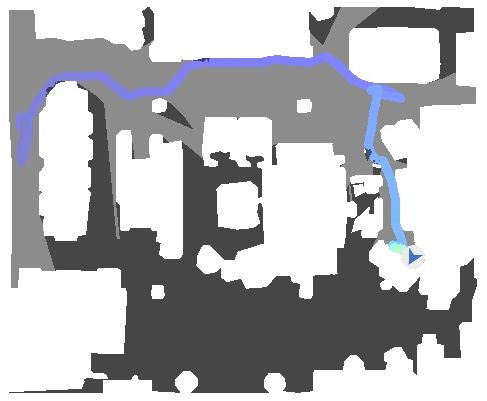} &
\includegraphics[height=0.17\linewidth]{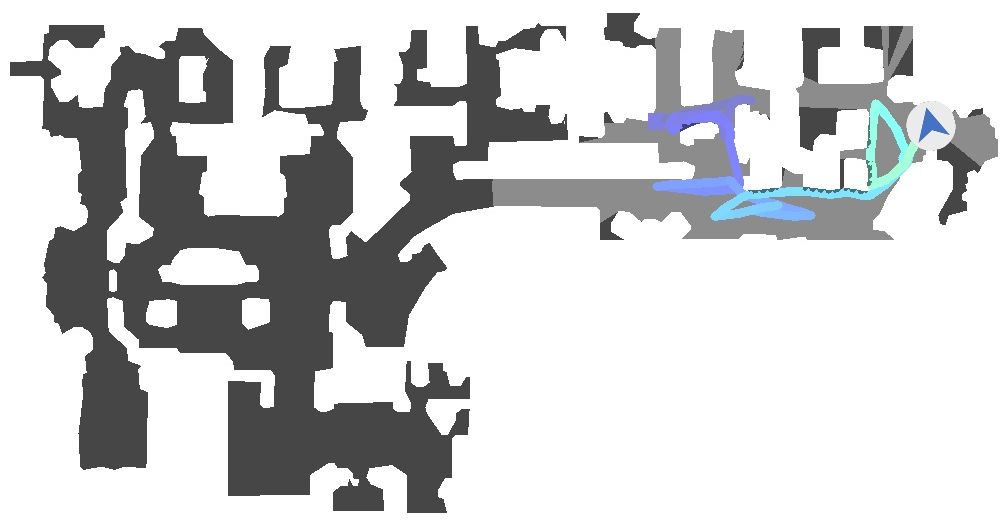} \\
\addlinespace[0.08cm]
\rotatebox{90}{\parbox[t]{0.8in}{\hspace*{\fill}\textbf{Coverage}\hspace*{\fill}}} & &
\includegraphics[height=0.17\linewidth]{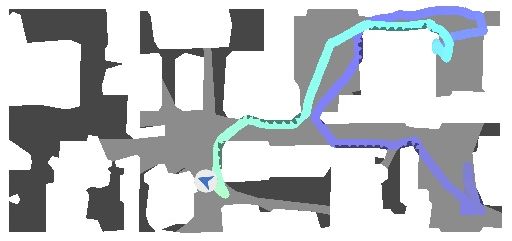} &
\includegraphics[height=0.17\linewidth]{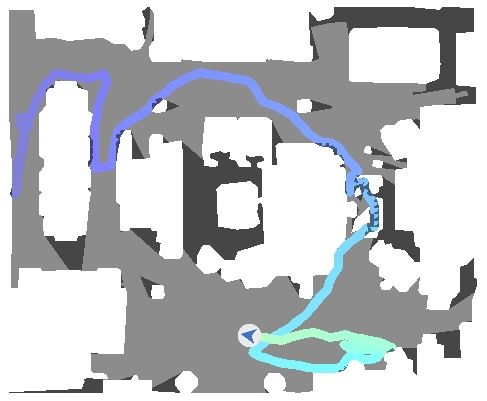} &
\includegraphics[height=0.17\linewidth]{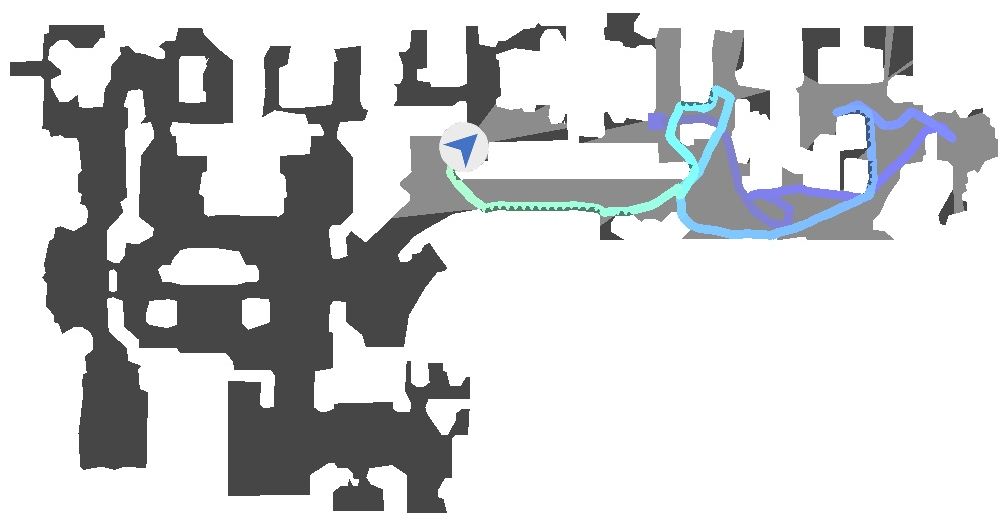} \\
\addlinespace[0.08cm]
\rotatebox{90}{\parbox[t]{0.8in}{\hspace*{\fill}\textbf{Anticipation}\hspace*{\fill}}} & &
\includegraphics[height=0.17\linewidth]{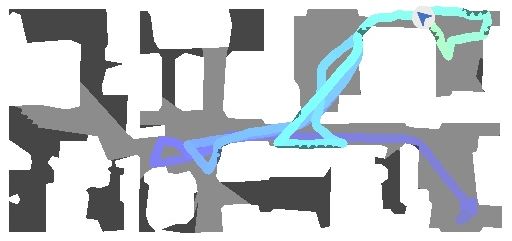} &
\includegraphics[height=0.17\linewidth]{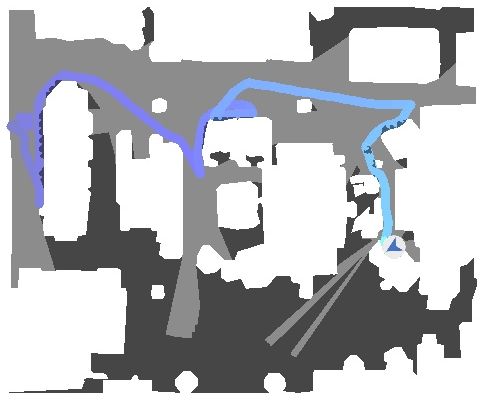} &
\includegraphics[height=0.17\linewidth]{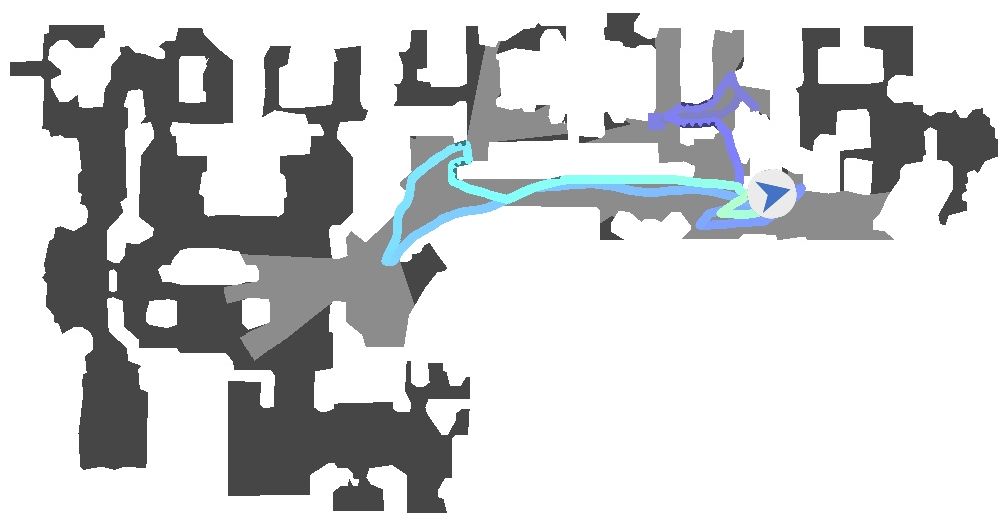} \\
\addlinespace[0.08cm]
\rotatebox{90}{\parbox[t]{0.8in}{\hspace*{\fill}\textbf{Impact (Grid)}\hspace*{\fill}}} & &
\includegraphics[height=0.17\linewidth]{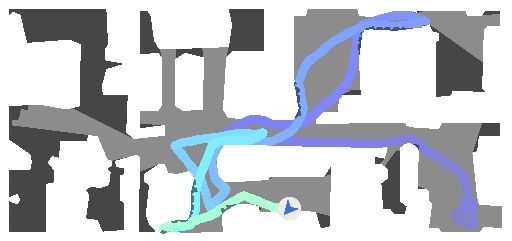} &
\includegraphics[height=0.17\linewidth]{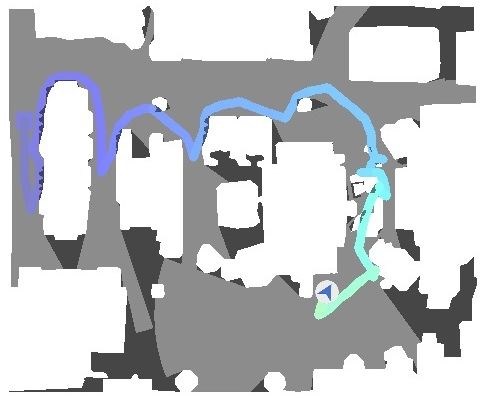} &
\includegraphics[height=0.17\linewidth]{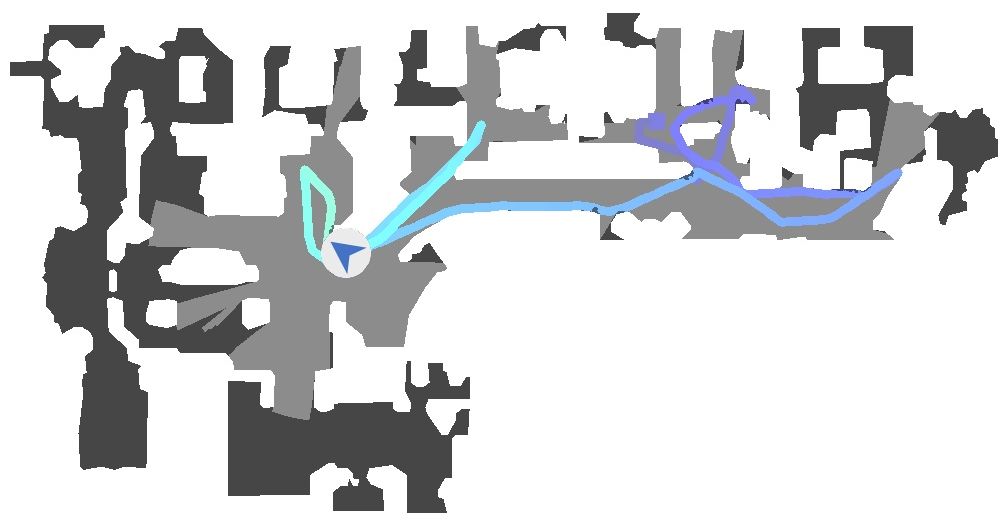} \\
\addlinespace[0.08cm]
\rotatebox{90}{\parbox[t]{0.8in}{\hspace*{\fill}\textbf{Impact (\gls{dme})}\hspace*{\fill}}} & &
\includegraphics[height=0.17\linewidth]{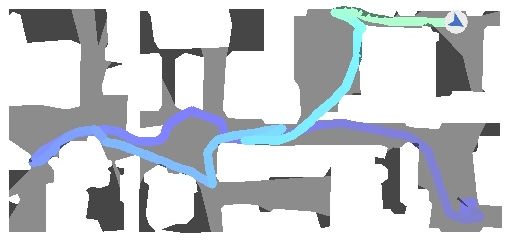} &
\includegraphics[height=0.17\linewidth]{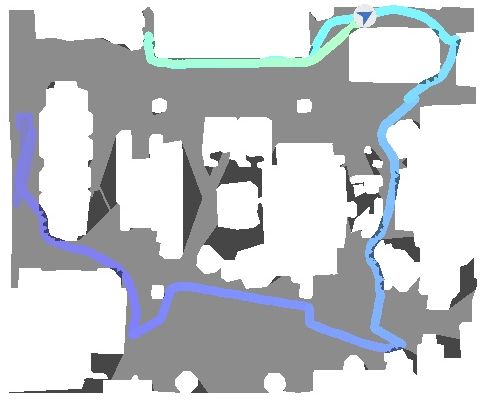} &
\includegraphics[height=0.17\linewidth]{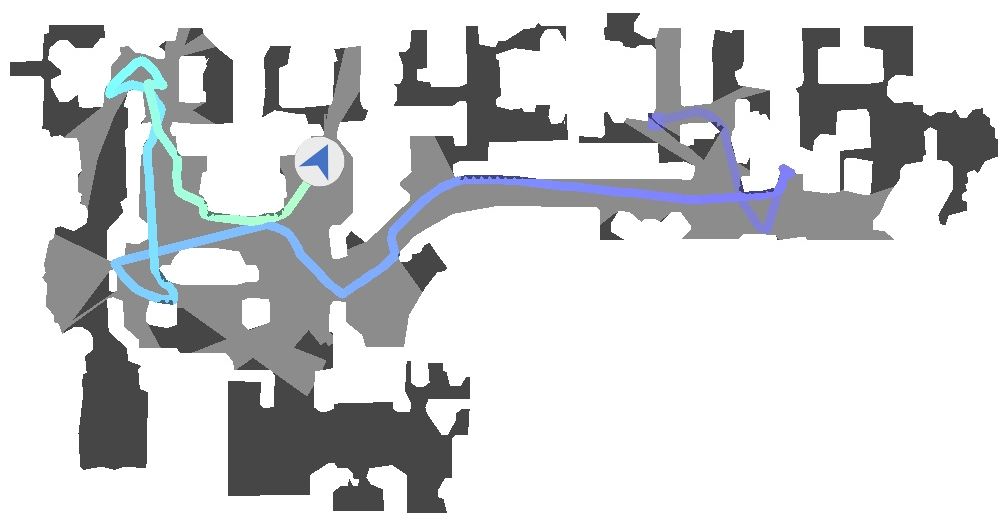} \\
\end{tabular}
\caption{Qualitative results. For each model, we report three exploration episodes on Gibson and \acrshort{mp3d} datasets for $T=500$. The exploration capabilities of the Impact-based models are higher than the baselines, in particular in larger environments.}
\label{fig:fig3_impact}
\end{figure*}

In Table \ref{tab:tab2_impact} and  \ref{tab:tab3_impact}, we compare the Impact (Grid) and Impact (\gls{dme}) agents with the baseline agents previously described on the considered datasets. For each model and each split, we test 5 different random seeds and report the mean result for each metric. For the sake of readability, we do not report the standard deviations for the different runs, which we quantify in around 1.2\% of the mean value reported.

As can be seen, results achieved by the two proposed impact-based agents are constantly better than those obtained by the competitors, both for $T=500$ and $T=1000$.
It is worth noting that our intrinsic impact-based reward outperforms strong extrinsic rewards that exploit information computed using the ground-truth layout of the environment.
Moreover, the different implementations chosen for the pseudo-counts affect final performance, with Impact (\gls{dme}) bringing the best results in terms of \acrshort{as} and Impact (Grid) in terms of \acrshort{iou} metrics.
From the results, it also emerges that, although the proposed implementations for the pseudo-count in Eq. \ref{eq:global_reward} lead to comparable results in small environments as those contained in Gibson and \acrshort{mp3d} Val, the advantage of using \gls{dme} is more evident in large, complex environments as those in \acrshort{mp3d} Test.

In Fig. \ref{fig:fig3_impact}, we report some qualitative results displaying the trajectories and the area seen by different agents in the same episode. Also from a qualitative point of view, the benefits given by the proposed reward in terms of exploration trajectories and explored areas are easy to identify.

\subsection{Point Goal Navigation Results} 
One of the main advantages of training deep modular agents for embodied exploration is that they easily adapt to perform downstream tasks, such as \acrlong{pointnav} \cite{savva2019habitat}. Recent literature \cite{chaplot2019learning,ramakrishnan2020occupancy} has discovered that hierarchical agents trained for exploration are competitive with state-of-the-art architecture tailored for PointGoal navigation and trained with strong supervision for $2.5$ billion frames \cite{wijmans2019dd}. Additionally, the training time and data required to learn the policy are much more limited (2 to 3 orders of magnitude smaller). 
In Table \ref{tab:tab3_gallerie}, we report the results obtained using two different settings. The \textit{noise-free pose sensor} setting is the standard benchmark for \acrlong{pointnav} in Habitat \cite{savva2019habitat}. In the \textit{noisy pose sensor} setting, instead, the pose sensor readings are noisy, and thus the agent position must be estimated as the episode progresses.
We consider four main metrics: the average distance to the goal achieved by the agent (\textbf{$\mathsf{\acrshort{d2g}}$}) and three success-related metrics. The success rate (\textbf{$\mathsf{\acrshort{sr}}$}) is the fraction of episodes terminated within $0.2$ meters from the goal, while the \textbf{$\mathsf{\acrshort{spl}}$} and \acrlong{sspl} (\textbf{$\mathsf{\acrshort{sspl}}$}) weigh the distance from the goal with the length of the path taken by the agent in order to penalize inefficient navigation.
As can be seen, the two proposed agents outperform the main competitors from the literature: \gls{ans} \cite{chaplot2019learning} and \acrshort{occant} \cite{ramakrishnan2020occupancy} (for which we report both the results from the paper and the official code release).

\begin{table}[t]
\footnotesize
\centering
\setlength{\tabcolsep}{.3em}
\resizebox{.9\linewidth}{!}{
\begin{tabular}{lc cccc c >{\color{black}}c >{\color{black}}c >{\color{black}}c >{\color{black}}c}
\toprule
& & \multicolumn{4}{c}{\textbf{Noise-free Pose Sensor}} & & \multicolumn{4}{c}{\textbf{Noisy Pose Sensor}} \\
\cmidrule{3-6} \cmidrule{8-11} 
\textbf{Model} & & \textbf{$\mathsf{\acrshort{d2g}}$} $\downarrow$ & \textbf{$\mathsf{\acrshort{sr}}$} $\uparrow$ & \textbf{$\mathsf{\acrshort{spl}}$} $\uparrow$ & \textbf{$\mathsf{\acrshort{sspl}}$} $\uparrow$ & & \textbf{$\mathsf{\acrshort{d2g}}$} $\downarrow$ & \textbf{$\mathsf{\acrshort{sr}}$} $\uparrow$ & \textbf{$\mathsf{\acrshort{spl}}$} $\uparrow$ & \textbf{$\mathsf{\acrshort{sspl}}$} \\ 
\midrule
\acrshort{ans}~\cite{chaplot2019learning} & & - & 0.950 & 0.846 & - & & - & - & - & - \\
\acrshort{occant}~\cite{ramakrishnan2020occupancy} & & - & 0.930 & 0.800 & - & & - & - & - & - \\
\acrshort{occant}~\cite{ramakrishnan2020occupancy}\tablefootnote{\url{https://github.com/facebookresearch/OccupancyAnticipation}} & & - & - & \underline{0.911} & - & & - & - & - & - \\
\midrule
Curiosity & & \textbf{0.238} & \textbf{0.970} & \textbf{0.914} & \textbf{0.899} & & 0.302 & 0.861 & 0.822 & \underline{0.890} \\
Coverage & & \underline{0.240} & \textbf{0.970} & 0.909 & \underline{0.895} & & 0.288 & 0.827 & 0.788 & 0.886 \\
Anticipation & & 0.285 & 0.965 & 0.906 & 0.892 & & 0.309 & 0.885 & 0.835 & 0.884 \\
\midrule
\textbf{Impact (Grid)} & & 0.252 & \underline{0.969} & 0.908 & 0.894 & & \textbf{0.226} & \textbf{0.923} & \textbf{0.867} & \textbf{0.893} \\ 
\textbf{Impact (\acrshort{dme})}  & & 0.264 & 0.967 & 0.907 & \underline{0.895} & & \underline{0.276} & \underline{0.913} & \underline{0.859} & \textbf{0.893} \\ 
\midrule
\textit{DD-PPO}~\cite{wijmans2019dd} & & - & \textit{0.967} & \textit{0.922} & - & & - & - & - & - \\
\bottomrule
\end{tabular}
}
\caption{\acrlong{pointnav} results on the validation subset of the Gibson dataset. Underlined denotes second best.
} 
\label{tab:tab3_gallerie}
\end{table}

When comparing with our baselines in the noise-free setting, the overall architecture design allows for high-performance results, as the reward influences map estimation only marginally. In fact, in this setting, the global policy and the pose estimation module are not used, as the global goal coincides with the episode goal coordinates, and the agent receives oracle position information. Thus, good results mainly depend on the effectiveness of the mapping module.
Instead, in the noisy setting, the effectiveness of the reward used during training influences navigation performance more significantly. In this case, better numerical results originate from a better ability to estimate the precise pose of the agent during the episode.

For completeness, we also compare with the results achieved by DD-PPO \cite{wijmans2019dd}, a method trained with reinforcement learning for the \gls{pointnav} task on $2.5$ billion frames, $500$ times more than the frames used to train our agents.

\subsection{Real-World Deployment}
As agents trained in realistic indoor environments using the Habitat simulator are adaptable to real-world deployment \cite{kadian2020sim2real}, we also deploy the proposed approach on a LoCoBot robot\footnote{\url{https://locobot-website.netlify.com}}. We employ the PyRobot interface \cite{murali2019pyrobot} to deploy code and trained models on the robot.
To enable the adaptation to the real-world environment, there are some aspects that must be taken into account during training. As a first step, we adjust the simulation in order to reproduce realistic actuation and sensor noise. To that end, we adopt the noise model proposed in \cite{chaplot2019learning} based on Gaussian Mixture Models fitting real-world noise data acquired from a LoCoBot. Additionally, we modify the parameters of the RGB-D sensor used in simulation to match those of the RealSense camera mounted on the robot. Specifically, we change the camera resolution and field of view, the range of depth information, and the camera height. Finally, it is imperative to prevent the agent from learning simulation-specific shortcuts and tricks. For instance, the agent may learn to slide along the walls due to imperfect dynamics in simulation \cite{kadian2020sim2real}. To prevent the learning of such dynamics, we employ the \textit{bump} sensor provided by Habitat and block the agent whenever it is in contact with an obstacle.
When deployed in the real world, our agent is able to explore the environment without getting stuck or bumping into obstacles.
% In the video accompanying the submission, we report exploration samples taken from our real-world deployment.

% \begin{savequote}[75mm]
% In life, we should explore options outside our comfort zone.
% \qauthor{Dani Alves}
% \end{savequote}

\hyphenation{speak-er}

\chapter[Exploration and Recounting]{Explore and Explain: \\ \Large Embodied Exploration and Recounting}
\label{chap:ex2}
\blfootnote{This Chapter is related to the publication ``R. Bigazzi \etal, Explore and Explain: Self-Supervised Navigation and Recounting, ICPR 2020'' \cite{bigazzi2020explore}. See the list of Publications on page \pageref{publications} for more details.}
\RemoveLabels

\lettrine[lines=1]{\textcolor{SchoolColor}{I}}{n} Chapter \ref{chap:focus} we have studied novel exploration methods for embodied agents in photo-realistic environments, in this chapter we investigate the ability of a simple exploration agent to communicate its perception in an intelligent way.

Speaking about how intelligent robots that could autonomously walk and talk were imagined a few decades ago, people used to think about \acrlong{ai} exclusively as a fictional feature, as the available machines they interacted with were purely reactive and showed no form of autonomy. Nowadays, intelligent systems are everywhere, with \gls{dl} being the main engine of the \gls{ai} revolution.
With the research in the \gls{eai} field, the capabilities of robotic agents were significantly improved especially in visual navigation and instruction following \cite{anderson2018vision} tasks. At the same time, tasks at the intersection of \gls{cv} and \acrfull{nlp} are of particular interest to the community, with \gls{ic} being one of the most active areas \cite{karpathy2015deep,anderson2018bottom,cornia2020meshed}. By describing the content of an image or a video, captioning models can bridge the gap between the black-box architecture and the user.

\begin{figure}[t]
    \centering
    \includegraphics[width=\linewidth]{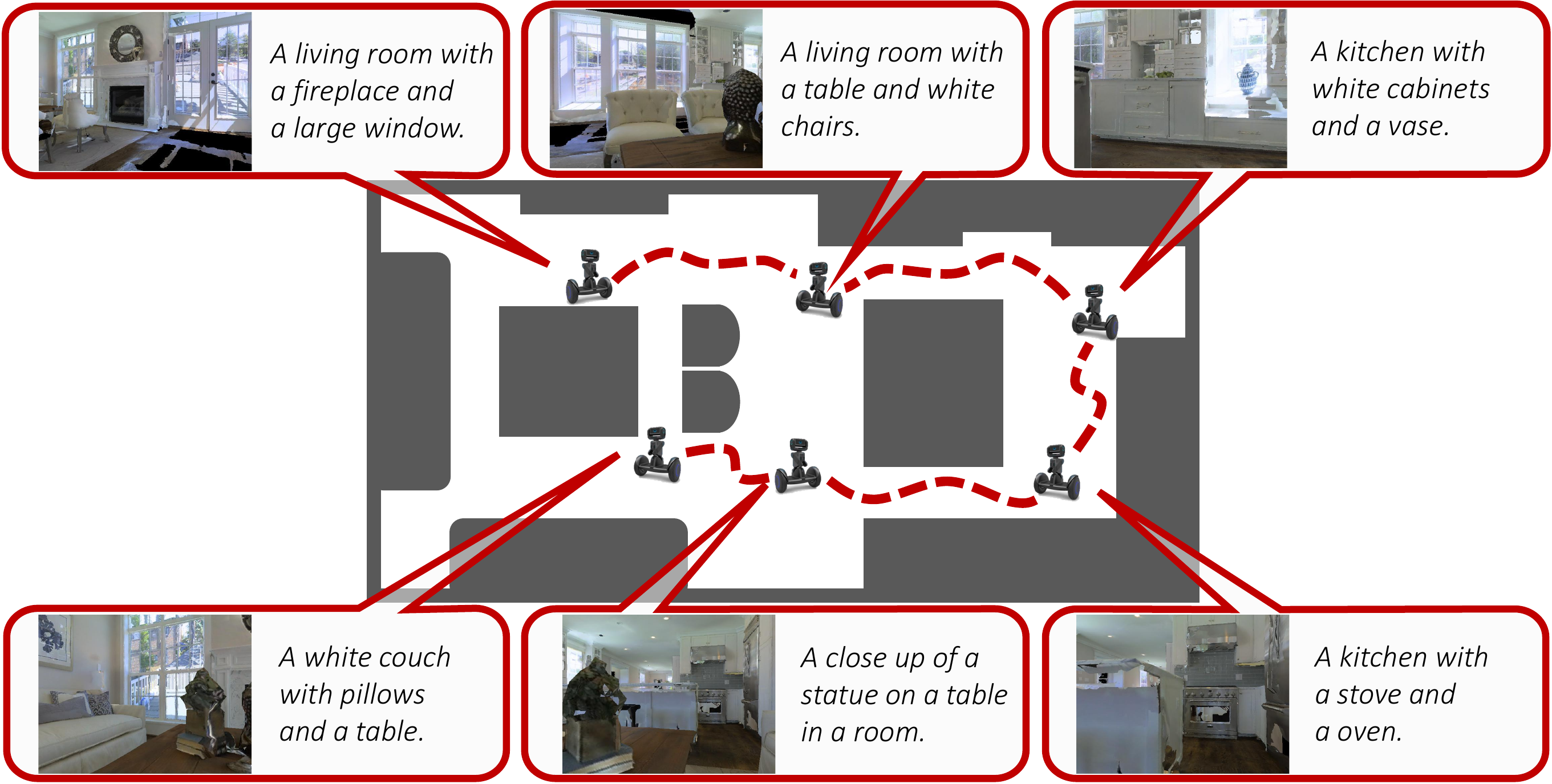}
    \caption{We propose a novel setting in which an embodied agent performs joint curiosity-driven exploration and explanation in unseen environments. While navigating the environment, the agent must produce informative descriptions of what it sees, providing a means of interpreting its internal state.}
    \label{fig:fig1_ex2}
\end{figure}

In this work, we propose a new task at the intersection of \gls{eai}, \gls{cv}, and \acrshort{nlp}, and aim to create a robot that can navigate through a new environment and describe what it sees.
We call this new task \textit{Explore and Explain} since it tackles the problem of joint exploration and captioning (Fig. \ref{fig:fig1_ex2}). In this schema, the agent needs to perceive the environment around itself, navigate it driven by an exploratory goal, and describe salient objects and scenes in natural language. Beyond navigating the environment and translating visual cues into natural language, the agent also needs to identify appropriate moments to perform the explanation step.

\AddLabels

It is worthwhile to mention that both exploration and explanation feature significant challenges. 
Effective exploration without any previous knowledge of the environment can not exploit a reference trajectory and the agent cannot be trained with classic methods from reinforcement learning \cite{wijmans2019dd}.
To overcome this problem, we design a self-supervised exploration module that is driven solely by curiosity toward the new environment. In this setting, rewards are more sparse than in traditional setups and encourage the agent to explore new places and interact with the environment.

While we are motivated by recent works incorporating curiosity in Atari and other exploration games \cite{agrawal2015learning,pathak2017curiosity,burda2018large}, the effectiveness of a curiosity-based approach in a photo-realistic, indoor environment has not been tested extensively. Some preliminary studies \cite{ramakrishnan2020exploration} suggest that curiosity struggles with embodied exploration. In this work, we show that a simple modification of the reward function can lead to striking improvements in the exploration of unseen environments.

Additionally, we encourage the agent to produce a description of what it sees throughout the navigation. In this way, we match the agent's internal state (the measure of curiosity) with the variety and relevance of the generated captions. Such matching offers a proxy for the desirable by-product of interpretability. In fact, by looking at the caption produced, the user can more easily interpret the navigation and perception capabilities of the agent, and the motivations of the actions it takes \cite{cornia2019smart}. In this sense, our work is related to goal-driven Explainable \acrshort{ai}, \ie~the ability of autonomous agents to explain their actions and the reasons leading to their decisions \cite{anjomshoae2019explainable}.

\thispagestyle{nosection}

Previous work on \gls{ic} has mainly focused on recurrent neural networks. However, the rise of Transformer \cite{vaswani2017attention} and the great effectiveness shown by the use of self-attention have motivated a shift towards recurrent-free architectures. Our captioning algorithm builds upon the importance of fully-attentive networks for \gls{ic} and incorporates self-attention both during the encoding of the image features and in the decoding phase. This also allows for a reduction in computational requirements.

Finally, to bridge exploration and recounting, our model can count on a novel speaker policy, which regulates the speaking rate of our captioner using information coming from the agent perception. We call our architecture \ours, from the name of the task: \textit{Explore and Explain}.

Our main contributions are as follows.
We propose a new setting for \acrshort{eai}, \textit{Explore and Explain} in which the agent must jointly deal with two challenging tasks: exploration and captioning of unseen environments.
We devise a novel solution involving curiosity for exploration. Thanks to curiosity, we can learn an efficient policy that can easily generalize to unseen environments.
We apply a captioning algorithm exclusively to indoor environments for robotic exploration.

\section{Proposed Method}
\label{sec:method_ex2}
\begin{landscape}
\begin{figure*}[t]
    \centering
    \includegraphics[width=0.95\linewidth]{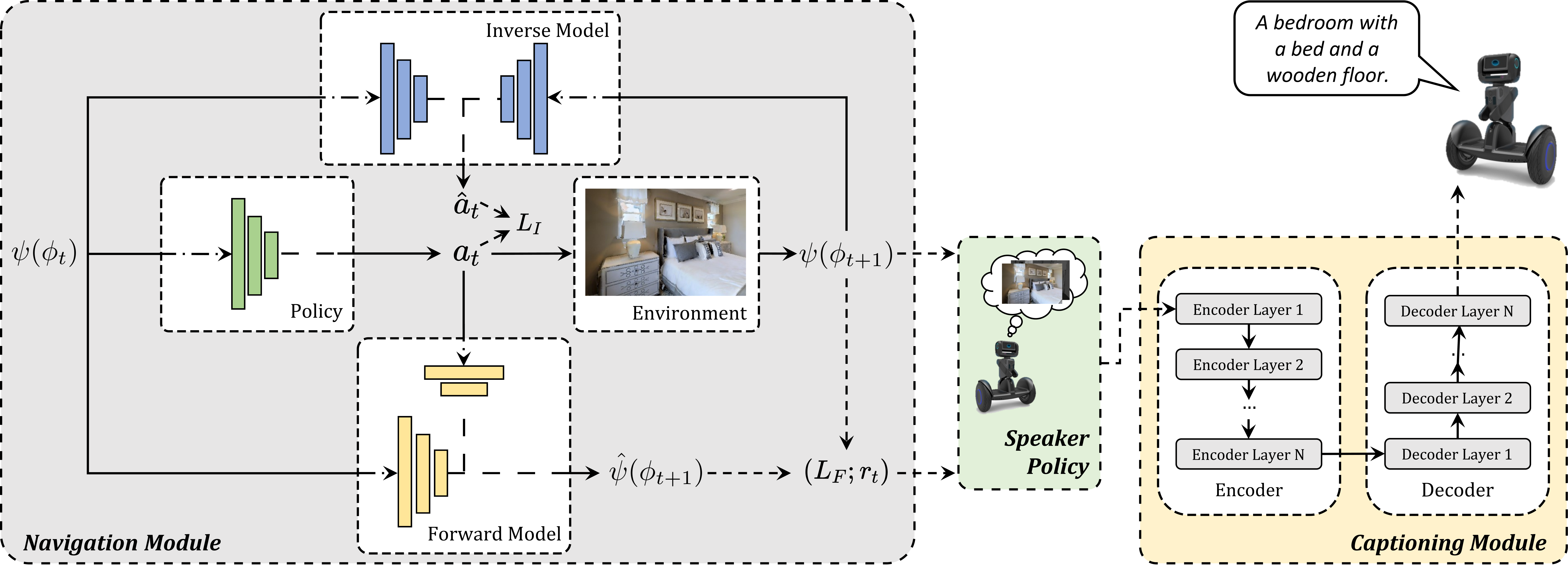}
    \caption{Overview of our \ours~framework for navigation and captioning. Our model is composed of three main components: a navigation module which is in charge of exploring the environment, a captioning module that produces a textual sentence describing the agent point of view, and a speaker policy that connects the previous modules and activates the captioning component based on the information collected during the navigation.}
    \label{fig:fig2_ex2}
\end{figure*}
\end{landscape}

The proposed method consists of three principal parts: a navigation module, a speaker policy, and a captioning module. The last two components constitute the speaker module, which is used to explain the agent's first-person point of view. The explanation is elicited by our speaker module basing on the information gathered during the navigation. Our architecture is depicted in Fig. \ref{fig:fig2_ex2} and detailed in the following sections. Note that, in this work, differently from the previous chapter we are investigating a navigation method that does not build a map along the way but predicts directly the actions to perform from the input.

\subsection{Navigation Module}
\label{sec:method_nav_ex2}
The navigation policy takes care of the agent displacement inside the environment. At each timestep $t$, the agent acquires an observation $\phi_t$ from the surroundings, performs an action $a_t$, and gets the consequent observation $\phi_{t+1}$. In this setting, the moves available to the agent are atomic actions such as \textit{rotate 15 degrees} and \textit{step ahead}.
Our navigation module consists of three main components: a feature embedding network, a forward model, and an inverse model. The discrepancy of the predictions of dynamics models with the actual observation is measured by a reward signal $r_t$, which is then used to stimulate the agent to move towards more informative states.

\tit{Embedding Network}
At each timestep $t$, the agent observes the environment and gathers $\phi_t = (\phi_t^{rgb}, \phi_t^{d})$. This observation corresponds to the raw RGB-D pixels coming from the forward-facing camera of the agent. Yet, raw pixels are not optimal to encode the visual information \cite{burda2018large}. For this reason, we employ a \acrlong{cnn} (\acrshort{cnn}) $\psi$ to encode a more efficient and compact representation of the surrounding environment. We call this embedded representation $\psi(\phi_t)$. To ensure that the features observed by the agent are stable throughout the training, we do not change the set of parameters of the \gls{cnn} $\theta_\psi$ during training. This approach is shown to be efficient for generic curiosity-based agents \cite{burda2018large}.

\tit{Forward Dynamics Model}
Given an agent with policy $\pi(\psi(\phi_t); \theta_\pi)$, represented by a neural network with parameters $\theta_\pi$, the selected action at timestep $t$ is given by:
\begin{equation}
    a_t \sim \pi \Big(\psi(\phi_t); \theta_\pi \Big).
\end{equation}
After executing the chosen action, the agent can observe a new visual stimulus\ $\phi(x_{t+1})$. The problem of predicting the next observation given the current input and action to be performed can be defined as a forward dynamics problem:
\begin{equation}
    \hat\psi(\phi_{t+1}) = f\Big(\psi(\phi_t), a_t; \theta_F \Big),
\end{equation}
where $\hat\phi(x_{t+1})$ is the predicted visual embedding for the next observation $x_{t+1}$ and $f$ is the forward dynamics model with parameters $\theta_F$.
The forward model is trained to minimize the following loss function:
\begin{equation}
    L_F%\Big(\hat\phi(x_{t+1}), \phi(x_{t+1}) \Big)
    = \frac{1}{2} \left\| \hat\psi(\phi_{t+1}) - \psi(\phi_{t+1}) \right\|^2_2 
\label{eq:forward_loss}
\end{equation}

\tit{Inverse Dynamics Model}
Given two consecutive observations $(\phi_{t}, \phi_{t+1})$, the inverse dynamics model aims to predict the action performed at timestep $t$:
\begin{equation}
    \hat a_t = g \Big(\psi(\phi_{t}), \psi(\phi_{t+1}); \theta_I \Big),
\end{equation}
where $\hat a_t$ is the predicted estimate for the action $a_t$ and $g$ is the inverse dynamics model with parameters $\theta_I$. In our work, the inverse model $g$ predicts a probability distribution over the possible actions and it is optimized to minimize the cross-entropy loss with the ground-truth action $a_t$ performed in the previous timestep:
\begin{equation}
    L_I = {y_t \log \hat a_t},
\label{eq:inverse_loss}
\end{equation}
where $y_t$ is the one-hot representation for $a_t$.

\tit{Curiosity-Driven Exploration}
The agent exploration policy $\pi(\psi(\phi_{t}); \theta_\pi)$ is trained to maximize the expected sum of
rewards:
\begin{equation}
    \max_{\theta_\pi} \mathbb{E}_{\pi (\psi(\phi_{t}); \theta_\pi)} \left[ \sum_t r_t \right],
\label{eq:reward}
\end{equation}
where the exploration reward $r_t$ at timestep $t$, also called surprisal \cite{achiam2017surprise}, is given by our forward dynamics model:
\begin{equation}
    r_t = \frac{\eta}{2} \left\| f\big(\psi(\phi_{t}), a_t \big) - \psi(\phi_{t+1}) \right\|^2_2 ,
\label{eq:surprisal}
\end{equation}
with $\eta$ being a scaling factor.
The overall optimization problem can be written as a composition of Eq. \ref{eq:forward_loss}, \ref{eq:inverse_loss}, and \ref{eq:reward}:
\begin{equation}
    \min_{\theta_\pi, \theta_F, \theta_I} \bigg[
        - \lambda \mathbb{E}_{\pi (\psi(\phi_{t}); \theta_\pi)} \Big[ \sum_t r_t \Big]
        + \beta L_F 
        + (1 - \beta) L_I
    \bigg]
\label{eq:optimization}
\end{equation}
where $\lambda$ weights the importance of the intrinsic reward signal \textit{w.r.t.} the policy loss, and $\beta$ balances the contributions of the forward and inverse models. 

\tit{Penalty for Repeated Actions}
\label{ssec:penalty_ex2}
To encourage diversity in our policy, we devise a  penalty that triggers after the agent has performed the same move for $\tilde t$ timesteps. This prevents the agent from always picking the same action and encourages the exploration of different combinations of atomic actions.

We can thus rewrite the surprisal in Eq. \ref{eq:surprisal} as:
\begin{equation}
    r_t = \frac{\eta}{2} \left\| f\big(\psi(\phi_{t}), a_t \big) - \psi(\phi_{t+1}) \right\|^2_2 - p_t ,
\label{eq:final_reward}
\end{equation}
where $p_t$ is the penalty at timestep $t$. In the simplest formulation, $p_t$ can be modeled with a scalar that is either $0$ or equal to a constant $\tilde p$, after an action has been repeated $\tilde t$ times.

\subsection{Speaker Policy}
\label{sec:policy_ex2}
As the navigation proceeds, new observations $\phi_t$ are acquired and rewards $r_t$ are obtained at each timestep. Based on these, a speaker policy can be defined, that activates the captioning module. Different types of information from the environment and the navigation module allow for defining different policies. In this work, we consider three policies, namely: object-driven, depth-driven, and curiosity-driven. 
\tit{Object-Driven Policy}
Given the RGB component of the observation $\phi_t$, relevant objects can be recognized. When at least a minimum number $\mathsf{O}$ of such objects is observed, the speaker policy triggers the captioning module. The idea behind this policy is to let the captioner describe the scene only when objects that allow connoting the different views are present.
\tit{Depth-Driven Policy}
Given the depth component of the observation $\phi_t$, the speaker policy activates the captioner when the mean depth value perceived $\mathsf{D}$ is above a certain threshold. This way, the captioner is triggered only depending on the distance of the agent from generic objects, regardless of their semantic category.
\tit{Curiosity-Driven Policy}
Given the surprisal reward defined as in Eq. \ref{eq:surprisal} and possibly cumulated over multiple timesteps, $\mathsf{S}$, the speaker policy triggers the captioner when $\mathsf{S}$ is above a certain threshold. This policy is independent of the type of information perceived from the environment but is instead closely related to the navigation module. Thus, it helps to match the agent's internal state with the generated captions more explicitly than the other policies.

\subsection{Captioning Module}
When the speaker policy activates, a captioning module is in charge of producing a description in natural language given the current RGB observation $\phi^{rgb}_t$. Following recent literature on the topic, we here employ a visual encoder based on image regions \cite{ren2017faster}, and a decoder that models the probability of generating one word given previously generated ones. In contrast to previous captioning approaches based on recurrent networks, we propose to use a fully-attentive model for both the encoding and the decoding stage, building on the Transformer model \cite{vaswani2017attention}.

\tit{Region Encoder}
Given a set of features from image regions $R = \{r_1, ..., r_N\}$ extracted from the agent visual view, our encoder applies a stack of self-attentive and linear projection operations. As the former be seen as convolutions on a graph, the role of the encoder can also be interpreted as that of learning visual relationships between image regions. The self-attention operator $S$ builds upon three linear projections of the input set, which are treated as queries, keys and values for an attention distribution. Stacking region features $R$ in matrix form, the operator can be defined as follows: 

\begin{align}
    S(R) = \text{Attention}(W_q R, W_k R, W_v R),\\ \nonumber
\text{Attention}(Q, K, V)=\operatorname{softmax}\left(\frac{Q K^{T}}{\sqrt{d}}\right) V.
\label{eq:attention}
\end{align}
The output of the self-attention operator is a new set of elements $S(R)$, with the same cardinality as $R$, in which each element of $R$ is replaced with a weighted sum of the values, \ie~of linear projections of the input.

Following the structure of the Transformer model, the self-attention operator $S$ is followed by a position-wise feed-forward layer, and each of these two operators is encapsulated within a residual connection and a layer norm operation. Multiple layers of this kind are then applied in a stack fashion to obtain the final encoder.

\tit{Language Decoder}
The output of the encoder module is a set of region encodings $\tilde{R}$ with the same cardinality of $R$. We employ a fully-attentive decoder which is conditioned on both previously generated words and region encodings and is in charge of generating the next tokens of the output caption. The structure of our decoder follows that of the Transformer \cite{vaswani2017attention}, and thus relies on self-attentive and cross-attentive operations.

Given a partially decoded sequence of words $W = \{w_0, w_1, ..., w_\tau\}$, each represented as a one-hot vector, the decoder applies a self-attention operation in which $W$ is used to build queries, keys and values. To ensure the causality of this sequence encoding process, we purposely mask the attention operator so that each word can only be conditioned to its left-hand sub-sequence, \ie~word $w_t$ is conditioned on $\{w_{t'}\}_{t' \leq t}$ only. 
Afterwards, a cross-attention operator is applied between $W$ and $\tilde{R}$ to condition words on regions, as follows:
\begin{align}
    C(W, \tilde{R}) = \text{Attention}(W_q W, W_k \tilde{R}, W_v \tilde{R}).
\end{align}

As in the Transformer model, after a self-attention and a cross-attention stage, a position-wise feed-forward layer is applied, and each of these operators is encapsulated within a residual connection and a layer norm operation. Finally, our decoder stacks together multiple decoder layers, helping to refine the understanding of the textual input.

Overall, the decoder takes as input word vectors, and the $t$-th element of its output sequence encodes the prediction of a word at time $t+1$, conditioned on $\{w_t\}_{\le t}$. After taking a linear projection and a softmax operation, this encodes a probability over words in the dictionary. During training, the model is trained to predict the next token given previous ground-truth words; during decoding, we iteratively sample a predicted word from the output distribution and feed it back to the model to decode the next one, until the end of the sequence is reached. Following the usual practice in \gls{ic} literature, the model is trained to predict an end-of-sequence token to signal the end of the caption.

\section{Experimental Setup}
\label{sec:experiments_ex2}

\subsection{Dataset}
\label{ssec:3_matterport}
The main testbed for this work is \acrlong{mp3d} \cite{chang2017matterport3d}, a photo-realistic dataset of indoor environments. Some of the buildings in the dataset contain outdoor components like swimming pools or gardens, raising the difficulty of the exploration task. The dataset is split into $61$ scenes for
training, $11$ for validation, and $18$ for testing. It also provides instance segmentation annotations that we use to evaluate the captioning module. Overall, the dataset is annotated with $40$ different semantic categories. For both training and testing, we use the episodes provided by Habitat Simulator \cite{savva2019habitat} for the \acrlong{pointnav} task, employing only the starting point of each episode. The size of the training set amounts to a total of $5$M episodes, while the test set is composed of $1008$ episodes.

\subsection{Evaluation Protocol}
\label{sec:protocol_ex2}
\tit{Navigation Module}
To quantitatively evaluate the navigation module, we do not use quantitative metrics as the model employs a map-less approach. Instead, we use a curiosity-based metric: we extract the sum of the surprisal values defined in Eq. \ref{eq:surprisal} every $20$ steps performed by the agent, and then we compute the average over the number of test episodes.
% Qualitative results compare the top-down view of the environment and show the path performed by the diverse policies in the same episode.

\tit{Captioning Module}
Standard captioning methods are usually evaluated by comparing each generated caption against the corresponding ground-truth sentences. However, in this setting, only the information on which objects are contained in the scene is available, thanks to the semantic annotations provided by the \acrlong{mp3d} dataset. Therefore, to evaluate the performance of our captioning module, we define two different metrics: a soft coverage measure that assesses how the predicted caption covers all the ground-truth objects, and a diversity score that measures the diversity in terms of described objects of two consecutively generated captions.

In detail, for each caption generated according to the speaker policy, we compute the soft coverage measure between the ground-truth set of semantic categories and the set of nouns in the caption. Given a predicted caption, we firstly extract all nouns $\bm{n}$ from the sentence and we compute the optimal assignment between them and the set of ground-truth categories $\bm{c}^*$, using distances between word vectors and the Hungarian algorithm \cite{kuhn1955hungarian}. We then define an intersection score between the two sets as the sum of assignment profits. Our coverage measure is computed as the ratio of the intersection score and the number of ground-truth semantic classes:
\begin{equation}
    \mathsf{Cov}(\bm{n}, \bm{c}^*) = \frac{\text{I}(\bm{n}, \bm{c}^*)}{\#\bm{c}^*},
\end{equation}
where $\text{I}(\cdot, \cdot)$ is the intersection score, and the $\#$ operator represents the cardinality of the set of ground-truth categories.

Since images may contain small objects which not necessarily should be mentioned in a caption describing the overall scene, we define a variant of the coverage measure by thresholding over the minimum object area. In this case, we consider $\bm{c}^*$ as the set of objects whose overall areas are greater than the threshold.

For the diversity measure, we consider the sets of nouns extracted from two consecutively generated captions, indicated as $\bm{n}_t$ and $\bm{n}_{t+1}$, and we define a soft intersection over union score between the two sets of nouns. Also in this case, we compute the intersection score between the two sets using word distances and the Hungarian algorithm to find the optimal assignment. Recalling
that set union can be expressed in function of an intersection, the final diversity score is computed by subtracting the intersection over union score from $1$ (\ie~the Jaccard distance between the two sets):
\begin{equation}
    \mathsf{Div}(\bm{n}_t, \bm{n}_{t+1}) = 1 - \frac{\text{I}(\bm{n}_t, \bm{n}_{t+1})}{\#\bm{n}_t + \#\bm{n}_{t+1} - \text{I}(\bm{n}_t, \bm{n}_{t+1})},
\end{equation}
where $\text{I}(\cdot, \cdot)$ is the intersection score previously defined, and the $\#$ operator represents the cardinality of the sets of nouns.

We evaluate the diversity of generated captions with respect to the three speaker policies described in Sec. \ref{sec:policy_ex2} and consider different thresholds for each policy (\ie~number of objects, mean depth value, and surprisal score). For each speaker policy and selected threshold, the agent is triggered a different number of times thus generating a variable number of captions during the episode. We define the agent's overall \acrlong{loq} ($\mathsf{\acrshort{loq}}$) as the number of times it is activated by the speaker policy according to a given threshold. In the experiments, we report the \acrlong{loq} values averaged over the test episodes.

\subsection{Implementation Details}
\tit{Navigation Module}
Navigation agents are trained using only visual inputs, with each observation converted to grayscale, cropped, and re-scaled to an $84 \times 84$ size. A stack of four historical observations $[\phi_{t-3}, \phi_{t-2}, \phi_{t-1}, \phi_t]$ is used for training in order to model temporal dependencies. We adopt PPO \cite{schulman2017proximal} as learning algorithm and employ Adam \cite{kingma2015adam} as optimizer. The learning rate for all networks is set to $10^{-4}$ and the length of rollouts is equal to $128$. For each rollout, we make 3 optimization epochs.
The features $\psi(\phi_t)$ used by the forward and backward dynamics networks are $512$-dimensional and are obtained using a randomly initialized convolutional network $\psi$ with fixed weights $\theta_\psi$, following the approach in \cite{burda2018large}.

The model is trained using the splits described in Sec. \ref{ssec:3_matterport}, stopping the training after $10000$ updates of the agent. The length of an exploration episode is $1000$ steps. In our experiments, we set the parameters reported in Eq. \ref{eq:optimization} to $\lambda=0.1$ and $\beta=0.2$, respectively.
Concerning the penalty $p_t$ given to the agent to stimulate diversity (Eq. \ref{eq:final_reward}), we set $p_t = \tilde p = 0.01$ after the same action is repeated for $\tilde{t}=5$ times. 

\tit{Speaker Policy}
For the object-driven policy, we use the instance segmentation annotations provided by the Matterport3D simulator. For this policy, we select $15$ of the $40$ semantic categories in the dataset, discarding the contextual ones, which would not be discriminative for the different views acquired by the agent, as for example \textit{wall}, \textit{floor}, and \textit{ceiling}. This way, we can better evaluate the effect of the policy without it being affected by the performance of an underlying object detector of recognizing objects in the agent's current view.
Also for the depth-driven policy, we obtain the depth information of the current view from the Matterport3D simulator, averaging the depth values to extract a single score. 
In the curiosity-driven policy, we consider the sum of surprisal scores extracted over the last 20 steps, obtained by the agent during navigation.

\tit{Captioning Module}
To represent image regions, we use Faster R-CNN finetuned on the Visual Genome dataset \cite{ren2017faster,anderson2018bottom}, thus obtaining a $2048$-dimensional feature vector for each region. To represent words, we use one-hot vectors and linearly project them to the input dimensionality of the model, $d$. We also employ sinusoidal positional encodings \cite{vaswani2017attention} to represent word positions inside the sequence and sum the two embeddings before the first encoding layer. 
In both the region encoder and language decoder, we set the dimensionality $d$ of each layer to $512$, the number of heads to $8$, and the dimensionality of the inner feed-forward layer to $2048$. We use dropout with keep probability $0.9$ after each attention layer and after position-wise feed-forward layers. 

Following standard practice in image captioning \cite{rennie2017self,anderson2018bottom}, we train our model in two phases using image-caption pairs coming from the COCO dataset \cite{lin2014microsoft}. Firstly, the model is trained with cross-entropy loss to predict the next token given previous ground-truth words. Then, we further optimize the sequence generation using reinforcement learning employing a variant of the self-critical sequence training \cite{rennie2017self} on sequences sampled using beam search \cite{anderson2018bottom}. 
Pretraining with cross-entropy loss is done using the learning rate scheduling strategy defined in \cite{vaswani2017attention} with a warmup equal to $10000$ iterations. Then, during finetuning with reinforcement learning, we use the \acrshort{cider}-D score \cite{vedantam2015cider} as reward and a fixed learning rate equal to $5^{-6}$. We train the model using the Adam optimizer \cite{kingma2015adam} and a batch size of $50$. During \acrshort{cider}-D optimization and caption decoding, we use beam search with a beam size equal to $5$. 
To compute coverage and diversity metrics and for extracting nouns from predicted captions, we use the spaCy \acrshort{nlp} toolkit\footnote{\url{https://spacy.io/}}. We use GloVe word embeddings \cite{pennington2014glove} to compute word similarities between nouns and semantic class names.

\begin{table}[t]
\centering
\setlength{\tabcolsep}{.45em}
\resizebox{0.75\linewidth}{!}{
\begin{tabular}{lcc}
\toprule
\textbf{Navigation Module} &  &  $\mathsf{Surprisal}$ \\
\midrule
Random Exploration & & 0.333 \\
\midrule
\ours w/o Penalty for repeated actions (RGB only) & & 0.193 \\
\ours w/o Penalty for repeated actions (Depth only) & & 0.361 \\
\ours w/o Penalty for repeated actions (RGB + Depth) & & 0.439 \\
\midrule
\textbf{\ours} & & \textbf{0.697} \\
\bottomrule
\end{tabular}
}
\caption{Surprisal scores for different navigation policies obtained during the agent exploration of the environment. The final model achieves a higher score than the baselines.
}
\label{tab:tab1_ex2}
\end{table}

\begin{figure}[t]
\centering
\footnotesize
\setlength{\tabcolsep}{.6em}
\begin{tabular}{ccc:cc}
\textbf{Random Exploration} & \textbf{w/o Penalty} & & & \textbf{\ours} \\
\addlinespace[0.2cm]
\includegraphics[width=0.22\linewidth]{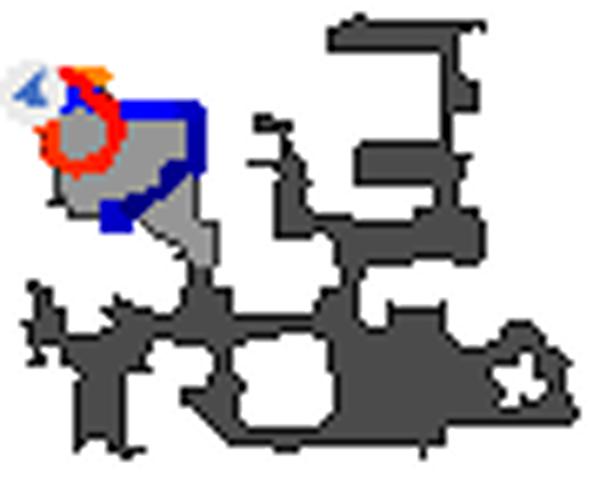} &
\includegraphics[width=0.22\linewidth]{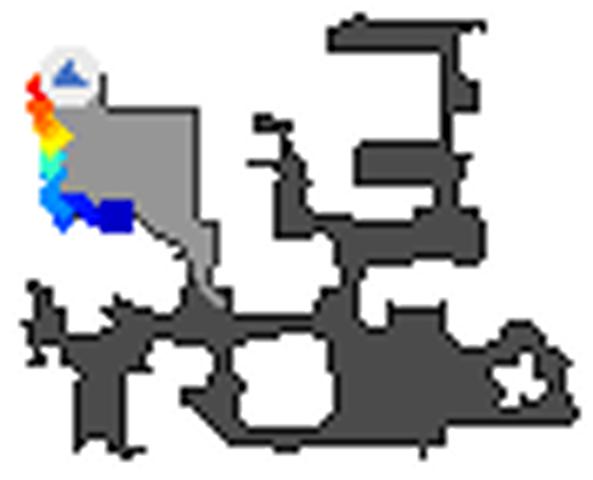} & & &
\includegraphics[width=0.22\linewidth]{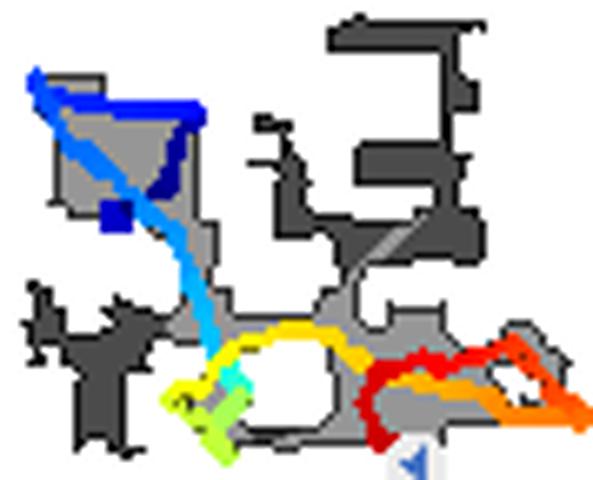} \\
\includegraphics[width=0.22\linewidth]{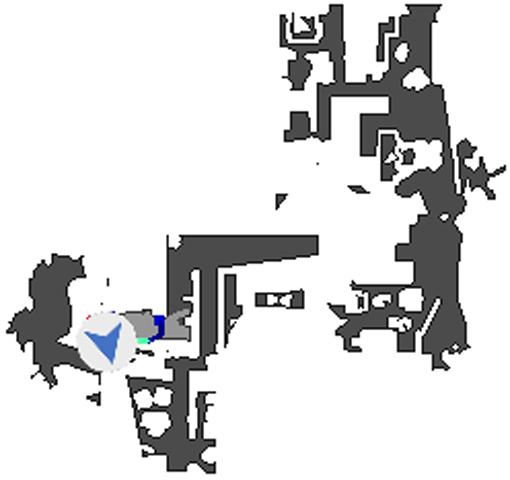} &
\includegraphics[width=0.22\linewidth]{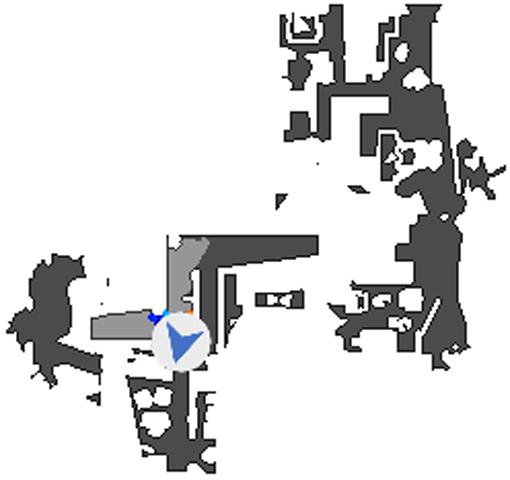} & & &
\includegraphics[width=0.22\linewidth]{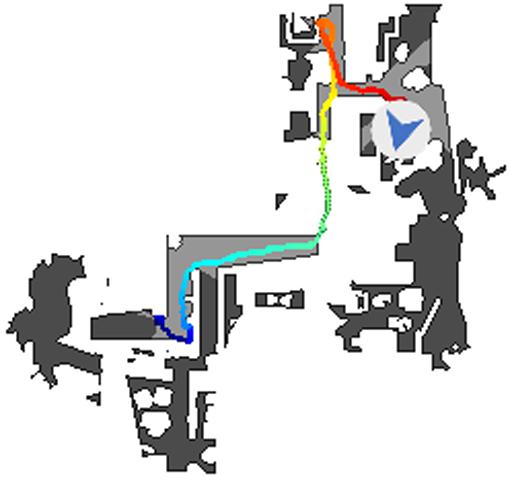} \\
\includegraphics[width=0.22\linewidth]{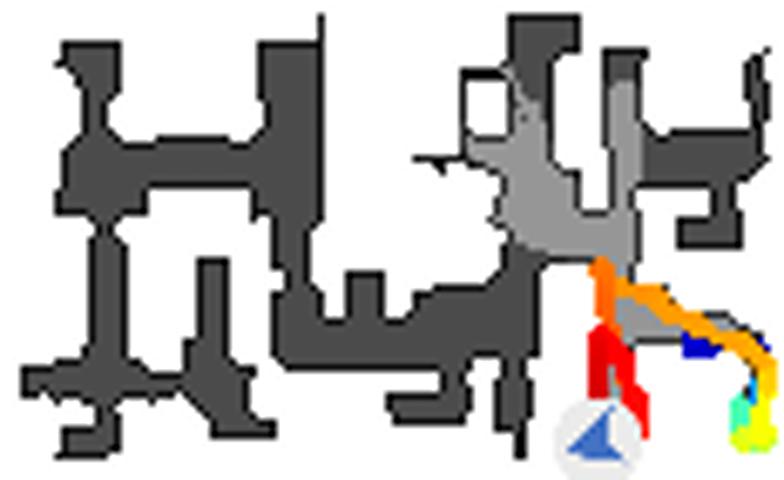} &
\includegraphics[width=0.22\linewidth]{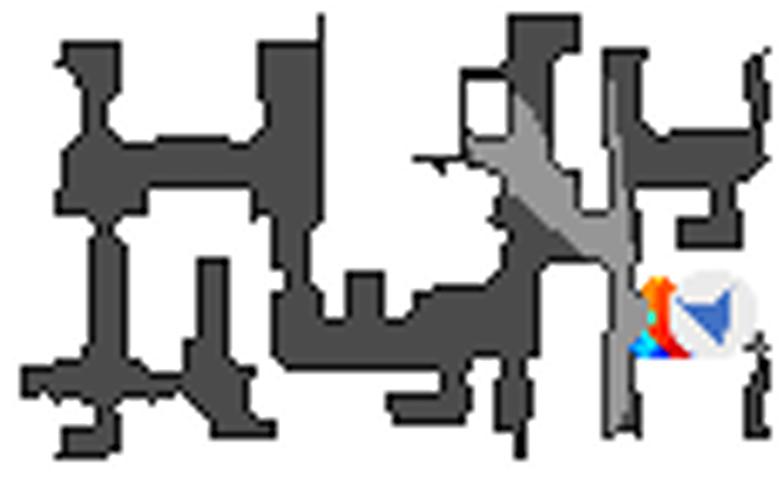} & & &
\includegraphics[width=0.22\linewidth]{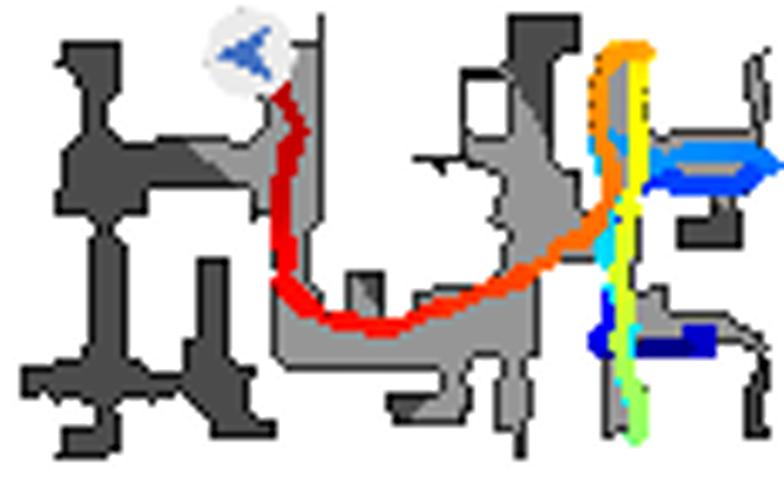} \\
\end{tabular}
\caption{Qualitative results of the agent trajectories in sample navigation episodes.}
\label{fig:fig3_ex2}
\end{figure}

\section{Experimental Results}

\subsection{Exploration Results}
As defined in Sec. \ref{sec:protocol_ex2}, we evaluate the performance of our navigation agents by computing the average surprisal score over test episodes.
Results are reported in Table \ref{tab:tab2_ex2} and show that our complete method (\ours) outperforms all other variants, achieving a significantly greater surprisal score than our method without penalty. In particular, the final performance greatly benefits from using both visual modalities (RGB and depth), instead of using a single visual modality to represent the scene.
Notably, random exploration (\eg~sampling $a_t$ from a uniform distribution over the available actions at each timestep $t$) proves to be a strong baseline for this task, performing better than our single-modality RGB agent. Nonetheless, our final agent greatly outperforms the baselines, scoring $0.364$ and $0.258$ above the random policy and the vanilla curiosity-based agent respectively.

\tit{Qualitative Analysis}
In Fig. \ref{fig:fig3_ex2}, we report some top-down views from the testing scenes, together with the trajectory from three different navigation agents: the random baseline, our approach without the penalty for repeated action described in Sec. \ref{sec:method_nav_ex2}, and our full model.
We notice that the agent without penalty usually remains in the starting area and thus has some difficulties in exploring the whole environment.
Instead, our complete model demonstrates better results as it is able to explore a much wider area within the environment.
Thus, we conclude that the addition of a penalty for repeated actions in the final reward function is of central importance when it comes to stimulating the agent toward the exploration of regions far from the starting point.

\subsection{Captioning Results}
Here, we provide quantitative and qualitative results for our captioning module that is helped by speaker policy. In fact, the captioning module generates a description of the first-person view of the agent when the policy is triggered.
For better readability, results are reported in two tables, Table \ref{tab:tab2_ex2} and Table \ref{tab:tab3_ex2}, and are discussed below.

\begin{table*}[!t]
\centering
\setlength{\tabcolsep}{.8em}
\resizebox{.85\linewidth}{!}{
\begin{tabular}{r l ccccc}
\toprule
\multicolumn{2}{l}{\textbf{$\mathsf{\textbf{Object}}$ $\mathsf{(O\geq1)}$}} \\
\multicolumn{2}{c}{\normalsize$\mathsf{\acrshort{loq}=43.3}$} &  \textbf{$\mathsf{\acrshort{cov}}_{>1\%}$} & \textbf{$\mathsf{\acrshort{cov}}_{>3\%}$} & \textbf{$\mathsf{\acrshort{cov}}_{>5\%}$} & \textbf{$\mathsf{\acrshort{cov}}_{>10\%}$} & \textbf{$\mathsf{\acrshort{div}}$}\\
\midrule
& \textbf{\ours}~(1 lay.) & 0.468 & 0.564 & 0.623 & 0.720 & \textbf{0.394} \\
& \textbf{\ours}~(2 lay.) & \textbf{0.485} & \textbf{0.579} & \textbf{0.637} & \textbf{0.727} & 0.368 \\
& \textbf{\ours}~(3 lay.) & 0.474 & 0.558 & 0.612 & 0.701 & 0.372 \\
& \textbf{\ours}~(6 lay.) & 0.456 & 0.550 & 0.609 & 0.706 & 0.386 \\
\midrule
\multicolumn{2}{l}{\textbf{$\mathsf{\textbf{Object}}$ $\mathsf{(O\geq3)}$}} \\
\multicolumn{2}{c}{\normalsize$\mathsf{\acrshort{loq}=27.4}$} &  \textbf{$\mathsf{\acrshort{cov}}_{>1\%}$} & \textbf{$\mathsf{\acrshort{cov}}_{>3\%}$} & \textbf{$\mathsf{\acrshort{cov}}_{>5\%}$} & \textbf{$\mathsf{\acrshort{cov}}_{>10\%}$} & \textbf{$\mathsf{\acrshort{div}}$}\\
\midrule
& \textbf{\ours}~(1 lay.) & 0.400 & 0.519 & 0.593 & 0.713 & \textbf{0.377} \\
& \textbf{\ours}~(2 lay.) & \textbf{0.416} & \textbf{0.534} & \textbf{0.607} & \textbf{0.721} & 0.349 \\
& \textbf{\ours}~(3 lay.) & 0.384 & 0.497 & 0.571 & 0.691 & 0.350 \\
& \textbf{\ours}~(6 lay.) & 0.387 & 0.502 & 0.576 & 0.696 & 0.363 \\
\midrule
\multicolumn{2}{l}{\textbf{$\mathsf{\textbf{Object}}$ $\mathsf{(O\geq5)}$}} \\
\multicolumn{2}{c}{\normalsize$\mathsf{\acrshort{loq}=15.8}$} &  \textbf{$\mathsf{\acrshort{cov}}_{>1\%}$} & \textbf{$\mathsf{\acrshort{cov}}_{>3\%}$} & \textbf{$\mathsf{\acrshort{cov}}_{>5\%}$} & \textbf{$\mathsf{\acrshort{cov}}_{>10\%}$} & \textbf{$\mathsf{\acrshort{div}}$}\\
\midrule
& \textbf{\ours}~(1 lay.) & 0.356 & 0.479 & 0.560 & 0.702 & \textbf{0.373} \\
& \textbf{\ours}~(2 lay.) & \textbf{0.373} & \textbf{0.497} & \textbf{0.577} & \textbf{0.713} & 0.340 \\
& \textbf{\ours}~(3 lay.) & 0.347 & 0.467 & 0.546 & 0.688 & 0.338 \\
& \textbf{\ours}~(6 lay.) & 0.348 & 0.468 & 0.549 & 0.691 & 0.352 \\

\midrule\multicolumn{2}{l}{\textbf{$\mathsf{\textbf{Depth}}$ $\mathsf{(D>0.25)}$}} \\
\multicolumn{2}{c}{\normalsize$\mathsf{\acrshort{loq}=38.5}$} &  \textbf{$\mathsf{\acrshort{cov}}_{>1\%}$} & \textbf{$\mathsf{\acrshort{cov}}_{>3\%}$} & \textbf{$\mathsf{\acrshort{cov}}_{>5\%}$} & \textbf{$\mathsf{\acrshort{cov}}_{>10\%}$} & \textbf{$\mathsf{\acrshort{div}}$}\\
\midrule
& \textbf{\ours}~(1 lay.) & 0.448 & 0.548 & 0.613 & 0.723 & \textbf{0.371} \\
& \textbf{\ours}~(2 lay.) & \textbf{0.463} & \textbf{0.562} & \textbf{0.625} & \textbf{0.730} & 0.341 \\
& \textbf{\ours}~(3 lay.) & 0.427 & 0.524 & 0.588 & 0.700 & 0.349 \\
& \textbf{\ours}~(6 lay.) & 0.433 & 0.532 & 0.600 & 0.705 & 0.360\\
\midrule
\multicolumn{2}{l}{\textbf{$\mathsf{\textbf{Depth}}$ $\mathsf{(D>0.5)}$}} \\
\multicolumn{2}{c}{\normalsize$\mathsf{\acrshort{loq}=31.1}$} &  \textbf{$\mathsf{\acrshort{cov}}_{>1\%}$} & \textbf{$\mathsf{\acrshort{cov}}_{>3\%}$} & \textbf{$\mathsf{\acrshort{cov}}_{>5\%}$} & \textbf{$\mathsf{\acrshort{cov}}_{>10\%}$} & \textbf{$\mathsf{\acrshort{div}}$}\\
\midrule
& \textbf{\ours}~(1 lay.) & 0.434 & 0.536 & 0.603 & 0.719 & \textbf{0.359} \\
& \textbf{\ours}~(2 lay.) & \textbf{0.449} & \textbf{0.550} & \textbf{0.612} & \textbf{0.726} & 0.330 \\
& \textbf{\ours}~(3 lay.) & 0.413 & 0.511 & 0.577 & 0.695 & 0.335 \\
& \textbf{\ours}~(6 lay.) & 0.420 & 0.519 & 0.585 & 0.701 & 0.346 \\
\midrule
\multicolumn{2}{l}{\textbf{$\mathsf{\textbf{Depth}}$ $\mathsf{(D>0.75)}$}} \\
\multicolumn{2}{c}{\normalsize$\mathsf{\acrshort{loq}=14.8}$} &  \textbf{$\mathsf{\acrshort{cov}}_{>1\%}$} & \textbf{$\mathsf{\acrshort{cov}}_{>3\%}$} & \textbf{$\mathsf{\acrshort{cov}}_{>5\%}$} & \textbf{$\mathsf{\acrshort{cov}}_{>10\%}$} & \textbf{$\mathsf{\acrshort{div}}$}\\
\midrule
& \textbf{\ours}~(1 lay.) & 0.412 & 0.513 & 0.583 & 0.708 & \textbf{0.355} \\
& \textbf{\ours}~(2 lay.) & \textbf{0.425} & \textbf{0.525} & \textbf{0.595} & \textbf{0.715} & 0.325 \\
& \textbf{\ours}~(3 lay.) & 0.394 & 0.491 & 0.559 & 0.685 & 0.330 \\
& \textbf{\ours}~(6 lay.) & 0.399 & 0.497 & 0.566 & 0.691 & 0.339\\
\bottomrule
\end{tabular}

}
\caption{\Acrlong{cov} and \acrlong{div} results for different versions of our captioning module. Results are reported for our object-based and depth-based speaker policies using different thresholds to determine the agent's \acrlong{loq} inside the episode. The model using 2 Transformer layers returns the best results.} 
\label{tab:tab2_ex2}
\end{table*}
\begin{table*}[!t]
\centering
\setlength{\tabcolsep}{.8em}
\resizebox{.85\linewidth}{!}{
\begin{tabular}{r l ccccc}
\toprule
\multicolumn{2}{l}{\textbf{$\mathsf{\textbf{Curiosity}}$ $\mathsf{(S>0.7)}$}} \\
\multicolumn{2}{c}{\normalsize$\mathsf{\acrshort{loq}=27.2}$} & \textbf{$\mathsf{\acrshort{cov}}_{>1\%}$} & \textbf{$\mathsf{\acrshort{cov}}_{>3\%}$} & \textbf{$\mathsf{\acrshort{cov}}_{>5\%}$} & \textbf{$\mathsf{\acrshort{cov}}_{>10\%}$} & \textbf{$\mathsf{\acrshort{div}}$}\\
\midrule
& \textbf{\ours}~(1 lay.) & 0.438 & 0.539 & 0.604 & 0.719 & \textbf{0.370} \\
& \textbf{\ours}~(2 lay.) & \textbf{0.453} & \textbf{0.552} & \textbf{0.617} & \textbf{0.726} & 0.340 \\
& \textbf{\ours}~(3 lay.) & 0.418 & 0.514 & 0.578 & 0.694 & 0.348 \\
& \textbf{\ours}~(6 lay.) & 0.425 & 0.523 & 0.588 & 0.703 & 0.356 \\
\midrule
\multicolumn{2}{l}{\textbf{$\mathsf{\textbf{Curiosity}}$ $\mathsf{(S>0.85)}$}} \\
\multicolumn{2}{c}{\normalsize$\mathsf{\acrshort{loq}=18.2}$} & \textbf{$\mathsf{\acrshort{cov}}_{>1\%}$} & \textbf{$\mathsf{\acrshort{cov}}_{>3\%}$} & \textbf{$\mathsf{\acrshort{cov}}_{>5\%}$} & \textbf{$\mathsf{\acrshort{cov}}_{>10\%}$} & \textbf{$\mathsf{\acrshort{div}}$}\\
\midrule
& \textbf{\ours}~(1 lay.) & 0.433 & 0.530 & 0.597 & 0.716 & \textbf{0.373} \\
& \textbf{\ours}~(2 lay.) & \textbf{0.448} & \textbf{0.545} & \textbf{0.611} & \textbf{0.724} & 0.342 \\
& \textbf{\ours}~(3 lay.) & 0.413 & 0.506 & 0.571 & 0.691 & 0.350 \\
& \textbf{\ours}~(6 lay.) & 0.421 & 0.515 & 0.581 & 0.699 & 0.360\\
\midrule
\multicolumn{2}{l}{\textbf{$\mathsf{\textbf{Curiosity}}$ $\mathsf{(S>1.0)}$}} \\
\multicolumn{2}{c}{\normalsize$\mathsf{\acrshort{loq}=6.4}$} & \textbf{$\mathsf{\acrshort{cov}}_{>1\%}$} & \textbf{$\mathsf{\acrshort{cov}}_{>3\%}$} & \textbf{$\mathsf{\acrshort{cov}}_{>5\%}$} & \textbf{$\mathsf{\acrshort{cov}}_{>10\%}$} & \textbf{$\mathsf{\acrshort{div}}$}\\
\midrule
& \textbf{\ours}~(1 lay.) & 0.434 & 0.532 & 0.597 & 0.717 & \textbf{0.380} \\
& \textbf{\ours}~(2 lay.) & \textbf{0.448} & \textbf{0.545} & \textbf{0.610} & \textbf{0.723} & 0.349 \\
& \textbf{\ours}~(3 lay.) & 0.413 & 0.506 & 0.570 & 0.690 & 0.361 \\
& \textbf{\ours}~(6 lay.) & 0.422 & 0.518 & 0.583 & 0.702 & 0.364\\
\bottomrule
\end{tabular}

}
\caption{\Acrlong{cov} and \acrlong{div} results for different versions of our captioning module. Results are reported for our curiosity-based speaker policy using different thresholds to determine the agent's \acrlong{loq} inside the episode.} 
\label{tab:tab3_ex2}
\end{table*}

\tit{Speaker Policy}
Among the three different policies, the object-driven speaker performs the best in terms of coverage and diversity. In particular, setting a low threshold ($\mathsf{O}\geq1$) provides the highest scores. At the same time, the agent tends to speak more often, which is desirable in a visually rich environment. As the threshold for $\mathsf{O}$ gets higher, performances get worse. This indicates that, as the number of objects in the scene increases, there are many details that the captioner cannot describe. The same applies to the depth-driven policy: while the agent tends to describe well items that are closer, it experiences some troubles when facing an open space with more distant objects ($\mathsf{D} \geq 0.75$).

Instead, our curiosity-driven speaker shows a more peculiar behavior: as the threshold grows, results get better in terms of diversity, while the coverage scores are quite stable (only $-0.005\%$ in terms of $\mathsf{Cov}_{>1\%}$). 
It is also worth mentioning that our curiosity-based speaker can be adopted in any kind of environment, as the driving metric is computed from the raw RGB-D input. The same does not apply in an object-driven policy, since the agent needs semantic information.
Further, the curiosity-driven policy employs a learned metric, hence being more related to the exploration module.

\begin{figure*}[t]
    \centering
    \includegraphics[width=.9\linewidth]{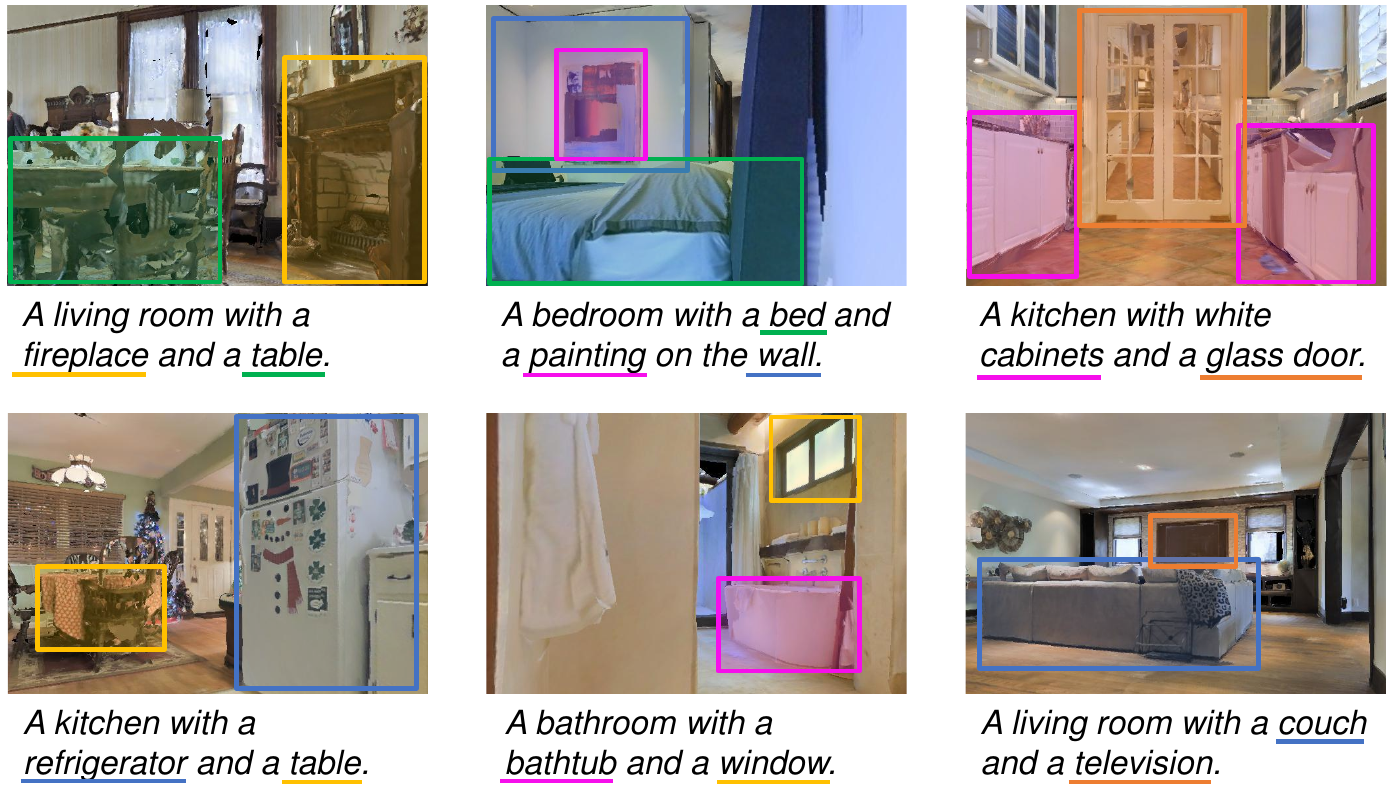}
    \caption{Sentences generated on sample images extracted from \ours~navigation trajectories. For each image, we report the relevant objects present on the scene and we underline their mentions in the caption.}
    \label{fig:fig4_ex2}
\end{figure*}

From all these observations, we can conclude that curiosity not only helps train navigation agents but also represents an important metric when bridging cross-modal components in embodied agents.

\tit{Captioning Module}
When evaluating the captioning module, we compare the performance using a different number of encoding and decoding layers. As it can be seen from Table \ref{tab:tab2_ex2} and \ref{tab:tab3_ex2}, the captioning model achieves the best results when composed of $2$ layers for coverage and $1$ layer for diversity. While this is in contrast with traditional Transformer-based models \cite{vaswani2017attention}, which employ $6$ or more layers, it is in line with recent research on image captioning \cite{cornia2020meshed}, which finds it beneficial to adopt fewer layers. At the same time, a more lightweight network can possibly be embedded in many embodied agents, thus being more appropriate for our task.

\tit{Qualitative Analysis}
We report some qualitative results for \ours~in Fig. \ref{fig:fig4_ex2}. To ease visualization, we underline the items mentioned by the captioner in the sentence and highlight them with a bounding box of the same color in the corresponding input image. Our agent can explain the scene perceived from a first-person, egocentric point of view. We can notice that \ours~identifies all the main objects in the environment and produces a suitable description even when the view is partially occluded or the object presents artifacts due to the 3D model reconstruction.

\chapter[Efficient Exploration and Smart Scene Description]{ Efficient Exploration \\ and Smart Scene Description}
\label{chap:eds}
\blfootnote{This Chapter is related to the publication ``R. Bigazzi \etal, Embodied Agents for Efficient Exploration and Smart Scene Description, ICRA 2023'' \cite{bigazzi2023embodied}. See the list of Publications on page \pageref{publications} for more details.}
\RemoveLabels

\lettrine[lines=1]{\textcolor{SchoolColor}{T}}{he} path explored with the work described in Chapter \ref{chap:ex2}, adopting language to facilitate human understanding of the behavior and perception of a robotic agent, poses as a first step towards fluid interaction between robotic agents and humans. 
Nevertheless, even with the introduction of \emph{Explore and Explain}, the task of joint exploration and scene description, and the implementation of \ours~architecture, there is still a long way to go before this ambitious objective becomes a reality. In fact, \ours~could be improved in each one of the three main components of the method, namely the navigation module, the speaker policy, and the captioning module. 

For example, a significant improvement in the exploration capabilities of the agent could be enabled by the adoption of a map-based navigation policy, in this way the agent is capable of remembering areas already observed. In this regard, a plethora of exploration rewards and strategies is available in the representation learning community, and generally, those strategies could be adapted to work on board smart autonomous agents to improve their perception of the surrounding.

At the same time, apart from the depth-based and the object-based speaker policies, the curiosity-based policy was specifically designed for a curiosity-based navigation policy, going against the modularity of the proposed method.
Instead, on the captioning side, the advent of foundation models like CLIP \cite{radford2021learning} in the last few years brought significant advances to the methods available in the literature. 

Furthermore, the absence of a metric designed to evaluate agents on \emph{Explore and Explain} was a major lack.

We present a revised pipeline for an autonomous agent for efficient exploration and mapping of unseen environments, providing user-understandable representations of the perceived environment and avoiding unnecessary repetitions.

\AddLabels

The key contributions of this work are twofold. First, we combine state-of-the-art approaches for image captioning and visual exploration to tackle \emph{Explore and Explain} task with the aim of improving human understanding of robotic perception. Second, we devise a novel metric, called \acrlong{eds} (\textbf{$\mathsf{\acrshort{eds}}$}), that evaluates the exploration and the ability to cover objects in the environment avoiding repetitions. We extensively test the performance of the proposed approach and validate the value of \textbf{$\mathsf{\acrshort{eds}}$} on both Gibson \cite{xia2018gibson} and \acrlong{mp3d} \cite{chang2017matterport3d} datasets.
Furthermore, while our approach is trained and evaluated in simulation, the proposed architecture is designed for the final deployment on a real robotic platform.

\section{Proposed Method}
\begin{landscape}
\begin{figure*}
    \centering
    \includegraphics[width=0.85\linewidth]{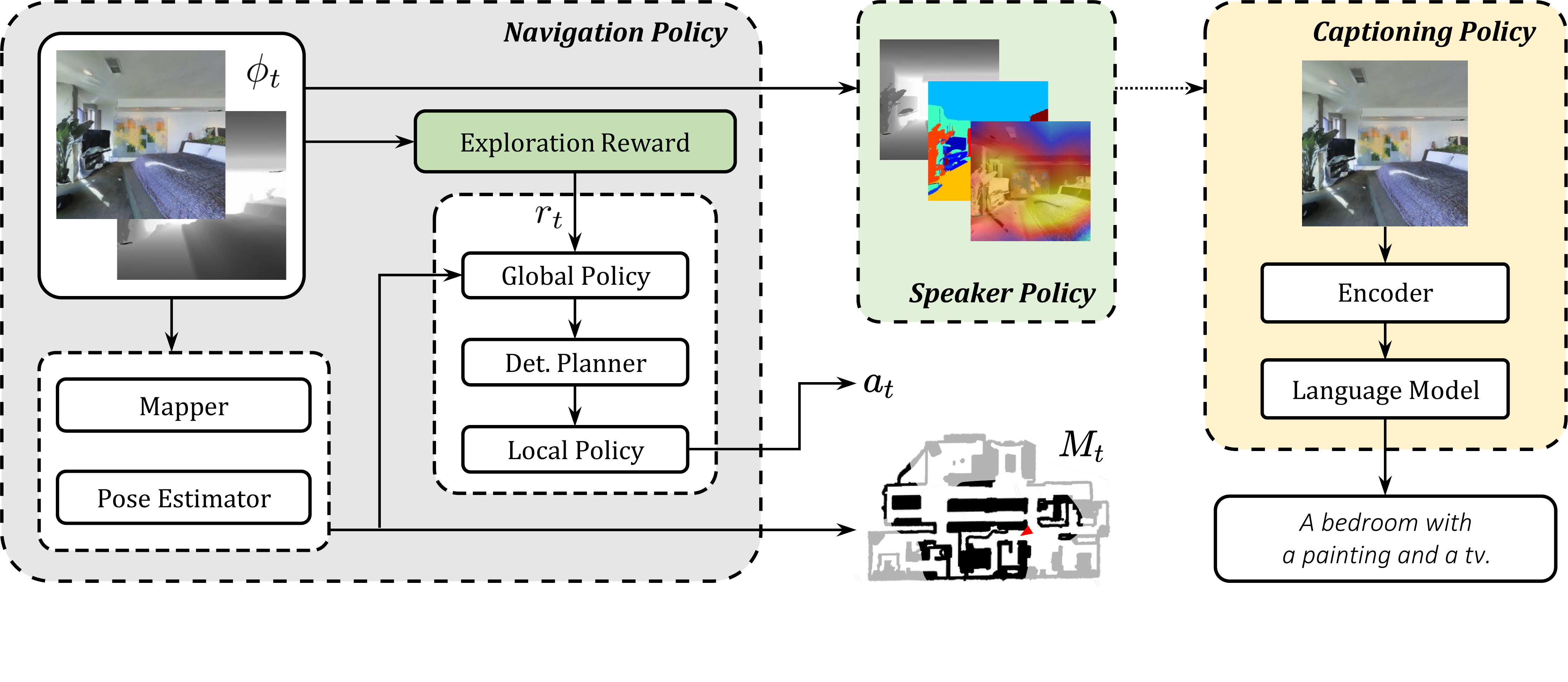}
    \captionsetup{justification=centering}
    \caption{Overview of the proposed approach for smart scene description, comprising a navigation module, a speaker policy, and a captioning module.}
    \label{fig:fig1_eds}
\end{figure*}
\end{landscape}

Our proposed architecture remains composed of three main components: a navigation module, in charge of the exploration, a captioning module, that describes interesting scenes, and the speaker policy that decides when the captioner should be activated. An overview of the updated complete architecture is shown in Fig. \ref{fig:fig1_eds}.

\subsection{Navigation Module}
\label{sec:navigation_method_eds}
The exploration capabilities of the agent are strictly dependent on the performance of the navigation module, therefore relying on a proper navigation approach is of fundamental importance. Following recent literature on embodied visual navigation \cite{chaplot2019learning,ramakrishnan2020occupancy,ramakrishnan2022poni}, we adopt a hierarchical policy coupled with a learned neural occupancy mapper and a pose estimator.
The hierarchical policy sets long and short-term navigation goals, while the neural mapper builds an occupancy grid map representation of the environment and the pose estimator locates the agent on such map.

\tit{Mapper}
In order to track explored and unexplored regions of the environment over time, the mapper is a fundamental component. In this work, as presented in Chapter \ref{chap:focus}, we use a neural-based mapper that allows inferring region occupancy beyond the observable area in front of the agent, facilitating the planning phase \cite{ramakrishnan2020occupancy}. The output of the mapper is a $W \times W \times 2$ global map of the environment $M_t$ that keeps track of the non-traversable space in its first channel and the area explored by the agent in the second one.
The mapper processes the RGB-D observation $\phi_t=(\phi^{rgb}_t, \phi^{d}_t)$ coming from the agent and predicts a $V \times V \times 2$ egocentric local map $m_t$ representing the state in front of the agent. More details on how the local map is output are described in Section \ref{ssec:mapper_impact}. 
At every timestep $t$, the local map $m_t$ is transformed using the estimated pose of the agent $\omega_t = (x_t,y_t,\theta_t)$ and registered to the global map $M_t$ with a moving average. The global map is initially empty and is built incrementally with the exploration of the environment.

\tit{Pose Estimator}
Relying on a global map requires a robust pose estimator in order to build geometrically coherent and precise maps. Indeed, an inaccurate pose estimate would rapidly diverge from the ground-truth pose, and loop closure is inapplicable if previous knowledge of the environment is not available. Furthermore, directly using the pose sensor of the robot is not sufficient since sensor noise, slipping wheels, and collisions with obstacles would not be taken into consideration.
The adopted approach uses the difference between consecutive pose sensor readings $\Delta \omega'_{t}=\omega'_{t}-\omega'_{t-1}$ as a first estimate of the motion of the agent, where $\omega'_t=(x'_t, y'_t, \theta'_t)$, with $(x'_t,y'_t)$ being the coordinates on the map, and $\theta'_t$ the orientation of the agent. In order to correct eventual inaccuracies, we use local maps $m_{t}$, $m_{t-1}$ extracted from the respective observations as feedback. The local map $m_{t-1}$ is rototranslated with respect to the current position of the agent using $\Delta \omega'_{t}$. Transformed $m_{t-1}$ and $m_t$ are concatenated and fed to a \acrshort{cnn} to output a corrected displacement $\Delta \omega_{t}$. At every timestep, $\Delta \omega_{t}$ is used to compute the pose of the agent with respect to the pose at the previous step:
\begin{equation}
    \omega_{t} = \omega_{t-1} + \Delta \omega_{t} \quad \text{where} \quad \omega_t = (x_t, y_t, \theta_t).
\label{eq:pose_estimator}
\end{equation}
%Without loss of generality, we consider the starting position of the agent $\omega_0=(0,0,0)$ corresponding to the center of the global map $m_t$.
Without loss of generality, we consider the agent starting from $\omega_0=(0,0,0)$, \ie~the center of the global map $M_t$.

\begin{figure*}
    \centering
    \footnotesize
    \resizebox{\linewidth}{!}{
        \setlength{\tabcolsep}{.15em}
        \begin{tabular}{ccc}
             \textbf{Curiosity} & 
             \textbf{Coverage} & 
             \textbf{Anticipation} \\
             \includegraphics[width=0.30\textwidth]{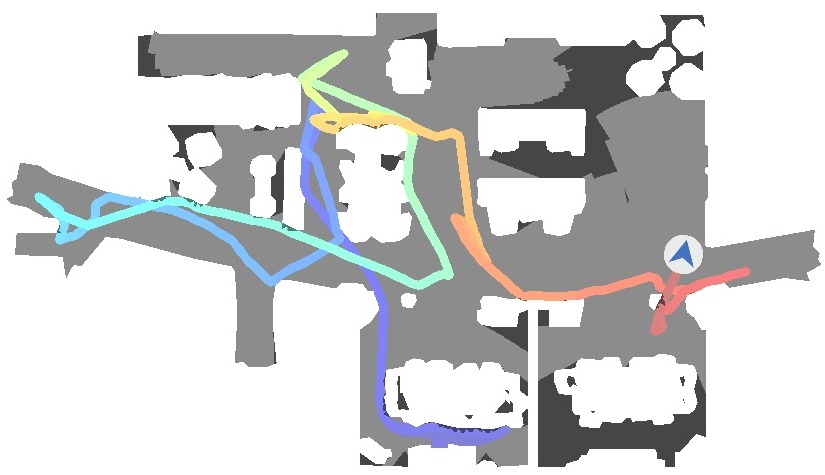} &
             \includegraphics[width=0.30\textwidth]{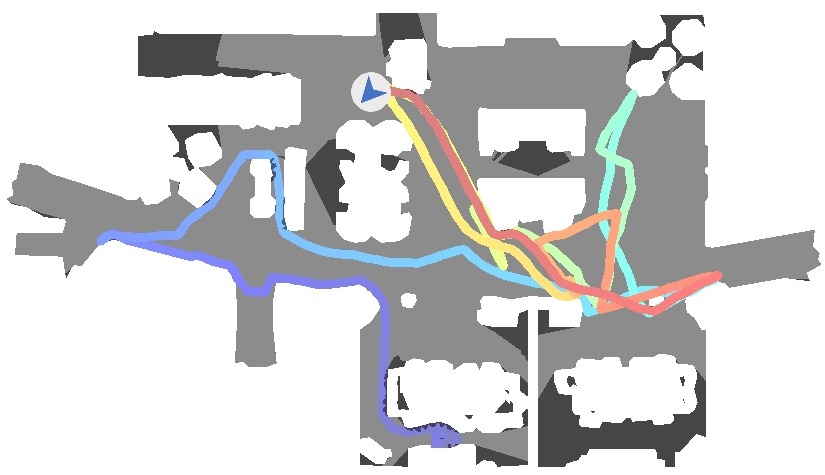} &
             \includegraphics[width=0.30\textwidth]{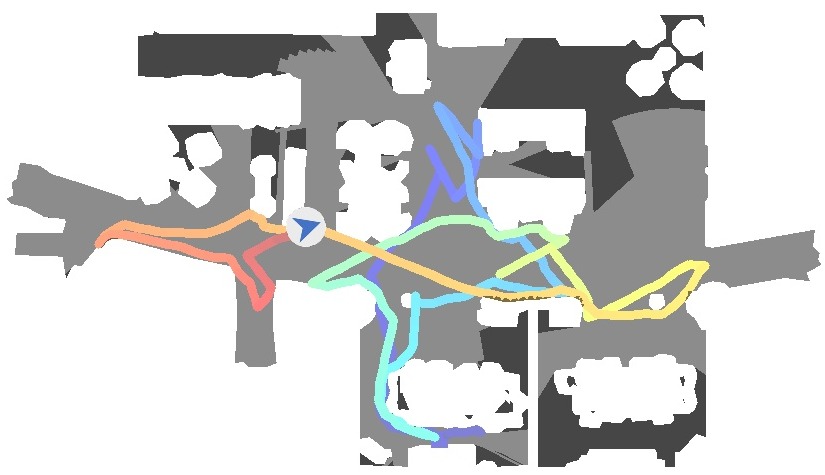} \\
        \end{tabular}
        }
    \resizebox{0.64\linewidth}{!}{
        \setlength{\tabcolsep}{.15em}
        \begin{tabular}{cc}
             \textbf{Impact (Grid)} & 
             \textbf{Impact (\acrshort{dme})} \\
             \includegraphics[width=0.30\textwidth]{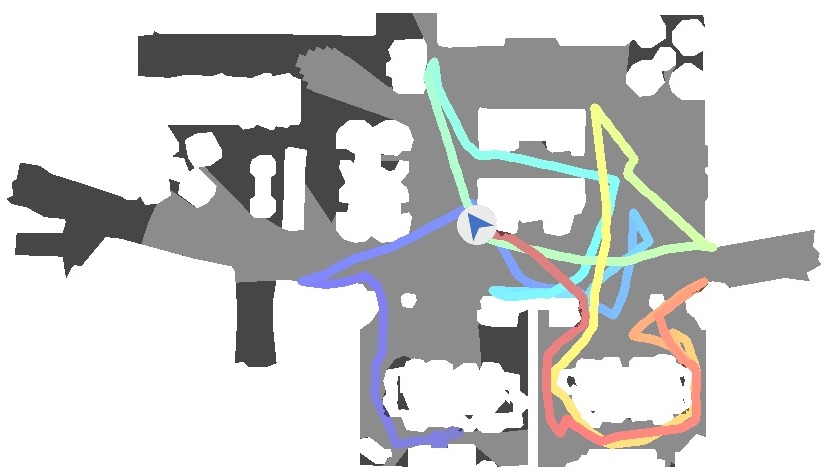} &
             \includegraphics[width=0.30\textwidth]{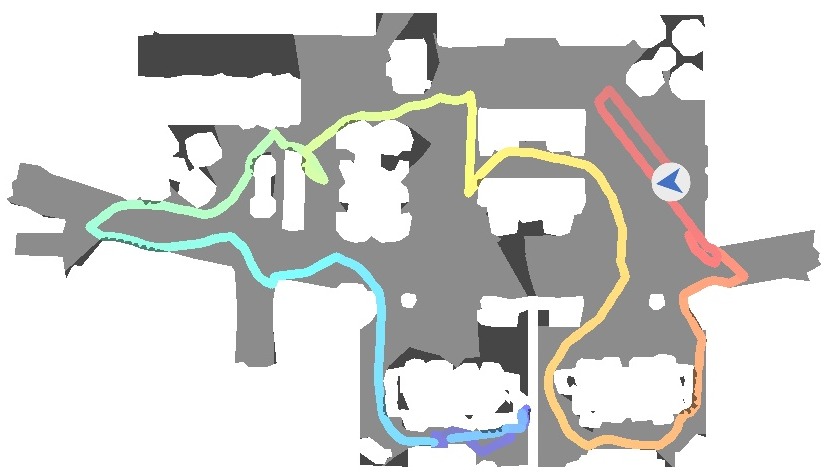} \\
        \end{tabular}
    }
    \caption{Qualitative exploration trajectories of different navigation agents on the same episode.}
    \label{fig:fig2_eds}
\end{figure*}

\tit{Navigation Policy}
The navigation policy adopts a hierarchical structure as used in 
embodied literature \cite{chaplot2019learning,ramakrishnan2020occupancy,ramakrishnan2022poni}. Specifically, the navigation policy comprehends three modules: a high-level global policy, a deterministic planner, and an atomic local policy. 
In this way, the hierarchical policy considers both high-level and low-level concepts like moving across rooms and avoiding obstacles.
The implementation of the navigation policy follows the same architecture presented in Section \ref{ssec:navpolicy_impact}.

The global policy takes as input an enriched version of the current global map $m_t$ and outputs the global goal $g_t$, that is a coordinate to be reached. The global policy is trained with reinforcement learning using the global reward $r^{global}_t$.

The planner consists of an A* algorithm. It uses the global map to plan a path towards the global goal and samples a local goal $l_t$ within $1.25$m from the position of the agent. 

The local policy takes as input the current RGB observation $\phi^{rgb}_t$ as well as the relative displacement of the local goal $l_t$ from agent's position $\omega_t$, and predicts the atomic action needed to reach the local goal. The output of the local policy corresponds to one of the following atomic actions: \textit{move forward 0.25m}, \textit{turn left 10\textdegree}, and \textit{turn right 10\textdegree}. This policy is trained with a reward $r^{local}_t$ that encourages the decrease in the geodesic distance between the agent and the local goal.

\subsection{Exploration Rewards}
We compare various global exploration rewards such as curiosity \cite{pathak2017curiosity}, coverage \cite{chaplot2020object}, anticipation \cite{ramakrishnan2020occupancy}, and impact \cite{bigazzi2022impact}. All the considered methods obtain the reward by exploiting visual input sensors only.
Exemplar exploration trajectories resulting from the different rewards are reported in Fig. \ref{fig:fig2_eds}.

\tit{Curiosity}
The curiosity reward follows the same paradigm presented in Chapter \ref{chap:ex2}, but in this case, the penalty presented in Section\ref{ssec:penalty_ex2} is not used. When using the curiosity-based reward the navigation policy adopts two additional neural networks that learn the environment dynamics. The forward model trained to predict the encoding of the future RGB observation given the encoding of the current observation and action and the inverse model trained to infer the action $a_t$ performed between consecutive observations $(\phi^{rgb}_{t},\phi^{rgb}_{t+1})$. These models are trained by minimizing the following losses:
\begin{equation}
    L_{F} = \frac{1}{2} \left\| \hat\psi(\phi^{rgb}_{t+1}) - \psi(\phi^{rgb}_{t+1}) \right\|^2_2  \quad\mathrm{and}\quad 
    L_{I} = {y_t \log \hat a_t},
\label{eq:curiosity_losses}
\end{equation}
where $\hat\psi(\phi^{rgb})$ and $\psi(\phi^{rgb})$ denote predicted and ground-truth encodings of the observation $\phi^{rgb}$, $y$ is the one-hot encoding of the ground-truth action $a$, and $\hat{a}$ is the predicted action probability distribution.
The global reward for the curiosity-driven model is given by the error of the forward dynamics model prediction during the navigation:
\begin{equation}
        r^{global}_t = \frac{\eta}{2} \| \hat\psi(\phi^{rgb}_{t+1}) - \psi(\phi^{rgb}_{t+1}) \|^2_2,
\label{eq:curiosity_reward}
\end{equation}
where $\eta$ is a normalizing term set to $0.01$.

\tit{Coverage}
The coverage-based reward maximizes the information gathered at each timestep, being the number of objects or landmarks reached or area seen. In this work, we consider the area seen definition, as proposed in \cite{chaplot2019learning}:
\begin{equation}
    r^{global}_t = \text{AS}_{t} - \text{AS}_{t-1},
\label{eq:coverage_reward}
\end{equation}
where AS indicates the number of pixels explored in the ground truth map.

\tit{Anticipation}
The occupancy anticipation reward \cite{ramakrishnan2020occupancy} aims to maximize accuracy in the prediction of the map including occluded unseen areas, \ie,%. In other words:
\begin{align}
    r^{global}_t = \text{Acc}(M_t, M^*) - \text{Acc}(M_{t-1}, M^*), \\
    \text{Acc}(M, M^*) = \sum_{i=1}^{W^2}\sum_{j=1}^2\mathbbmm{1}[M_{ij} = M^*_{ij}],
\label{eq:occant_reward}
\end{align}
where $M$ is the predicted global map, $M^*$ is the ground truth global map, and $\mathbbmm{1}[\cdot]$ is the indicator function.

\tit{Impact}
The impact reward encourages actions that modify agent's internal representation of the environment, with impact at timestep $t$ that is measured as the $l_2$-norm of the encodings of the two consecutive states $\psi(s_{t})$ and $\psi(s_{t+1})$. However, using the formulation of impact as it is, could lead to trajectory cycles with high impact but low exploration. To overcome this issue, Raileanu \etal \cite{raileanu2020ride} uses the state visitation count $N(s_t)$ to scale the reward. Unfortunately in our setting the concept of the visitation count is not directly applicable, due to the continuous space of the photo-realistic environment. Hence, we adopt and evaluate the impact-based methods proposed in \cite{bigazzi2022impact}. Such methods formalize a pseudo-visitation count $\widehat{N}(s_t)$ in continuous environments with two different approaches: grid and density model estimation. The final global reward for the impact-driven model becomes:
\begin{equation}
    r^{global}_t = \left\| \psi(\phi^{rgb}_{t+1}) - \psi(\phi^{rgb}_{t}) \right\|_2\Big/\sqrt{\widehat{N}(s_t)},
\label{eq:impact_reward_dme}
\end{equation}
where $\psi(\phi^{rgb}_t)$ and $\widehat{N}(s_t)$ are the encoding and the estimated pseudo-visitation count at timestep $t$.

\subsection{Captioning Module}\label{ssec:cap_module_eds}
The goal of the captioning module is that of modeling an autoregressive distribution probability $p(\bm{w}_t|\bm{w}_{\tau<t}, V)$, where $V$ is an image captured from the agent and $\{\bm{w}_t\}_t$ is the sequence of words comprising the generated caption. This is usually achieved by training a language model conditioned on visual features to mimic ground-truth descriptions. 
For multimodal fusion, we employ an encoder-de-coder Transformer \cite{vaswani2017attention} architecture. Each layer of the encoder employs \gls{msa} and feed-forward layers, while each layer of the decoder employs \gls{msca} and feed-forward layers. For enabling text generation, sequence-to-sequence attention masks are employed in each self-attention layer of the decoder. 

To obtain the set of visual features $V$ for an image, we employ a visual encoder that is pretrained to match vision and language (\ie~CLIP \cite{radford2021learning}). Compared to using features extracted from object detectors \cite{anderson2018bottom,zhang2021vinvl}, 
our strategy is beneficial in terms of both computational efficiency and feature quality. 
The visual descriptors $V=\{ \bm{v}_i \}_{i=1}^N$ are encoded via bi-directional attention in the encoder, while the token embeddings of the caption $W = \{ \bm{w}_i \}_{i=1}^L$ are inputs of the decoder, where $N$ and $L$ indicate the number of visual embeddings and caption tokens, respectively. The overall network operates according to the following schema: 
\begin{align}
    \text{encoder} \quad \quad & \bm{\tilde{v}}_i = \text{\acrshort{msa}}(\bm{v}_i, V) \nonumber \\
  \addlinespace[0.08cm]
    \text{decoder} \quad \quad & 
    \begin{aligned}
    O_{\bm{w}_i} &= \text{\acrshort{msca}}(\bm{w}_i, \tilde{V}, \{ \bm{w}_t \}_{t=1}^i), \\
    \end{aligned}
\end{align}
where $O$ is the network output, $\text{\acrshort{msa}}(\bm{x}, Y)$ is a self-attention with $\bm{x}$ mapped to query and $Y$ mapped to key-values, and $\text{\acrshort{msca}}(\bm{x}, Y, Z)$ indicates a self-attention with $\bm{x}$ as query and $Z$ as key-values, followed by cross-attention with $\bm{x}$ as query and $Y$ as key-values. We omit feed-forward layers and the dependency between consecutive layers for ease of notation.

\begin{figure} [t]
    \centering
    \footnotesize
    \setlength{\tabcolsep}{.2em}
    \resizebox{0.9\linewidth}{!}{
    \begin{tabular}{cccc}
    \textbf{Original Image} & \textbf{Depth Map} & \textbf{Objects} & \textbf{Visual Activation} \\
    \addlinespace[0.05cm]
     \includegraphics[width=0.25\columnwidth]{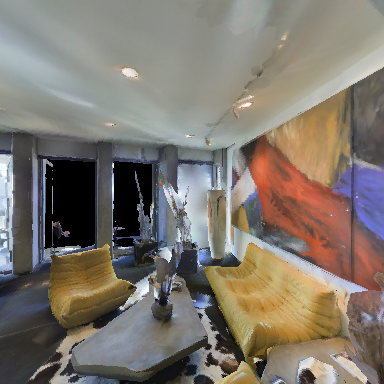} &
     \includegraphics[width=0.25\columnwidth]{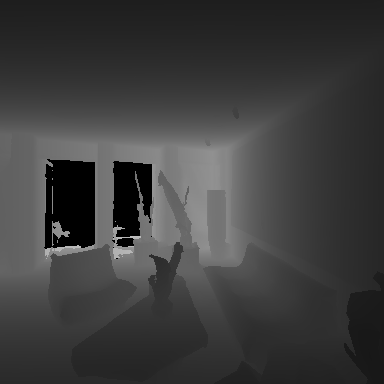} &
     \includegraphics[width=0.25\columnwidth]{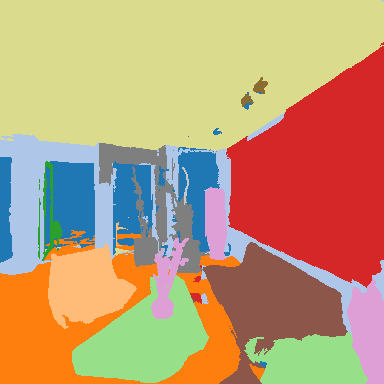} &
     \includegraphics[width=0.25\columnwidth]{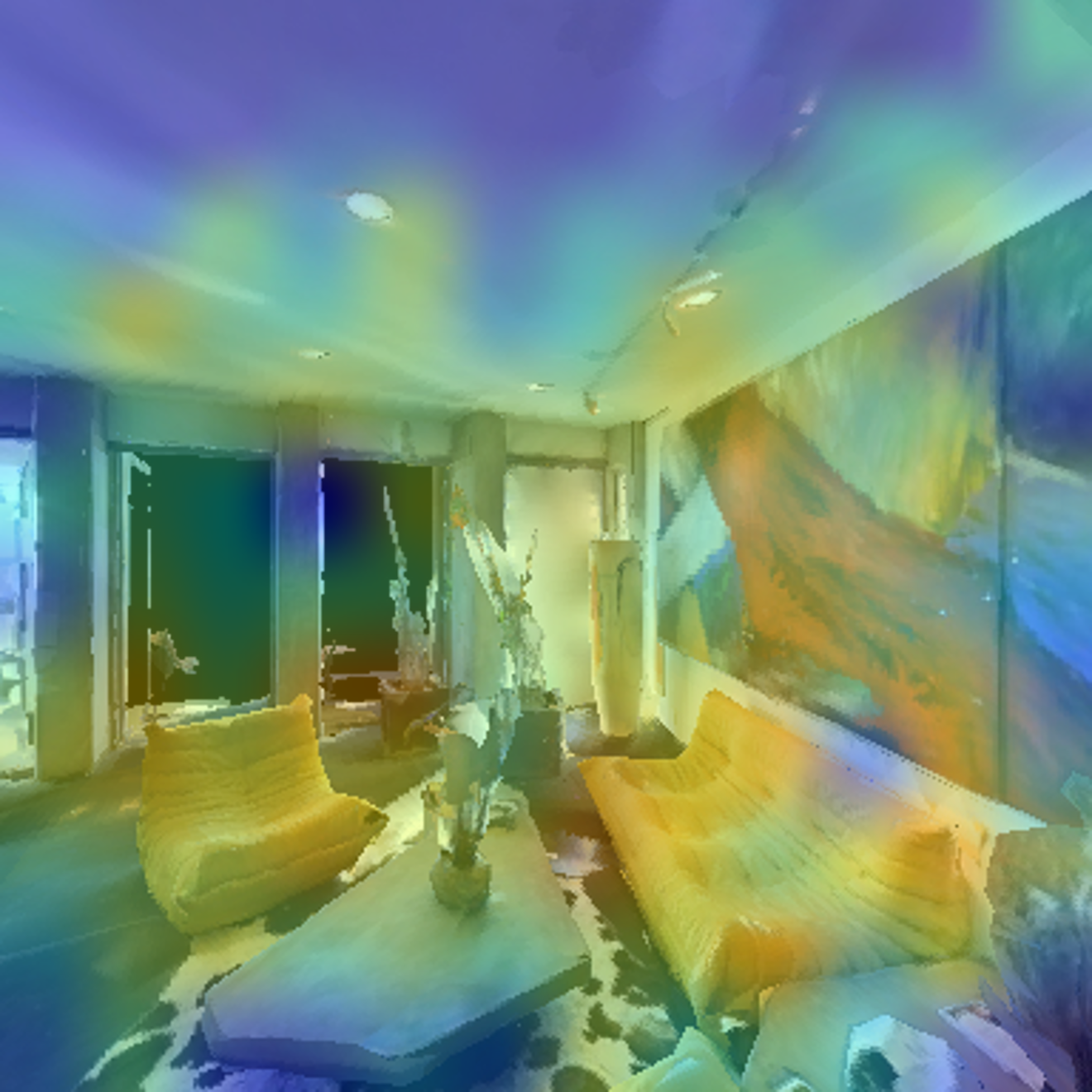} \\
    \end{tabular}
    }
    \caption{A sample of agent observation and corresponding images used by the speaker policy to trigger the captioner.}
    \label{fig:fig3_eds}
\end{figure}

\subsection{Speaker Policy}
While exploring the environment, the agent sees various RGB observations $\phi^{rgb}_t$. Even if the agent is navigating efficiently, the majority of the observations will be overlapped with each other, and the same objects will be observed at multiple consecutive timesteps. Since the agent should describe only relevant scenes during exploration and avoid uninformative captions or unnecessary repetitions, a criterion for making the captioner generate a description becomes necessary.
Similarly to \cite{bigazzi2020explore}, we use a speaker policy responsible for triggering the captioner. We compare three approaches that exploit different modalities and can be used regardless of the methods used for the other components of architecture: a depth-based policy, an object-based policy, and a visual activation-based policy. 

An example of the considered modalities for the same observation is reported in Fig. \ref{fig:fig3_eds}.

\tit{Depth-based Policy}
The depth-driven policy uses the current depth observation $\phi^{d}_t$ and computes the mean depth value. The captioner is activated if the mean depth value is above a predetermined threshold $\mathsf{D}$. High mean depth values indicate a larger area observed by the agent, and potentially, a richer scene to be described. In fact, when the field of view of the agent is occluded by an obstacle, the mean depth value of the observation $\phi^{d}_t$ is typically low. 

\tit{Object-based Policy}
Considering that the description of the scene will concentrate on relevant objects, the object-driven policy uses the number of relevant objects in the RGB observation $\phi^{rgb}_t$ to decide if the captioner should generate the description. Specifically, the captioner is triggered only if at least a number $\mathsf{O}$ of objects are being observed in the scene since using observations with multiple objects allows a larger variety of generated captions.

\tit{Visual Activation-based Policy}
Another possible strategy to implement the speaker policy entails exploiting the activation maps of the same visual encoder used by the captioner (which is a CLIP-like \cite{radford2021learning} encoder, as detailed in Sec. \ref{ssec:cap_module_eds}). Such a speaker policy is more closely related to the captioning module and provides a means to interpret the image regions that are more relevant to the agent. In particular, in this work, we consider a \acrshort{cnn}-based visual encoder and thus take the feature maps from the last convolutional block, projected into a $d$-dimensional vector. This vector is then averaged, and the speaker policy is activated if its average is above a certain threshold $\mathsf{A}$, thus indicating the presence of sufficient semantic content in the image.

\section{Experimental Setup}

\subsection{Implementation Details}

\tit{Navigation Module}
All the exploration models are trained for $5\text{M}$ frames on Gibson Dataset \cite{xia2018gibson} environments using Habitat simulator \cite{savva2019habitat}. The evaluation is performed using the  test split of \acrlong{mp3d} (\acrshort{mp3d}) dataset \cite{chang2017matterport3d} and the validation split of Gibson tiny dataset, because they contain object annotations that are used to evaluate the generated captions.

The RGB-D input to the components of the navigator is resized to $128\times128$ pixels, and the global map size is $W=2001$ for \acrshort{mp3d} and $W=961$ and Gibson environments. The size of the local map predicted by the mapper is $V=101$, and each pixel in the maps describes a $5 \times 5$ cm\textsuperscript{2} of the environment. Regarding the global policy, the global goal is sampled every $N_G=25$ timesteps. 
Both the global and local policies are trained using PPO algorithm \cite{schulman2017proximal} with a learning rate of $2.5\times10^{-4}$, while the mapper and the pose estimator use a learning rate of $10^{-3}$. Episode length is set to $500$ and $1000$ for the training and evaluation phases, respectively.

\tit{Speaker Policy}
The performance of the navigation module, which is able to move efficiently and capture interesting observations most of the time made it necessary to choose appropriate thresholds for the activation of the policy. Moreover, since the \acrshort{mp3d} dataset has richer object annotations and larger environments than the Gibson dataset, we compare two different sets of threshold values depending on the evaluation dataset for the depth- and object-based policies for triggering the captioner. In particular, the threshold values for \acrshort{mp3d} dataset are $\mathsf{D}=(1.0, 2.0, 3.0)$ and $\mathsf{O}=(1, 3, 5)$ for depth- and object-based policies. Depth and object thresholds are $\mathsf{D}=(1.0, 1.5, 2.0)$ and $\mathsf{O}=(1, 2, 3)$ for Gibson dataset. On the other hand, for the activation-based criterion, we use the same set of threshold values for both the evaluation datasets, \ie~$\mathsf{A}=(4.5, 5.0, 5.5)$.

\tit{Captioning Module}
As training and evaluation dataset, we employ COCO \cite{lin2014microsoft} following the splits defined in \cite{karpathy2015deep}. To improve the generalization abilities of the model, we also train a variant on a combination of 35.7M images taken from both human-collected datasets (\ie~COCO \cite{lin2014microsoft}) and web-collected sources (\ie~SBU \cite{ordonez2011im2text}, CC3M \cite{sharma2018conceptual}, CC12M \cite{changpinyo2021conceptual}, WIT \cite{srinivasan2021wit}, and a subset of YFCC100M \cite{thomee2016yfcc100m}).

We consider three configurations of the captioner, varying the number of decoding layers $l$, model dimensionality $d$, and number of attention heads $h$: Tiny ($l=3$, $d=384$, $h=6$), Small ($l=6$, $d=512$, $h=8$), and Base ($l=12$, $d=768$, $h=12$). For all models, we employ CLIP-ViT-L/14 \cite{radford2021learning} as visual feature extractor and three layers in the visual encoder. To assess the effectiveness of CLIP-based features, we also consider a variant of the Tiny model that employs region-based visual features, extracted from Faster R-CNN \cite{ren2017faster,anderson2018bottom}.
We train all captioning variants with cross-entropy loss using LAMB \cite{you2019large} as optimizer. We employ the learning rate scheduling strategy proposed in \cite{vaswani2017attention}, with a warmup of 6000 iterations and a batch size equal to 1080. We additionally finetune the models with the SCST strategy \cite{rennie2017self}, by using the Adam optimizer \cite{kingma2015adam}, a fixed learning rate equal to $5\times 10^{-6}$, and a batch size of 80.

\subsection{Evaluation Protocol}
As the task considered requires both exploration and description capabilities, for evaluation we propose to use both exploration and captioning-specific metrics, and a specific score devised for the task.

\tit{Navigation Module}
As for the performance of the navigation module, we express them in terms of metrics that are commonly used for evaluating embodied exploration agents. In particular, we consider the \acrlong{iou} between the ground-truth map of the environment and the map reconstructed by the agent (\textbf{$\mathsf{\acrshort{iou}}$}), the extent of correctly mapped area (\ie~the map accuracy \textbf{$\mathsf{\acrshort{acc}}$}), and the extent of environment area visited by the agent (\ie~the area seen \textbf{$\mathsf{\acrshort{as}}$}), both expressed in $m^2$. 

\tit{Captioning Module} 
For evaluating the performance of the captioner on the COCO dataset, we consider the standard image captioning metrics $\mathsf{\acrshort{bleu}}\text{-}\mathsf{4}$ \cite{papineni2002bleu}, $\mathsf{\acrshort{meteor}}$ \cite{banerjee2005meteor}, $\mathsf{\acrshort{rouge}}$ \cite{lin2004rouge}, $\mathsf{\acrshort{cider}}$ \cite{vedantam2015cider}, and $\mathsf{\acrshort{spice}}$ \cite{spice2016}.

\tit{Episode Description Score}
Different from standard captioning settings, where the ground truth captions are available for the images, in our setting such information is not available. However, the considered 3D environments datasets come with annotations of the objects in the scene, which can be exploited for performance evaluation. In particular, based on the objects in the scene, we use the soft-\acrlong{cov} score (\textbf{$\mathsf{\acrshort{cov}}$}) and the \acrlong{div} score (\textbf{$\mathsf{\acrshort{div}}$}) as presented in Chapter \ref{chap:ex2}, to evaluate the ability of the agent to mention all the relevant objects in the scene and to produce interesting, non-repetitive descriptions, respectively.
The first is computed by considering the intersection score between the set of nouns in the produced caption and the set of categories of the relevant objects in the scene. Note that in this work, by `relevant object' we mean those whose area covers at least 10\% of the total image area, and thus, can be more useful to identify a scene.
The latter score is defined as the \acrlong{iou} between the sets of nouns mentioned in two consecutively generated captions.

Additionally, we measure the agent's overall loquacity (\textbf{$\mathsf{Loq}$}) as the number of times it is activated by the speaker policy, normalized by the episode length. In other words, the loquacity can be seen as the inverse of the average number of navigation steps between two consecutive captions. Moreover, we resort to the recently-proposed \acrlong{clips} ($\mathsf{\acrshort{clips}}$) \cite{hessel2021clipscore}, in its unpaired definition.

Finally, to evaluate an overall system on each episode of the proposed task, we define an ad hoc score to measure the concept coverage of the generated descriptions, which is an important aspect of the task.
The proposed \acrlong{eds} (\textbf{$\mathsf{\acrshort{eds}}$}) reflects the ability of the robot to produce sufficient descriptions in strategic moments so that the maximum amount of information collected in the environment is covered. The rationale is that it should capture the ability of the agent to mention all the relevant landmarks (objects and rooms) when needed, without unnecessary repetitions. This makes the description more useful and interesting.
The score is defined as:
\begin{equation}
    \mathsf{\acrshort{eds}} = \overline{\mathsf{\acrshort{clips}}} \cdot \mathsf{\acrshort{iou}}(\textbf{N}, \textbf{O}) \cdot \text{\%}\mathsf{\acrshort{as}},
\end{equation}
where $\overline{\mathsf{\acrshort{clips}}}$ is the mean of the CLIP scores of all the captions produced during the episode. Moreover, $\textbf{N}$ is the list of nouns in all the captions produced during the episode, and $\textbf{O}$ is the list of objects in the environment. The \acrlong{iou} operator $\mathsf{\acrshort{iou}}$ is implemented via the Jonker-Volgenant linear assignment algorithm \cite{jonker1987shortest}. Finally, \%$\mathsf{\acrshort{as}}$ is the percentage of the total environment area visited by the agent.
At the dataset level, the $\mathsf{\acrshort{eds}}$ is given by the average of the scores obtained in the dataset episodes.

\begin{table}[t]
\centering
\setlength{\tabcolsep}{.35em}
\resizebox{0.95\linewidth}{!}{
\begin{tabular}{l c ccc c ccc}
\toprule
 & & \multicolumn{3}{c}{\textbf{Gibson Val}} & & \multicolumn{3}{c}{\textbf{MP3D Test}} \\
\cmidrule{3-5} \cmidrule{7-9}
\textbf{Model} & & 
$\mathsf{\acrshort{iou}}$ & $\mathsf{\acrshort{acc}}$ & $\mathsf{\acrshort{as}}$ & & 
$\mathsf{\acrshort{iou}}$ & $\mathsf{\acrshort{acc}}$ & $\mathsf{\acrshort{as}}$ \\
\midrule
Curiosity~\cite{ramakrishnan2020exploration} 
& & 0.528 & 66.19 & 102.59 
& & 0.368 & 130.34 & 186.67\\
Coverage~\cite{chaplot2019learning} 
& & 0.608 & 73.69 & 102.66
& & 0.417 & 146.16 & 195.03 \\
Anticipation~\cite{ramakrishnan2020occupancy} 
& & 0.706 & 81.13 & 102.22
& & 0.494 & 157.02 & 177.14 \\
Impact (Grid)~\cite{bigazzi2022impact} 
& & \textbf{0.738} & \textbf{82.91} & 104.16 
& & \textbf{0.519} & 164.26 & 185.13 \\
Impact (\acrshort{dme})~\cite{bigazzi2022impact} 
& & 0.694 & 79.47 & \textbf{105.03}
& & 0.496 & \textbf{167.58} & \textbf{205.02} \\
\bottomrule
\end{tabular}
}
% \captionsetup{justification=centering}
\caption{Navigation results on Gibson Val and \acrshort{mp3d} Test. Impact (DME) model achieves the best results in terms of \acrlong{as}.}
% \label{tab:navigation}
\label{tab:tab1_eds}
\end{table}
\begin{table}[t]
\centering
\setlength{\tabcolsep}{.25em}
\resizebox{\linewidth}{!}{
\begin{tabular}{lc ccccccc}
\toprule
 & & \makecell{\textbf{Train Ims}} & $\mathsf{\acrshort{bleu}}\text{-}\mathsf{4}$ & $\mathsf{\acrshort{meteor}}$ & $\mathsf{\acrshort{rouge}}$ & $\mathsf{\acrshort{cider}}$ & $\mathsf{\acrshort{spice}}$ \\
\midrule
Region-based$^\text{tiny}$ & & 112k & 37.7 & 28.3 & 57.6 & 124.8 & 21.9 \\
CLIP-based$^\text{tiny}$ & & 112k & 40.6 & 30.0 & 59.9 & 139.4 & 23.9 \\
CLIP-based$^\text{small}$ & & 112k & 40.9 & 30.4 & 60.1 & 141.5 & 24.5 \\
CLIP-based$^\text{base}$ & & 112k & 41.4 & 30.2 & 60.2 & 142.0 & 24.0 \\
CLIP-based$^\text{base}$ & & 35.7M & \textbf{42.9} & \textbf{31.4} & \textbf{61.5} & \textbf{149.6} & \textbf{25.0} \\
\bottomrule
\end{tabular}
}
% \captionsetup{justification=centering}
\caption{Captioning results on the COCO test set. CLIP-based$^\text{base}$ achieves the best results on all the episode description metrics.}
% \label{tab:captioning}
\label{tab:tab2_eds}
\end{table}

\section{Experimental Results}

\subsection{Navigation Results}
First, we compare the different exploration approaches alone on the \acrshort{mp3d} and Gibson datasets. The results of this analysis are reported in Table \ref{tab:tab1_eds}.
The best agent in terms of the area seen (\textbf{$\mathsf{\acrshort{as}}$}) is the impact-based method using density model estimation. In particular, this approach is able to efficiently explore both Gibson and \acrshort{mp3d} datasets, giving its best in large environments. In fact, the small $0.87 \text{m}^2$ margin over the second best approach on Gibson becomes $9.99 \text{m}^2$ in the larger \acrshort{mp3d} environments. Moreover, this method is still competitive in terms of \textbf{$\mathsf{\acrshort{iou}}$}, being also the best in terms of \textbf{$\mathsf{\acrshort{acc}}$} on the \acrshort{mp3d} test split. In light of these results, we use the impact-based navigator with \acrshort{dme} as the navigator of the overall approach.

\subsection{Captioning Results}
Then, we evaluate the performance of the captioner alone on the COCO dataset. The results of this analysis are reported in Table \ref{tab:tab2_eds}. It can be observed that the CLIP-based variants are the best-performing ones, with a noticeable advantage over the region-based captioner. This confirms the representative power of CLIP features. The Base variant has also been trained on additional image-caption pairs from web-collected sources, which further increases its performance.
It is worth mentioning that these results are in line with those of state-of-the-art captioners (\eg \cite{li2020oscar,zhang2021vinvl}).
In light of these results, we use the CLIP-based Base variant as the captioner of the overall approach.

\begin{table}[t]
\setlength{\tabcolsep}{.3em}
\resizebox{\linewidth}{!}{
\begin{tabular}{lcc cccc c cccc}
\toprule
& & & \multicolumn{4}{c}{\textbf{COCO Only}} & & \multicolumn{4}{c}{\textbf{COCO + Web-Collected}} \\
\cmidrule{3-7} \cmidrule{9-12}
 & $\mathsf{\acrshort{loq}}$ & & $\mathsf{\acrshort{cov}}$ & $\mathsf{\acrshort{div}}$ & $\mathsf{\acrshort{clips}}$ & $\mathsf{\acrshort{eds}}$  & & $\mathsf{\acrshort{cov}}$ & $\mathsf{\acrshort{div}}$ & $\mathsf{\acrshort{clips}}$ & $\mathsf{\acrshort{eds}}$ \\
\midrule
\textbf{Always} & 100.00 & & 0.864 & 0.352 & 0.670 & 0.119 & & 0.862 & 0.348 & 0.692 & 0.120 \\
\midrule
\textbf{Depth} \\
\hspace{0.3cm}$\mathsf{D\geq1.0}$ & 83.26 & & 0.868 & 0.335 & 0.670 & 0.140 & & 0.865 & 0.338 & 0.690 & 0.140 \\
\hspace{0.3cm}$\mathsf{D\geq1.5}$ & 55.24 & & 0.871 & 0.323 & 0.664 & 0.203 & & 0.868 & 0.330 & 0.683 & 0.204 \\
\hspace{0.3cm}$\mathsf{D\geq2.0}$ & 27.38 & & 0.793 & 0.293 & 0.629 & 0.250 & & 0.780 & 0.304 & 0.650 & 0.257 \\
\midrule
\textbf{Object} \\
\hspace{0.3cm}$\mathsf{O\geq1}$ & 41.73 & & 0.793 & 0.314 & 0.663 & 0.222 & & 0.784 & 0.332 & 0.682 & 0.225 \\
\hspace{0.3cm}$\mathsf{O\geq2}$ & 21.55 & & 0.703 & 0.289 & 0.645 & 0.219 & & 0.697 & 0.307 & 0.664 & 0.220 \\
\hspace{0.3cm}$\mathsf{O\geq3}$ &  7.58 & & 0.416 & 0.232 & 0.549 & 0.107 & & 0.410 & 0.260 & 0.561 & 0.105 \\
\midrule
\textbf{Activation} \\
\hspace{0.3cm}$\mathsf{A\geq4.5}$ & 87.79 & & 0.866 & 0.340 & 0.672 & 0.134 & & 0.864 & 0.343 & 0.691 & 0.134 \\
\hspace{0.3cm}$\mathsf{A\geq5.0}$ & 51.13 & & 0.828 & 0.349 & 0.674 & 0.223 & & 0.827 & 0.348 & 0.691 & 0.220 \\
\hspace{0.3cm}$\mathsf{A\geq5.5}$ &  2.20 & & 0.133 & 0.153 & 0.455 & 0.038 & & 0.140 & 0.153 & 0.464 & 0.040 \\
\bottomrule
\end{tabular}
}
\centering
\captionsetup{justification=centering}
\caption{Episode description results on Gibson tiny validation set.} 
\label{tab:tab3_eds}
% \label{tab:ex2_results_gibson}
\end{table}
\begin{table}[t]
\setlength{\tabcolsep}{.3em}
\resizebox{\linewidth}{!}{
\begin{tabular}{lcc cccc c cccc}
\toprule
& & & \multicolumn{4}{c}{\textbf{COCO Only}} & & \multicolumn{4}{c}{\textbf{COCO + Web-Collected}} \\
\cmidrule{3-7} \cmidrule{9-12}
 & $\mathsf{\acrshort{loq}}$ & & $\mathsf{\acrshort{cov}}$ & $\mathsf{\acrshort{div}}$ & $\mathsf{\acrshort{clips}}$ & $\mathsf{\acrshort{eds}}$  & & $\mathsf{\acrshort{cov}}$ & $\mathsf{\acrshort{div}}$ & $\mathsf{\acrshort{clips}}$ & $\mathsf{\acrshort{eds}}$ \\
\midrule
 \textbf{Always} & 100.00 & & 0.768 & 0.363 & 0.648 & 0.172 & & 0.771 & 0.348 & 0.687 & 0.179 \\
\midrule
\textbf{Depth} \\
\hspace{0.3cm}$\mathsf{D\geq1.0}$ & 89.05 & & 0.765 & 0.352 & 0.648 & 0.180 & &
                                              0.767 & 0.341 & 0.687 & 0.187 \\
\hspace{0.3cm}$\mathsf{D\geq2.0}$ & 45.06 & & 0.751 & 0.317 & 0.637 & 0.155 & & 
                                              0.750 & 0.311 & 0.668 & 0.160 \\
\hspace{0.3cm}$\mathsf{D\geq3.0}$ & 15.98 & & 0.317 & 0.161 & 0.338 & 0.030 & & 
                                              0.317 & 0.151 & 0.360 & 0.031 \\
\midrule
\textbf{Object} \\
\hspace{0.3cm}$\mathsf{O\geq1}$ & 75.82 & & 0.754 & 0.340 & 0.635 & 0.190 & &
                                            0.756 & 0.333 & 0.670 & 0.196 \\
\hspace{0.3cm}$\mathsf{O\geq3}$ & 46.57 & & 0.700 & 0.310 & 0.605 & 0.168 & & 
                                            0.701 & 0.310 & 0.634 & 0.172 \\
\hspace{0.3cm}$\mathsf{O\geq5}$ & 19.90 & & 0.616 & 0.255 & 0.533 & 0.106 & & 
                                            0.614 & 0.254 & 0.553 & 0.107 \\
\midrule
\textbf{Activation} \\
\hspace{0.3cm}$\mathsf{A\geq4.5}$ & 82.05 & & 0.765 & 0.348 & 0.641 & 0.106 & & 
                                              0.767 & 0.337 & 0.676 & 0.107 \\
\hspace{0.3cm}$\mathsf{A\geq5.0}$ & 46.28 & & 0.754 & 0.350 & 0.636 & 0.153 & & 
                                              0.757 & 0.341 & 0.667 & 0.158 \\
\hspace{0.3cm}$\mathsf{A\geq5.5}$ &  1.28 & & 0.325 & 0.118 & 0.347 & 0.015 & & 
                                              0.328 & 0.116 & 0.362 & 0.016 \\
\bottomrule
\end{tabular}
}
\centering
\captionsetup{justification=centering}
\caption{Episode description results on MP3D test set.} 
\label{tab:tab4_eds}
% \label{tab:ex2_results_mp3d}
\end{table}

\subsection{Episode Description Results} 
\begin{figure} [t]
    \centering
    \includegraphics[width=0.8\linewidth]{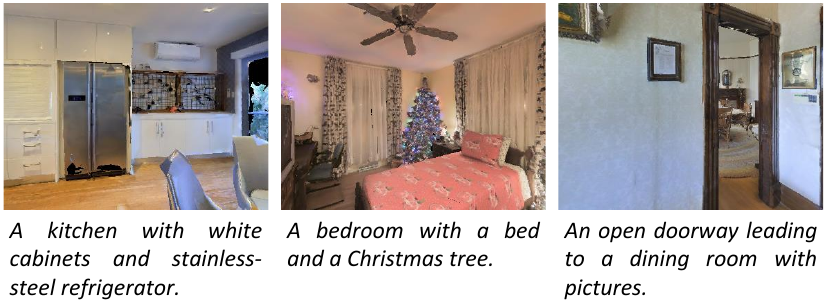}
    \captionsetup{justification=centering}
    \caption{Sample observations and corresponding captions generated by our model.}
    \label{fig:fig4_eds}
\end{figure}

Finally, we compare variants of the overall approach using different speaking policies with different threshold values, and use as reference a dummy policy according to which the captioning module is always activated. The results are reported in Tables \ref{tab:tab3_eds} and \ref{tab:tab4_eds}. It can be noticed that the captioner trained on web-collected sources performs better that the variant trained on COCO only in terms of all metrics, suggesting its superior generalization capabilities and thus, suitability to be employed in an embodied setting. However, to evaluate on the overall task, the proposed $\mathsf{\acrshort{eds}}$ score is more informative than the other metrics, which can nonetheless be used in combination with the $\mathsf{\acrshort{eds}}$ to gain additional insights on the agents' behaviour. In fact, the values of all metrics but the $\mathsf{\acrshort{eds}}$ are comparable in both datasets, while the $\mathsf{\acrshort{eds}}$ is on average higher on Gibson: this is due to the fact that Gibson has on average smaller and less cluttered spaces, which can be more easily fully explored (higher values of the $\mathsf{\acrshort{cov}}$ on Gibson confirm this intuition). This trend is further confirmed by the fact that on the Gibson dataset, the speaking policy must ensure the \acrfull{loq} being in a specific range (roughly between 20 and 80) to obtain the best $\mathsf{\acrshort{eds}}$ scores, while on the wider spaces of \acrshort{mp3d}, speaking policies ensuring a higher $\mathsf{\acrshort{loq}}$ lead to better performance. Qualitative examples of the output of our approach on selected observations are reported in Fig. \ref{fig:fig4_eds}.

\subsection{Real-World Deployment}
Using exploration agents trained on the photo-realistic environments of the Habitat simulator and general-purpose captioners allows the deployment of our approach to the real world. In this respect, we use a LoCobot platform \cite{locobot}. For the deployment, the captioner is left untouched, whilst we modify the camera parameters of the navigator, such as camera height, \acrlong{fov}, and depth sensor range to match the real-world setting. Furthermore, the deployed agent is trained by adding noise models fitted to mimic the LoCobot camera noise over the observations retrieved from the simulator. As the last step, we apply the correction presented by \cite{bigazzi2021out} to correct noisy real-world depth observations. We test the agent exploration and description in a real-world apartment.

\chapter[Embodied Agents in Changing Environments]{Spot the Difference: \\ \Large Embodied Agents in Changing Environments}
\label{chap:sd}
\blfootnote{This Chapter is related to the publication ``F. Landi \etal, Spot the Difference: A Novel Task for Embodied Agents in Changing Environments, ICPR 2022'' \cite{landi2022spot}. See the list of Publications on page \pageref{publications} for more details.}
\RemoveLabels

\lettrine[lines=1]{\textcolor{SchoolColor}{T}}{he} work described in previous chapters follows the same paradigm frequently used in \gls{eai}, the one of starting every embodied navigation episode without any previously acquired knowledge of the environment. However, now imagine you have just bought a personal robot, and you ask it to bring you a cup of tea. It will start roaming around the house while looking for the cup. It probably will not come back until some minutes, as it is new to the environment. After the robot knows your house, instead, you expect it to perform navigation tasks much faster, exploiting its previous knowledge of the environment while adapting to possible changes in objects, people, and furniture positioning.
In fact, usually in the literature on \gls{eai}, the agent is initialized in a completely unknown environment. We believe that this choice is not supported by real-world experience, where information about the environment can be stored and reused for future tasks.
Nevertheless, as agents are likely to stay in the same place for long periods, such information may be outdated and inconsistent with the actual layout of the environment. Therefore, the agent also needs to discover those differences during navigation.

In this work, we introduce a new task for \acrshort{eai}, which we name \emph{Spot the Difference}. In the proposed setting, the agent must identify all the differences between an outdated map of the environment and its current state, a challenge that combines visual exploration using monocular images and embodied navigation with spatial reasoning. To succeed in this task, the agent needs to develop efficient exploration policies to focus on likely changed areas while exploiting priors about objects of the environment. We believe that this task could be useful to train agents that will need to deal with changing environments.

\begin{figure}[t]
\centering
\scriptsize
\setlength{\tabcolsep}{.15em}
\begin{tabular}{cccccc}
\textbf{Original Map} & & & \multicolumn{3}{c}{\textbf{Sample Manipulated Maps}} \\
\addlinespace[0.12cm]
\includegraphics[width=0.22\linewidth]{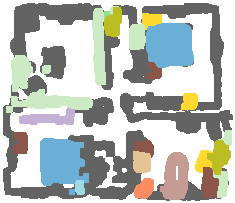} & & &
\includegraphics[width=0.22\linewidth]{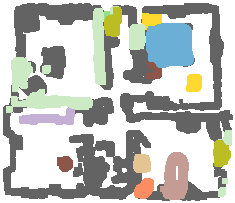} &
\includegraphics[width=0.22\linewidth]{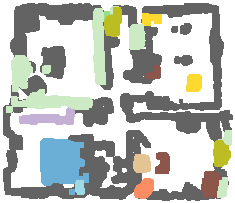} &
\includegraphics[width=0.22\linewidth]{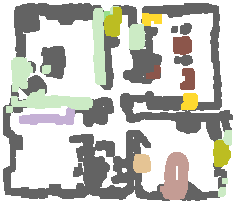} \\
\includegraphics[width=0.22\linewidth]{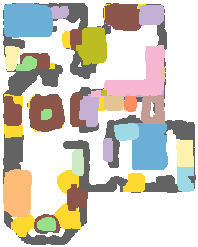} & & &
\includegraphics[width=0.22\linewidth]{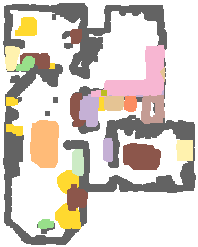} &
\includegraphics[width=0.22\linewidth]{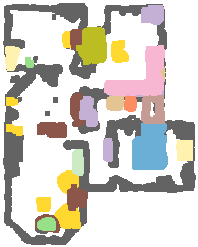} &
\includegraphics[width=0.22\linewidth]{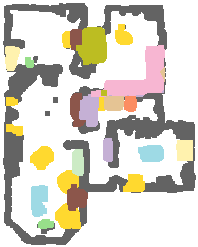} \\
\includegraphics[width=0.22\linewidth]{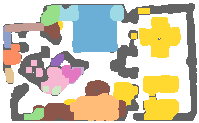} & & &
\includegraphics[width=0.22\linewidth]{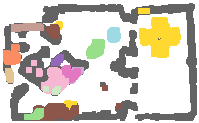} &
\includegraphics[width=0.22\linewidth]{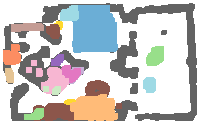} &
\includegraphics[width=0.22\linewidth]{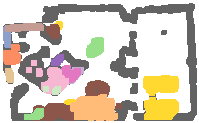} \\
\addlinespace[0.12cm]
\multicolumn{6}{c}{\includegraphics[width=0.9\linewidth]{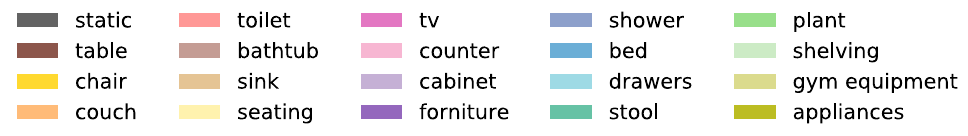}}
\end{tabular}
\caption{Generation of alternative states of an environment: original and sample manipulated semantic maps.}
\label{fig:fig1_sd}
\end{figure}

\AddLabels

Recent work on \acrshort{eai} \cite{chen2019learning,luperto2020robot,ramakrishnan2020exploration,karkus2021differentiable,mayo2021visual}, considers building a representation of the state of the environment to increase the performance on both exploration and down-stream tasks. 
Unfortunately, if the environment changes over time, the agent needs to rebuild a full representation from scratch and cannot count on an efficient policy to update its internal representation of the environment. In the following, we simulate the natural evolution of an environment and design a specific policy to navigate in changing environments seamlessly.  

Due to the high cost of 3D acquisitions from the real world, existing datasets of photo-realistic 3D spaces \cite{chang2017matterport3d,xia2018gibson} do not contain different layouts for the same environment. In this work, we create a reproducible setup to generate alternative layouts for an environment. We semi-automatically build a dataset of 2D semantics occupancy maps in which the objects can be removed and rearranged while the area and the position of architectural elements do not change (Fig. \ref{fig:fig1_sd}). 
In the proposed setting, the agent is deployed in an interactive 3D environment and provided with a map from our produced dataset, which represents the information retained while performing tasks in a past state of the environment.

\thispagestyle{nosection}

To train agents that can deal with changing environments efficiently, we develop a novel reward function and an approach for navigation aiming at finding relevant differences between the previous layout of the environment and the current one. Our method is based on \acrlong{ans} paradigm proposed in \cite{chaplot2019learning} and \cite{ramakrishnan2020occupancy}. Differently from previous proposals, it can read and update the given map to identify relevant differences in the environment in the form of their projections on the map.
Experimental results show that our approach performs better than existing state-of-the-art architectures for exploration in our newly-proposed task. We also compare with different baselines and evaluate our agent in terms of the percentage of area seen in the environment, percentage of discovered differences, and metric curves at varying exploration time budgets.
The new dataset, together with our code and pretrained models, is released publicly\footnote{https://github.com/aimagelab/spot-the-difference}.

\section{Spot the Difference Task}
At the beginning of an episode, the agent is spawned in a 3D environment and is given a prebuilt occupancy map $M$, representing its spatial knowledge of the environment, \ie~a previous state of the environment that is now obsolete:
\begin{equation}
    M = (m_{ij}) \in [0,1], \quad 0 \leq i,j < W, 
\end{equation}
where $m_{ij}$ represents the probability of finding an obstacle at coordinates $(i,j)$.
The task entails exploring the current environment to recognize all the differences with respect to the state in which $M$ was computed, in the form of free and occupied space.
To accomplish the task, the agent should build a correct occupancy map of the current environment starting from $M$, recognizing and focusing on parts that are likely to change (\eg, the middle of wide rooms rather than tight corridors).

For every episode of \textit{Spot the Difference}, the agent is given a time budget of $T$ timesteps. At time $t=0$, the agent holds the starting map representation $M$. At each timestep $t$, the map is updated depending on the current observation to obtain $M_t$. Whenever the agent discovers a new object or a new portion of free space, the internal representation of the map changes accordingly. The goal is to gather as much information as possible about changes in the environment by the end of the episode. To measure the agent performance, we compare the final map $M_T$ produced by the agent with the ground-truth occupancy map $M^*$.
In this sense, the paradigm we adopt is the one of knowledge reuse starting from partial knowledge.

\begin{table}[t!]
\centering
\setlength{\tabcolsep}{.4em}
\resizebox{\linewidth}{!}{
\begin{tabular}{lccccc}
\toprule
\textbf{Dataset Split} & & \textbf{Semantic Classes} & \textbf{Scans} & \textbf{Generated \acrlongpl{som}} & \textbf{Episodes} \\
\midrule
\acrshort{mp3d} Train  & & 42 & 58 & 2070 & $\approx 4.5\text{M}$ \\
\acrshort{mp3d} Val  & & 42 & 9 & 160 & 320 \\
\acrshort{mp3d} Test  & & 42 & 14 & 260 & 610 \\
Gibson Val  & & 20 & 5 & 130 & 450 \\
\bottomrule
\end{tabular}
}
\captionsetup{justification=centering}
\caption{Number of manipulated maps generated per dataset split.}
\label{tab:dataset_detail}
\end{table}

\section{Dataset Creation} 
\label{ssec:dataset_sd}
In this section, we describe the newly-proposed dataset that we create to enable research on \emph{Spot the Difference}.

\subsection{Semantic Occupancy Map}
Given a 3D environment, we place the agent in a free navigable location with heading $\theta=0\degree$ (facing eastward).
We assume that the input consists of a depth image and a semantic image and that the camera intrinsics $K$ are known.
To build the \gls{som} of an environment, we project each semantic pixel of the acquired scene into a 2-dimensional top-down map:
given a pixel with image coordinates $(i,j)$ and depth value $d_{i,j}$, we first recover its coordinates $(x,y,z)$ with respect to the agent position. 
Then, we compute the corresponding $(u,v)$ pixel in map through an orthographic projection, using the information about the agent position and heading:
\begin{equation}
    \begin{bmatrix}
        x \\ y \\ z
    \end{bmatrix}
     = d_{i,j} K^{-1}
     \begin{bmatrix}
         i \\ j \\ 1
     \end{bmatrix}
     \text{,}\quad\text{and}\quad
     \begin{bmatrix}
         u \\ v \\ 0 \\ 1
     \end{bmatrix}
     = P_v
     \begin{bmatrix}
         x \\ y \\ z \\ 1
     \end{bmatrix} .
\end{equation}
We perform the same operation after rotating the agent by $\Delta_\theta=30\degree$ until we perform a span from $0\degree$ to $180\degree$. To cover the whole scene, we repeat this procedure placing the agent at a distance of $0.5m$ from the previous capture point, following the axis directions. The agent elevation is instead kept fixed. During this step, we average the results of subsequent observations of overlapping portions of space.

After the acquisition, we obtain a \gls{som} with $C$ channels, where each pixel corresponds to a $5cm \times 5cm$ portion of space in the 3D environment. For each channel $c \in \{0, ..., C\}$, the map values represent the probability that the corresponding portion of space is occupied by an object of semantic class $c$.

\subsection{Multiple SOMs for Each Environment}
The \glspl{som} obtained in the previous step can be seen as one possible layout for the corresponding 3D environments.
In order to create a dataset with different states (\ie~different layouts) of the same environment, instead of manipulating the real-world 3D scenes (changing the furniture position, removing chairs, \etc), we propose to modify the \gls{som} to create a set of plausible and different layouts for the environment.

First, we isolate the objects belonging to each semantic category by using an algorithm for connected component labeling \cite{grana2010optimized,bolelli2019spaghetti,allegretti2019optimized}. Then, we sample a subset of objects to be deleted from the map and a subset of objects to be re-positioned in a different free location on the map. During sampling, we consider categories that have a high probability of being displaced or removed in the real world and ignore non-movable semantic categories such as \textit{fireplaces}, \textit{columns}, and \textit{stairs}.
After this step, we obtain a new \gls{som} representing a possible alternative state for the environment, which could be very different from the one in which the 3D acquisition was taken. Sample manipulated maps can be found in Fig. \ref{fig:fig1_sd} and Fig \ref{fig:sup_maps}.

\subsection{Dataset Details}
To generate alternative \glspl{som}, we start from the \acrlong{mp3d} (\acrshort{mp3d}) dataset of spaces \cite{chang2017matterport3d}, which comprises $90$ different building scans, and is enriched with dense semantic annotations.
We consider each floor in the building and compute the \gls{som} for that floor. For each map, we create 10 alternative versions of that same environment. In this step, we discard the floors that have few semantic objects (\eg,~empty rooftops) or that are not fully navigable by the agent. As a result, we retain $249$ floors belonging to $81$ different buildings, thus generating a total of $2490$ different semantic occupancy maps for these floors.  Finally, we split the dataset into train, validation, and test subsets.

As an additional testbed, we also build a set of out-of-domain maps ($13$ floors from $5$ spaces) taken from the Gibson dataset \cite{xia2018gibson}, enriched with semantic annotations from \cite{armeni20193d} and manipulated as done for the \acrshort{mp3d} dataset. For each \gls{som}, multiple episodes are generated by selecting different starting points.
More information about our dataset can be found in Table \ref{tab:dataset_detail}.

\tit{Semantic Classes Division}
The generation of semantic maps for each floor of each scene produces $2001 \times 2001 \times 43$ maps for the \acrshort{mp3d} dataset and $961 \times 961 \times 21$ maps for the Gibson dataset. The last channel of every map registers the explorable space, so it is ignored for the creation of the dataset and is concatenated, as it is, to the manipulated map obtained at the end of the semi-automatic dataset creation process.

We divide the semantic channels of the maps depending on the possible actions that can be performed on the connected components in that channel. We identify four types of classes: \textit{No Operation}, \textit{Removal}, \textit{Displacement}, and \textit{Overlap Removal}. A list of semantic categories with their classification is reported at the end of this Chapter in Table \ref{tab:semantic_classes_mp3d} for the \acrshort{mp3d} dataset and in Table \ref{tab:semantic_classes_gibson} for the Gibson dataset.
\textit{No Operation} classes are left untouched, and correspond to non-movable objects, such as \textit{wall}, \textit{stairs}, and \textit{columns}; the connected component of the \textit{Removal} classes can be removed; those in the \textit{Displacement} classes can be either removed or relocated in other free spaces in the environment; and \textit{Overlap Removal} components are removed if connected components removed or displaced in other channels overlap with them, \eg,~if a \textit{sofa} is removed, every instance of \textit{cushion} overlapping with that \textit{sofa} will be removed as well because it is supposed to be on top of it.

\begin{figure*}[t]
\centering
\scriptsize
\setlength{\tabcolsep}{.2em}
\resizebox{\linewidth}{!}{
\begin{tabular}{cccccc}
\textbf{Original Map} & & \textbf{Manipulated Map 1} & \textbf{Difference Map 1}  & \textbf{Manipulated Map 2}  & \textbf{Difference Map 2} \\
\addlinespace[0.12cm]
\includegraphics[width=0.18\linewidth]{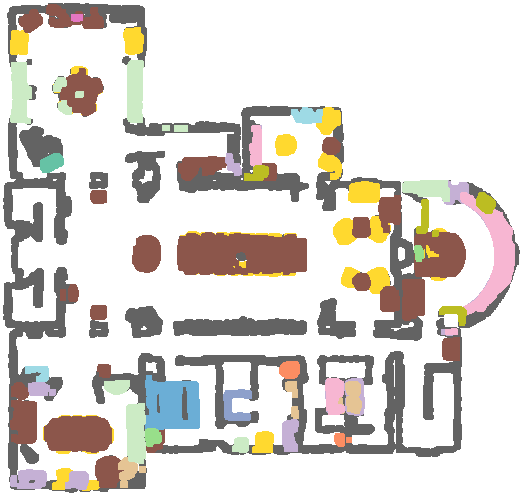} & & 
\includegraphics[width=0.18\linewidth]{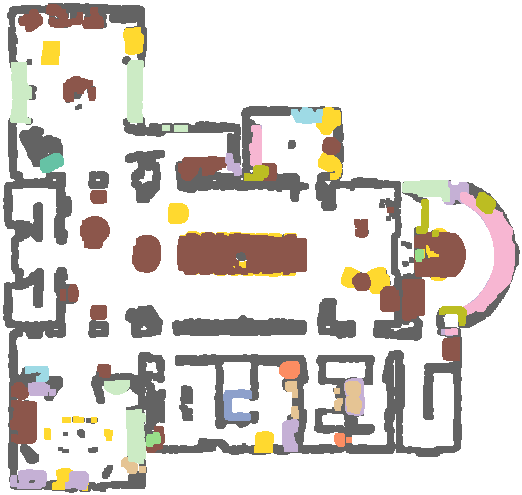} &
\includegraphics[width=0.18\linewidth]{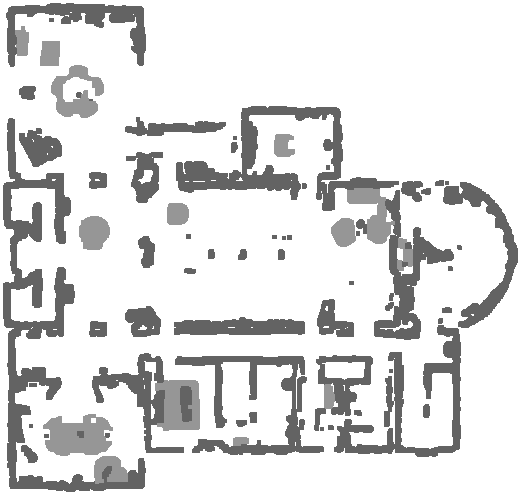} &
\includegraphics[width=0.18\linewidth]{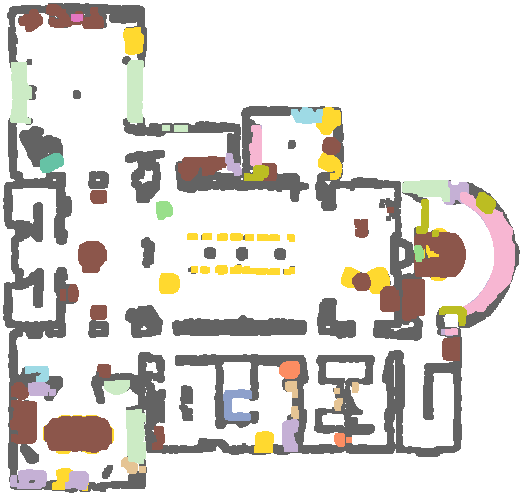} &
\includegraphics[width=0.18\linewidth]{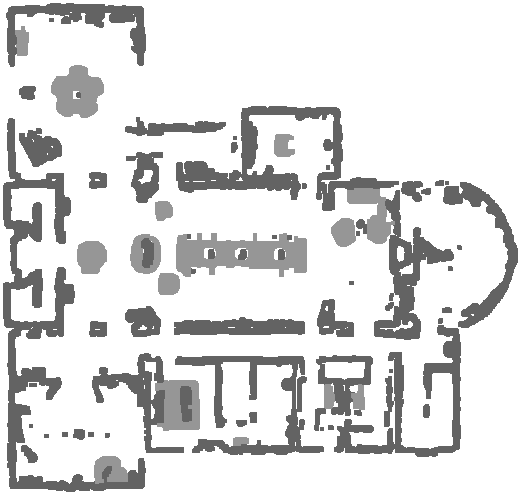} \\
\addlinespace[0.05cm]
\includegraphics[width=0.18\linewidth]{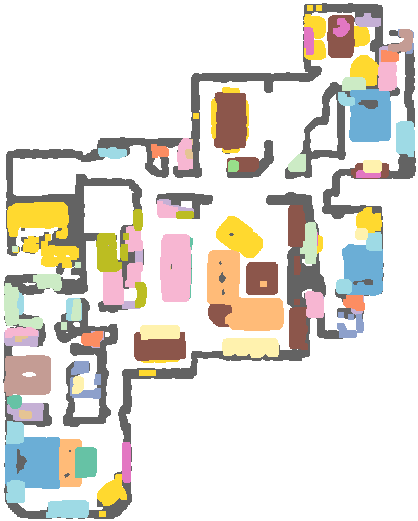} & & 
\includegraphics[width=0.18\linewidth]{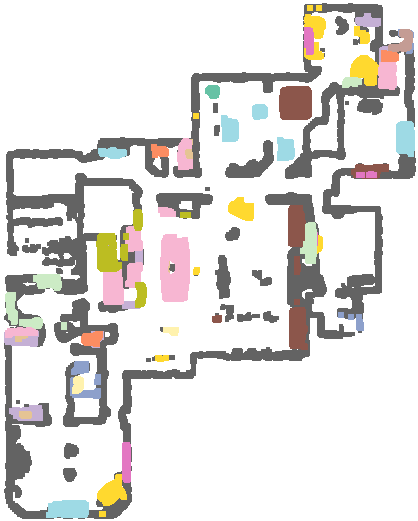} &
\includegraphics[width=0.18\linewidth]{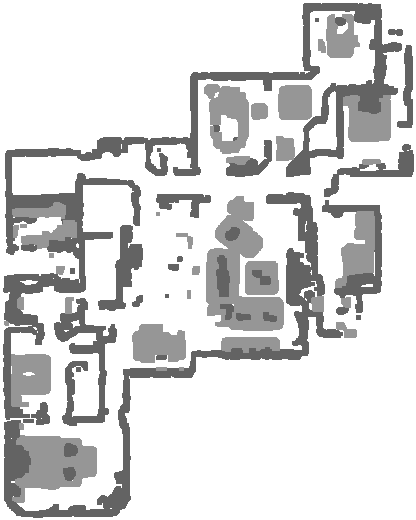} &
\includegraphics[width=0.18\linewidth]{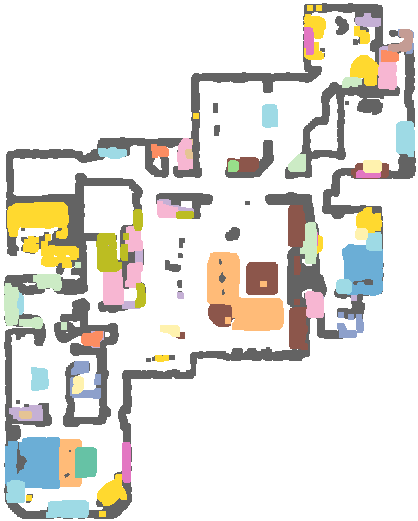} &
\includegraphics[width=0.18\linewidth]{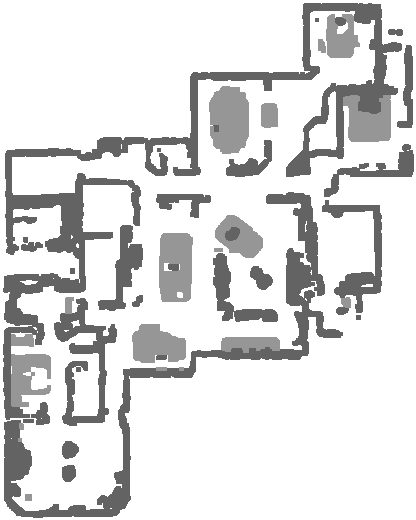} \\
\addlinespace[0.05cm]
\includegraphics[width=0.18\linewidth]{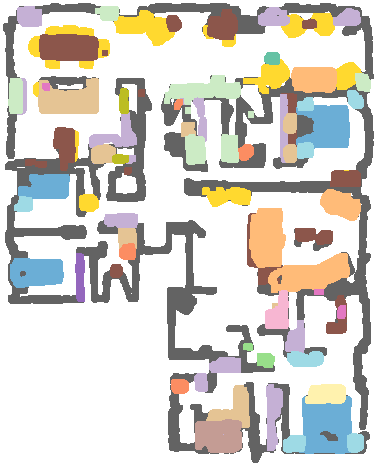} & & 
\includegraphics[width=0.18\linewidth]{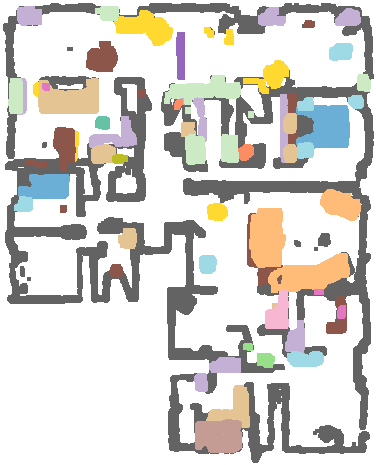} &
\includegraphics[width=0.18\linewidth]{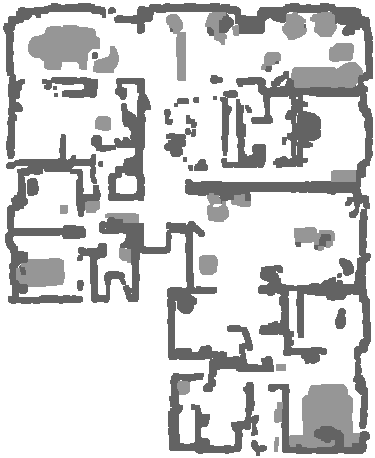} &
\includegraphics[width=0.18\linewidth]{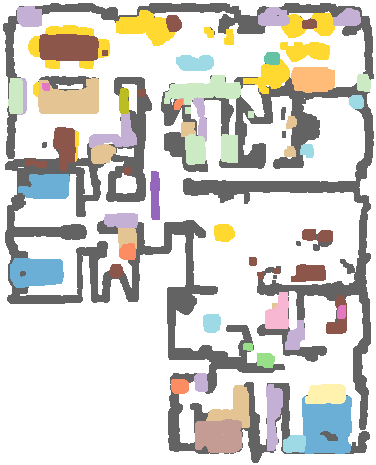} &
\includegraphics[width=0.18\linewidth]{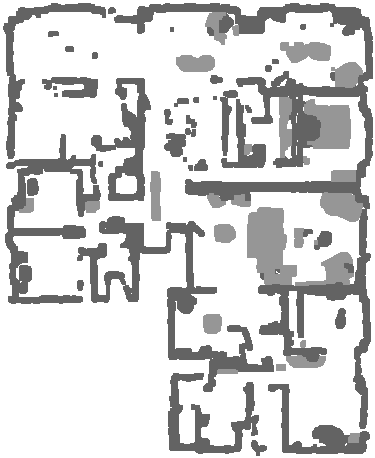} \\
\addlinespace[0.05cm]
\includegraphics[width=0.18\linewidth]{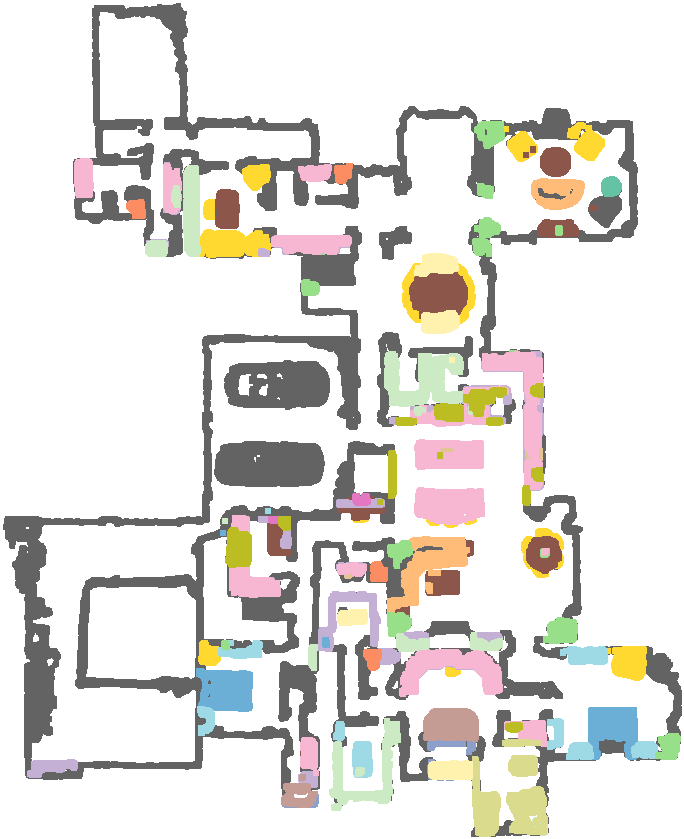} & & 
\includegraphics[width=0.18\linewidth]{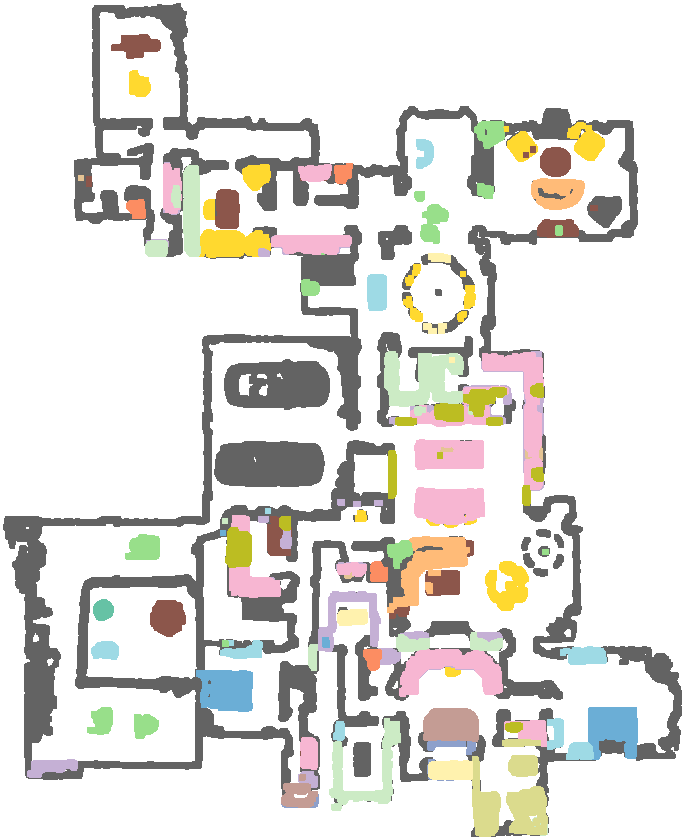} &
\includegraphics[width=0.18\linewidth]{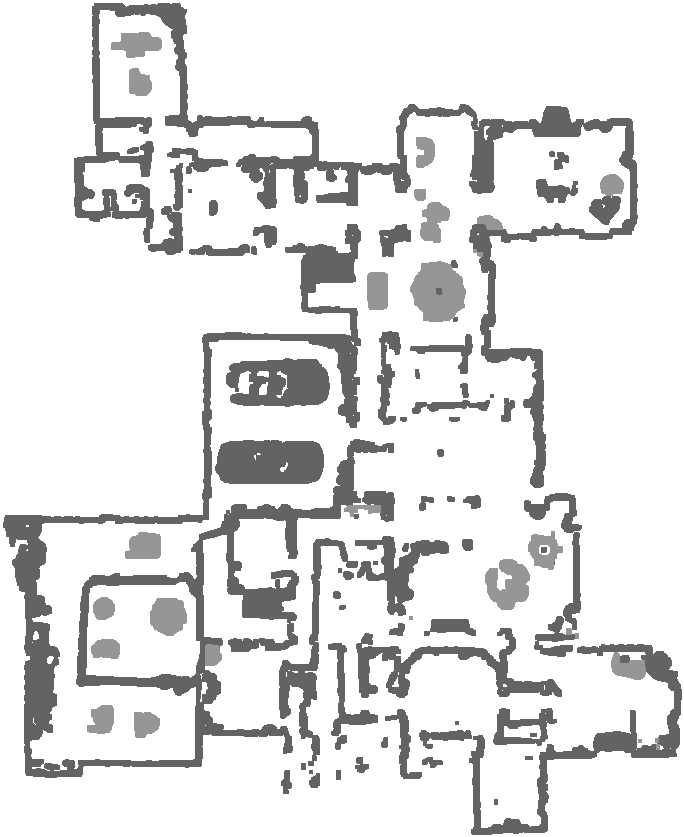} &
\includegraphics[width=0.18\linewidth]{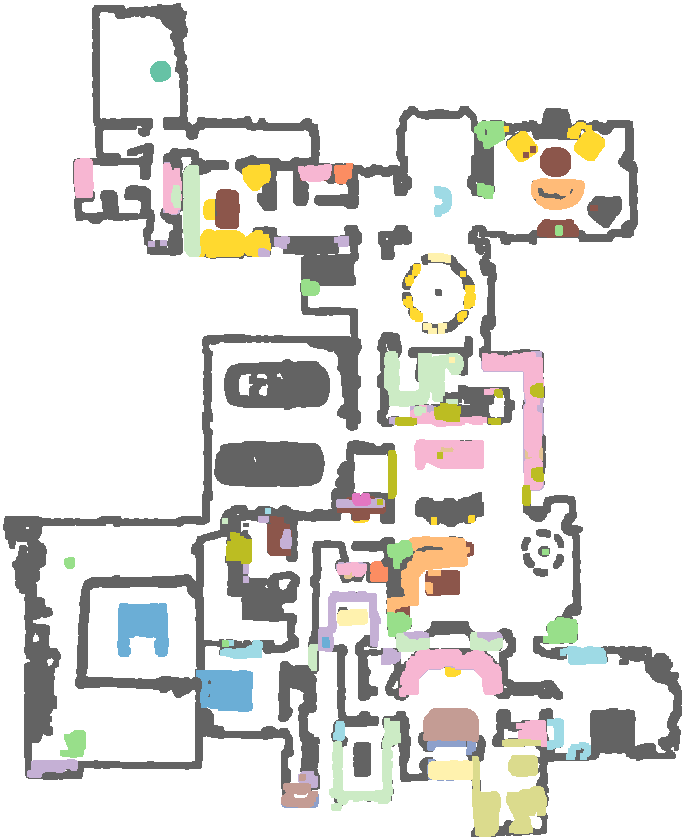} &
\includegraphics[width=0.18\linewidth]{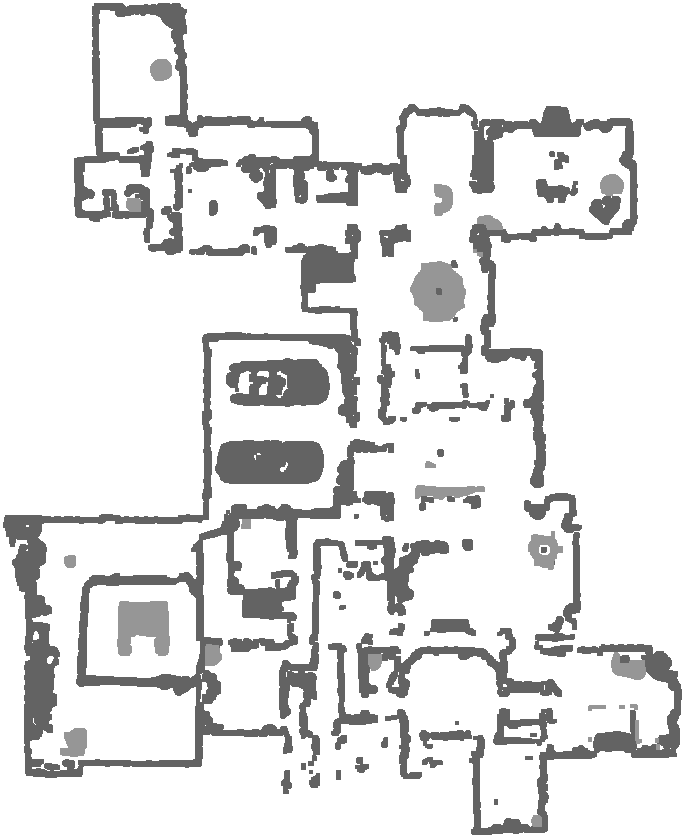} \\
\addlinespace[0.05cm]
\includegraphics[width=0.18\linewidth]{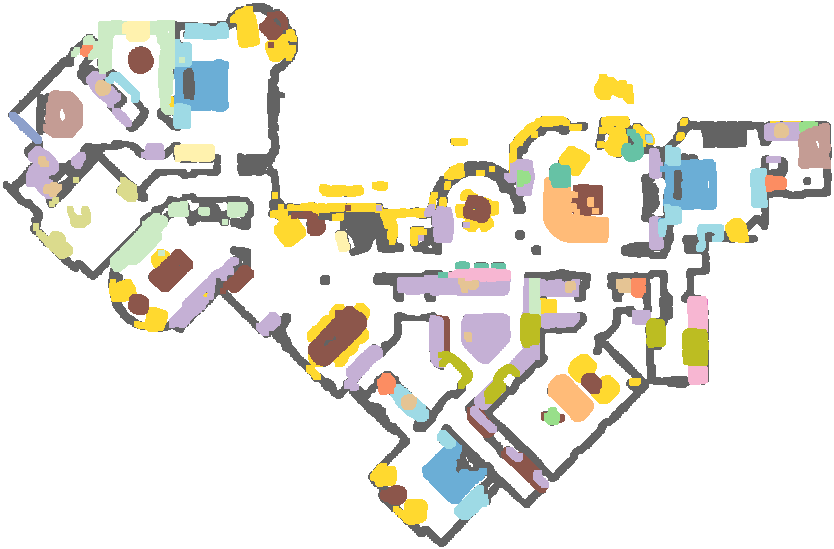} & & 
\includegraphics[width=0.18\linewidth]{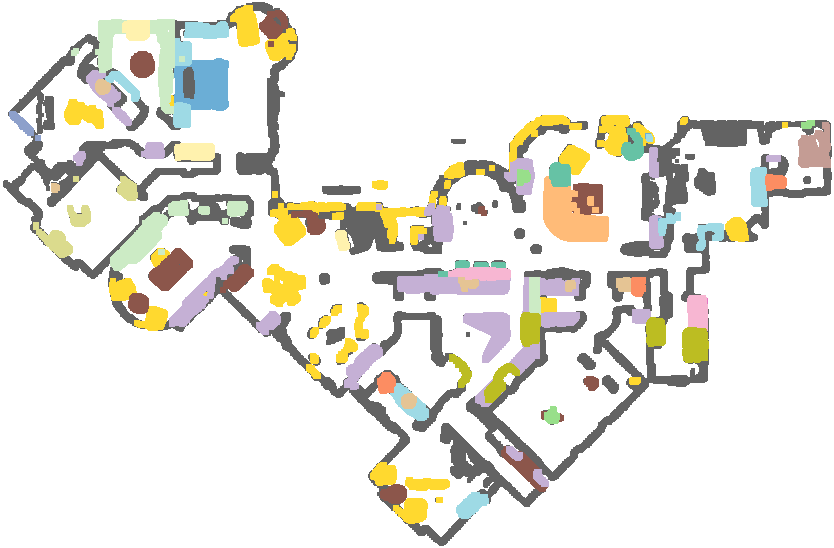} &
\includegraphics[width=0.18\linewidth]{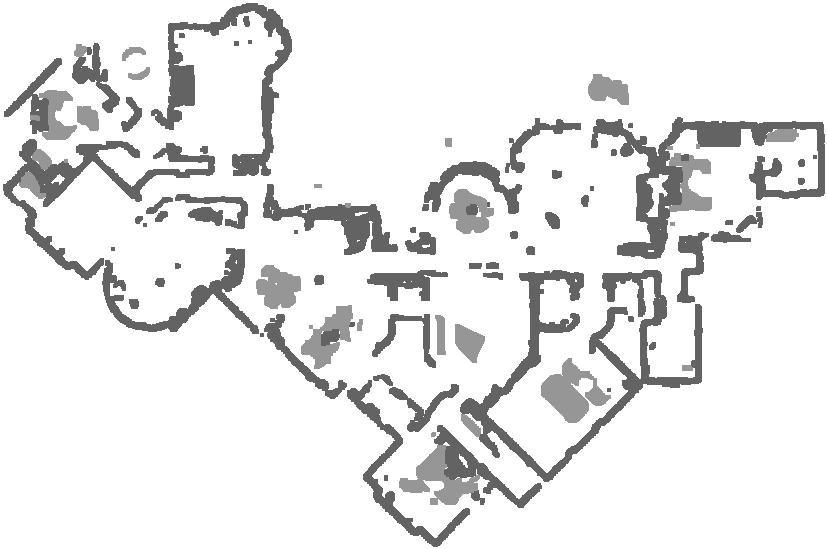} &
\includegraphics[width=0.18\linewidth]{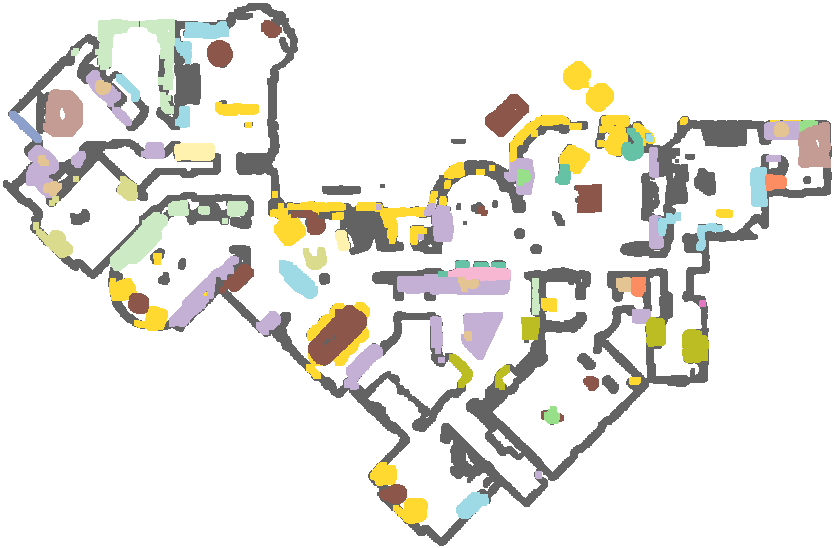} &
\includegraphics[width=0.18\linewidth]{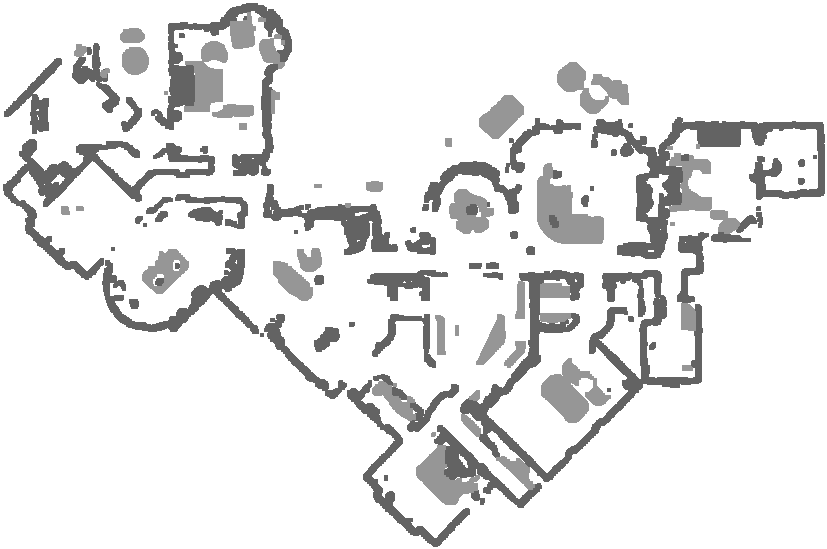} \\
\addlinespace[0.12cm]
\multicolumn{6}{c}{\includegraphics[width=0.9\linewidth]{figures/6_spot_the_difference/legend_f1.pdf}} \\
\end{tabular}
}
\caption{Generation of alternative states of an environment: original and sample manipulated semantic maps with relative difference maps.}
\label{fig:sup_maps}
\end{figure*}

Fig. \ref{fig:fig1_sd} and Fig \ref{fig:sup_maps}, we report some examples of manipulated semantic maps with relative difference maps obtained by applying our semi-automatic procedure.

\tit{Episode Creation}
For the creation of the episodes of our dataset, we use the starting positions of the exploration dataset for \acrshort{mp3d}, and of the \acrlong{pointnav} dataset for Gibson Tiny. After the episodes located on floors with few semantic objects or that are not fully navigable by the agent are discarded, we associate one of the alternative versions of the ground-truth semantic map with each episode. 
We use the same scene partitioning as it is adopted by the existing datasets for embodied exploration and \acrlong{pointnav} on \acrlong{mp3d} and Gibson Tiny \cite{chang2017matterport3d,xia2018gibson}. 
For the validation and test splits of the \acrshort{mp3d} dataset and the validation split of the Gibson dataset, we create new episodes with random sampled starting positions so that the number of episodes on every floor is at least $10$ and fix the number of episodes per floor to a multiple of $10$. We report a detailed list of scans, selected floors, and the number of episodes per scan in Tables  \ref{tab:sup_gibsonval}, \ref{tab:sup_mp3dtrain}, \ref{tab:sup_mp3dval}, and \ref{tab:sup_mp3dtest} at the end of this Chapter.

\section{Proposed Method}
\label{sec:method_sd} 
\begin{landscape}
\begin{figure*}[t!]
    \centering
    \includegraphics[width=\linewidth]{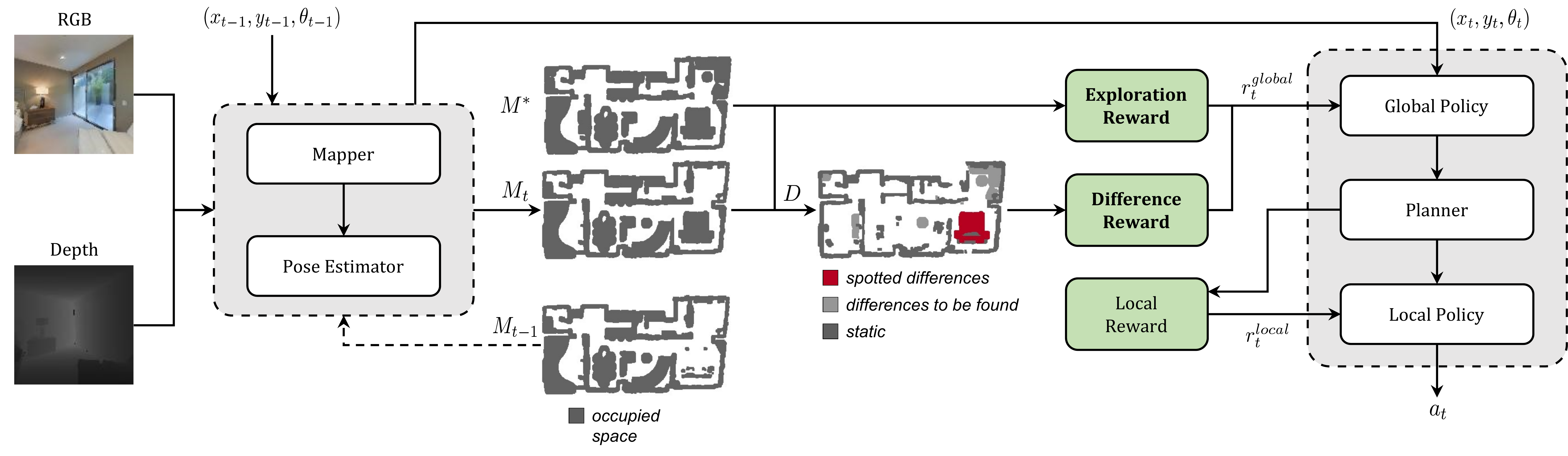}
    \captionsetup{justification=centering}
    \caption{Overview of the proposed approach for navigation in changing environments.}
    \label{fig:fig2_sd}
\end{figure*}
\end{landscape}

Our model for embodied navigation in changing environments comprises three major components: a mapper module, a pose estimator, and a navigation policy (which, in turn, consists of a global policy, a planner, and a local policy). The implementation details of the modules of the architecture are described in Section \ref{sec:method_impact}. An overview of the proposed architecture is shown in Fig. \ref{fig:fig2_sd} and described in the following section.
Although the data we provide is enriched with semantic labels, our agent does not make use of such information directly. This is in line with current architectures for embodied exploration that we choose as competitors.

\subsection{Mapper}
The mapper module takes as inputs an RGB observation $\phi^{rgb}_t$ and the corresponding depth image $\phi^d_t$,  representing the first-person view of the agent at timestep $t$, and outputs the agent-centric occupancy map $m_t$ of a $V\times V$ region in front of the camera. Differently from the work described in Chapter \ref{chap:focus}, each pixel in $m_t$ corresponds to a $25mm \times25mm$ portion of space and consists of two channels containing the probability of that cell being occupied and explored, respectively.
The computed agent-centric occupancy map $m_t$ is then registered in the global occupancy map $M_{t-1}$ coming from the previous timestep to obtain $M_t$ using the pose of the agent $(x_t, y_t, \theta_t)$.

\subsection{Pose Estimator}
The agent can move across the environment using three actions: \textit{go forward 0.25m}, \textit{turn left 10\degree}, \textit{turn right 10\degree}. Since each action may produce a different outcome because of physical interactions with the environment (\eg,~bumping into a wall) or noise in the actuation system, the pose estimator is used to estimate the real displacement made at every timestep.

We estimate the agent displacement $(\Delta x_t,\Delta y_t,\Delta \theta_t)$ at timestep $t$ by using two consecutive RGB and depth observations, as well as the agent-centric occupancy maps $(m_{t-1}, m_t)$ computed by the mapper at $t-1$ and $t$.
The actual agent position $(x_t,y_t,\theta_t)$ is computed iteratively as:
\begin{equation}
    (x_t,y_t,\theta_t) = (x_{t-1},y_{t-1},\theta_{t-1}) + (\Delta x_t,\Delta y_t,\Delta \theta_t).
    \label{eq:pose_sd}
\end{equation}
We assume that the agent starting position is the triple $(x_0,y_0,\theta_0) = (0,0,0)$.

\subsection{Navigation Policy}
The sampling of atomic actions for the exploration relies on a three-component hierarchical policy.
The first component is the global policy, which samples a long-term global goal on the map.
Details on the navigation policy can be found in Section \ref{ssec:navpolicy_impact}.
The global policy outputs a probability distribution over discretized locations of the global map. We sample the global goal from this distribution and then transform it in $(x,y)$ global coordinates.
The second component is a planner module, which employs the A* algorithm to decode a local goal on the map. The local goal is an intermediate point, within \textit{0.25m} from the agent, along the trajectory towards the global goal.
The last element of our navigation module is the local policy, which decodes the series of atomic actions taking the agent towards the local goal.
In particular, the local policy is an \acrshort{rnn} decoding the atomic action $a_t$ to execute at every timestep.
The reward $r^{local}_t$ given to the local policy is proportional to the reduction in the Euclidean distance $d$ between the agent position and the current local goal.

Following the hierarchical structure, a global goal is sampled every $\eta$ timesteps. A new local goal is computed if a new global goal is sampled, if the previous local goal is reached, or if the local goal location is known to be not traversable.

\subsection{Exploiting Past Knowledge for Efficient Navigation}
The global policy is trained using a two-term reward. The first term encourages exhaustive exploration and is proportional either to the increase of coverage \cite{chen2019learning} or to the increase of anticipated map accuracy as in \cite{ramakrishnan2020occupancy}. Intuitively, the agent strives to maximize the portion of the seen area and thus maximizes the knowledge gathered during exploration. 
Moreover, since we consider a setting where a significant amount of knowledge is already available to the agent, we add a reward term to guide the agent towards meaningful points of the map. These correspond to the coordinates where major changes are likely to happen.

Given the occupancy map of the agent $M_t$, the true occupancy map for the same environment $M^*$, and a time budget of $T$ timesteps for exploration, we aim to minimize the following, for $0<t\leq T$:
\begin{equation}
    D = \sum \mathbbmm{1} [M_t \neq M^{*} ]
\end{equation}
In other words, we want to maximize the number of pixels in the online reconstructed map $M_t$ that the agent correctly shifts from free to occupied (and vice-versa) during exploration. This leads to the reward term for difference discovery:
\begin{equation}
    r_{\text{diff}} = \sum \mathbbmm{1} [M_t = M^{*} ] - \sum \mathbbmm{1} [M_{t-1} = M^{*}].
\end{equation}

The proposed reward term is designed to encourage navigation towards areas in the map that are more likely to contain meaningful differences (\eg,~rooms containing more objects that can be displaced or removed from the scene). Additionally, an agent trained with this reward will tend to avoid difficult spots that are likely to produce a mismatch in terms of the predicted occupancy maps. This is because errors in the mapping phase would result in a negative reward.

To train our model, we combine a reward promoting exploration and a more specific reward on found differences to exploit semantic clues in the environment:
\begin{equation}
    \label{eq:rglobal_sd}
    r^{global}_t = \beta_1 r_{\text{exp}} + \beta_2 r_{\text{diff}}
\end{equation}
where $r_\text{exp}$ is the reward term encouraging task-agnostic exploration (such as cov-erage-based or anticipation-based rewards, as described in the next section), and $\beta_1$ and $\beta_2$ are two coefficients weighing the importance of the two elements.

\section{Experimental Setup}

\subsection{Evaluation Protocol}
To evaluate the performance in \textit{Spot the Difference}, we consider three main classes of metrics. First, we consider the percentage of navigable area in the environment seen by the agent during the episode (\%$\mathsf{\acrshort{as}}$). Then, we evaluate the percentage of elements that have been correctly detected as changed in the occupancy map ($\mathsf{\acrshort{acc}}$) and the pixel-wise \acrlong{iou} for the \textit{changed} occupancy map elements ($\mathsf{\acrshort{iou}}$). Besides, we evaluate the task as a two-class problem and compute the $\mathsf{\acrshort{iou}}$ score for objects that were added in place of free space ($\mathsf{\acrshort{iou}}_+$) and for objects that were deleted during the map creation ($\mathsf{\acrshort{iou}_-}$). In addition, to evaluate the performance independently from the exploration capability, we propose to compute the metrics only on the portion of space that the agent actually visited
($\mathsf{\acrshort{macc}}$, $\mathsf{\acrshort{miou}}$, $\mathsf{\acrshort{miou}_+}$, and $\mathsf{\acrshort{miou}_-}$).

\subsection{Implementation Details}
We conduct our experiment using Habitat \cite{savva2019habitat}, a popular platform for \acrshort{eai} in photo-realistic indoor environments \cite{xia2018gibson,chang2017matterport3d}.
The agent observations are $128 \times 128$ RGB-D images from the environment.
The learning algorithm adopted for training is PPO \cite{schulman2017proximal}. The learning rate is $10^{-3}$ for the mapper and $2.5\times10^{-4}$ for the other modules.
Every model is trained for $\approx6.5$M frames using Adam optimizer \cite{kingma2015adam}.
A global goal is sampled every $\eta=25$ timesteps. The local and global policies are updated, respectively, every $\eta$ and $20\times \eta$ timesteps, and the mapper is updated every $4\times \eta$ timesteps.
The size of the local map is $V=101$, while the global map size is set to $W=2001$ for \acrshort{mp3d} and to $W=961$ for Gibson.
The reward coefficients $\{\beta_1, \beta_2\}$ are set to $\{1, 10^{-2}\}$ and $\{1, 10^{-1}\}$ when the exploration reward is based on coverage and anticipation reward, respectively.
% 0.05^2
The length of each episode is fixed to $T=1000$ timesteps.

\subsection{Competitors and Baselines}
We consider the competitors that do not require additional modules to compute the reward signal. The competitors and the variants of the proposed method are evaluated on two different setups: one where the agent position is predicted by the agent (as in Eq. \ref{eq:pose_sd}), and one where it has access to oracle coordinates:

\vspace{5pt}
\noindent\textsc{Difference Reward (\acrshort{dr})}: an exploration policy that maximizes the correctly predicted changes between $M$ and $M^*$. This corresponds to setting $\beta_1=0$ and $\beta_2=1$ in Eq. \ref{eq:rglobal_sd}.

\vspace{5pt}
\noindent\textsc{Coverage Reward (\acrshort{cr})}: an agent that explores the environment with an exploration policy that maximizes the covered area and builds the occupancy map as it goes, as in \cite{ramakrishnan2020occupancy}.

\vspace{5pt}
\noindent\textsc{Anticipation Reward (\acrshort{ar})}: an agent that explores the environment with an exploration policy that maximizes the covered area and the correctly anticipated values in the occupancy map built as it goes,  from \cite{ramakrishnan2020occupancy}.

\vspace{5pt}
\noindent\textsc{Occupancy Anticipation (\acrshort{occant})}: we also compare with the agent presented by Ramakrishnan \etal \cite{ramakrishnan2020occupancy} using the available pretrained models, referenced to as \textit{\acrshort{occant}}. Note that \textit{\acrshort{occant}} was trained on the Gibson dataset for the standard exploration task and without any prior map. Thus, it is not directly comparable with the other methods considered. We include it to gain insights into the performance of an off-the-shelf agent on our task.

\vspace{5pt}
\noindent Our proposed approach consists of an agent trained with the combination of the difference reward with the coverage reward (\textit{\acrshort{cr}+\acrshort{dr}}) or with the anticipation reward (\textit{\acrshort{ar}+\acrshort{dr}}).

\begin{table*}[t]
\centering
\setlength{\tabcolsep}{.4em}
\resizebox{.9\linewidth}{!}{
\begin{tabular}{lc ccccccccc c ccccccccc}
\toprule
 & & \multicolumn{9}{c}{\textbf{Estimated Localization}} \\
\cmidrule{3-11}
 & & \%$\mathsf{\acrshort{as}}$ & $\mathsf{\acrshort{acc}}$ & $\mathsf{\acrshort{iou}}_{+}$ & $\mathsf{\acrshort{iou}}_{-}$ & $\mathsf{\acrshort{iou}}$ & $\mathsf{m\acrshort{acc}}$ & $\mathsf{m\acrshort{iou}}_{+}$ & $\mathsf{m\acrshort{iou}}_{-}$ & $\mathsf{m\acrshort{iou}}$ \\
\midrule
\textbf{\acrshort{occant}}   & & 52.1 & 26.2 & 13.4 & 6.1 & 11.5 & 51.1 & 19.1 & 8.3 & 15.8 \\
\midrule
\textbf{\acrshort{dr}}   & & 49.4 & 29.3 & 15.3 & 8.7 & 13.9 & 59.7 & 23.1 & 11.9 & 20.2 \\
\textbf{\acrshort{ar}} & & 43.8 & 30.6 & 19.7 & 12.9 & 18.8 & 72.5 & 36.8 & 18.4 & 32.7 \\
\textbf{\acrshort{cr}} & & \textbf{53.2} & 33.1 & 18.1 & 9.6 & 16.1 & 65.2 & 26.4 & 12.7 & 22.6 \\
\midrule
\textbf{\acrshort{ar}+\acrshort{dr}}  & & 51.4 & 34.5 & 20.9 & 12.0 & 19.3 & 71.5 & 33.9 & 16.2 & 30.0 \\
\textbf{\acrshort{cr}+\acrshort{dr}}  & & 52.3 & \textbf{37.8} & \textbf{24.2} & \textbf{14.8} & \textbf{22.7} & \textbf{76.2} & \textbf{39.1} & \textbf{19.8} & \textbf{34.8} \\ 

\midrule

 & & \multicolumn{9}{c}{\textbf{Oracle Localization}}\\
\cmidrule{3-11}
& & \%$\mathsf{\acrshort{as}}$ & $\mathsf{\acrshort{acc}}$ & $\mathsf{\acrshort{iou}}_{+}$ & $\mathsf{\acrshort{iou}}_{-}$ & $\mathsf{\acrshort{iou}}$ & $\mathsf{m\acrshort{acc}}$ & $\mathsf{m\acrshort{iou}}_{+}$ & $\mathsf{m\acrshort{iou}}_{-}$ & $\mathsf{m\acrshort{iou}}$\\
\midrule
\textbf{\acrshort{occant}}   & & 49.0 & 35.6 & 26.5 & 16.1 & 24.8 & 77.8 & 49.2 & 23.6 & 43.2 \\
\midrule
\textbf{\acrshort{dr}}   & & 48.6 & 37.4 & 27.2 & 18.4 & 26.5 & 80.1 & 49.8 & 27.4 & 45.8 \\
\textbf{\acrshort{ar}} & & 43.6 & 32.5 & 23.2 & 17.5 & 23.0 & 78.7 & 47.5 & 26.7 & 44.5 \\
\textbf{\acrshort{cr}} & & \textbf{52.8} & 39.2 & 29.6 & 18.8 & 28.0 & 78.5 & 51.0 & 26.6 & 45.7 \\
\midrule
\textbf{\acrshort{ar}+\acrshort{dr}}  & & 51.4 & 37.8 & 27.3 & 18.0 & 26.2 & 79.3 & 48.9 & 25.8 & 44.4 \\
\textbf{\acrshort{cr}+\acrshort{dr}}  & & 51.8 & \textbf{40.3} & \textbf{29.2} & \textbf{19.2} & \textbf{28.1} & \textbf{82.1} & \textbf{50.4} & \textbf{26.9} & \textbf{46.2} \\                        
\bottomrule
\end{tabular}
}
\caption{Experimental results on \acrshort{mp3d} test set. The agent incorporating the proposed reward term for discovered differences outperforms the competitors on the main metrics for the novel Spot the Difference task. Our model \acrshort{cr}+\acrshort{dr} achieves the best results on all the metrics except for \%$\mathsf{\acrshort{as}}$.
} 
\label{tab:tab1_sd}
% \label{tab:mp3d_results}
\end{table*}

\begin{figure}[!ht]
\centering
\scriptsize
\setlength{\tabcolsep}{.2em}
\begin{tabular}{cccc}
& & \textbf{Cumulative $\mathsf{\acrshort{acc}}$} & \textbf{Cumulative $\mathsf{\acrshort{iou}}$} \\
\rotatebox{90}{\parbox[t]{1.5in}{\hspace*{\fill}\textbf{Estimated Localization}\hspace*{\fill}}} & & 
\includegraphics[width=0.42\linewidth]{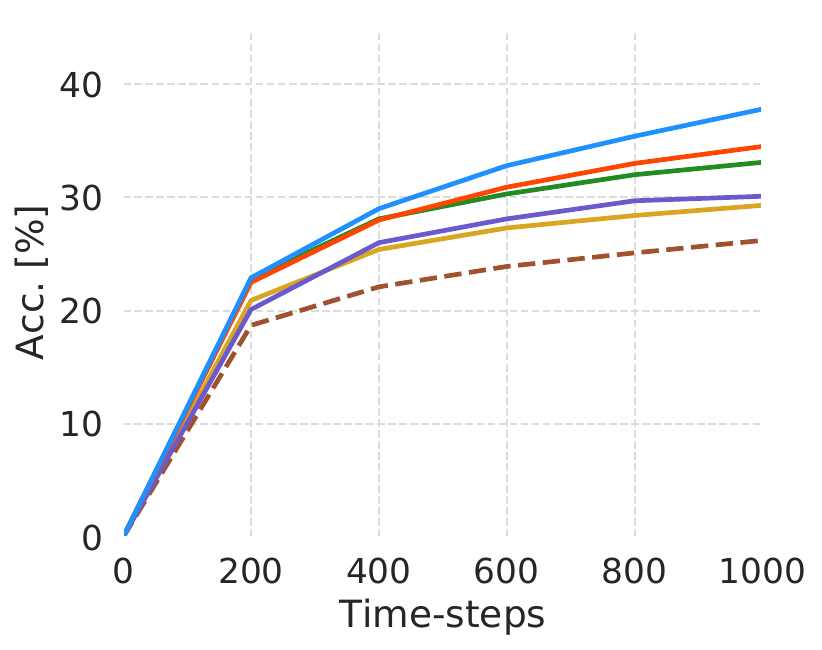} & \includegraphics[width=0.42\linewidth]{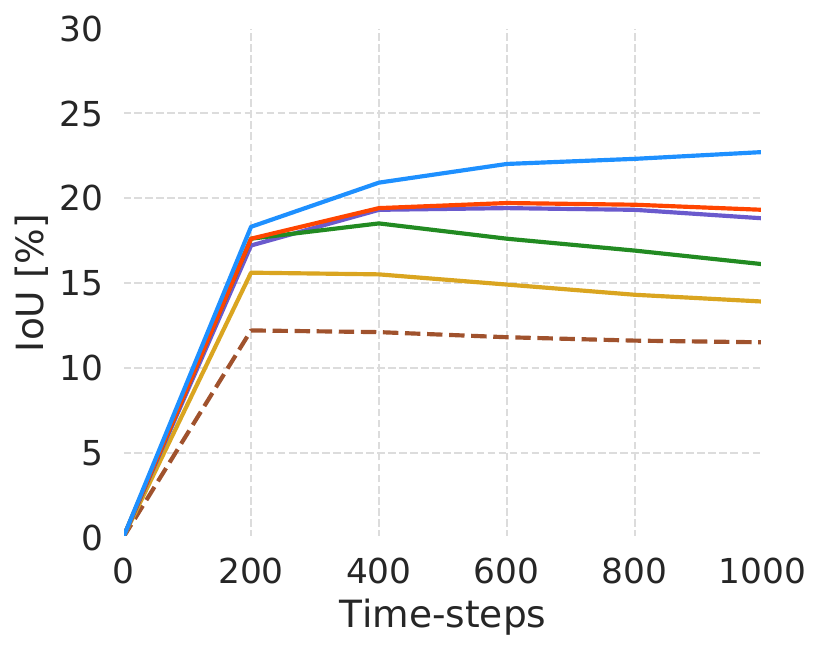} \\ 
\rotatebox{90}{\parbox[t]{1.5in}{\hspace*{\fill}\textbf{Oracle Localization}\hspace*{\fill}}} & &
\includegraphics[width=0.42\linewidth]{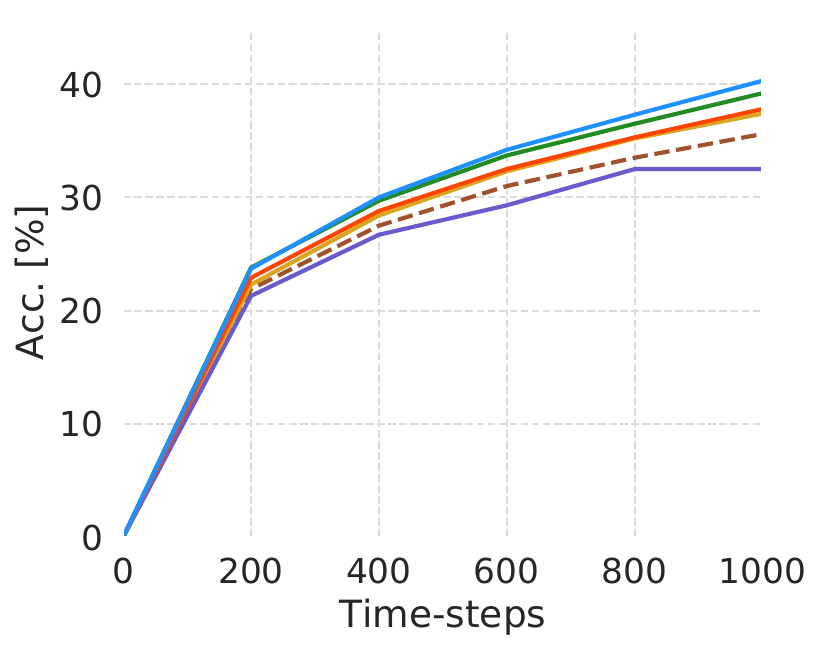} & \includegraphics[width=0.42\linewidth]{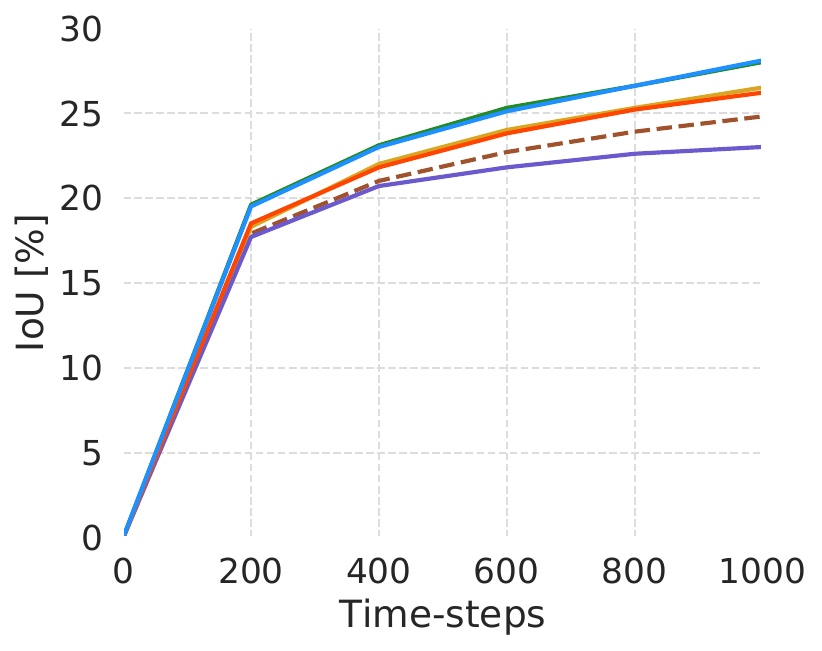} \\
& & \multicolumn{2}{c}{\includegraphics[width=0.9\linewidth]{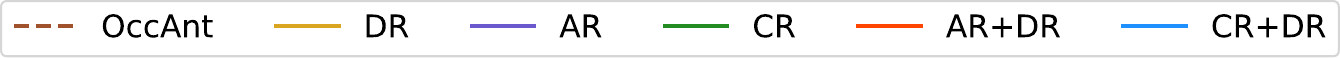}} \\
\end{tabular}
\caption{Value of accuracy and IoU for the different models at varying timesteps on the \acrshort{mp3d} test set.}
\label{fig:fig3_sd}
\end{figure}

\begin{table}[t]
\centering
\setlength{\tabcolsep}{.4em}
\resizebox{.9\linewidth}{!}{
\begin{tabular}{lc ccccccccc}
\toprule
& & \multicolumn{9}{c}{\textbf{Estimated Localization}} \\
\cmidrule{3-11}
& & \%$\mathsf{\acrshort{as}}$ & $\mathsf{\acrshort{acc}}$ & $\mathsf{\acrshort{iou}}_{+}$ & $\mathsf{\acrshort{iou}}_{-}$ & $\mathsf{\acrshort{iou}}$ & $\mathsf{m\acrshort{acc}}$ & $\mathsf{m\acrshort{iou}}_{+}$ & $\mathsf{m\acrshort{iou}}_{-}$ & $\mathsf{m\acrshort{iou}}$\\
\midrule
\textbf{\acrshort{occant}} & & 86.2 & 49.8 & 11.9 & 7.2 & 10.4 & 58.0 & 12.3 & 7.5 & 10.8 \\
\midrule
\textbf{\acrshort{dr}}   & & \textbf{86.2} & 53.2 & 13.2 & 8.5 & 11.7 & 63.7 & 13.9 & 8.8 & 12.3 \\
\textbf{\acrshort{ar}} & & 75.3 & 51.5 & 21.4 & 16.6 & 20.4 & 72.7 & 25.8 & 17.3 & 23.3 \\
\textbf{\acrshort{cr}} & & 85.9 & 57.6 & 16.7 & 11.9 & 15.4 & 71.3 & 18.6 & 12.3 & 16.7 \\
\midrule
\textbf{\acrshort{ar}+\acrshort{dr}} & & 83.4 & 58.7 & 20.0 & 14.9 & 19.0 & 75.8 & 23.0 & 15.6 & 21.1 \\
\textbf{\acrshort{cr}+\acrshort{dr}} & & 82.1 & \textbf{60.1} & \textbf{24.0} & \textbf{19.0} & \textbf{23.1} & \textbf{78.5} & \textbf{27.8} & \textbf{19.9} & \textbf{25.9} \\
\midrule
 & & \multicolumn{9}{c}{\textbf{Oracle Localization}}\\
\cmidrule{3-11}
& & \%$\mathsf{\acrshort{as}}$ & $\mathsf{\acrshort{acc}}$ & $\mathsf{\acrshort{iou}}_{+}$ & $\mathsf{\acrshort{iou}}_{-}$ & $\mathsf{\acrshort{iou}}$ & $\mathsf{m\acrshort{acc}}$ & $\mathsf{m\acrshort{iou}}_{+}$ & $\mathsf{m\acrshort{iou}}_{-}$ & $\mathsf{m\acrshort{iou}}$\\
\midrule
\textbf{\acrshort{occant}} & & 81.6 & 60.1 & \textbf{32.1} & 21.2 & 29.2 & 78.7 & \textbf{39.6} & 22.2 &  \textbf{34.1} \\
\midrule
\textbf{\acrshort{dr}} & & \textbf{86.1} & \textbf{65.2} & 30.1 & \textbf{24.1} & \textbf{29.9} & 81.1 & 36.0 & 25.2 & 33.8 \\
\textbf{\acrshort{ar}} & & 74.1 & 53.8 & 27.9 & 21.9 & 27.2 & 77.0 & 35.4 & 23.5 & 32.7 \\
\textbf{\acrshort{cr}} & & 84.0 & 62.2 & 30.6 & 22.1 & 28.8 & 79.5 & 36.1 & 23.3 & 32.8 \\
\midrule
\textbf{\acrshort{ar}+\acrshort{dr}}  & & 83.2 & 63.2 & 29.6 & 23.8 & 29.1 & \textbf{81.6} & 35.8 & 25.1 & 33.7 \\
\textbf{\acrshort{cr}+\acrshort{dr}}  & & 82.6 & 63.8 & 30.3 & \textbf{24.1} & 29.5 & \textbf{81.6} & 36.1 & \textbf{25.5} & 34.0 \\
\bottomrule
\end{tabular}
}
% \captionsetup{justification=centering}
\caption{Experimental results on Gibson validation set. Our model \acrshort{cr}+\acrshort{dr} achieves the best results in the setting of estimated localization while being competitive when using oracle localization.
} 
\label{tab:tab2_sd}
% \label{tab:gibson_results}
\end{table}

\begin{figure}[t]
\centering
\scriptsize
\setlength{\tabcolsep}{.2em}
\begin{tabular}{cccc}
& & \textbf{Cumulative $\mathsf{\acrshort{acc}}$} & \textbf{Cumulative $\mathsf{\acrshort{iou}}$} \\
\rotatebox{90}{\parbox[t]{1.5in}{\hspace*{\fill}\textbf{Estimated Localization}\hspace*{\fill}}} & & 
\includegraphics[width=0.42\linewidth]{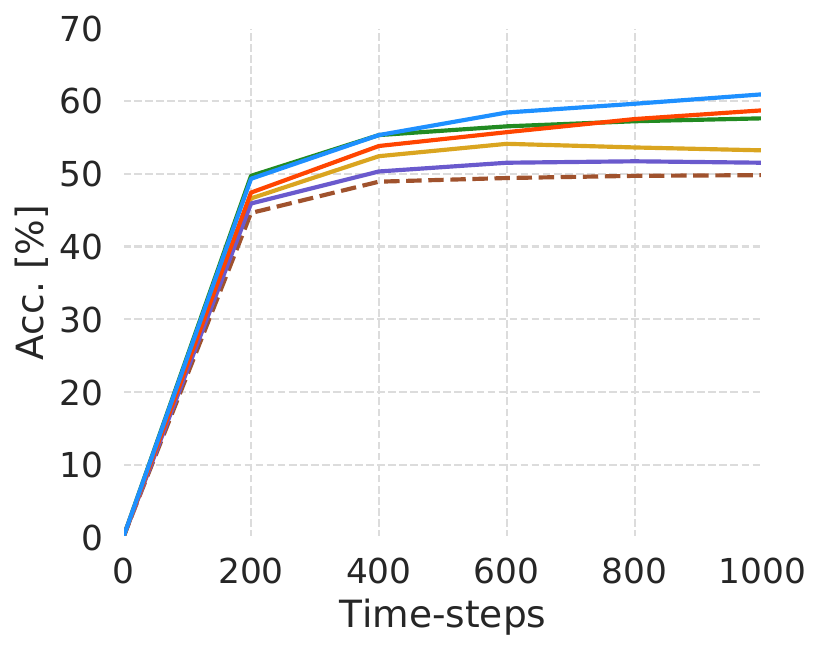} & \includegraphics[width=0.42\linewidth]{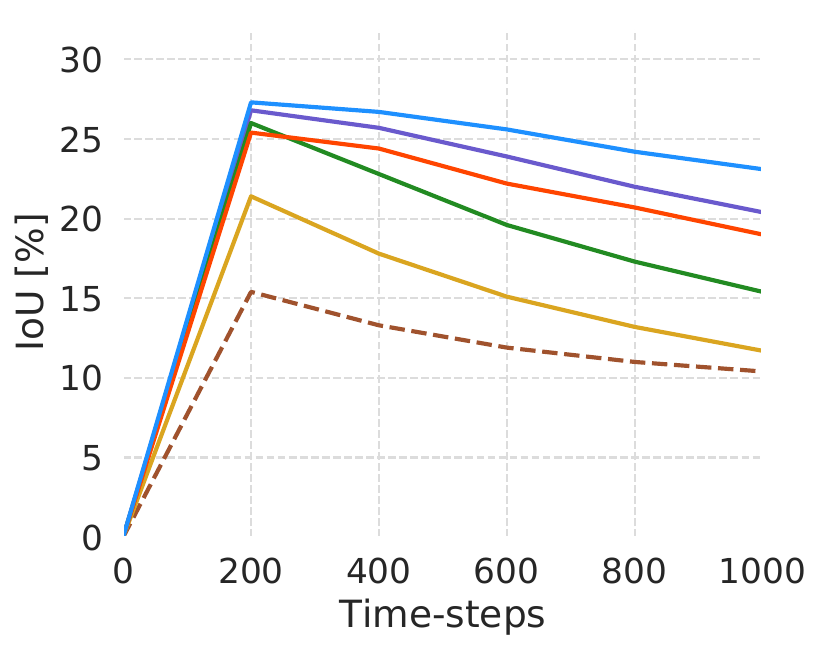} \\ 
\rotatebox{90}{\parbox[t]{1.5in}{\hspace*{\fill}\textbf{Oracle Localization}\hspace*{\fill}}} & &
\includegraphics[width=0.42\linewidth]{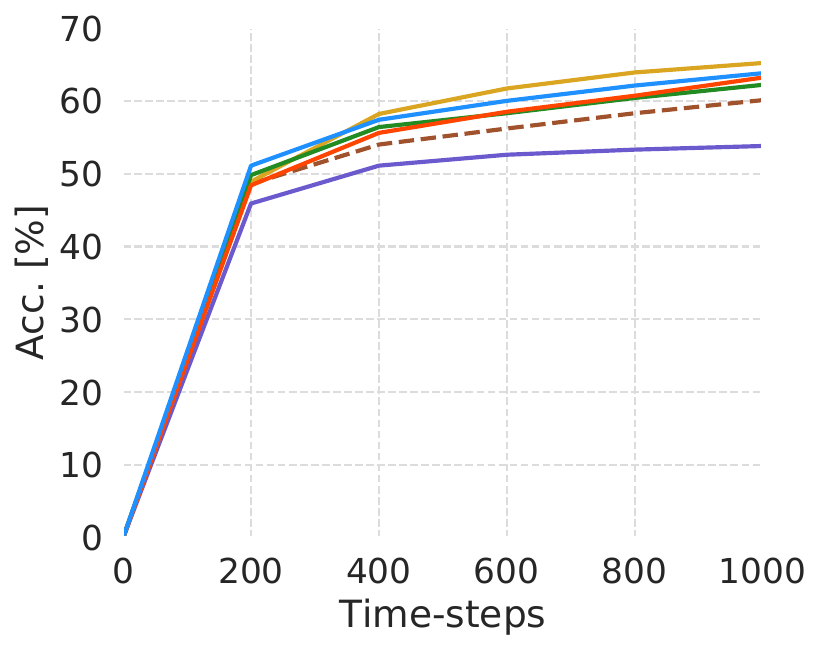} & \includegraphics[width=0.42\linewidth]{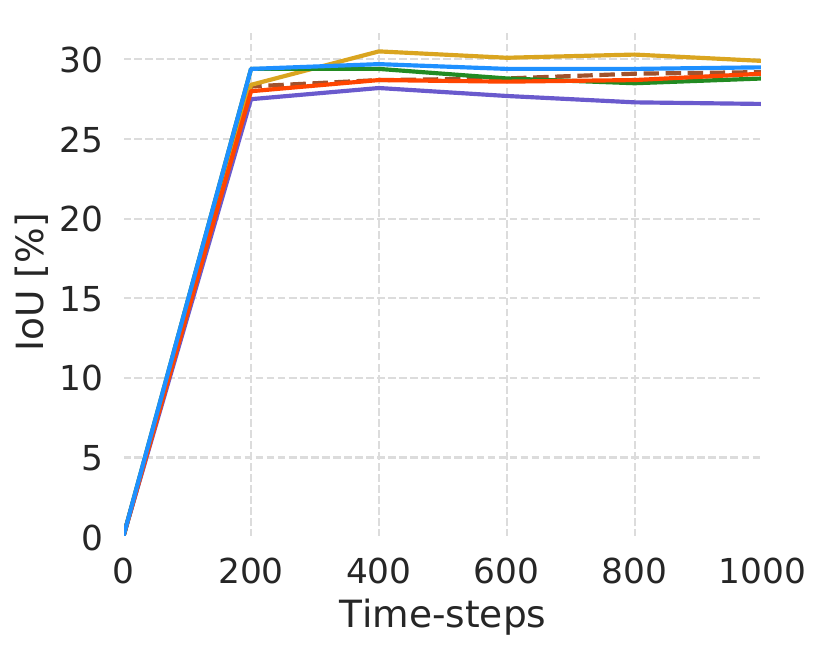} \\
& & \multicolumn{2}{c}{\includegraphics[width=0.9\linewidth]{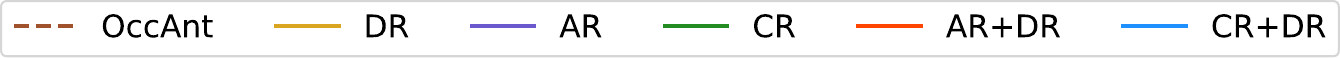}} \\
\end{tabular}
\caption{Value of accuracy and IoU for the different models at varying timesteps on the Gibson validation set.}
\label{fig:curves_gibson_val}
\end{figure}

\section{Experimental Results}
\label{sec:experiments_sd}

\tit{Results on \acrshort{mp3d} dataset}
As a first testbed, we evaluate the different agents on the \acrshort{mp3d} \textit{Spot the Difference} test set.
We report the results for this experiment in Table \ref{tab:tab1_sd}.

We observe that the agent combining a reward based on coverage and our reward based on the differences in the environment (\textit{\acrshort{cr}+\acrshort{dr}}) performs best on all the pixel-based metrics and places second in terms of percentage of seen area. It is worth noting that, even if the results in terms of the area seen are not as high as the ones obtained by the \textit{\acrshort{cr}} agent, the addition of our Difference Reward helps the agent to focus on more relevant parts, and thus, it can discover more substantial differences. Additionally, predictions are more accurate and more precise, as indicated by the $4.7\%$ and $6.6\%$ improvements in terms of $\mathsf{\acrshort{acc}}$ and $\mathsf{\acrshort{iou}}$ with respect to the \textit{\acrshort{cr}} competitor. Instead, a reward based on differences alone is not sufficient to promote good exploration. In fact, although the \textit{\acrshort{dr}} agent outperforms the \textit{\acrshort{cr}} and \textit{\acrshort{ar}} agents on some metrics, our reward alone does not provide as much improvement as when combined with rewards encouraging exploration (as for \textit{\acrshort{cr}+\acrshort{dr}} and \textit{\acrshort{ar}+\acrshort{dr}}).

Even in the oracle localization setup, the \textit{\acrshort{cr}+\acrshort{dr}} agent achieves the best results. Interestingly, the gap with the \textit{\acrshort{cr}} agent decreases to $1.1\%$ and $0.1\%$ in terms of $\mathsf{\acrshort{acc}}$ and $\mathsf{\acrshort{iou}}$, respectively. This is because our \textit{\acrshort{cr}+\acrshort{dr}} agent learns to sample trajectories that can be performed more efficiently and without accumulating a high positioning error. For this reason, the performance boost given by the oracle localization is lower. For both setups, our \textit{\acrshort{cr}+\acrshort{dr}} agent outperforms the state-of-the-art \textit{\acrshort{occant}} agent for exploration on all the metrics.

Finally, in Fig. \ref{fig:fig3_sd}, we plot different values of $\mathsf{\acrshort{acc}}$ and $\mathsf{\acrshort{iou}}$ over different timesteps during the episodes. This way, we can evaluate the whole exploration trend, and not only its final point.
We can observe that the proposed models incorporating the difference reward outperform the competitors. In particular, the \textit{\acrshort{cr}+\acrshort{dr}} agent scores first by a significant margin. The performance gap can be noticed even in the first half of the episode and tends to grow with the number of steps.

\begin{figure}[!t]
\centering
\scriptsize
\setlength{\tabcolsep}{.3em}
\resizebox{\linewidth}{!}{
\begin{tabular}{cc}
\textbf{\acrshort{cr}} & \textbf{\acrshort{cr}+\acrshort{dr}} \\
\addlinespace[0.12cm]
\includegraphics[width=0.4\linewidth]{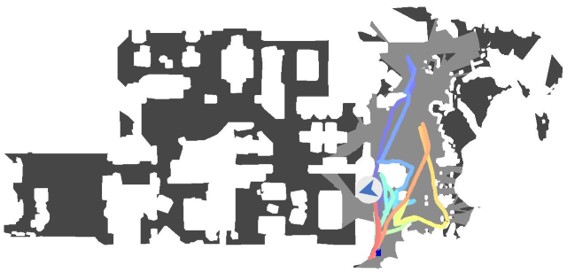} & \includegraphics[width=0.4\linewidth]{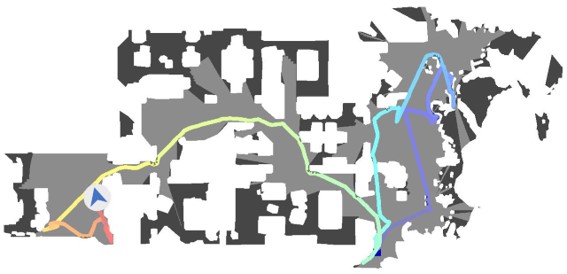} \\ 
\includegraphics[width=0.4\linewidth]{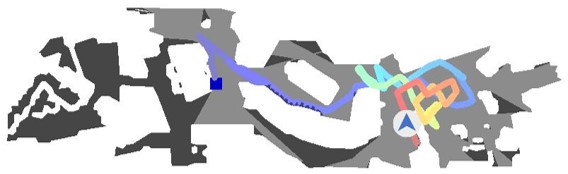} & \includegraphics[width=0.4\linewidth]{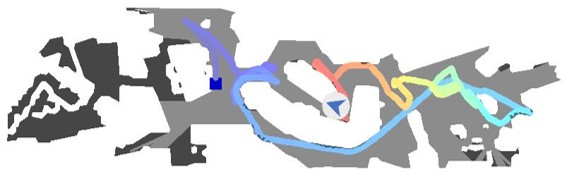} \\ 
\addlinespace[0.2cm]
\multicolumn{2}{c}{\includegraphics[width=0.7\linewidth]{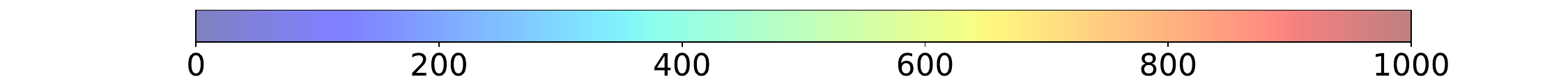}}
\end{tabular}
}
\caption{Exploration trajectories of the \acrshort{cr} and \acrshort{cr}+\acrshort{dr} agents on sample \acrshort{mp3d} test episodes.}
\label{fig:navigation}
\end{figure}

\begin{figure}[!t]
\centering
\scriptsize
\setlength{\tabcolsep}{.8em}
\resizebox{\linewidth}{!}{
\begin{tabular}{cccc}
\textbf{Starting Map} &\textbf{\acrshort{cr}} & \textbf{\acrshort{cr}+\acrshort{dr}} & \textbf{Ground-truth Map} \\
\addlinespace[0.12cm]
\includegraphics[width=0.21\linewidth]{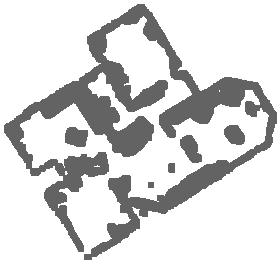} &
\includegraphics[width=0.21\linewidth]{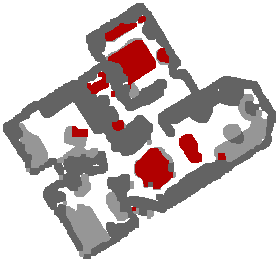} &
\includegraphics[width=0.21\linewidth]{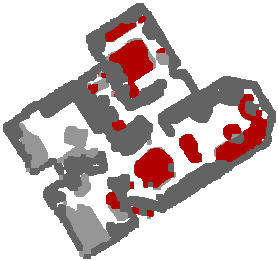} & 
\includegraphics[width=0.21\linewidth]{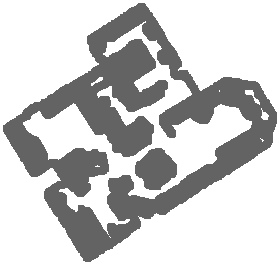} \\
\includegraphics[width=0.21\linewidth]{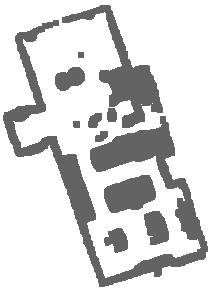} &
\includegraphics[width=0.21\linewidth]{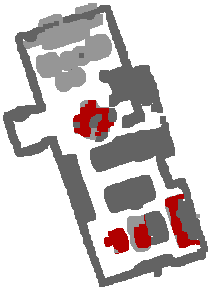} &
\includegraphics[width=0.21\linewidth]{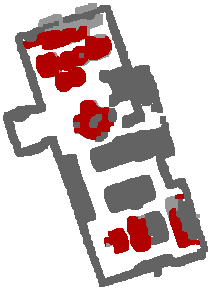} & 
\includegraphics[width=0.21\linewidth]{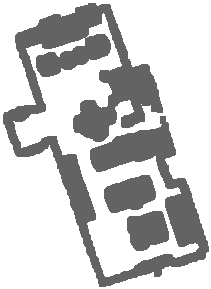} \\
\includegraphics[width=0.21\linewidth]{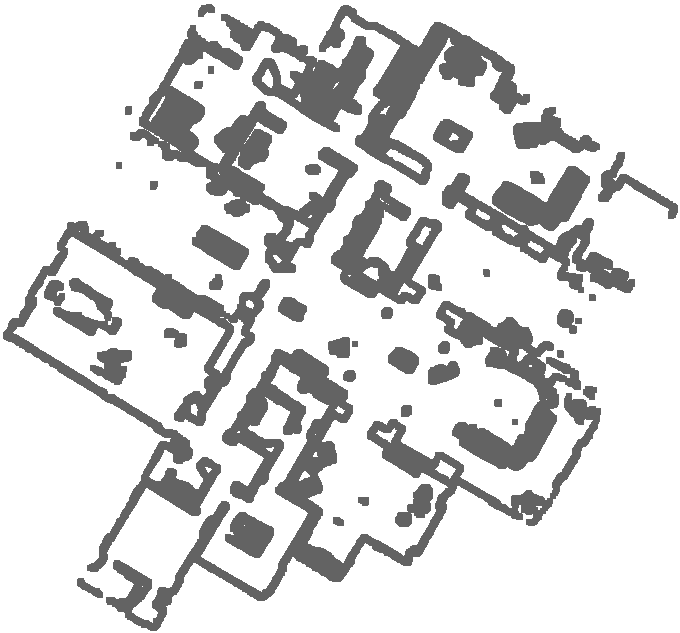} &
\includegraphics[width=0.21\linewidth]{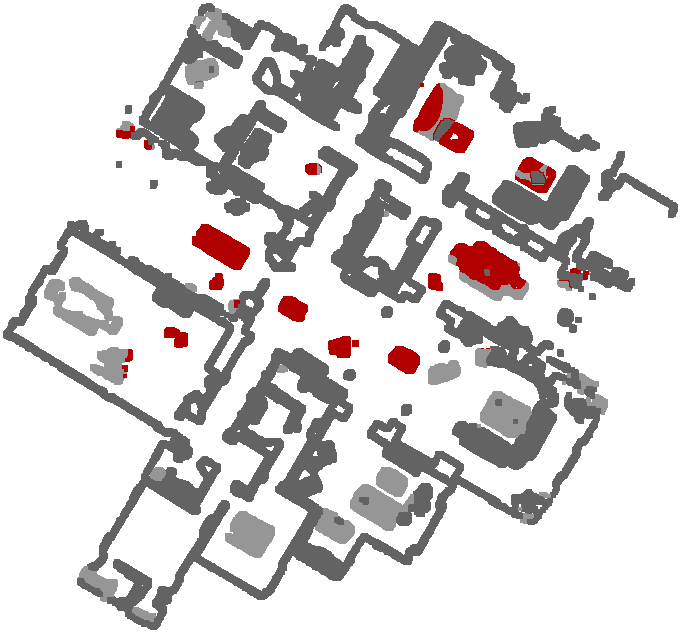} &
\includegraphics[width=0.21\linewidth]{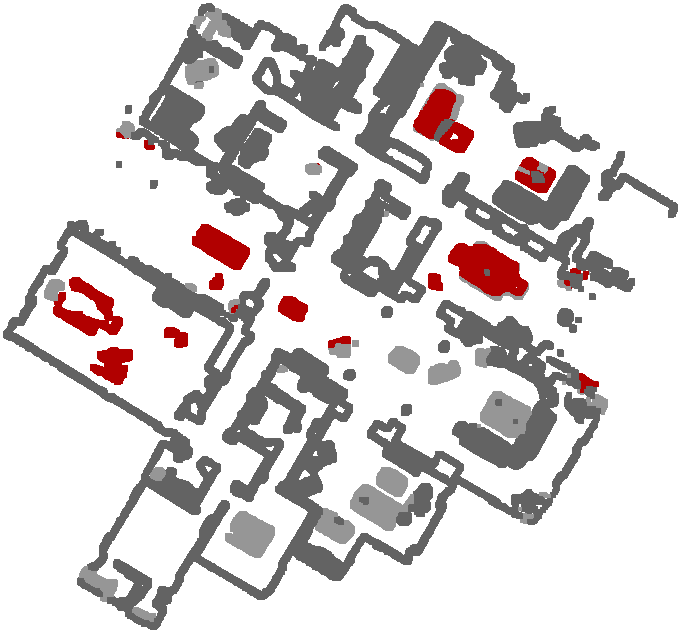} & 
\includegraphics[width=0.21\linewidth]{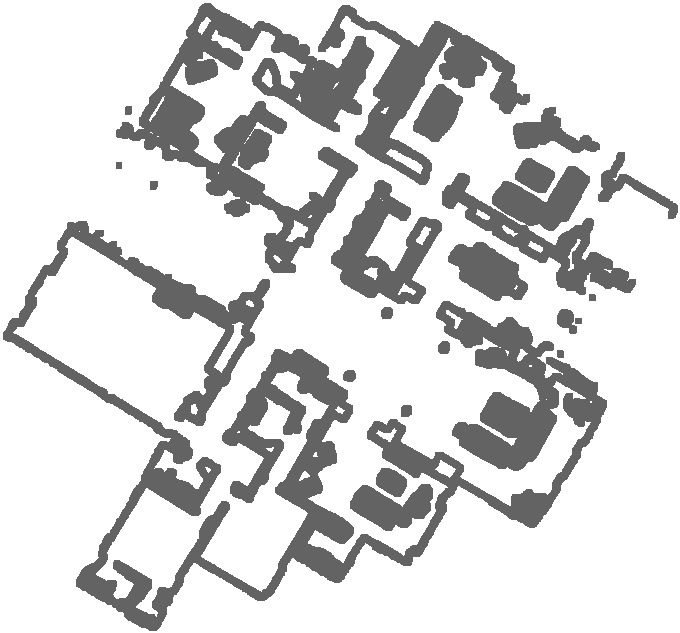} \\
\includegraphics[width=0.21\linewidth]{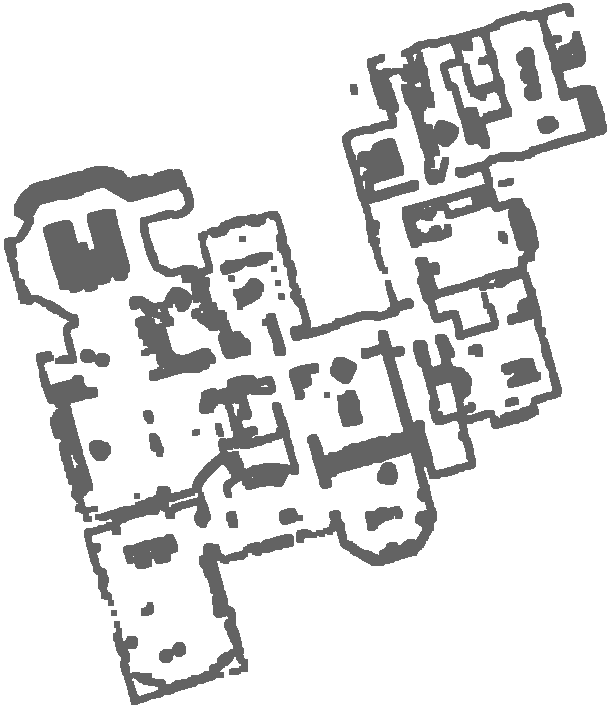} &
\includegraphics[width=0.21\linewidth]{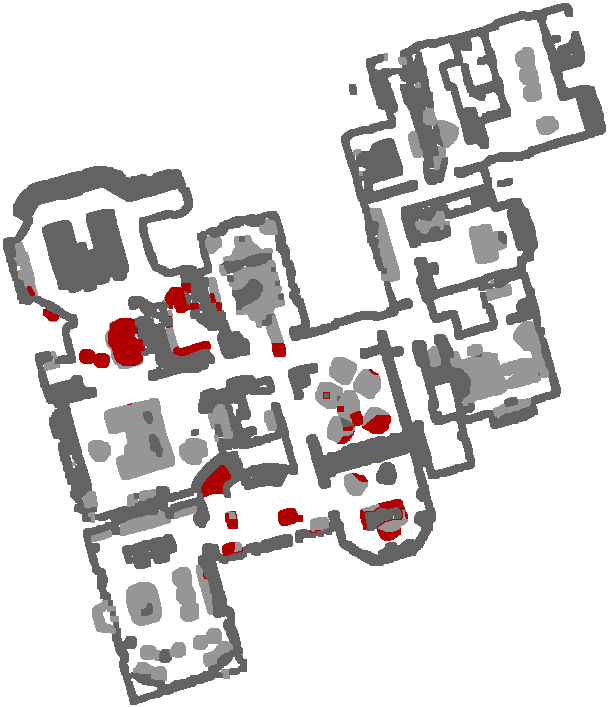} &
\includegraphics[width=0.21\linewidth]{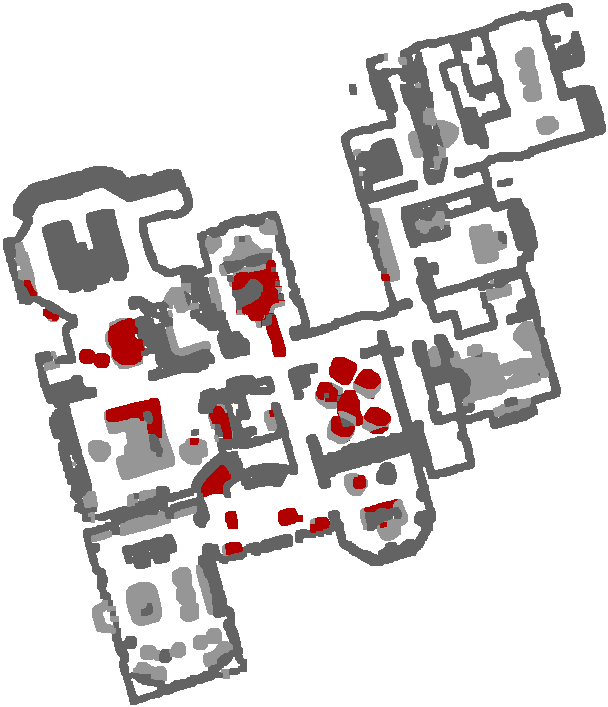} & 
\includegraphics[width=0.21\linewidth]{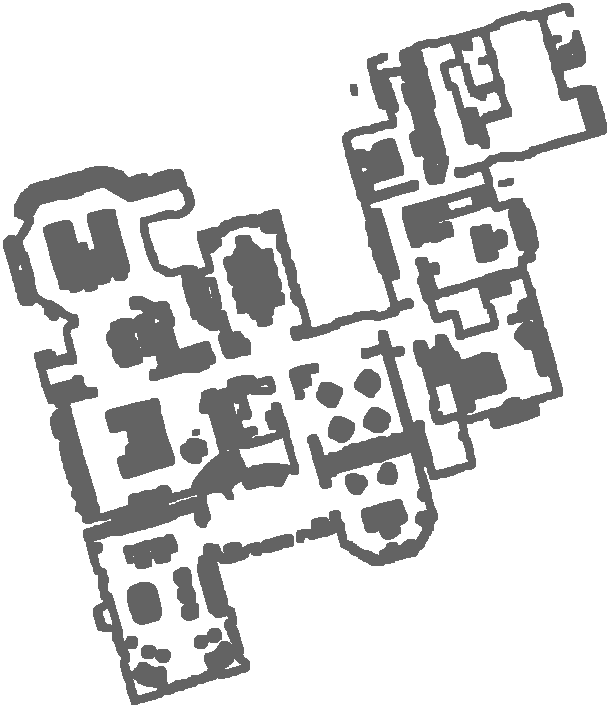} \\
\includegraphics[width=0.21\linewidth]{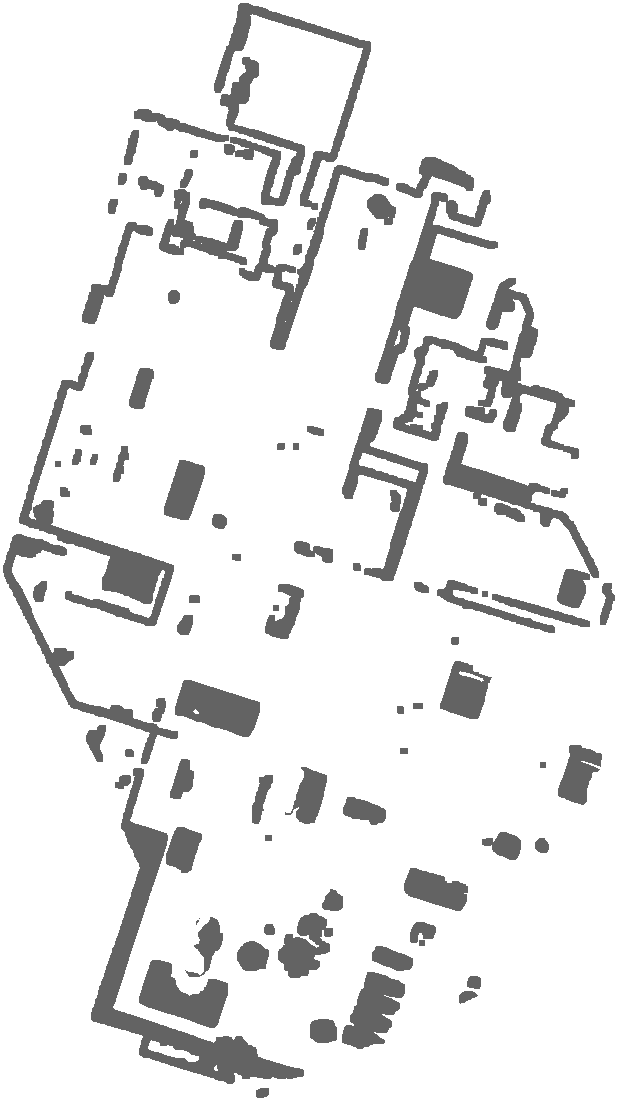} &
\includegraphics[width=0.21\linewidth]{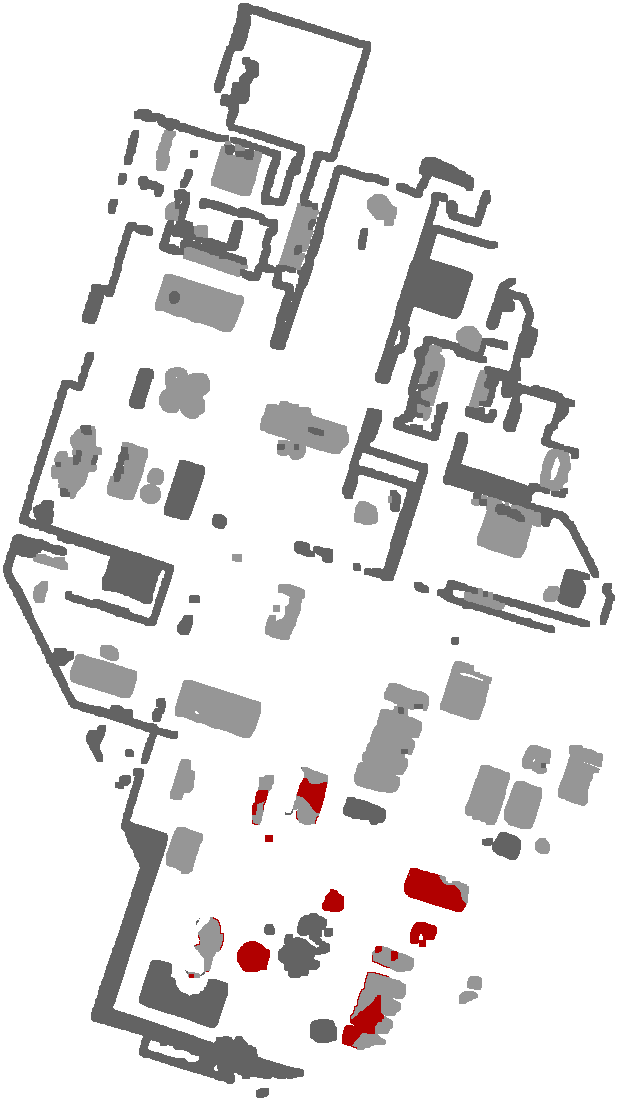} &
\includegraphics[width=0.21\linewidth]{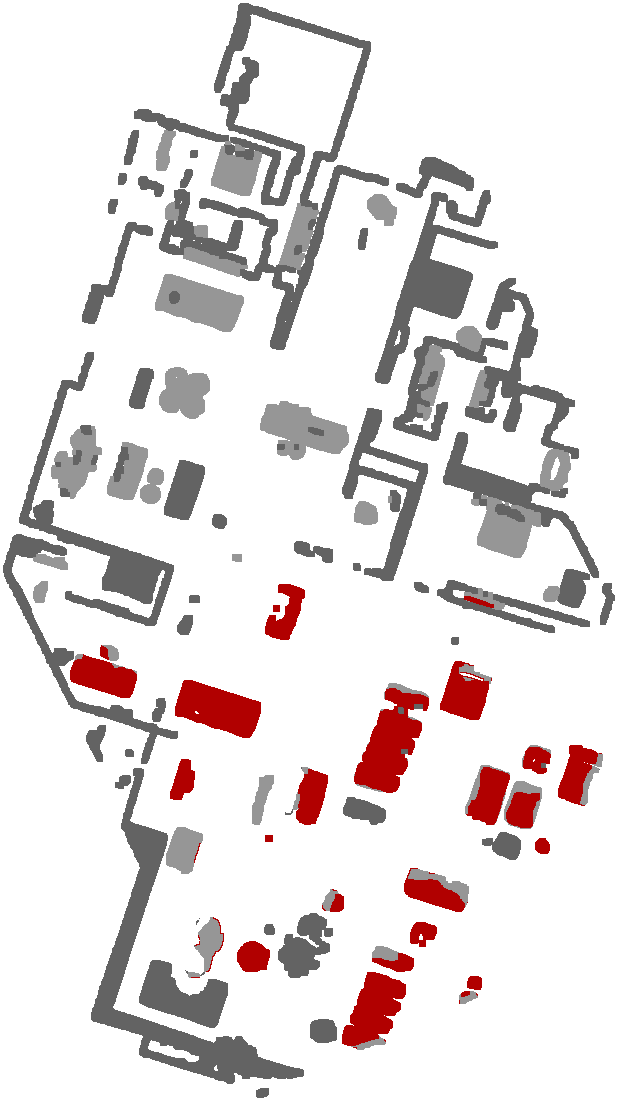} & 
\includegraphics[width=0.21\linewidth]{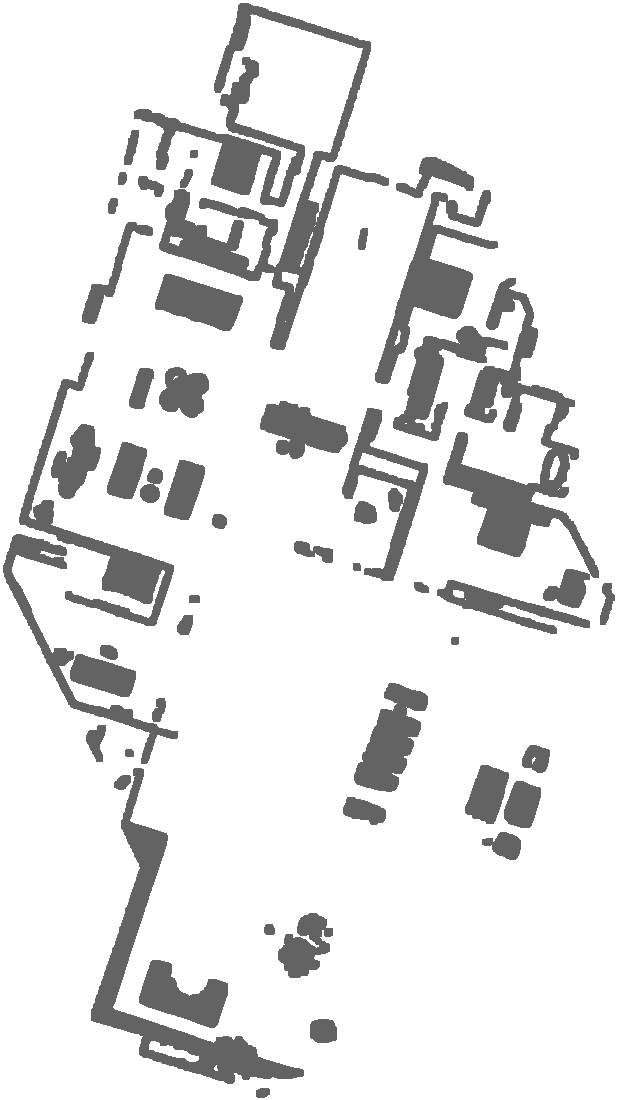} \\
\addlinespace[0.12cm]
\multicolumn{4}{c}{\includegraphics[width=0.95\linewidth]{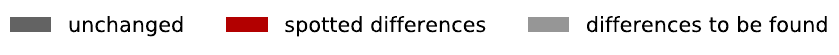}}
\end{tabular}
}
\caption{Qualitative results comparing the performances of the \acrshort{cr} and \acrshort{cr}+\acrshort{dr} agents for different episodes.}
\label{fig:fig4_sd}
\end{figure}

\tit{Results on Gibson dataset}
The environments from the Gibson dataset \cite{xia2018gibson} are generally smaller than those in \acrshort{mp3d}, and thus, they can be explored more easily and exhaustively. We report the results for this experiment in Table \ref{tab:tab2_sd}.
Also in this experiment, the \textit{\acrshort{cr}+\acrshort{dr}} agent performs best on all the metrics but the percentage of the area seen. Although \textit{\acrshort{cr}+\acrshort{dr}} explores $3.8\%$ of the environment less than the \textit{\acrshort{cr}} agent, it still overcomes the competitor by $2.5\%$ and $7.7\%$ in terms of $\mathsf{\acrshort{acc}}$ and $\mathsf{\acrshort{iou}}$. The \textit{\acrshort{ar}+\acrshort{dr}} agent is the second-best in terms of $\mathsf{\acrshort{acc}}$. The \textit{\acrshort{occant}} agent, instead, is competitive in terms of area seen but achieves low $\mathsf{\acrshort{acc}}$ and $\mathsf{\acrshort{iou}}$ metrics. 
In the setting of exploration with oracle localization, the agent using only the proposed difference reward (\textit{\acrshort{dr}}) performs the best on almost all the metrics. We can conclude that, for small environments, and given an optimal localization system, our reward alone is sufficient to surpass the competitors on \textit{Spot the Difference}.

In Fig. \ref{fig:curves_gibson_val}, we report the plots of different values of $\mathsf{\acrshort{acc}}$ and $\mathsf{\acrshort{iou}}$ for the \acrshort{mp3d} and Gibson validation sets, respectively. In these plots, we show how $\mathsf{\acrshort{acc}}$ and $\mathsf{\acrshort{iou}}$ vary at different timesteps during the episodes for the various methods.

\tit{Qualitative Results}
In Fig. \ref{fig:fig4_sd}, we report some qualitative results. 
Starting from the left-most column, we present the starting map given to the agent as the episode begins, the results achieved by the \textit{\acrshort{cr}} agent, those of the proposed \textit{\acrshort{cr}+\acrshort{dr}} agent, and the ground-truth map. The differences that the agents have correctly identified during the episode are highlighted in red. As it can be seen, the \textit{\acrshort{cr}+\acrshort{dr}} agent can identify more differences than the \textit{\acrshort{cr}} counterpart, even in small environments (top row). As the size of the environments grows (bottom three rows), the performance gap increases and the \textit{\acrshort{cr}+\acrshort{dr}} agent outperforms its competitor.
Moreover, we report sample exploration trajectories for both the \acrshort{cr} and the \acrshort{cr}+\acrshort{dr} agents with estimated localization in Fig. \ref{fig:navigation}. These confirm the competitive exploration capabilities of our proposed agent.

\section{Future Directions}
Our method exploits outdated information about the current environment to improve the exploration capabilities of the agent. However, the focus of this work is on pure occupation, ignoring semantic information. For future work, we expect to include semantic reasoning in the agent's pipeline, assuming that additional information could boost the performance.
With the proposed dataset, we enable a series of possible embodied tasks that imply dynamic environments and incorporate available past knowledge.

\begin{table*}[t]
\begin{minipage}[t]{.5\linewidth}
\centering
\setlength{\tabcolsep}{.35em}
\footnotesize
\vspace{1cm}
\resizebox{.85\linewidth}{!}{
\begin{tabular}{llc}
\toprule
\multicolumn{3}{c}{\textbf{\acrshort{mp3d}}} \\
\midrule
\textbf{Index} & \textbf{Category} & \textbf{Action} \\
\midrule
0 & Void & No Operation \\
1 & Wall & No Operation \\
2 & Floor & No Operation \\
3 & Chair & Displacement \\
4 & Door & No Operation \\
5 & Table & Displacement \\
6 & Picture & No Operation \\
7 & Cabinet & Removal \\
8 & Cushion & Overlap Removal \\
9 & Window & No Operation \\
10 & Sofa & Displacement \\
11 & Bed & Displacement \\
12 & Curtain & No Operation \\
13 & Chest of Drawers & Displacement \\
14 & Plant & Displacement \\
15 & Sink & Empty \\
16 & Stairs & No Operation \\
17 & Ceiling & No Operation \\
18 & Toilet & Removal \\
19 & Stool & Displacement \\
20 & Towel & Overlap Removal \\
21 & Mirror & No Operation \\
22 & TV Monitor & Removal \\
23 & Shower & Removal \\
24 & Column & No Operation \\
25 & Bathtub & Removal \\
26 & Counter & Removal \\
27 & Fireplace & No Operation \\
28 & Lighting & No Operation \\
29 & Beam & No Operation \\
30 & Railing & No Operation \\
31 & Shelving & Removal \\
32 & Blinds & No Operation \\
33 & Gym Equipment & Displacement \\
34 & Seating & Removal \\
35 & Board Panel & No Operation \\
36 & Furniture & Displacement \\
37 & Appliances & Removal \\
38 & Clothes & Overlap Removal \\
39 & Objects & Overlap Removal \\
40 & Misc & Overlap Removal \\
41 & Unlabeled & No Operation \\
\bottomrule
\end{tabular}
}
\caption{\acrshort{mp3d} semantic categories per channel index.}
\label{tab:semantic_classes_mp3d}
\end{minipage}
\begin{minipage}[t]{.5\linewidth}
\centering
\setlength{\tabcolsep}{.35em}
\footnotesize
\vspace{1.5cm}
\resizebox{.8\linewidth}{!}{
\begin{tabular}{llc}
\toprule
\multicolumn{3}{c}{\textbf{Gibson}} \\
\midrule
\textbf{Index} & \textbf{Category} & \textbf{Action} \\
\midrule
0 & Chair & Displacement \\
1 & Couch & Displacement \\
2 & Potted Plant & Removal \\
3 & Bed & Displacement \\
4 & Toilet & Removal \\
5 & TV & Removal \\
6 & Dining Table & Displacement \\
7 & Oven & Removal \\
8 & Sink & Removal \\
9 & Refrigerator & Removal \\
10 & Book & Overlap Removal \\
11 & Clock & Removal \\
12 & Vase & Removal \\
13 & Cup & Overlap Removal \\
14 & Bottle & Overlap Removal \\
15 & Bench & Removal \\
16 & Appliances & Removal \\
17 & Objects & Overlap Removal \\
18 & Misc & Overlap Removal \\
19 & Void & No Operation \\
\bottomrule
\end{tabular}
}
\caption{Gibson semantic categories per channel index.}
\label{tab:semantic_classes_gibson}
% \centering
% \footnotesize
% \setlength{\tabcolsep}{.35em}
% \vspace{0.3cm}
% \resizebox{.85\linewidth}{!}{
% \begin{tabular}{lll}
% \toprule
% \multicolumn{3}{c}{\textbf{Hyperparameters}} \\
% \midrule
% \textbf{Module} & \textbf{Name} &  \textbf{Value} \\
% \midrule

% Global Policy   & map size (G)      & $240$ \\
%                 & lr                & $2.5\times10^{-4}$ \\
%                 & max\_grad\_norm   & $0.5$ \\
% \addlinespace[0.2cm]
% Local Policy    & forward step      & $0.25m$ \\
%                 & turn angle        & $10\degree$ \\
%                 & hidden size       & $256$ \\
%                 & lr                & $2.5\times10^{-4}$ \\
% \addlinespace[0.2cm]
% Mapper          & batch size        & $32$ \\
%                 & map scale         & $0.05cm^2$ \\
%                 & map size          & $101$ \\
%                 & lr                & $10^{-3}$ \\
%                 & momentum          & $0.9$ \\
%                 & max grad norm     & $0.5$ \\
% \addlinespace[0.2cm]
% PPO             & clip param        & $0.2$ \\
%                 & entropy coef      & $10^{-3}$ \\
%                 & eps               & $10^{-5}$ \\
%                 & gamma             & $0.99$ \\
%                 & n.~mini-batches   & $4$ \\
%                 & n.~epochs         & $4$ \\
%                 & tau               & $0.95$ \\
%                 & using gae         & True \\
%                 & value loss coef   & $0.5$ \\
% \bottomrule
% \end{tabular}
% }
% \caption{List of hyperparameters.}
% \label{tab:hyperparams}
\footnotesize
\setlength{\tabcolsep}{.35em}
\vspace{0.3cm}
\resizebox{0.7\linewidth}{!}{
\begin{tabular}{lccc}
\toprule
\multicolumn{4}{c}{\textbf{Gibson Validation}} \\
\midrule
\textbf{Scan} & & \textbf{Floors} & \textbf{\# Episodes} \\
\midrule
Wiconisco & & 1,2 & 90 \\
Corozal & & 0,2,4 & 90 \\
Collierville & & 0,1,2 & 80 \\
Markleeville & & 0,1 & 90 \\
Darden & & 0,1,2 & 100 \\
\midrule
\textbf{Total:} 5 & & 13 & 450 \\
\bottomrule
\end{tabular}
}
\caption{Gibson validation scans and floors, with relative number of episodes for \textit{Spot the Difference}.}
\label{tab:sup_gibsonval}
\end{minipage}
\end{table*}

\begin{table*}[t]
\begin{minipage}[t]{.5\linewidth}
\centering
\setlength{\tabcolsep}{.35em}
\footnotesize
\vspace{0.01cm}
\resizebox{0.85\linewidth}{!}{
\begin{tabular}{lccc}
\toprule
\multicolumn{4}{c}{\textbf{\acrshort{mp3d} Train}} \\
\midrule
\textbf{Scan} & & \textbf{Floors} & \textbf{\# Episodes} \\
\midrule
HxpKQynjfin & & 0 & 81967 \\
gTV8FGcVJC9 & & 0,1,2,3,4,6,10,11 & 77186 \\
29hnd4uzFmX & & 0,1,2,3 & 81967 \\
5LpN3gDmAk7 & & 0,1,2,3 & 81885 \\
SN83YJsR3w2 & & 0,1,2,3,7,8,10,12 & 81438 \\
VzqfbhrpDEA & & 0,1,3,6 & 81641 \\
D7N2EKCX4Sj & & 0,1,2,3,5,6 & 81830 \\
5q7pvUzZiYa & & 0,1,2,3,4 & 81967 \\
ac26ZMwG7aT & & 0,1 & 81967 \\
r47D5H71a5s & & 0,1 & 81965 \\
Pm6F8kyY3z2 & & 0 & 81967 \\
8WUmhLawc2A & & 0,1,2 & 81967 \\
82sE5b5pLXE & & 0,1,2 & 80682 \\
mJXqzFtmKg4 & & 0,1,2 & 81967 \\
i5noydFURQK & & 0,1 & 81120 \\
V2XKFyX4ASd & & 0,1,2,3,4,5,7 & 81129 \\
759xd9YjKW5 & & 0,1,2,3 & 81913 \\
r1Q1Z4BcV1o & & 0 & 81812 \\
S9hNv5qa7GM & & 0,1 & 81967 \\
1LXtFkjw3qL & & 0,1,2,3,4,5,6 & 81967 \\
PuKPg4mmafe & & 0 & 81940 \\
EDJbREhghzL & & 0,1,3 & 64755 \\
ur6pFq6Qu1A & & 0,1 & 81967 \\
B6ByNegPMKs & & 0 & 81951 \\
b8cTxDM8gDG & & 0,1,2,8,11 & 73307 \\
17DRP5sb8fy & & 0 & 81967 \\
YmJkqBEsHnH & & 0 & 80780 \\
ULsKaCPVFJR & & 0,1,2 & 81967 \\
XcA2TqTSSAj & & 0,2,3,5,6,8,9,11,12 & 60679 \\
sKLMLpTHeUy & & 0,1,2,4 & 79736 \\
ZMojNkEp431 & & 0,1,2 & 81967 \\
e9zR4mvMWw7 & & 0,1,2 & 80193 \\
JeFG25nYj2p & & 0,1 & 81967 \\
uNb9QFRL6hY & & 1,4,5,6 & 59613 \\
p5wJjkQkbXX & & 0,1,2,3 & 81967 \\
Vvot9Ly1tCj & & 0,3 & 78115 \\
E9uDoFAP3SH & & 0,1,5,6 & 81914 \\
qoiz87JEwZ2 & & 0,1,2,3 & 81967 \\
VFuaQ6m2Qom & & 0,1,2,4,5,6 & 81758 \\
VLzqgDo317F & & 0,1,2 & 81396 \\
kEZ7cmS4wCh & & 0,1,2,3,7 & 69135 \\
7y3sRwLe3Va & & 0,1,2,5 & 81386 \\
VVfe2KiqLaN & & 0,1,2 & 81967 \\
2n8kARJN3HM & & 0,1,2,4 & 81076 \\
PX4nDJXEHrG & & 0,1,2,3,4,5 & 79151 \\
Uxmj2M2itWa & & 0,1,3,4 & 49942 \\
pRbA3pwrgk9 & & 0,2,3,7,9,11 & 53295 \\
cV4RVeZvu5T & & 0,1,2,3 & 81038 \\
sT4fr6TAbpF & & 0 & 81625 \\
GdvgFV5R1Z5 & & 0 & 81967 \\
JF19kD82Mey & & 0,1,2 & 81927 \\
JmbYfDe2QKZ & & 0,1 & 81489 \\
s8pcmisQ38h & & 0,1,2 & 80428 \\
1pXnuDYAj8r & & 0,1,2,5 & 81901 \\
jh4fc5c5qoQ & & 0,1,2 & 81967 \\
vyrNrziPKCB & & 0,1,3,4,7 & 81388 \\
aayBHfsNo7d & & 0,1,2 & 81693 \\
rPc6DW4iMge & & 0,1,3,4 & 80296 \\
\midrule
\textbf{Total:} 58 & & 207 & 4581881 \\
\bottomrule
\end{tabular}
}
\captionsetup{justification=centering}
\caption{\acrshort{mp3d} train scans and floors.} 
\label{tab:sup_mp3dtrain}
\end{minipage}
\begin{minipage}[t]{.5\linewidth}
\centering
\setlength{\tabcolsep}{.35em}
\footnotesize
\vspace{2.1cm}
\resizebox{0.75\linewidth}{!}{
\begin{tabular}{lccc}
\toprule
\multicolumn{4}{c}{\textbf{\acrshort{mp3d} Validation}} \\
\midrule
\textbf{Scan} & & \textbf{Floors} & \textbf{\# Episodes} \\
\midrule
2azQ1b91cZZ & & 0,1 & 40 \\
8194nk5LbLH & & 0 & 40 \\
EU6Fwq7SyZv & & 0 & 30 \\
QUCTc6BB5sX & & 1 & 20 \\
TbHJrupSAjP & & 0,1,2 & 30 \\
Z6MFQCViBuw & & 0 & 40 \\
oLBMNvg9in8 & & 0,1,2,3 & 50 \\
x8F5xyUWy9e & & 0,1 & 30 \\
zsNo4HB9uLZ & & 0 & 40 \\
\midrule
\textbf{Total:} 9 & & 16 & 320 \\
\bottomrule
\end{tabular}
}
\centering
\caption{\acrshort{mp3d} validation scans and floors, with relative number of episodes for \textit{Spot the Difference}.}
\label{tab:sup_mp3dval}
\setlength{\tabcolsep}{.35em}
\footnotesize
\vspace{0.3cm}
\resizebox{0.75\linewidth}{!}{
\begin{tabular}{lccc}
\toprule
\multicolumn{4}{c}{\textbf{\acrshort{mp3d} Test}} \\
\midrule
\textbf{Scan} & & \textbf{Floors} & \textbf{\# Episodes} \\
\midrule
2t7WUuJeko7 & & 0 & 50 \\
5ZKStnWn8Zo & & 0,1 & 50 \\
RPmz2sHmrrY & & 0 & 50 \\
UwV83HsGsw3 & & 0,1,2,3 & 50 \\
WYY7iVyf5p8 & & 0,2 & 30 \\
YFuZgdQ5vWj & & 1 & 10 \\
YVUC4YcDtcY & & 0 & 50 \\
fzynW3qQPVF & & 0,1 & 50 \\
jtcxE69GiFV & & 0,1 & 40 \\
pa4otMbVnkk & & 0,1 & 50 \\
q9vSo1VnCiC & & 0 & 50 \\
rqfALeAoiTq & & 0,2 & 20 \\
wc2JMjhGNzB & & 0,1 & 50 \\
yqstnuAEVhm & & 0,1,2 & 60 \\
\midrule
\textbf{Total:} 14 & & 26 & 610 \\
\bottomrule
\end{tabular}
}
\centering
\caption{\acrshort{mp3d} test scans and floors, with relative number of episodes for \textit{Spot the Difference}.}
\label{tab:sup_mp3dtest}
\end{minipage}
\end{table*}

\chapter[Navigation in the Real World]{Out of the Box: \\ \Large Embodied Navigation in the Real World}
\label{chap:out}
\blfootnote{This Chapter is related to the publication ``R. Bigazzi \etal, Out of the Box: Embodied Navigation in the Real World, CAIP 2021'' \cite{bigazzi2021out}. See the list of Publications on page \pageref{publications} for more details.}
\RemoveLabels

\lettrine[lines=1]{\textcolor{SchoolColor}{F}}{ollowing} the previous chapters where the contributions are mainly related to the work in simulation devising novel methods or tasks for \gls{eai}. In this chapter and in the following one, our efforts are targeting real-world aspects of \gls{eai}. 
\gls{eai} has recently attracted a lot of attention from the vision and learning communities. Nevertheless, the majority of current research focuses on the creation of rich and complex architectures that are trained in simulation, using large amounts of data. Thanks to powerful simulating platforms \cite{deitke2020robothor,savva2019habitat,xia2018gibson}, the Embodied AI community could achieve nearly perfect results on tasks such as \acrfull{pointnav} \cite{wijmans2019dd}. However, current research is still in the first mile of the race for the creation of intelligent and autonomous agents. Naturally, the next milestones involve bridging the gap between simulated platforms (in which the training takes place) and the real world \cite{kadian2020sim2real}. 
In this work, we aim to design a robot that can navigate in unknown, real-world environments \cite{cascianelli2016robust}.

We ask ourselves a simple research question: \textit{can the agent transfer the skills acquired in simulation to a more realistic setting?} To answer this question, we devise a new experimental setup in which models learned in simulation are deployed on a LoCoBot \cite{locobot}. Previous work on \acrshort{sim2real} adaptability from the Habitat simulator \cite{savva2019habitat} has focused on a setting where the real-world environment was matched with a corresponding simulated environment to test the \acrshort{sim2real} metric gap. To that end, Kadian \etal \cite{kadian2020sim2real} carry on a 3D acquisition of the environment specifically built for robotic experiments. Here, we assume a setting in which the final user cannot count on the technology/expertise required to make a 3D scan. This experimental setup is more challenging for the agent, as it cannot count on semantic priors on the environment acquired in simulation. Moreover, while \cite{kadian2020sim2real} employs large boxes as obstacles, our testing scene contains real-life objects with complicated shapes such as desks, office chairs, and doors.

\AddLabels

\begin{landscape}
\begin{figure}[t!]
    \centering
    \includegraphics[width=0.9\linewidth]{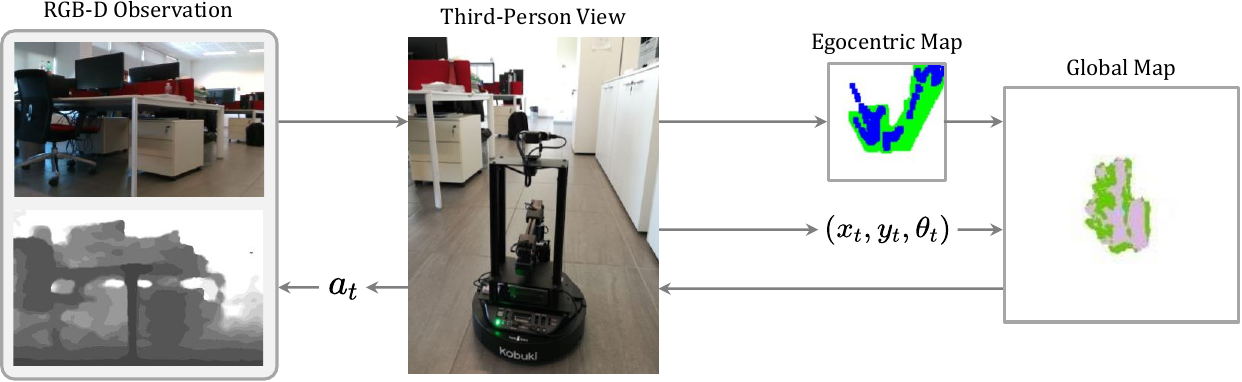}
    \caption{We deploy a state-of-art navigation architecture on a LoCoBot and test it in a realistic, office-like environment. Our model exploits egocentric and global occupancy maps to plan a route towards the goal.}
    \label{fig:fig1_out}
\end{figure}
\end{landscape}

Our agent builds on a recent model proposed by Ramakrishnan \etal \cite{ramakrishnan2020occupancy} for the PointNav task. As a first step, we research the optimal setup to train the agent in simulation. We find out that default options (tailored for simulated tasks) are not optimal for real-world deployment: for instance, the simulated agents often exploit imperfections in the simulator physics to slide along the walls. As a consequence, deployed agents tend to get stuck when trying to replicate the same sliding dynamic. By enforcing a more strict interaction with the environment, it is possible to avoid such shortcomings in the locomotor policy. Secondly, we employ the software library PyRobot \cite{murali2019pyrobot} to create a transparent interface with the LoCoBot: thanks to PyRobot, the code used in simulation can be seamlessly deployed on the real-world agent by changing only a few lines of code. Finally, we test the navigation capabilities of the trained model on a real scene: we create a set of navigation episodes in which goals are defined using relative coordinates. While previous tests were mainly made in robot-friendly scenarios (often consisting of a single room), we test our model, which we call LoCoNav, in a realistic type of environment: obstacles are represented by common office furniture such as desks, chairs, cupboards; the floor is uneven as there are gaps between floor tiles that make actuation noisy and very position-dependent, and there are multiple rooms that must be accessed through doorways (Fig.\ref{fig:fig1_out}). Thanks to our experiments, we show that models trained in simulation can adapt to real unseen environments. By making our code and models publicly available, we hope to motivate further research on \acrshort{sim2real} adaptability and deployment in the real world of agents trained on the Habitat simulator. Our code and models are available publicly\footnote{https://github.com/aimagelab/LoCoNav}.

\section{Real-World Navigation with Habitat}
In this section, we describe our out-of-the-box navigation robot. First, we describe the baseline architecture and its training procedure that takes place in the Habitat simulator \cite{savva2019habitat}. Then we present our LoCoNav agent, which builds upon the baseline and implements various modules to enable real-world navigation.

\subsection{Baseline Architecture}
We draw inspiration from the occupancy anticipation agent \cite{ramakrishnan2020occupancy} to design our baseline architecture. The model consists of three main parts: a mapper, a pose estimator, and a hierarchical policy, which we describe in the following. More details on their implementation is contained in Section \ref{sec:method_impact}.

\tit{Mapper}
The mapper is responsible for producing an occupancy map of the environment, which is then employed by the agent as an auxiliary representation during navigation. Following the work presented in Chapters \ref{chap:focus}, \ref{chap:eds}, and \ref{chap:sd}, we use two different types of maps at each timestep $t$: the local map $m_t$ that depicts the portion of the environment immediately in front of the agent, and the global map $M_t$ that captures the area of the environment already visited by the agent. The global map of the environment $M_t$ is blank at $t=0$ and it is built in an incremental way. Each map has two channels, identifying the free/occupied and the explored/unexplored space, respectively; each pixel contains the state of a $5$cm $\times$ $5$cm area.
The mapper module takes as input the RGB and depth observations $(\phi^{rgb}_t, \phi^d_t)$ at time $t$ and produces the agent-centric local map $m_t \in [0,1]^{2 \times V \times V }$. 

As described in previous chapters, at each timestep $m_t$ is registered to the global map $M_t \in [0,1]^{2 \times W \times W}$, with $W > V$, using the agent's position and heading in the environment $(x_t, y_t, \theta_t)$. %$(x^M_t, y^M_t, \theta_t)$.

\tit{Pose Estimator}
While the agent navigates towards the goal, the interactions with the environment are subject to noise and errors, so that, for instance, the action \textit{go forward $25$cm} might not result in a real displacement of $25$cm. That could happen for a variety of reasons: bumping into an obstacle, slipping on the terrain, or simple actuation noise. The pose estimator is responsible for avoiding such positioning mistakes and keeps track of the agent pose in the environment at each timestep $t$. This module computes the relative displacement $(\Delta x_t, \Delta y_t, \Delta\theta_t)$ caused by the action selected by the agent at time $t$. It takes as input the RGB-D observations $(\phi{rgb}r_t, \phi^d_t)$ and $(\phi^{rgb}_{t-1}, \phi^d_{t-1})$ retrieved at time $t$ and $t-1$, and the egocentric maps $m_t$ and $m_{t-1}$, and outputs the displacement $(\Delta x_t,\Delta y_t,\Delta \theta_t)$
The estimated pose of the agent at time $t$ is given by:
\begin{equation}
    (x_t, y_t, \theta_t) = (x_{t-1}, y_{t-1}, \theta_{t-1}) + (\Delta x_t,\Delta y_t,\Delta \theta_t).
\end{equation}

\tit{Navigation Policy}
The baseline navigation policy is defined by a hierarchical design. The highest-level component of our policy is the global policy. The global policy selects a long-term goal on the global map, which we call global goal $g_t$. A new global goal is sampled every $\eta$ timesteps during training and is set to the navigation goal during deployment and test.
The middle-level component of our hierarchical policy is the planner. After the global goal is set, an A* planner decodes the next local goal within $0.25$m from the agent and on the trajectory toward the global goal. A new local goal is sampled if at least one of the following three conditions verifies: a new global goal is sampled by the global policy, the previous local goal is reached, or the local goal is known to be in an occupied area.
Finally, the local policy performs the low-level navigation and decodes the series of actions to perform. The actions available to the agents are \textit{go forwards $25$cm} and \textit{turn $15$\textdegree}. The local policy samples an atomic action $a_t$ at each timestep $t$.

\subsection{Training in Simulation}
The baseline architecture described in the previous lines is trained in simulation using Habitat \cite{savva2019habitat} and 3D scans from the Gibson dataset of spaces \cite{xia2018gibson}. The mapper is trained with a binary cross-entropy loss using the ground-truth occupancy maps of the environment, obtained as described in \cite{ramakrishnan2020occupancy}. The navigation policy is trained using reinforcement learning. We choose PPO \cite{schulman2017proximal} as training algorithm. The global policy receives a reward signal equal to the increase in terms of anticipated map accuracy \cite{ramakrishnan2020occupancy}:
\begin{equation}
    r_t^{\text{\textit{global}}} = \text{Acc}(M_t, M^*) - \text{Acc}(M_{t-1}, M^*) ,
\end{equation}
where $M_{t}$ and $M_{t-1}$ represent the global occupancy maps computed at time $t$ and $t-1$ respectively, and $^* \in [0,1]^{2 \times W \times W}$ is the ground-truth global map.
The map accuracy is defined as: 
\begin{equation}
    \text{Acc}(M,M^*) = \sum_{i = 1}^{W^2}{\sum_{j=1}^{2}{\mathbbmm{1}[M_{ij} = M^*_{ij}]}},
\end{equation}
where $\mathbbmm{1}[\cdot]$ is an indicator function that returns one if the condition $[\cdot]$ is true and zero otherwise.
The local policy is trained using a reward that encourages the decrease in the euclidean distance between the agent and the local goal while penalizing collisions with obstacles:
\begin{equation}
    r_t^{local} = d_t - d_{t-1} - \alpha \cdot \text{\textit{bump}}_t,
    \label{eq:r_t^local_out}
\end{equation}
where $d_t$ and $d_{t-1}$ are the euclidean distances to the local goal at times $t$ and $t-1$, $\text{\textit{bump}}_t \in \{0,1\}$ identifies a collision at time $t$ and $\alpha$ regulates the contributions of the collision penalty.
The training procedure described in this section exploits the experience collected throughout $6.5$ million exploration frames.

\subsection{LoCoNav: Adapting for Real World}
\label{loconav}
The baseline architecture described above is trained in simulation and achieves state-of-art results on embodied exploration and navigation \cite{ramakrishnan2020occupancy}. The reality, however, poses some major challenges that need to be addressed to achieve good real-world performances. For instance, uneven ground might give rise to errors and noise in the actuation phase. To overcome this and other discrepancies between simulated and real environments, we design LoCoNav: an agent that leverages the availability of powerful simulating platforms during training but is tailored for real-world use. In this section, we describe the main characteristics of the LoCoNav design.
We deploy our architecture on a LoCoBot \cite{locobot} and use PyRobot \cite{murali2019pyrobot} for seamless code integration.

\tit{Prevent your Agent from Learning Tricks}
All simulations are imperfect. One of the main objectives when training an agent for real-world use in simulation is to prevent it from learning simulator-specific tricks instead of the basic navigation skills. During training, we observed that the agent tends to hit the obstacles instead of avoiding them. This behavior is given by the fact that the simulator allows the agent to slide towards its direction even if it is in contact with an obstacle as if there were no friction at all. Unfortunately, this ideal situation does not fit the real world, as the agent needs to actively rotate and head towards a free direction every time it bumps into an obstacle. To replicate the realistic \textit{sticky} behavior of surfaces, we check the \textit{bump}$_t$ flag before every step. If a collision is detected, we prevent the agent from moving forward. As a result, our final agent is more cautious about any form of collision. 

\begin{table}[!t]
    \centering
    \setlength{\tabcolsep}{.3em}
    \resizebox{\linewidth}{!}{
    \begin{tabular}{lcccccc}
    \toprule
         & & \textbf{Height} & \textbf{RGB \acrshort{fov}} & \textbf{Depth \acrshort{fov}} & \textbf{Depth Range} & \textbf{Obst. Height Thr.}\\
    \midrule
        \textbf{Default for Simulation} & & 1.25 & \textbf{H}: 90, \textbf{V}: 90 & \textbf{H}: 90, \textbf{V}: 90 & [0.0, 10.0] & [0.2, 1.5]\\
        \textbf{LoCoNav} & & 0.60 & \textbf{H}: 70, \textbf{V}: 90 & \textbf{H}: 57, \textbf{V}: 86 & [0.0, 5.00] & [0.3, 0.6]\\
    \bottomrule
    \end{tabular}
    }
    \captionsetup{justification=centering}
    \caption{List of hyperparameters changes for Sim2Real transfer.}
    \label{tab:tab1_out}
    % \label{tab:param}
\end{table}

\tit{Sensor and Actuation Noise}
Another important discrepancy between simulation and real world is the difference in the sensor and actuation systems. Luckily, the Habitat simulator allows for great customization of input-output dynamics, thus being very convenient for our goal.
In order to train a model that is more resilient to the camera noise, we apply a Gaussian Noise Model on the RGB observations and a Redwood Noise Model \cite{choi2015robust} on the depth observations. Unfortunately, the LoCoBot RealSense camera still presents various artifacts and regions with missing depth values. For that reason, we need to restore the observation retrieved from the depth camera before using it in our architecture. To that end, we apply the hole-filling algorithm described in \cite{telea2004image}, followed by the application of a median filter.

Regarding the actuation noise, we find out that the use of the incremental pose estimator (employed in the occupancy anticipation model and described in our baseline architecture) is not optimal, especially when combined with the actuation noise typical of real-world scenarios. Luckily, we can count on more precise and reliable information coming from the LoCoBot actuation system. By checking the actual rotation of each wheel at every timestep, the robot can update its position step by step. We adapt the odometry sensor of the LoCoBot platform to be compliant with our architecture. To that end, the pose returned by the sensor is converted by resetting it with respect to its state at the beginning of the episode. We name $\hat{\omega}_0 = (\hat{x}_0, \hat{y}_0, \hat{\theta}_0)$ the coordinate triplet given by the odometry sensor at $t=0$. We then define:
\begin{equation}
    \mathbf{A} = \begin{pmatrix}
            \mathbf{R}_0 & \mathbf{t}_0\\
            \mathbf{0} & 1 
            \end{pmatrix}
            = 
            \begin{pmatrix}
            \cos{\hat{\theta}_0} & -\sin{\hat{\theta}_0} & \hat{x}_0\\
            \sin{\hat{\theta}_0} & \cos{\hat{\theta}_0} & \hat{y}_0\\
            0 & 0 & 1 
            \end{pmatrix}.
\end{equation}
Let us define $\tilde{\omega}_t$ as the augmented position vector $(\hat{x}_t, \hat{y}_t, 1)$ containing the agent position at each step $t$. We compute the relative position of the robot as:
\begin{equation}
    \bar{\omega}_t = \mathbf{A}^{-1}\tilde{\omega}_t, \qquad \theta_t = \hat{\theta}_t - \hat{\theta}_0
\end{equation}
where $\bar{\omega}_t = (x_t, y_t, 1)$ and $\theta_t$ relatively contain the position and the orientation of the agent after the conversion to episode coordinates. In fact, the relative position and heading are given by $\omega_t = ({x}_t, {y}_t, {\theta}_t)$.

Note that, for $t=0$, $\omega_0 = ({x}_0, {y}_0, {\theta}_0) = (0,0,0)$.

\tit{Hyperparameters}
Finally, we noticed that typical hyperparameters employed in simulation do not match the real robot characteristics. For instance, the camera height is set to $1.25$m in previous works, but the RealSense camera on the LoCoBot is placed only $0.6$m from the floor. During the adaptation to the real-world robot, we change some hyperparameters to align the observation characteristics of the simulated and the real world and to match real robot constraints. These parameters are listed in Table \ref{tab:tab1_out}.

\section{Experiments}
\subsection{Testing Setup} We run multiple episodes in the real environment, in which the agent needs to navigate from a starting point A to a destination B. The goal is specified by using relative coordinates (in meters) with respect to the agent's starting position and heading. Although the agent knows the position of its destination, it has no prior knowledge of the surrounding environment. Because of this, it cannot immediately plan a direct route to the goal and must check for obstacles and walls before stepping ahead. After each run, we reset the agent memory so that it cannot retain any information from previous episodes.
We design five different navigation episodes that take place in three different office rooms and the corridor connecting them (Fig. \ref{fig:fig3_out}). For each episode, we run different trials with different configurations: obstacles are added/moved, or people are sitting/standing in the room. In total, we run $50$ different experiments, resulting in more than $10$ hours of real-world testing.

\subsection{Evaluation Protocol} An episode is considered successful if the agent sends a specific \textit{stop} signal within $0.2m$ from the goal. This threshold corresponds to the radius of the robot base. For every navigation episode, we also track the number of steps and the time required to reach the goal. Since the absolute number of steps is not comparable among different episodes, we ask human users to control the LoCoBot and complete each navigation path via a remote interface (we report human performance in Table \ref{tab:fig2_out}). We then normalize these measures using this information so that results close to $1.0$ indicate human-like performances. We provide absolute and normalized length and time for each episode, as well as the popular $\mathsf{\acrshort{spl}}$ metric (Success rate weighted by inverse Path Length). We employ a slightly modified version of the $\mathsf{\acrshort{spl}}$, in which the normalization is made basing on the number of steps and not on the effective path length to penalize purposeless rotations. Additionally, we set a boolean flag for each episode that signals whether the robot has bumped into an obstacle, and we report the average \acrlong{br} ($\mathsf{\acrshort{br}}$). We also report the \acrlong{hfr} ($\mathsf{\acrshort{hfr}}$) as the fraction of episodes terminated if the agent gets stuck and cannot proceed, or if the episode length exceeds the limit of $300$ steps.

\begin{table}[!t]
\centering
\setlength{\tabcolsep}{.45em}
\resizebox{0.5\linewidth}{!}{
\begin{tabular}{cc ccc}
\toprule          
\textbf{Path} & & $\mathsf{Length}$ [m] & $\mathsf{Time}$ [s] & $\mathsf{Steps}$ \\
\midrule
\textbf{A} & & 3.80 & 124 & 23  \\
\textbf{B} & & 6.75 & 239 & 45  \\
\textbf{C} & & 5.95 & 223 & 43  \\
\textbf{D} & & 6.55 & 217 & 42  \\
\textbf{E} & & 4.20 & 227 & 33  \\
\bottomrule
\end{tabular}
}
\captionsetup{justification=centering}
\caption{Path-specific information, as obtained with human supervision.}
\label{tab:fig2_out}
% \label{tab:results}
\end{table}
\begin{figure}[t]
    \centering
    \footnotesize
    \includegraphics[width=0.7\linewidth]{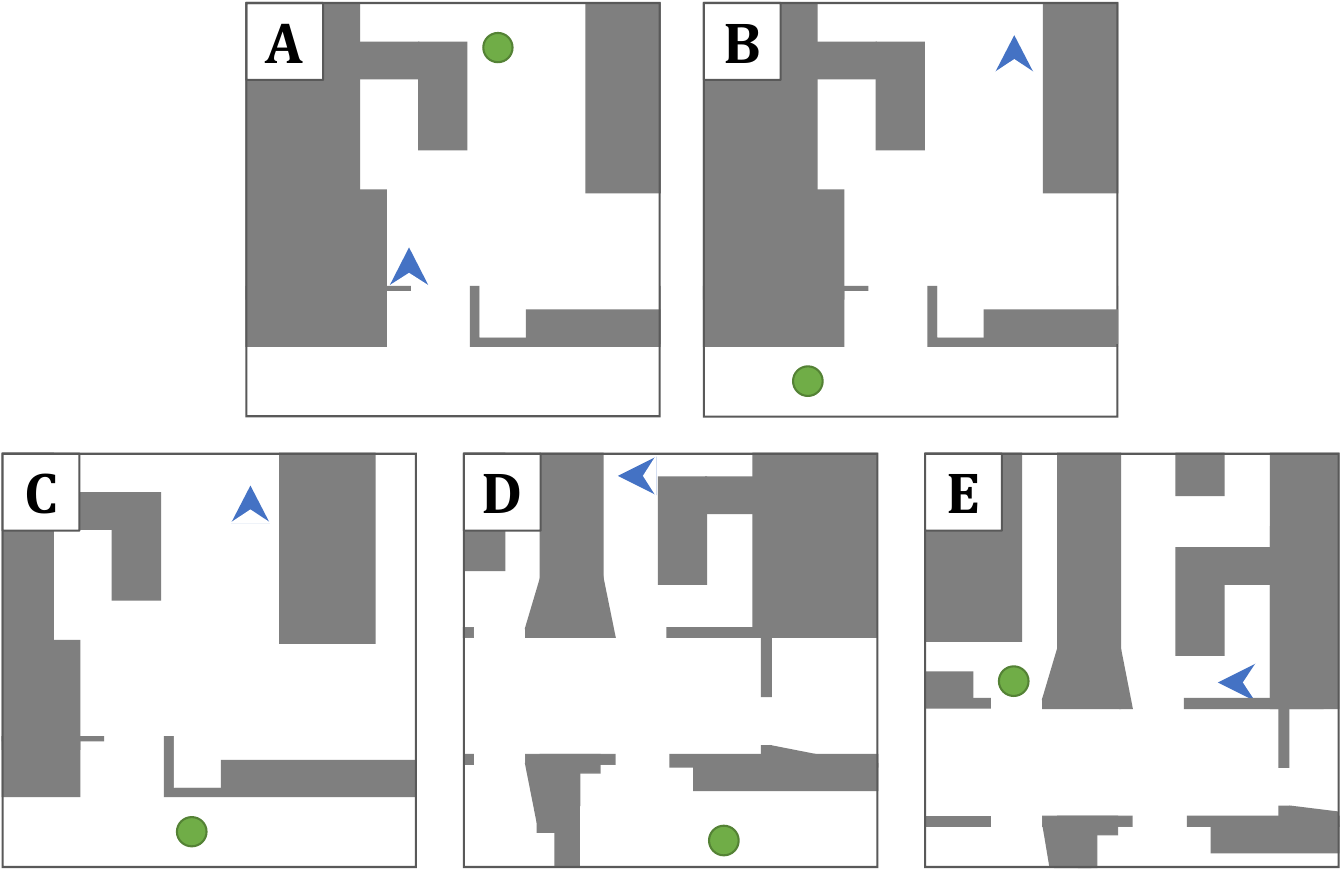}
    \captionsetup{justification=centering}
    \caption{Layout of the navigation episodes.}
    \label{fig:fig2_out}
\end{figure}

\subsection{Real-World Navigation}
\begin{figure}[t!]
    \centering
    \includegraphics[width=\linewidth]{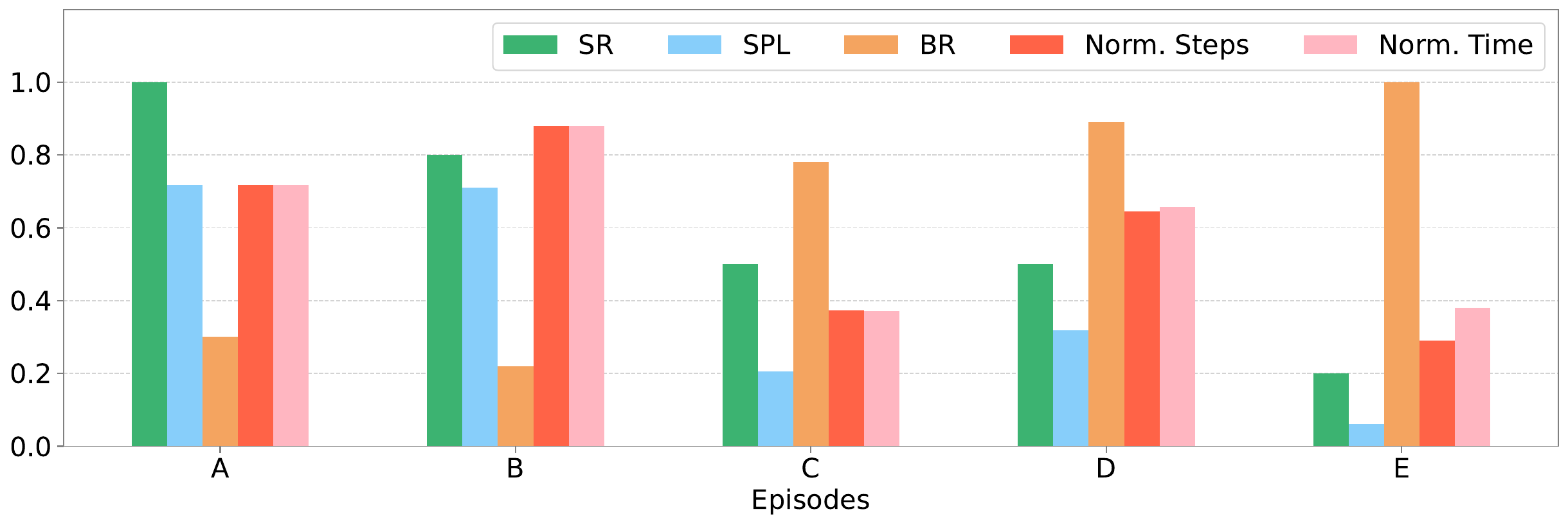}
    \captionsetup{justification=centering}
    \caption{Comparison of the main navigation metrics on different episodes.}
    \label{fig:fig3_out}
\end{figure}

In this experiment, we test our robot on five different realistic navigation paths (Fig. \ref{fig:fig2_out}). We report the numerical results for these experiments in Table \ref{tab:tab3_out}, and we plot the main metrics in Fig. \ref{fig:fig3_out} to allow for a better visualization of navigation results across different episodes. When a path is contained in a single room (A), the agent achieves optimal results, as it always stops within the success threshold from the goal. The number of steps is slightly higher than the minimum required by the episode ($33$ instead of $23$), but this overhead is necessary as the agent must rotate and ``look around'' to build a decent map of the surrounding before planning a route to the goal. Paths that involve going outside the room and navigating different spaces (B, C, D, E) are fairly complicated, but the agent can generally terminate the episode without hard failures. When the shortest path to the goal leads to a wall or a dead-end, the agent needs to find an alternative way to circumvent this obstacle (\eg~a door). This leads to a higher episode length because the robot must dedicate some time to general exploration of the surroundings. Finally, we find out that the most challenging scenario for our LoCoNav is when reaching the goal requires to get out of a room and then enter a door immediately after, on the same side of the corridor (as in E). Since the robot sticks to the shortest path, the low parallax prevents it from identifying the second door correctly. Even in these cases, a bit of general exploration helps to solve the problem.

\begin{table}[t]
\centering
\setlength{\tabcolsep}{.2em}
\resizebox{\linewidth}{!}{
\begin{tabular}{cc cccc cc cc }
\toprule          
$\textbf{Path}$ & & $\mathsf{\acrshort{sr}}$ $\uparrow$ & $\mathsf{\acrshort{spl}}$ $\uparrow$ & $\mathsf{\acrshort{hfr}}$ $\downarrow$ & $\mathsf{\acrshort{br}}$ $\downarrow$
& $\mathsf{Abs.~Steps}$ & $\mathsf{Norm.~Steps}$ $\uparrow$ & $\mathsf{Abs.~Time}$ & $\mathsf{Norm.~Time}$ $\uparrow$ \\
\midrule
\textbf{A} & & 1.0 & 0.718 &  0.0 & 0.30 & 32.70$\pm$1.73 & 0.717$\pm$0.033 & 176.11$\pm$10.39 & 0.718$\pm$0.031 \\
\textbf{B} & & 0.8 & 0.711 & 0.10 & 0.22 & 51.67$\pm$1.72 & 0.880$\pm$0.027 & 273.70$\pm$8.24 & 0.879$\pm$0.030 \\
\textbf{C} & & 0.5 & 0.205 & 0.10 & 0.78 & 123.44$\pm$10.66 & 0.374$\pm$0.034 & 631.15$\pm$50.09 & 0.372$\pm$0.036 \\
\textbf{D} & & 0.5 & 0.318 & 0.10 & 0.89 & 65.67$\pm$3.90 & 0.645$\pm$0.037 & 344.00$\pm$20.08 & 0.657$\pm$0.038 \\
\textbf{E} & & 0.2 & 0.060 & 0.40 & 1.00 & 135.17$\pm$29.97 & 0.290$\pm$0.049 & 722.76$\pm$162.01 & 0.38$\pm$0.066 \\
\midrule
\textbf{Mean} & & 0.6 & 0.402 & 0.14 & 0.60 & \textbf{-} & 0.608$\pm$0.036 & \textbf{-} & 0.617$\pm$0.034 \\
\bottomrule
\end{tabular}
}
\captionsetup{justification=centering}
\caption{Navigation results. Numbers after $\pm$ denote the standard error of the mean.}
\label{tab:tab3_out}
% \label{tab:results}
\end{table}

\subsection{Discussion}
Overall, our experimental setup provides a challenging test-bed for real-world robots. We find out that failures are due to two main issues. First, when the agent must navigate to a different room, it has no access to a map representing the general layout of the environment. This prevents the robot from computing a general plan to reach the long-term goal and forces it to explore the environment before proceeding. If a map was given to the agent, this problem would have been greatly alleviated. A second problem arises when the goal is close in terms $(x,y)$ coordinates but is physically placed in an adjacent room. To solve this problem, one could decompose the navigation between rooms in a multi-goal problem where neighboring nodes are closer. In this way, it is possible to reduce a complex navigation episode in simpler sub-episodes (like A or B), in which our agent has proved to be successful.

% \begin{savequote}[75mm]
% Another inspiring quote for this chapter
% \qauthor{Another author maybe}
% \end{savequote}

\chapter[Navigation at the Art Gallery]{Embodied Navigation at the Art Gallery}
\label{chap:gallerie}
\blfootnote{This Chapter is related to the publication ``R. Bigazzi \etal, Embodied Navigation at the Art Gallery, ICIAP 2021'' \cite{bigazzi2022embodied}. See the list of Publications on page \pageref{publications} for more details.}
\RemoveLabels

\lettrine[lines=1]{\textcolor{SchoolColor}{I}}{n} this last chapter, we contribute to the literature on \gls{eai} by introducing a novel dataset for embodied navigation. 
In recent years, \acrshort{eai} has benefited from the introduction of rich datasets of 3D spaces and new tasks, ranging from exploration to Point Goal or Image Goal Navigation \cite{chang2017matterport3d,xia2018gibson}. Such availability of 3D data allows the training and deployment of modular embodied agents, also thanks to powerful simulation platforms \cite{savva2019habitat}. Despite the high number of available spaces, though, the topology and nature of the different scenes have low variance. Indeed, many environments represent apartments, offices, or houses. In this work, we take a different path and collect and introduce the 3D space of an art gallery.

Current agents for embodied exploration feature a modular approach \cite{chaplot2019learning,ramakrishnan2020occupancy}. While the agents are trained for embodied exploration using deep reinforcement learning, this hierarchical paradigm allows for great adaptability on downstream tasks. Hence, models trained to explore the Gibson dataset can solve \acrlong{pointnav} with satisfactory accuracy under the appropriate hypotheses. Furthermore, accurate and realistic simulating platforms such as Habitat \cite{savva2019habitat} facilitate the deployment in the real world of the trained agents \cite{kadian2020sim2real}. While agent architectures and simulating platforms are possible sources of improvement, there is a third important direction of research that regards the availability of 3D scenes to train and test the different agents. Indeed, the nature of the different environments influences the variety of tasks that the agent can learn and perform. 

In this work, we contribute to this third direction by collecting and presenting a previously unseen type of 3D space, \ie~a museum. This new environment for embodied exploration and navigation, named \acrlong{ag3d} (\acrshort{ag3d}), presents unique features when compared to flats and offices. First, the dimension of the rooms drastically increases, and the same goes for the size of the building itself. In our 3D model, some rooms are as big as $20 \times 15$ meters, while the floor hosting the art gallery spans a total of 2000 square meters. However, dimensions are not the only difference with current available 3D spaces. As a second factor, the presented gallery is incredibly rich in visual features, offering multiple paintings, sculptures, and rare objects of historical and artistic interest. Every item represents a unique point of interest, and this is in contrast to traditional scenes where all elements have approximately the same visual relevance. Finally, the museum has sparse occupancy information. Many agents count on depth information to plan short-term displacements. However, when placed in the middle of an open empty hall, the depth information is less informative. In our challenging 3D scene, the agent must learn to combine RGB and depth information and not be overconfident in immediately available knowledge on the occupancy map. All these challenges make our newly-proposed 3D space a valuable asset for current and future research.

\AddLabels

Together with the 3D model of the museum, we present a dataset for embodied exploration and navigation. For the navigation task, we annotate the position of most of the points of interest in the museum. Examples include numerous paintings, sculptures, and other relevant objects. Finally, we present an experimental analysis including the performance of existing architectures on this novel benchmark and a discussion of potential future research directions made possible by the presence of the collected 3D space.

\section{Art Gallery 3D Dataset}
\label{sec:dataset_gallerie}
Existing datasets for indoor navigation comprise 3D acquisitions of different typologies of buildings, ranging from private houses, that cover the majority of the scenes, to offices and shops. Nevertheless, the focus of these datasets is on private spaces and there is low variance in terms of dimension and contained objects. In fact, to the best of our knowledge, among the publicly available datasets, no acquired indoor environment is composed of large rooms with a low occupied/free space ratio as in a museum. To overcome this deficiency in current literature we release a new indoor dataset for exploration and navigation captured inside a museum environment, called \acrshort{ag3d}\footnote{The dataset has been collected at the Galleria Estense museum of Modena and can be found at \url{https://github.com/aimagelab/ag3d}.}.

\begin{figure}[t!]
\centering
\includegraphics[width=\linewidth]{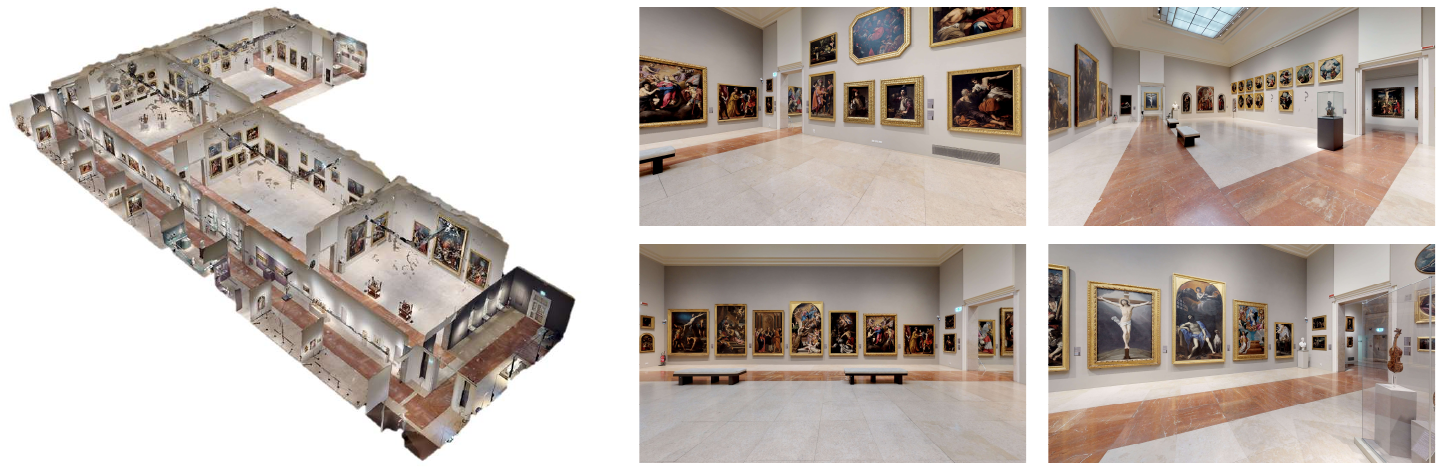}
\caption{On the left: a view of the 3D model of the acquired environment. On the right: images captured during the acquisition of the scene.}
\label{fig:fig1_gallerie}
\end{figure}

\tit{Acquisition}
To build the 3D model of the art gallery, we employ a Matterport camera\footnote{\url{https://matterport.com/it/cameras/pro2-3D-camera}} and related software. This technology is the same employed to collect Matterport3D and HM3D datasets of spaces \cite{chang2017matterport3d,ramakrishnan2021hm3d} and is particularly suitable to capture indoor photo-realistic environments.
We place the camera in the physical environment and capture a 360\textdegree~RGB-D image of the surrounding. Then, we repeat the same process after moving the camera approximately 1.5 meters away. Using consecutive panoramic acquisitions, the software is able to compute the 3D geometry of the space using depth information and the correspondences between the same keypoints in different acquisitions.
To capture the entire museum, we make 232 different scans. Thanks to the high number of acquisitions, we are able to reproduce fine geometric and visual details of the original space (see Fig. \ref{fig:fig1_gallerie}). The resulting 3D model consists of more than $1430$~m\textsuperscript{2} of navigable space.

\begin{figure}[t!]
\centering
\includegraphics[width=0.9\linewidth]{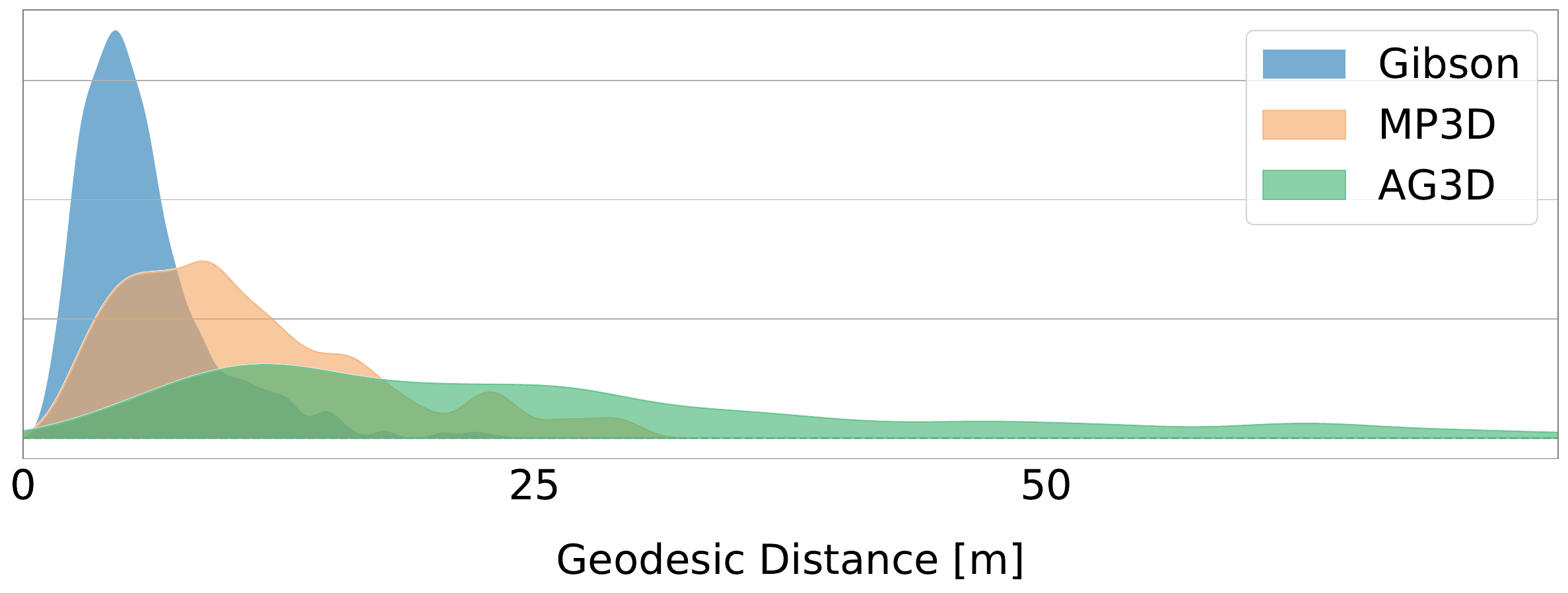}
\caption{Comparison of the distribution of the geodesic distances from starting position to goal position of the episode for different datasets.}
\label{fig:fig2_gallerie}
\end{figure}

\tit{Dataset Details}
The proposed dataset allows two different tasks: exploration and navigation. Episodes for the exploration task include starting position and orientation of the agent which are sampled uniformly over the entire navigable space. The navigation dataset, instead, extends traditional \acrlong{pointnav} where episodes are defined with a starting pose and a goal coordinate, including an additional final orientation vector. Conceptually, we can consider this setting as the link between \acrlong{pointnav} and ImageGoal navigation since the goal is to rotate the agent towards a precise objective/scene, specifying the goal using coordinates instead of an image. We name this new setting Point Goal$\mathcal{++}$ navigation (\POINTNAV). To create the navigation dataset we annotate 147 points of interest mostly consisting of paintings and statues. The annotated goal position is around 1 meter in front of the artwork and the goal orientation vector is directed to its center. For each point of interest, we define three episodes with different difficulties based on the geodesic distance between start and goal positions: easy ($<15$m), medium ($>15$m), and difficult ($>30$m). In particular, thanks to the dimension of the acquired environment, each difficult episode has a geodesic distance larger than the longest path of \acrlong{mp3d} and Gibson datasets. A comparison of the geodesic distance distribution of the episodes of various available \acrlong{pointnav} datasets is presented in Fig. \ref{fig:fig2_gallerie}. The introduction of \acrshort{ag3d} enables the evaluation of agents on long navigation episodes which were previously not possible and highlights the inaccuracy of components of the architecture that accumulate error over time.
The exploration task dataset contains 500k, 100, 1000 episodes respectively for training, validation, and test, while the \POINTNAV~dataset includes 411 annotated navigation episodes.

\section{Proposed Method}
\label{sec:architecture_gallerie}
\begin{landscape}
\begin{figure}[t!]
\centering
\includegraphics[width=0.8\linewidth]{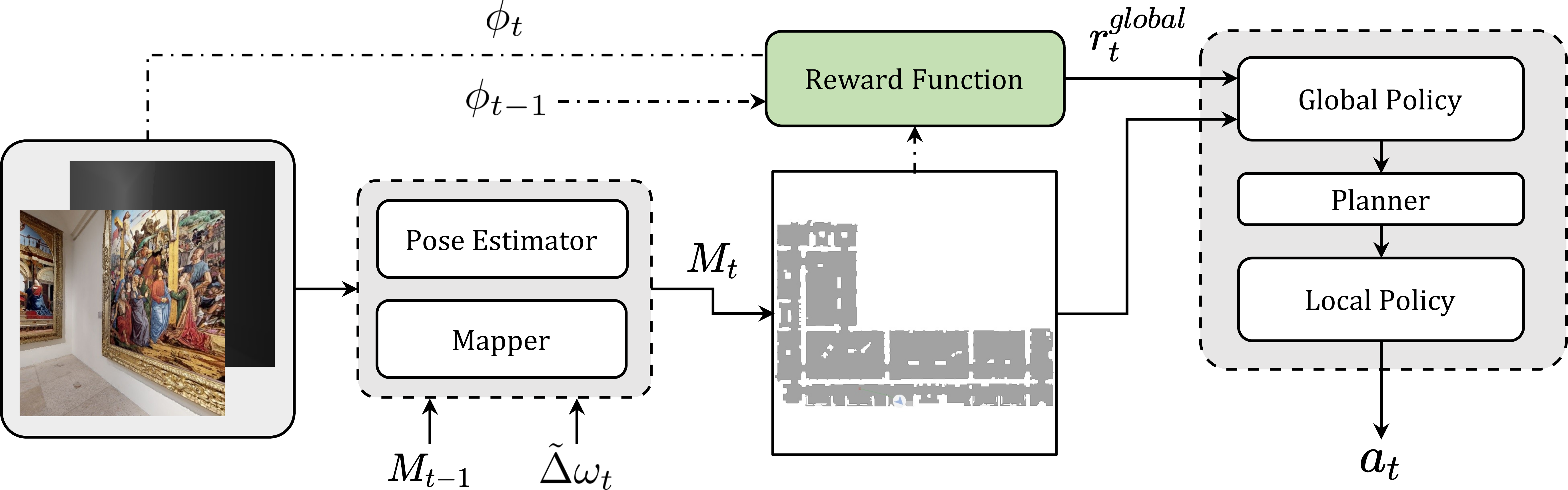}
\captionsetup{justification=centering}
\caption{Overall architecture of the models employed for exploration and navigation on \acrshort{ag3d} .}
\label{fig:fig3_gallerie}
\end{figure}
\end{landscape}

We provide an experimental analysis comparing recently proposed approaches on the devised environment, both for exploration and \POINTNAV~tasks. The evaluated methods are consistent with recent literature on \acrshort{eai} \cite{chaplot2019learning,ramakrishnan2020occupancy} and adopt an architecture shown in Fig. \ref{fig:fig3_gallerie}, which is composed of a neural mapper, a pose estimator, and a hierarchical navigation policy.
The mapper generates a representation of the environment while the agent moves, the pose estimator is in charge of locating the agent in the environment, and the policy is responsible for the movement capabilities of the agent. The core difference between the evaluated approaches resides in the navigation policy, as described in the following. For further details, see Section \ref{sec:method_impact}.

\subsection{Mapper}
The mapper module incrementally builds an occupancy grid map of the environment in parallel with the navigation task. At each timestep, the RGB-D observations $(s^{rgb}_t, s^d_t)$ coming from the visual sensors are processed to extract a $V \times V \times 2$ agent-centric map $m_t$ where the channels indicate, respectively, the occupancy and exploration state of the currently observed region, and each pixel of the map describes the state of an area of $5 \times 5$ cm. The mapper is not limited to predicting the occupancy map of the visible space but infers also occluded and not visible regions of the local map. More details on the architecture of the mapper can be found in Section \ref{ssec:mapper_impact}.
The global level map of the environment $M_t$ has a dimensionality of $W \times W \times 2$, where $W > V$, and is built using local maps $m_t$ step-by-step. At each timestep the pose of the agent $\omega_t$ is used to apply a rototranslation to the local map, then, the transformed local map is finally registered to the global map $M_t$ with a moving average.

\subsection{Pose Estimator}
In order to create a coherent representation of the environment during navigation, a precise and robust pose estimation needs to be achieved. To address problems like noise in the sensors and collisions with obstacles, we adopt a pose estimator and avoid the direct use of sensor readings. The pose estimator computes the pose of the agent $\omega_t = (x_t, y_t, \theta_t)$ where $(x_t, y_t)$ and $\theta_t$ are its position and orientation in the internal representation of the environment. 

Specifically, the output of this module is the displacement $\Delta \omega_t$ caused by the agent's last action. The input of the pose estimator is the difference between consecutive readings of the pose sensor $(\widetilde{\omega}_{t-1},\widetilde{\omega}_t)$, but such measure could be noisy and needs to be adjusted. To do so, we use consecutive local maps $(m_{t-1},m_{t})$ coming from the mapper as feedback. At each timestep $\Delta \omega_t$ is used to compute the pose of the robot $\omega_{t}$:

\begin{equation}
    \omega_t = \omega_{t-1} + \Delta \omega_t,
    \label{eq:estimate_displacement}
\end{equation}
where we assume $\omega_0 = (0, 0, 0)$ without loss of generality and $\omega_0$ corresponds to the center of the map $M_t$ with the agent facing north.

\subsection{Navigation Policy}
The navigation policy is the module that determines the movement of the agent in the environment. Its hierarchical design is required in order to allow the agent to uncouple high-level navigation concepts, such as navigating through different rooms, and low-level concepts, like obstacle avoidance. The navigation policy is defined by a three-component module consisting of a global farsighted policy, a deterministic planner, and a local policy for atomic action inference. The architecture of such modules is presented in Section \ref{ssec:navpolicy_impact}.

The global policy is the high-level component of the navigation policy and is responsible for extracting a long-term goal on the global map $g_t$.
The global policy is trained with reinforcement learning using PPO \cite{schulman2017proximal} to maximize different rewards used in literature. 
In the experiments, we employ and compare different reward methods, namely Coverage, Anticipation, and Curiosity. The Coverage reward \cite{chaplot2019learning,ramakrishnan2020exploration} maximizes the information gathered at each timestep, expressed in terms of the area seen. The Anticipation reward \cite{ramakrishnan2020occupancy} is defined by comparing the predicted local occupancy map with the ground truth considering also occluded areas. The Curiosity reward \cite{pathak2017curiosity} encourages the agent towards areas that maximize the prediction error of a model trained to predict future states, thus improving the learning of the dynamics of the environment.

Given the global goal on the map, the planner has the task of computing a short-term goal on the map that the agent should reach. We employ an A* algorithm on the global map $M_t$ to plan a path from the current position of the agent to the global goal and a local goal $l_t$ is computed on the obtained trajectory within a distance $D$ from the agent.

The local policy is the module that allows the movement of the agent in the environment and its objective is to reach the local goal $l_t$ determined by the planner. The input of the local policy, formed by the relative displacement from the position of the agent to the local goal $l_t$ and the current RGB observation $\phi^{rgb}_t$, is processed to compute an atomic action $a_t$. The available actions are \textit{move ahead 0.25m}, \textit{turn left 10\textdegree}, \textit{turn right 10\textdegree}, with the addition of a \textit{stop} action when performing the navigation task. During training with reinforcement learning, the reward of the local policy $r^{local}_t$ encourages the agent to reduce the distance from the local goal.

Following the hierarchical design, the global goal is sampled every $\eta$ timesteps, while the local goal is reset if a new global goal is sampled, if the previous local goal is found to be in an occupied area, or if the previous local goal has been reached.

\section{Experimental Setup}
\begin{figure}[t!]
\centering
\includegraphics[width=\linewidth]{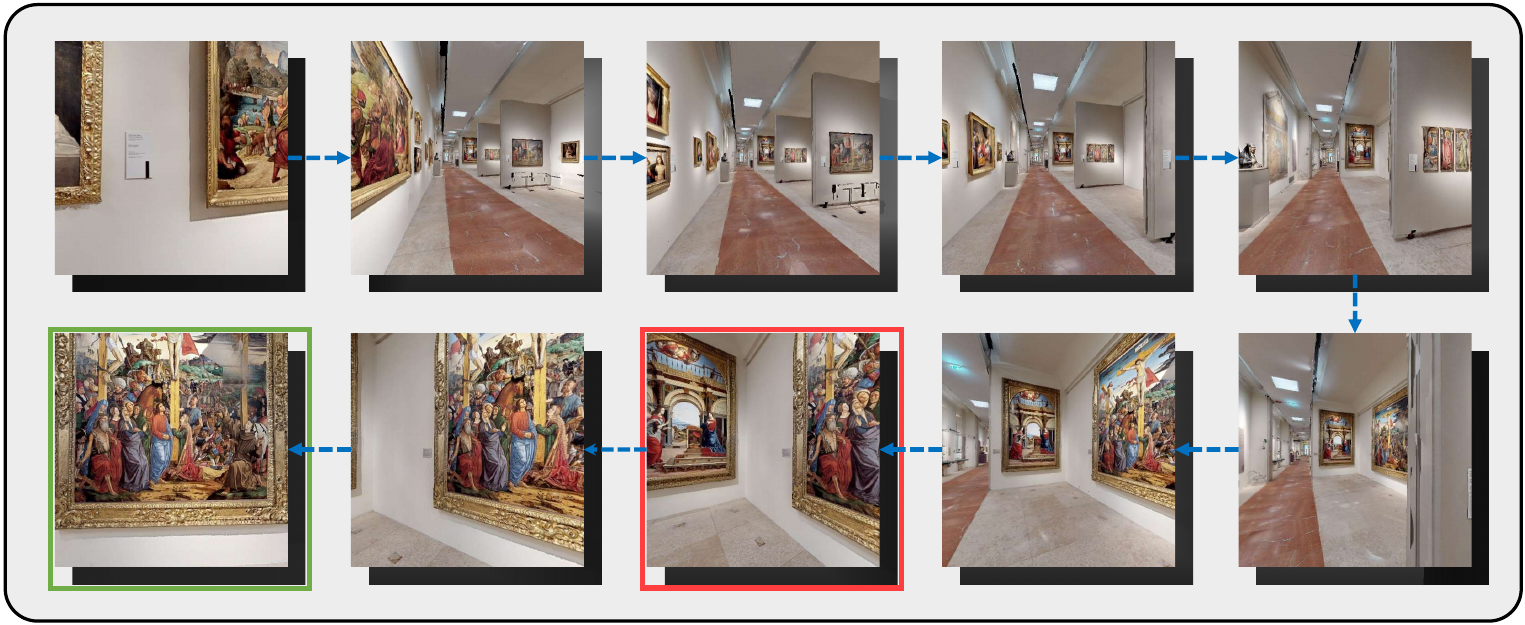}
\caption{An episode of \POINTNAV~in \acrshort{ag3d} where consecutive frames have a distance of 10 timesteps approximately. The red frame indicates the stop action in the traditional \acrshort{pointnav} task. The green frame corresponds to the stop action in \POINTNAV.}
\label{fig:fig4_gallerie}
\end{figure}

We perform some experiments on the proposed dataset comparing various models trained with different global rewards on another dataset, with models trained from scratch or finetuned on \acrshort{ag3d} on exploration and \POINTNAV~to evaluate the performance gap between these approaches and highlight the difference between the characteristics of \acrshort{ag3d} compared to other datasets. An example of an episode of Point Goal$\mathcal{++}$ navigation of \acrlong{ag3d} is shown in Fig. \ref{fig:fig4_gallerie}.

\subsection{Evaluation Protocol}

The baselines are trained with Coverage, Anticipation, and Curiosity rewards on the Gibson dataset for $\approx5$M frames corresponding to 12 \acrshort{gpu} days on a single NVIDIA V100. The best-performing approach among the baseline is also both trained from scratch and finetuned, but since high-quality textures and memory occupancy of \acrshort{ag3d} do not allow training with the same number of environments in parallel as Gibson, we trained the model from scratch on \acrshort{ag3d} with the same \acrshort{gpu} time for $\approx2.8$M frames, while the finetuned model is trained for $\approx1$M additional frames.

For the exploration task, we evaluate the following metrics: \acrlong{iou} ($\mathsf{\acrshort{iou}}$) between the map built at the end of the episode and the ground-truth map. $\mathsf{\acrshort{acc}}$ measures the correctly reconstructed map in m\textsuperscript{2}. $\mathsf{\acrshort{as}}$ indicates the area seen by the agent during exploration (in m\textsuperscript{2}). $\mathsf{\acrshort{fiou}}$, $\mathsf{\acrshort{oiou}}$, $\mathsf{\acrshort{fas}}$, and $\mathsf{\acrshort{oas}}$ measure, respectively, $\mathsf{\acrshort{iou}}$ and area seen for free and occupied portions of the environment. $\mathsf{\acrshort{te}}$ and $\mathsf{\acrshort{ae}}$ are the translation and angular error between estimated and ground-truth pose measured respectively in meters and degrees.
Point Goal++ navigation is evaluated considering these metrics: \acrlong{d2g} ($\mathsf{\acrshort{d2g}}$) and \acrlong{oe} ($\mathsf{\acrshort{oe}}$) are the mean geodesic distance to the goal and the mean orientation error at the end of the episode. The orientation error is computed considering the vector between the center of the artwork and the position of the agent as ground truth. \Acrlong{sr} ($\mathsf{\acrshort{sr}}$) is the percentage of episodes terminated successfully. In \POINTNAV~the agent needs to be within 0.2 meters of the goal and with an orientation error lower than 10 degrees. \acrlong{pgsr} ($\mathsf{\acrshort{pgsr}}$) and \acrlong{asr} ($\mathsf{\acrshort{asr}}$) consider only one component of $\mathsf{\acrshort{sr}}$; respectively $\mathsf{\acrshort{d2g}}$ and $\mathsf{\acrshort{oe}}$. $\mathsf{\acrshort{spl}}$ and \acrlong{sspl} ($\mathsf{\acrshort{sspl}}$) are \acrlongpl{sr} weighted on the length of the trajectory of the agent.

\subsection{Implementation Details}
The experiments are performed by extracting $128 \times 128$ RGB-D observations from the acquired 3D model using the Habitat simulator. The maximum length of the exploration episodes during training is set to $T=500$. Regarding the mapping process, we set $V=101$ and $W=2881$ for the local and global map dimensionalities. The action space grid $H \times H$ of the global policy is $240 \times 240$. The maximum distance of the local goal $l_t$ from the position of the agent is $D=0.5$m for exploration and $D=0.25$m for \POINTNAV.

\section{Experimental Results}

\begin{table}[t]
\footnotesize
\centering
\setlength{\tabcolsep}{.3em}
\resizebox{\linewidth}{!}{
\begin{tabular}{l cc ccccccccc}
\toprule
\textbf{Model} & \textbf{Training} & & $\mathsf{\acrshort{iou}}$ $\uparrow$ & $\mathsf{\acrshort{fiou}}$ $\uparrow$ & $\mathsf{\acrshort{oiou}}$ $\uparrow$ & $\mathsf{\acrshort{acc}}$ $\uparrow$ & $\mathsf{\acrshort{as}}$ $\uparrow$ & $\mathsf{\acrshort{fas}}$ $\uparrow$ & $\mathsf{\acrshort{oas}}$ $\uparrow$ & $\mathsf{\acrshort{te}}$ $\downarrow$ & $\mathsf{\acrshort{ae}}$ $\downarrow$ \\
\midrule
\textbf{Noise-Free}\\
\hspace{0.25cm}Anticipation~\cite{ramakrishnan2020occupancy} & Gibson & & 0.163 & 0.170 & 0.157 & 294.4 & 290.6 & 258.3 & 32.3 & 0.0 & 0.0\\
\hspace{0.25cm}Curiosity~\cite{pathak2017curiosity} & Gibson & & 0.175 & 0.184 & 0.166 & 317.9 & 317.5 & 281.7 & 35.8 & 0.0 & 0.0\\
\hspace{0.25cm}Coverage~\cite{chaplot2019learning,ramakrishnan2020exploration} & Gibson & & 0.214 & 0.237 & 0.191 & 403.1 & 384.3 & 341.1 & 43.2 & 0.0 & 0.0 \\
\hspace{0.25cm}Coverage~\cite{chaplot2019learning,ramakrishnan2020exploration} & \acrshort{ag3d} & & 0.219 & 0.239 & 0.200 & 400.6 & 354.6 & 316.6 & 38.0 & 0.0 & 0.0 \\
\hspace{0.25cm}Coverage~\cite{chaplot2019learning,ramakrishnan2020exploration} & Gibson+\acrshort{ag3d} & & \textbf{0.296} & \textbf{0.313} & \textbf{0.278} & \textbf{531.8} & \textbf{470.2} & \textbf{418.1} & \textbf{52.1} & 0.0 & 0.0 \\
\midrule
\textbf{Noisy}\\
\hspace{0.25cm}Anticipation~\cite{ramakrishnan2020occupancy} & Gibson & & 0.144 & 0.157 & \textbf{0.131} & 269.6 & 281.7 & 249.9 & 31.9 & \textbf{0.48} & \textbf{2.95}\\
\hspace{0.25cm}Curiosity~\cite{pathak2017curiosity} & Gibson & & 0.119 & 0.151 & 0.086 & 251.1 & 307.3 & 272.8 & 34.5 & 2.62 & 15.99\\
\hspace{0.25cm}Coverage~\cite{chaplot2019learning,ramakrishnan2020exploration} & Gibson & & 0.148 & 0.203 & 0.093 & 327.0 & 380.7 & 337.9 & 42.8 & 2.98 & 12.54\\
\hspace{0.25cm}Coverage~\cite{chaplot2019learning,ramakrishnan2020exploration} & \acrshort{ag3d} & & 0.144 & 0.200 & 0.088 & 320.0 & 356.4 & 317.3 & 39.2 & 2.65 & 13.35 \\
\hspace{0.25cm}Coverage~\cite{chaplot2019learning,ramakrishnan2020exploration} & Gibson+\acrshort{ag3d} & & \textbf{0.191} & \textbf{0.266} & 0.116 & \textbf{427.3} & \textbf{461.9} & \textbf{413.2} & \textbf{48.7} & 2.68 & 10.16 \\
\bottomrule
\end{tabular}
}
\caption{Exploration results over the 100 episodes of the \acrshort{ag3d} validation split in noise-free and noisy conditions. Using both Gibson and \acrshort{ag3d} during the training guarantees significantly higher performance.}
\label{tab:tab1_gallerie}
\end{table}

\subsection{Exploration Results} As a first experiment, in Table \ref{tab:tab1_gallerie} we compare the considered models on the exploration task on the \acrshort{ag3d} validation set. Each exploration episode has a length of $T=1000$ timesteps during which the agent has to disclose the initially unknown environment. Among the baselines trained only on the Gibson dataset, the Coverage-based model achieves the best results in terms of $\mathsf{\acrshort{iou}}$ and \acrlong{as} in both noise-free and noisy settings. The model trained with Coverage from scratch obtains competitive results even using fewer training frames (2.8M vs. 5M), showing the importance of adapting the models to \acrshort{ag3d}. This conclusion is supported by the fact that the model trained on Gibson and finetuned for 1M frames on \acrshort{ag3d} achieves the best results on noise-free and noisy settings, with a significant margin on the second-best model. In both settings, the performance gap in terms of \acrlong{as} (85.9m\textsuperscript{2} and 81.2m\textsuperscript{2}) and $\mathsf{\acrshort{iou}}$ (0.082 and 0.043) between the best models trained only on Gibson dataset and using \acrshort{ag3d} denotes the need of adapting the weight of the models to the different visual characteristics and occupation of \acrshort{ag3d}.

\begin{table}[t]
\footnotesize
\centering
\setlength{\tabcolsep}{.3em}
\resizebox{\linewidth}{!}{
\begin{tabular}{l cc cccccccc}
\toprule
\textbf{Model} & \textbf{Training} & & $\mathsf{\acrshort{spl}}$ $\uparrow$ & $\mathsf{\acrshort{sspl}}$ $\uparrow$ & $\mathsf{\acrshort{sr}}$ $\uparrow$ & $\mathsf{\acrshort{pgsr}}$ $\uparrow$ & $\mathsf{\acrshort{asr}}$ $\uparrow$ & $\mathsf{Steps}$ $\downarrow$ & $\mathsf{\acrshort{d2g}}$ $\downarrow$ & $\mathsf{\acrshort{oe}}$ $\downarrow$ \\
\midrule
\textbf{Noise-Free}\\
\hspace{0.25cm}Anticipation~\cite{ramakrishnan2020occupancy} & Gibson & & 0.697 & 0.780 & 0.803 & 0.873 & 0.808 & 364.3 & 4.131 & 12.2 \\
\hspace{0.25cm}Curiosity~\cite{pathak2017curiosity} & Gibson & & 0.625 & 0.706 & 0.732 & 0.803 & 0.732 & 416.4 & 7.934 & 17.0\\
\hspace{0.25cm}Coverage~\cite{chaplot2019learning,ramakrishnan2020exploration} & Gibson & & 0.760 & 0.838 & 0.876 & 0.954 & 0.883 & 314.6 & 0.700 & 5.2 \\
\hspace{0.25cm}Coverage~\cite{chaplot2019learning,ramakrishnan2020exploration} & \acrshort{ag3d} & & \textbf{0.805} & \textbf{0.875} & \textbf{0.898} & \textbf{0.973} & \textbf{0.908} & \textbf{270.3} & \textbf{0.268} & \textbf{4.8} \\
\hspace{0.25cm}Coverage~\cite{chaplot2019learning,ramakrishnan2020exploration} & Gibson+\acrshort{ag3d} & & 0.793 & 0.873 & 0.883 & 0.964 & 0.891 & 273.1 & 0.323 & 5.0 \\
\midrule
\textbf{Noisy}\\
\hspace{0.25cm}Anticipation~\cite{ramakrishnan2020occupancy} & Gibson & & 0.211 & 0.788 & 0.224 & 0.255 & 0.387 & 338.6 & 3.152 & 32.2 \\
\hspace{0.25cm}Curiosity~\cite{pathak2017curiosity} & Gibson & & 0.225 & 0.655 & 0.243 & 0.275 & 0.341 & 446.3 & 9.746 & 38.7 \\
\hspace{0.25cm}Coverage~\cite{chaplot2019learning,ramakrishnan2020exploration} & Gibson & & 0.228 & 0.783 & 0.243 & 0.260 & 0.392 & 348.6 & 3.165 & 34.1 \\
\hspace{0.25cm}Coverage~\cite{chaplot2019learning,ramakrishnan2020exploration} & \acrshort{ag3d} & & 0.235 & 0.832 & 0.248 & 0.273 & 0.445 & 306.2 & 2.420 & 28.8 \\
\hspace{0.25cm}Coverage~\cite{chaplot2019learning,ramakrishnan2020exploration} & Gibson+\acrshort{ag3d} & & \textbf{0.373} & \textbf{0.853} & \textbf{0.399} & \textbf{0.443} & \textbf{0.543} & \textbf{283.8} & \textbf{1.430} & \textbf{19.8} \\
\bottomrule
\end{tabular}
}
\caption{\POINTNAV~results on the \acrshort{ag3d} navigation episodes under noise-free and noisy settings. \acrshort{ag3d} is fundamental to reach the best results.}
\label{tab:tab2_gallerie}
\end{table}

\subsection{Navigation Results} 
Moving on to the navigation task, models trained on exploration substitute the global goal with a fixed goal specified by the navigation episode.
Experimental results on \POINTNAV, shown in Table \ref{tab:tab2_gallerie}, present a similar trend as on the exploration task. In fact, the Coverage model has the best results in terms of $\mathsf{\acrshort{spl}}$ and \acrlong{sr} related metrics among the models trained only on Gibson. Moving to the Coverage models trained on \acrshort{ag3d}, in the noise-free setting, the model trained from scratch achieves the best results even in comparison to the finetuned counterpart which is trained with more than double the total observations (2.8M vs 6M). This behavior can be explained by the performance of its mapper which is trained for more frames using visual observation from \acrshort{ag3d} (2.8M vs 1M) and extracts a more detailed map sacrificing robustness and generalization. Accordingly, in the noisy setting, the higher robustness of the Coverage-based model finetuned on \acrshort{ag3d} regains the first place with a noteworthy margin on the other models, while the Coverage model trained from scratch goes down to the second position in terms of $\mathsf{\acrshort{spl}}$ and $\mathsf{\acrshort{sr}}$. As in the case of the exploration task, the performance gap between models trained on Gibson and using \acrshort{ag3d} (0.045 and 0.145 for $\mathsf{\acrshort{spl}}$ in noise-free and noisy settings) stresses the importance of adapting the parameters to the features extracted from \acrshort{ag3d}. Moreover, it is worth noting that the gap of the best model from noise-free to noisy navigation (0.432 for $\mathsf{\acrshort{spl}}$) is a consequence of the length of the navigation episodes of \acrshort{ag3d}, and the difficulty of performing precise lengthy trajectories in the presence of noise. This is an interesting aspect that the \acrshort{ag3d} dataset offers for exploration in future works.

\chapter[Conclusions]{Conclusions}
\label{chap:conclusions}
\RemoveLabels

\lettrine[lines=1]{\textcolor{SchoolColor}{T}}{his} dissertation contributes to the research in \gls{eai}, aiming to foster future research on the topic and hoping to help researchers willing to work in this complex field. This final chapter starts with a section that summarizes the work included in this thesis, reviewing the contributions introduced in each chapter. In the following section, we include a discussion about the problems and lacks afflicting the current research in this field. After that, the possible future work and directions of research are described. To conclude this thesis, we present some final remarks and the activities carried out during the Ph.D. program.

\section{Contributions of the Thesis}

\subsection{Exploration with Intrinsic Motivation}
\label{sec:conclusion_impact}
In Chapter \ref{chap:focus}, we presented an impact-driven approach for robotic exploration in indoor environments. Differently from previous research that considered a setting with procedurally-generated environments with a finite number of possible states, we tackle a problem where the number of possible states is non-numerable. To deal with this scenario, we exploit a deep neural density model to compute a running pseudo-count of past states and use it to regularize the impact-based reward signal. The resulting intrinsic reward allows to efficiently train an agent for exploration even in absence of an extrinsic reward, including also remarkable results on the downstream task of navigation. The proposed agent stands out from the recent literature on embodied exploration in photo-realistic environments.

\subsection{Exploration and Recounting}
\label{sec:conclusion_ex2}
Chapter \ref{chap:ex2} presents a new setting for \gls{eai} that is composed of two tasks: exploration and captioning. The architecture of \ours~uses intrinsic rewards applied to exploration in a photo-realistic environment and a speaker module that generates captions. The captioner produces sentences according to a speaker policy that could be based on three metrics. The experiments show that \ours~is able to generalize to unseen environments in terms of exploration, while the speaker policy functions to filter the number of timesteps where the caption is actually generated.

\AddLabels

\subsection{Efficient Exploration and Smart Scene Description}
\label{sec:conclusion_eds}
In this work proposed in Chapter \ref{chap:eds}, we have improved the architecture presented in the previous chapter in both the exploration and the speaker module. Alongside these improvements, we devise a novel metric for the task that enables the evaluation and comparison of different baselines. This approach is a viable solution to gain insights into the perception and navigation capabilities of embodied agents. Moreover, the generalization capabilities of the modules adopted allow real-world deployment without major redesigns.

\subsection{Embodied Agents in Changing Environments}
\label{sec:conclusion_sd}
Chapter \ref{chap:sd} proposed \emph{Spot the Difference}: a new task for navigation agents in changing environments. In this setting, the agent has to find all variations that occurred in the environment with respect to an outdated occupancy map. Since current datasets of 3D spaces do not account for such variety, we collected a new dataset with different layouts for the same environment. 
We tested two exploration agents on this task and proposed a novel reward term to encourage the discovery of meaningful information during exploration. The proposed agent outperforms the competitors and can identify changes in the environment more efficiently.

\subsection{Navigation in the Real World}
\label{sec:conclusion_out}
In Chapter \ref{chap:out} we have presented LoCoNav, an out-of-the-box architecture for embodied navigation in the real world. After a phase of training using simulation, the implementation of the trained model on a real robotic platform needs to be performed. We present a series of techniques specifically designed for real-world deployment. Experiments are conducted in reality on challenging navigation paths and in a realistic office-like environment demonstrating the validity of our approach.

\subsection{Navigation at the Art Gallery}
\label{sec:conclusion_gallerie}
In this work contained in Chapter \ref{chap:gallerie}, we introduced \DATASETS, a photo-realistic 3D dataset designed for embodied exploration and navigation tasks. The dataset has been collected in an art gallery, which features larger and more uncluttered spaces compared to most of the environments available in commonly used benchmark datasets. For the \acrshort{pointnav} task, we propose a variant that is more suitable to the type of environment in the \DATASETS~dataset. The variant entails not only reaching the specified coordinates, as in standard \acrshort{pointnav} but also assuming a specified orientation. We also present an experimental comparison of exploration approaches on the devised dataset, which can serve as baselines for future research in museum-like environments.

\section{Key Future Directions}
After the description of the contributions presented in the thesis, this section describes what is still missing and what we think could be important deficiencies in the research on \gls{eai}.

One of the aspects that has been only slightly explored in the current literature on \gls{eai} is the capacity of the robots to reason about semantic concepts in the environment. Great effort has been poured into agents' ability to perform high-level navigation reaching objects and coordinates, or following textual instructions. These tasks are usually done by training models with massive use of data, instructions, and training time, focusing only on reaching the final goal. However, if we take \acrlong{objectnav} as a case study, the relations between different types of objects or commonsense knowledge has been hardly exploited to pursue the task. If the agent is requested to find a chair, there is a high probability that it will be found next to a table. Instead, the goal object is a bathtub, commonsense suggests finding a bathroom first. Unfortunately, these concepts and relations are usually not acquired or annotated in current datasets, or in case the dataset contains this type of annotations they are usually very noisy or not curated. Our expectations for the next years of research in \gls{eai} is to focus more on the reasoning of the agents, instead of working exclusively looking for higher numbers on the available datasets.  

Another major drawback is that in current research on \gls{eai}, the agent's interactions are limited. Usually, most tasks consist of just an input at the beginning of the episode and the agent does not interact with the surrounding environment for the entire duration of the episode. We think that before being able to reach the seamless interaction between robots and humans, ideally the episode should not be reset and the agents should be able to interact more with surrounding humans and other entities and be able to condition their behavior using these interactions.

Additionally, after having considered some of the work that has been published in the last few years \cite{karkus2021differentiable,zhao2021surprising,partsey2022mapping}, we would like to address the fact that taking inspiration from some traditional robotic approaches, also learning-based \gls{ai} research could flourish and produce improvements over previous literature. Even if this aspect could seem simple, we think that traditional robotic knowledge should be linked tighter with the research on this topic.

The last concern we would like to address is the actual realism of simulating platforms. After having discussed photo-realistic environments in Chapter \ref{chap:gallerie}, we believe that a major problem is the presence of artifacts, reconstruction errors, and unrealistic physics behaviors in such simulators. All these issues degrade the performances of agents trained in simulation when deployed in real-world settings. Improving the simulators in these aspects could return a significant boost to embodied agents' performance in the real world.

\section{Future Work}
Regarding the possible future directions of research, we are interested in tackling the issues presented in the previous section. A first step towards real semantic navigation could be giving the agent the ability to understand the semantic meaning of the surrounding environment, \ie being able to understand that a certain room is a kitchen, a bathroom, or another room, studying its visual appearance and the objects contained in it.
Moreover, observing the outstanding results achieved in the research on conversational agents with \glspl{llm}, an increasing interest in exploiting such models for robotic tasks has risen. For example, recent work has started leveraging \glspl{llm} for manipulation \cite{huang2022inner,liang2022code,ahn2022can} and navigation tasks \cite{huang2022visual,singh2022progprompt}. However, further research is still required to fully integrate \glspl{llm} capabilities on board of an embodied agent to consistently upgrade its reasoning. We aim to be part of such research by developing new approaches that combine aspects of both robotic and conversational agents. In fact, the advances brought by this dissertation can be further evolved by integrating the semantics contained in \acrlongpl{llm}, because while the presented work is focusing on the observation level, \glspl{llm} could be adopted as a general high-level controller.

In parallel, we also think that extending the duration of episodes and moving towards multi-goal navigation could be another way to move towards fluid robot-human interaction. In such a task, the robot should be able to request more information on the object to look for and be corrected when it goes the wrong way. The final goal of this change is the development of lifelong or never-resetting agents, while in the current setting, agents are reset every episode and the capacity to backtrack from errors is not explored.

\section{Final Remarks}
The efforts presented in this thesis have been published in international journals and conferences. 
For example, the work on intrinsic motivation presented in Chapter \ref{chap:focus} has been published in IEEE Robotics and Automation Letters and has been presented at IEEE International Conference on Robotics and Automation 2022, while the paper related to Chapter \ref{chap:eds} on efficient exploration and scene description has been accepted at IEEE International Conference on Robotics and Automation 2023. We aim to follow the track started with the research presented in the previous chapters and hope that \gls{eai} researchers will find it useful for their future work.
\RemoveLabels 

\setstretch{\dnormalspacing}

\pagestyle{plain}
\begin{spacing}{\dcompressedspacing}
    \label{publications}
    \bibliographystyle{apalike}
    \cleardoublepage
    \addcontentsline{toc}{chapter}{Bibliography}
    \bibliography{bibliography}

\begin{thebibliography}{}

\bibitem[Achiam and Sastry, 2017]{achiam2017surprise}
Achiam, J. and Sastry, S. (2017).
\newblock Surprise-based intrinsic motivation for deep reinforcement learning.
\newblock In {\em NeurIPS Workshops}.

\bibitem[Agrawal et~al., 2015]{agrawal2015learning}
Agrawal, P., Carreira, J., and Malik, J. (2015).
\newblock Learning to see by moving.
\newblock In {\em Proceedings of the IEEE/CVF International Conference on Computer Vision}.

\bibitem[Ahn et~al., 2022]{ahn2022can}
Ahn, M., Brohan, A., Brown, N., Chebotar, Y., Cortes, O., David, B., Finn, C., Gopalakrishnan, K., Hausman, K., Herzog, A., et~al. (2022).
\newblock Do as i can, not as i say: Grounding language in robotic affordances.
\newblock In {\em Proceedings of the Conference on Robot Learning}.

\bibitem[Allegretti et~al., 2019]{allegretti2019optimized}
Allegretti, S., Bolelli, F., and Grana, C. (2019).
\newblock Optimized block-based algorithms to label connected components on gpus.
\newblock {\em IEEE Transactions on Parallel and Distributed Systems}.

\bibitem[Anderson et~al., 2018a]{anderson2018evaluation}
Anderson, P., Chang, A., Chaplot, D.~S., Dosovitskiy, A., Gupta, S., Koltun, V., Kosecka, J., Malik, J., Mottaghi, R., Savva, M., et~al. (2018a).
\newblock On evaluation of embodied navigation agents.
\newblock {\em arXiv preprint arXiv:1807.06757}.

\bibitem[Anderson et~al., 2016]{spice2016}
Anderson, P., Fernando, B., Johnson, M., and Gould, S. (2016).
\newblock {SPICE: Semantic Propositional Image Caption Evaluation}.
\newblock In {\em Proceedings of the European Conference on Computer Vision}.

\bibitem[Anderson et~al., 2018b]{anderson2018bottom}
Anderson, P., He, X., Buehler, C., Teney, D., Johnson, M., Gould, S., and Zhang, L. (2018b).
\newblock {Bottom-up and top-down attention for image captioning and visual question answering}.
\newblock In {\em Proceedings of the IEEE/CVF Conference on Computer Vision and Pattern Recognition}.

\bibitem[Anderson et~al., 2021]{anderson2021sim}
Anderson, P., Shrivastava, A., Truong, J., Majumdar, A., Parikh, D., Batra, D., and Lee, S. (2021).
\newblock {Sim-to-Real Transfer for Vision-and-Language Navigation}.
\newblock In {\em Proceedings of the Conference on Robot Learning}.

\bibitem[Anderson et~al., 2018c]{anderson2018vision}
Anderson, P., Wu, Q., Teney, D., Bruce, J., Johnson, M., S{\"u}nderhauf, N., Reid, I., Gould, S., and van~den Hengel, A. (2018c).
\newblock Vision-and-language navigation: Interpreting visually-grounded navigation instructions in real environments.
\newblock In {\em Proceedings of the IEEE/CVF Conference on Computer Vision and Pattern Recognition}.

\bibitem[Anjomshoae et~al., 2019]{anjomshoae2019explainable}
Anjomshoae, S., Najjar, A., Calvaresi, D., and Fr{\"a}mling, K. (2019).
\newblock {Explainable agents and robots: Results from a systematic literature review}.
\newblock In {\em Proceedings of the International Conference on Autonomous Agents and Multiagent Systems}.

\bibitem[Armeni et~al., 2019]{armeni20193d}
Armeni, I., He, Z.-Y., Gwak, J., Zamir, A.~R., Fischer, M., Malik, J., and Savarese, S. (2019).
\newblock {3D Scene Graph: A structure for unified semantics, 3D space, and camera}.
\newblock In {\em Proceedings of the IEEE/CVF Conference on Computer Vision and Pattern Recognition}.

\bibitem[Banerjee and Lavie, 2005]{banerjee2005meteor}
Banerjee, S. and Lavie, A. (2005).
\newblock {METEOR: An automatic metric for MT evaluation with improved correlation with human judgments}.
\newblock In {\em Proceedings of the Annual Meeting of the Association for Computational Linguistics Workshops}.

\bibitem[Barraco et~al., 2022]{barraco2022camel}
Barraco, M., Stefanini, M., Cornia, M., Cascianelli, S., Baraldi, L., and Cucchiara, R. (2022).
\newblock {CaMEL: Mean Teacher Learning for Image Captioning}.
\newblock In {\em Proceedings of the International Conference on Pattern Recognition}.

\bibitem[Beattie et~al., 2016]{beattie2016deepmind}
Beattie, C., Leibo, J.~Z., Teplyashin, D., Ward, T., Wainwright, M., K{\"u}ttler, H., Lefrancq, A., Green, S., Vald{\'e}s, V., Sadik, A., et~al. (2016).
\newblock {DeepMind Lab}.
\newblock {\em arXiv preprint arXiv:1612.03801}.

\bibitem[Bellemare et~al., 2016]{bellemare2016unifying}
Bellemare, M., Srinivasan, S., Ostrovski, G., Schaul, T., Saxton, D., and Munos, R. (2016).
\newblock Unifying count-based exploration and intrinsic motivation.
\newblock In {\em Advances in Neural Information Processing Systems}.

\bibitem[Bellemare et~al., 2013]{bellemare2013arcade}
Bellemare, M.~G., Naddaf, Y., Veness, J., and Bowling, M. (2013).
\newblock The arcade learning environment: An evaluation platform for general agents.
\newblock {\em Journal of Artificial Intelligence Research}.

\bibitem[Bigazzi et~al., 2023]{bigazzi2023embodied}
Bigazzi, R., Cornia, M., Cascianelli, S., Baraldi, L., and Cucchiara, R. (2023).
\newblock {Embodied Agents for Efficient Exploration and Smart Scene Description}.
\newblock In {\em Proceedings of the IEEE International Conference on Robotics and Automation}.

\bibitem[Bigazzi et~al., 2022a]{bigazzi2022impact}
Bigazzi, R., Landi, F., Cascianelli, S., Baraldi, L., Cornia, M., and Cucchiara, R. (2022a).
\newblock {Focus on Impact: Indoor Exploration with Intrinsic Motivation}.
\newblock {\em IEEE Robotics and Automation Letters}.

\bibitem[Bigazzi et~al., 2022b]{bigazzi2022embodied}
Bigazzi, R., Landi, F., Cascianelli, S., Cornia, M., Baraldi, L., and Cucchiara, R. (2022b).
\newblock Embodied navigation at the art gallery.
\newblock In {\em Proceedings of the International Conference on Image Analysis and Processing}.

\bibitem[Bigazzi et~al., 2020]{bigazzi2020explore}
Bigazzi, R., Landi, F., Cornia, M., Cascianelli, S., Baraldi, L., and Cucchiara, R. (2020).
\newblock {Explore and Explain: Self-supervised Navigation and Recounting}.
\newblock In {\em Proceedings of the International Conference on Pattern Recognition}.

\bibitem[Bigazzi et~al., 2021]{bigazzi2021out}
Bigazzi, R., Landi, F., Cornia, M., Cascianelli, S., Baraldi, L., and Cucchiara, R. (2021).
\newblock {Out of the Box: Embodied Navigation in the Real World}.
\newblock In {\em Proceedings of the International Conference on Computer Analysis of Images and Patterns}.

\bibitem[Bircher et~al., 2016]{bircher2016receding}
Bircher, A., Kamel, M., Alexis, K., Oleynikova, H., and Siegwart, R. (2016).
\newblock {Receding Horizon ``Next–Best–View'' Planner for 3D Exploration}.
\newblock In {\em Proceedings of the IEEE International Conference on Robotics and Automation}.

\bibitem[Biswas, 2019]{biswas2019quest}
Biswas, J. (2019).
\newblock {The Quest For" Always-On" Autonomous Mobile Robots.}
\newblock In {\em Proceedings of the International Joint Conferences on Artificial Intelligence}.

\bibitem[Bolelli et~al., 2019]{bolelli2019spaghetti}
Bolelli, F., Allegretti, S., Baraldi, L., and Grana, C. (2019).
\newblock Spaghetti labeling: Directed acyclic graphs for block-based connected components labeling.
\newblock {\em IEEE Transactions on Image Processing}.

\bibitem[Brockman et~al., 2016]{brockman2016openai}
Brockman, G., Cheung, V., Pettersson, L., Schneider, J., Schulman, J., Tang, J., and Zaremba, W. (2016).
\newblock {OpenAI Gym}.
\newblock {\em arXiv preprint arXiv:1606.01540}.

\bibitem[Burda et~al., 2019]{burda2018large}
Burda, Y., Edwards, H., Pathak, D., Storkey, A., Darrell, T., and Efros, A.~A. (2019).
\newblock Large-scale study of curiosity-driven learning.
\newblock In {\em Proceedings of the International Conference on Learning Representations}.

\bibitem[Burda et~al., 2018]{burda2018exploration}
Burda, Y., Edwards, H., Storkey, A., and Klimov, O. (2018).
\newblock Exploration by random network distillation.
\newblock In {\em Proceedings of the International Conference on Learning Representations}.

\bibitem[Cartillier et~al., 2020]{cartillier2020semantic}
Cartillier, V., Ren, Z., Jain, N., Lee, S., Essa, I., and Batra, D. (2020).
\newblock {Semantic MapNet: Building Allocentric Semantic Maps and Representations from Egocentric Views}.
\newblock {\em arXiv preprint arXiv:2010.01191}.

\bibitem[Cascianelli et~al., 2016]{cascianelli2016robust}
Cascianelli, S., Costante, G., Bellocchio, E., Valigi, P., Fravolini, M.~L., and Ciarfuglia, T.~A. (2016).
\newblock A robust semi-semantic approach for visual localization in urban environment.
\newblock In {\em Proceedings of the IEEE International Smart Cities Conference}.

\bibitem[Chang et~al., 2017]{chang2017matterport3d}
Chang, A., Dai, A., Funkhouser, T., Halber, M., Niessner, M., Savva, M., Song, S., Zeng, A., and Zhang, Y. (2017).
\newblock {Matterport3D: Learning from RGB-D Data in Indoor Environments}.
\newblock In {\em Proceedings of the International Conference on 3D Vision}.

\bibitem[Changpinyo et~al., 2021]{changpinyo2021conceptual}
Changpinyo, S., Sharma, P., Ding, N., and Soricut, R. (2021).
\newblock {Conceptual 12M: Pushing Web-Scale Image-Text Pre-Training To Recognize Long-Tail Visual Concepts}.
\newblock In {\em Proceedings of the IEEE/CVF Conference on Computer Vision and Pattern Recognition}.

\bibitem[Chaplot et~al., 2019a]{chaplot2019learning}
Chaplot, D.~S., Gandhi, D., Gupta, S., Gupta, A., and Salakhutdinov, R. (2019a).
\newblock {Learning To Explore Using Active Neural SLAM}.
\newblock In {\em Proceedings of the International Conference on Learning Representations}.

\bibitem[Chaplot et~al., 2020]{chaplot2020object}
Chaplot, D.~S., Gandhi, D.~P., Gupta, A., and Salakhutdinov, R.~R. (2020).
\newblock {Object Goal Navigation using Goal-Oriented Semantic Exploration}.
\newblock In {\em Advances in Neural Information Processing Systems}.

\bibitem[Chaplot et~al., 2019b]{chaplot2019embodied}
Chaplot, D.~S., Lee, L., Salakhutdinov, R., Parikh, D., and Batra, D. (2019b).
\newblock {Embodied Multimodal Multitask Learning}.
\newblock {\em Proceedings of the International Joint Conferences on Artificial Intelligence}.

\bibitem[Chen et~al., 2021]{chen2021history}
Chen, S., Guhur, P.-L., Schmid, C., and Laptev, I. (2021).
\newblock History aware multimodal transformer for vision-and-language navigation.
\newblock In {\em Advances in Neural Information Processing Systems}.

\bibitem[Chen et~al., 2022a]{chen2022learning}
Chen, S., Guhur, P.-L., Tapaswi, M., Schmid, C., and Laptev, I. (2022a).
\newblock Learning from unlabeled 3d environments for vision-and-language navigation.
\newblock In {\em Proceedings of the European Conference on Computer Vision}.

\bibitem[Chen et~al., 2022b]{chen2022think}
Chen, S., Guhur, P.-L., Tapaswi, M., Schmid, C., and Laptev, I. (2022b).
\newblock Think global, act local: Dual-scale graph transformer for vision-and-language navigation.
\newblock In {\em Proceedings of the IEEE/CVF Conference on Computer Vision and Pattern Recognition}.

\bibitem[Chen et~al., 2019]{chen2019learning}
Chen, T., Gupta, S., and Gupta, A. (2019).
\newblock {Learning Exploration Policies for Navigation}.
\newblock In {\em Proceedings of the International Conference on Learning Representations}.

\bibitem[Choi et~al., 2015]{choi2015robust}
Choi, S., Zhou, Q.-Y., and Koltun, V. (2015).
\newblock Robust reconstruction of indoor scenes.
\newblock In {\em Proceedings of the IEEE/CVF Conference on Computer Vision and Pattern Recognition}.

\bibitem[Cornia et~al., 2020a]{cornia2019smart}
Cornia, M., Baraldi, L., and Cucchiara, R. (2020a).
\newblock {SMArT: Training Shallow Memory-aware Transformers for Robotic Explainability}.
\newblock In {\em Proceedings of the IEEE International Conference on Robotics and Automation}.

\bibitem[Cornia et~al., 2022]{cornia2021universal}
Cornia, M., Baraldi, L., Fiameni, G., and Cucchiara, R. (2022).
\newblock {Universal Captioner: Inducing Content-Style Separation in Vision-and-Language Model Training}.
\newblock {\em arXiv preprint arXiv:2111.12727}.

\bibitem[Cornia et~al., 2020b]{cornia2020meshed}
Cornia, M., Stefanini, M., Baraldi, L., and Cucchiara, R. (2020b).
\newblock {Meshed-Memory Transformer for Image Captioning}.
\newblock In {\em Proceedings of the IEEE/CVF Conference on Computer Vision and Pattern Recognition}.

\bibitem[Da~Silva et~al., 2018]{da2018autonomously}
Da~Silva, F.~L., Taylor, M.~E., and Costa, A. H.~R. (2018).
\newblock {Autonomously Reusing Knowledge in Multiagent Reinforcement Learning.}
\newblock In {\em Proceedings of the International Joint Conferences on Artificial Intelligence}.

\bibitem[Deitke et~al., 2020]{deitke2020robothor}
Deitke, M., Han, W., Herrasti, A., Kembhavi, A., Kolve, E., Mottaghi, R., Salvador, J., Schwenk, D., VanderBilt, E., Wallingford, M., et~al. (2020).
\newblock {RoboTHOR: An Open Simulation-to-Real Embodied AI Platform}.
\newblock In {\em Proceedings of the IEEE/CVF Conference on Computer Vision and Pattern Recognition}.

\bibitem[Deitke et~al., 2022]{deitke2022procthor}
Deitke, M., VanderBilt, E., Herrasti, A., Weihs, L., Salvador, J., Ehsani, K., Han, W., Kolve, E., Farhadi, A., Kembhavi, A., et~al. (2022).
\newblock Procthor: Large-scale embodied ai using procedural generation.
\newblock {\em Advances in Neural Information Processing Systems}.

\bibitem[Dinh et~al., 2016]{dinh2016density}
Dinh, L., Sohl-Dickstein, J., and Bengio, S. (2016).
\newblock Density estimation using real nvp.
\newblock {\em Proceedings of the International Conference on Learning Representations}.

\bibitem[Esser et~al., 2021]{esser2021taming}
Esser, P., Rombach, R., and Ommer, B. (2021).
\newblock Taming transformers for high-resolution image synthesis.
\newblock In {\em Proceedings of the IEEE/CVF Conference on Computer Vision and Pattern Recognition}.

\bibitem[Georgakis et~al., 2022]{georgakis2022uncertainty}
Georgakis, G., Bucher, B., Arapin, A., Schmeckpeper, K., Matni, N., and Daniilidis, K. (2022).
\newblock Uncertainty-driven planner for exploration and navigation.
\newblock In {\em Proceedings of the IEEE International Conference on Robotics and Automation}.

\bibitem[Gervet et~al., 2022]{gervet2022navigating}
Gervet, T., Chintala, S., Batra, D., Malik, J., and Chaplot, D.~S. (2022).
\newblock Navigating to objects in the real world.
\newblock {\em arXiv preprint arXiv:2212.00922}.

\bibitem[Gonz{\'a}lez-Banos and Latombe, 2002]{gonzalez2002navigation}
Gonz{\'a}lez-Banos, H.~H. and Latombe, J.-C. (2002).
\newblock {Navigation Strategies for Exploring Indoor Environments}.
\newblock {\em The International Journal of Robotics Research}.

\bibitem[Goodfellow et~al., 2014]{goodfellow2014generative}
Goodfellow, I., Pouget-Abadie, J., Mirza, M., Xu, B., Warde-Farley, D., Ozair, S., Courville, A., and Bengio, Y. (2014).
\newblock Generative adversarial nets.
\newblock {\em Advances in Neural Information Processing Systems}.

\bibitem[Grana et~al., 2010]{grana2010optimized}
Grana, C., Borghesani, D., and Cucchiara, R. (2010).
\newblock Optimized block-based connected components labeling with decision trees.
\newblock {\em IEEE Transactions on Image Processing}.

\bibitem[Guhur et~al., 2021]{guhur2021airbert}
Guhur, P.-L., Tapaswi, M., Chen, S., Laptev, I., and Schmid, C. (2021).
\newblock Airbert: In-domain pretraining for vision-and-language navigation.
\newblock In {\em Proceedings of the IEEE/CVF International Conference on Computer Vision}.

\bibitem[Gupta et~al., 2017]{gupta2017cognitive}
Gupta, S., Davidson, J., Levine, S., Sukthankar, R., and Malik, J. (2017).
\newblock Cognitive mapping and planning for visual navigation.
\newblock In {\em Proceedings of the IEEE/CVF Conference on Computer Vision and Pattern Recognition}.

\bibitem[He et~al., 2016]{he2016deep}
He, K., Zhang, X., Ren, S., and Sun, J. (2016).
\newblock Deep residual learning for image recognition.
\newblock In {\em Proceedings of the IEEE/CVF Conference on Computer Vision and Pattern Recognition}.

\bibitem[Hessel et~al., 2021]{hessel2021clipscore}
Hessel, J., Holtzman, A., Forbes, M., Bras, R.~L., and Choi, Y. (2021).
\newblock {CLIPScore: A Reference-free Evaluation Metric for Image Captioning}.
\newblock In {\em Proceedings of the Conference on Empirical Methods in Natural Language Processing}.

\bibitem[Ho et~al., 2019]{ho2019flow++}
Ho, J., Chen, X., Srinivas, A., Duan, Y., and Abbeel, P. (2019).
\newblock Flow++: Improving flow-based generative models with variational dequantization and architecture design.
\newblock In {\em Proceedings of the International Conference on Machine Learning}.

\bibitem[Holz et~al., 2010]{holz2010evaluating}
Holz, D., Basilico, N., Amigoni, F., and Behnke, S. (2010).
\newblock {Evaluating the Efficiency of Frontier-based Exploration Strategies}.
\newblock In {\em ISR and ROBOTIK}.

\bibitem[Horgan et~al., 2018]{horgan2018distributed}
Horgan, D., Quan, J., Budden, D., Barth-Maron, G., Hessel, M., Van~Hasselt, H., and Silver, D. (2018).
\newblock Distributed prioritized experience replay.
\newblock In {\em Proceedings of the International Conference on Learning Representations}.

\bibitem[Houthooft et~al., 2016]{houthooft2016vime}
Houthooft, R., Chen, X., Duan, Y., Schulman, J., De~Turck, F., and Abbeel, P. (2016).
\newblock {VIME: Variational Information Maximizing Exploration}.
\newblock In {\em Advances in Neural Information Processing Systems}.

\bibitem[Huang et~al., 2023]{huang2022visual}
Huang, C., Mees, O., Zeng, A., and Burgard, W. (2023).
\newblock Visual language maps for robot navigation.
\newblock In {\em Proceedings of the IEEE International Conference on Robotics and Automation}.

\bibitem[Huang et~al., 2019]{huang2019attention}
Huang, L., Wang, W., Chen, J., and Wei, X.-Y. (2019).
\newblock {Attention on Attention for Image Captioning}.
\newblock In {\em Proceedings of the IEEE/CVF International Conference on Computer Vision}.

\bibitem[Huang et~al., 2022]{huang2022inner}
Huang, W., Xia, F., Xiao, T., Chan, H., Liang, J., Florence, P., Zeng, A., Tompson, J., Mordatch, I., Chebotar, Y., et~al. (2022).
\newblock Inner monologue: Embodied reasoning through planning with language models.
\newblock In {\em Proceedings of the Conference on Robot Learning}.

\bibitem[Irshad et~al., 2021]{irshad2021hierarchical}
Irshad, M.~Z., Ma, C.-Y., and Kira, Z. (2021).
\newblock {Hierarchical Cross-Modal Agent for Robotics Vision-and-Language Navigation}.
\newblock In {\em Proceedings of the IEEE International Conference on Robotics and Automation}.

\bibitem[Jain et~al., 2019]{jain2019stay}
Jain, V., Magalhaes, G., Ku, A., Vaswani, A., Ie, E., and Baldridge, J. (2019).
\newblock Stay on the path: Instruction fidelity in vision-and-language navigation.
\newblock {\em arXiv preprint arXiv:1905.12255}.

\bibitem[Jonker and Volgenant, 1987]{jonker1987shortest}
Jonker, R. and Volgenant, A. (1987).
\newblock A shortest augmenting path algorithm for dense and sparse linear assignment problems.
\newblock {\em Computing}.

\bibitem[Kadian et~al., 2020]{kadian2020sim2real}
Kadian, A., Truong, J., Gokaslan, A., Clegg, A., Wijmans, E., Lee, S., Savva, M., Chernova, S., and Batra, D. (2020).
\newblock {Sim2Real Predictivity: Does evaluation in simulation predict real-world performance?}
\newblock {\em IEEE Robotics and Automation Letters}.

\bibitem[Karkus et~al., 2021]{karkus2021differentiable}
Karkus, P., Cai, S., and Hsu, D. (2021).
\newblock {Differentiable SLAM-net: Learning Particle SLAM for Visual Navigation}.
\newblock In {\em Proceedings of the IEEE/CVF Conference on Computer Vision and Pattern Recognition}.

\bibitem[Karpathy and Fei-Fei, 2015]{karpathy2015deep}
Karpathy, A. and Fei-Fei, L. (2015).
\newblock Deep visual-semantic alignments for generating image descriptions.
\newblock In {\em Proceedings of the IEEE/CVF Conference on Computer Vision and Pattern Recognition}.

\bibitem[Karras et~al., 2019]{karras2019style}
Karras, T., Laine, S., and Aila, T. (2019).
\newblock A style-based generator architecture for generative adversarial networks.
\newblock In {\em Proceedings of the IEEE/CVF Conference on Computer Vision and Pattern Recognition}.

\bibitem[Kempka et~al., 2016]{kempka2016vizdoom}
Kempka, M., Wydmuch, M., Runc, G., Toczek, J., and Ja{\'s}kowski, W. (2016).
\newblock {ViZDoom: A doom-based ai research platform for visual reinforcement learning}.
\newblock In {\em Proceedings of the IEEE Conference on Computational Intelligence and Games}.

\bibitem[Kingma and Ba, 2015]{kingma2015adam}
Kingma, D. and Ba, J. (2015).
\newblock Adam: a method for stochastic optimization.
\newblock In {\em Proceedings of the International Conference on Learning Representations}.

\bibitem[Kingma and Welling, 2013]{kingma2013auto}
Kingma, D.~P. and Welling, M. (2013).
\newblock Auto-encoding variational bayes.
\newblock In {\em Proceedings of the International Conference on Learning Representations}.

\bibitem[Klyubin et~al., 2005]{klyubin2005empowerment}
Klyubin, A.~S., Polani, D., and Nehaniv, C.~L. (2005).
\newblock Empowerment: A universal agent-centric measure of control.
\newblock In {\em CEC}.

\bibitem[Krantz et~al., 2020]{krantz2020beyond}
Krantz, J., Wijmans, E., Majumdar, A., Batra, D., and Lee, S. (2020).
\newblock {Beyond the Nav-Graph: Vision-and-Language Navigation in Continuous Environments}.
\newblock In {\em Proceedings of the European Conference on Computer Vision}.

\bibitem[Ku et~al., 2020]{ku2020room}
Ku, A., Anderson, P., Patel, R., Ie, E., and Baldridge, J. (2020).
\newblock Room-across-room: Multilingual vision-and-language navigation with dense spatiotemporal grounding.
\newblock In {\em Proceedings of the Conference on Empirical Methods in Natural Language Processing}.

\bibitem[Kuhn, 1955]{kuhn1955hungarian}
Kuhn, H.~W. (1955).
\newblock {The Hungarian method for the assignment problem}.
\newblock {\em Naval Research Logistics Quarterly}.

\bibitem[Landi et~al., 2021a]{landi2021multimodal}
Landi, F., Baraldi, L., Cornia, M., Corsini, M., and Cucchiara, R. (2021a).
\newblock {Multimodal Attention Networks for Low-Level Vision-and-Language Navigation}.
\newblock {\em Computer Vision and Image Understanding}.

\bibitem[Landi et~al., 2021b]{landi2021working}
Landi, F., Baraldi, L., Cornia, M., and Cucchiara, R. (2021b).
\newblock {Working Memory Connections for LSTM}.
\newblock {\em Neural Networks}.

\bibitem[Landi et~al., 2019]{landi2019embodied}
Landi, F., Baraldi, L., Corsini, M., and Cucchiara, R. (2019).
\newblock Embodied vision-and-language navigation with dynamic convolutional filters.
\newblock {\em Proceedings of the British Machine Vision Conference}.

\bibitem[Landi et~al., 2022]{landi2022spot}
Landi, F., Bigazzi, R., Cornia, M., Cascianelli, S., Baraldi, L., and Cucchiara, R. (2022).
\newblock Spot the difference: A novel task for embodied agents in changing environments.
\newblock In {\em Proceedings of the International Conference on Pattern Recognition}.

\bibitem[Li et~al., 2021]{li2021igibson}
Li, C., Xia, F., Mart{\'\i}n-Mart{\'\i}n, R., Lingelbach, M., Srivastava, S., Shen, B., Vainio, K., Gokmen, C., Dharan, G., Jain, T., et~al. (2021).
\newblock igibson 2.0: Object-centric simulation for robot learning of everyday household tasks.
\newblock In {\em Proceedings of the Conference on Robot Learning}.

\bibitem[Li et~al., 2020]{li2020oscar}
Li, X., Yin, X., Li, C., Zhang, P., Hu, X., Zhang, L., Wang, L., Hu, H., Dong, L., Wei, F., et~al. (2020).
\newblock {Oscar: Object-semantics aligned pre-training for vision-language tasks}.
\newblock In {\em Proceedings of the European Conference on Computer Vision}.

\bibitem[Liang et~al., 2023]{liang2022code}
Liang, J., Huang, W., Xia, F., Xu, P., Hausman, K., Ichter, B., Florence, P., and Zeng, A. (2023).
\newblock Code as policies: Language model programs for embodied control.
\newblock In {\em Proceedings of the IEEE International Conference on Robotics and Automation}.

\bibitem[Lin, 2004]{lin2004rouge}
Lin, C.-Y. (2004).
\newblock Rouge: A package for automatic evaluation of summaries.
\newblock In {\em Proceedings of the Annual Meeting of the Association for Computational Linguistics Workshops}.

\bibitem[Lin et~al., 2014]{lin2014microsoft}
Lin, T.-Y., Maire, M., Belongie, S., Hays, J., Perona, P., Ramanan, D., Doll{\'a}r, P., and Zitnick, C.~L. (2014).
\newblock {Microsoft COCO: Common Objects in Context}.
\newblock In {\em Proceedings of the European Conference on Computer Vision}.

\bibitem[Liu et~al., 2021]{liu2021cptr}
Liu, W., Chen, S., Guo, L., Zhu, X., and Liu, J. (2021).
\newblock {CPTR: Full Transformer Network for Image Captioning}.
\newblock {\em arXiv preprint arXiv:2101.10804}.

\bibitem[Locobot, 2020]{locobot}
Locobot (2020).
\newblock {LoCoBot: An Open Source Low Cost Robot}.
\newblock \url{https://locobot-website.netlify.com}.

\bibitem[Lu et~al., 2017]{lu2017knowing}
Lu, J., Xiong, C., Parikh, D., and Socher, R. (2017).
\newblock Knowing when to look: Adaptive attention via a visual sentinel for image captioning.
\newblock In {\em Proceedings of the IEEE/CVF Conference on Computer Vision and Pattern Recognition}.

\bibitem[Lu et~al., 2018]{lu2018neural}
Lu, J., Yang, J., Batra, D., and Parikh, D. (2018).
\newblock {Neural Baby Talk}.
\newblock In {\em Proceedings of the IEEE/CVF Conference on Computer Vision and Pattern Recognition}.

\bibitem[Luperto et~al., 2020]{luperto2020robot}
Luperto, M., Antonazzi, M., Amigoni, F., and Borghese, N.~A. (2020).
\newblock Robot exploration of indoor environments using incomplete and inaccurate prior knowledge.
\newblock {\em Robotics and Autonomous Systems}.

\bibitem[Mac et~al., 2016]{mac2016heuristic}
Mac, T.~T., Copot, C., Tran, D.~T., and De~Keyser, R. (2016).
\newblock Heuristic approaches in robot path planning: A survey.
\newblock {\em Robotics and Autonomous Systems}.

\bibitem[Machado et~al., 2018]{machado2018revisiting}
Machado, M.~C., Bellemare, M.~G., Talvitie, E., Veness, J., Hausknecht, M., and Bowling, M. (2018).
\newblock Revisiting the arcade learning environment: Evaluation protocols and open problems for general agents.
\newblock {\em Journal of Artificial Intelligence Research}.

\bibitem[Mayo et~al., 2021]{mayo2021visual}
Mayo, B., Hazan, T., and Tal, A. (2021).
\newblock Visual navigation with spatial attention.
\newblock In {\em Proceedings of the IEEE/CVF Conference on Computer Vision and Pattern Recognition}.

\bibitem[Mohamed and Rezende, 2015]{mohamed2015variational}
Mohamed, S. and Rezende, D.~J. (2015).
\newblock Variational information maximisation for intrinsically motivated reinforcement learning.
\newblock In {\em Advances in Neural Information Processing Systems}.

\bibitem[Morad et~al., 2021]{morad2021embodied}
Morad, S.~D., Mecca, R., Poudel, R.~P., Liwicki, S., and Cipolla, R. (2021).
\newblock Embodied visual navigation with automatic curriculum learning in real environments.
\newblock {\em IEEE Robotics and Automation Letters}.

\bibitem[Murali et~al., 2019]{murali2019pyrobot}
Murali, A., Chen, T., Alwala, K.~V., Gandhi, D., Pinto, L., Gupta, S., and Gupta, A. (2019).
\newblock {PyRobot: An Open-source Robotics Framework for Research and Benchmarking}.
\newblock {\em arXiv preprint arXiv:1906.08236}.

\bibitem[Nardi and Stachniss, 2020]{nardi2020long}
Nardi, L. and Stachniss, C. (2020).
\newblock Long-term robot navigation in indoor environments estimating patterns in traversability changes.
\newblock In {\em Proceedings of the IEEE International Conference on Robotics and Automation}.

\bibitem[Niroui et~al., 2019]{niroui2019deep}
Niroui, F., Zhang, K., Kashino, Z., and Nejat, G. (2019).
\newblock {Deep Reinforcement Learning Robot for Search and Rescue Applications: Exploration in Unknown Cluttered Environments}.
\newblock {\em IEEE Robotics and Automation Letters}.

\bibitem[Oord et~al., 2016]{oord2016conditional}
Oord, A. v.~d., Kalchbrenner, N., Vinyals, O., Espeholt, L., Graves, A., and Kavukcuoglu, K. (2016).
\newblock Conditional image generation with pixelcnn decoders.
\newblock In {\em Advances in Neural Information Processing Systems}.

\bibitem[Ordonez et~al., 2011]{ordonez2011im2text}
Ordonez, V., Kulkarni, G., and Berg, T. (2011).
\newblock {Im2Text: Describing Images Using 1 Million Captioned Photographs}.
\newblock In {\em Advances in Neural Information Processing Systems}.

\bibitem[O{\ss}wald et~al., 2016]{osswald2016speeding}
O{\ss}wald, S., Bennewitz, M., Burgard, W., and Stachniss, C. (2016).
\newblock {Speeding-Up Robot Exploration by Exploiting Background Information}.
\newblock {\em IEEE Robotics and Automation Letters}.

\bibitem[Ostrovski et~al., 2017]{ostrovski2017count}
Ostrovski, G., Bellemare, M.~G., Oord, A., and Munos, R. (2017).
\newblock Count-based exploration with neural density models.
\newblock In {\em Proceedings of the International Conference on Machine Learning}.

\bibitem[Oudeyer and Kaplan, 2009]{oudeyer2009intrinsic}
Oudeyer, P.-Y. and Kaplan, F. (2009).
\newblock {What is intrinsic motivation? A typology of computational approaches}.
\newblock {\em Frontiers in Neurorobotics}.

\bibitem[Papineni et~al., 2002]{papineni2002bleu}
Papineni, K., Roukos, S., Ward, T., and Zhu, W.-J. (2002).
\newblock {BLEU: a method for automatic evaluation of machine translation}.
\newblock In {\em Proceedings of the Annual Meeting of the Association for Computational Linguistics}.

\bibitem[Parmar et~al., 2018]{parmar2018image}
Parmar, N., Vaswani, A., Uszkoreit, J., Kaiser, L., Shazeer, N., Ku, A., and Tran, D. (2018).
\newblock Image transformer.
\newblock In {\em Proceedings of the International Conference on Machine Learning}.

\bibitem[Partsey et~al., 2022]{partsey2022mapping}
Partsey, R., Wijmans, E., Yokoyama, N., Dobosevych, O., Batra, D., and Maksymets, O. (2022).
\newblock Is mapping necessary for realistic pointgoal navigation?
\newblock In {\em Proceedings of the IEEE/CVF Conference on Computer Vision and Pattern Recognition}.

\bibitem[Pathak et~al., 2017]{pathak2017curiosity}
Pathak, D., Agrawal, P., Efros, A.~A., and Darrell, T. (2017).
\newblock Curiosity-driven exploration by self-supervised prediction.
\newblock In {\em Proceedings of the International Conference on Machine Learning}.

\bibitem[Pathak et~al., 2019]{pathak2019self}
Pathak, D., Gandhi, D., and Gupta, A. (2019).
\newblock Self-supervised exploration via disagreement.
\newblock In {\em Proceedings of the International Conference on Machine Learning}.

\bibitem[Pennington et~al., 2014]{pennington2014glove}
Pennington, J., Socher, R., and Manning, C.~D. (2014).
\newblock {GloVe: Global Vectors for Word Representation}.
\newblock In {\em Proceedings of the Conference on Empirical Methods in Natural Language Processing}.

\bibitem[Poppi et~al., 2023]{poppi2023towards}
Poppi, S., Bigazzi, R., Rawal, N., Cornia, M., Cascianelli, S., Baraldi, L., and Cucchiara, R. (2023).
\newblock {Towards Explainable Embodied Navigation and Recounting}.
\newblock {\em Under Review}.

\bibitem[Qi et~al., 2020]{qi2020reverie}
Qi, Y., Wu, Q., Anderson, P., Wang, X., Wang, W.~Y., Shen, C., and Hengel, A. v.~d. (2020).
\newblock Reverie: Remote embodied visual referring expression in real indoor environments.
\newblock In {\em Proceedings of the IEEE/CVF Conference on Computer Vision and Pattern Recognition}.

\bibitem[Radford et~al., 2021]{radford2021learning}
Radford, A., Kim, J.~W., Hallacy, C., Ramesh, A., Goh, G., Agarwal, S., Sastry, G., Askell, A., Mishkin, P., Clark, J., Krueger, G., and Sutskever, I. (2021).
\newblock {Learning Transferable Visual Models From Natural Language Supervision}.
\newblock In {\em Proceedings of the International Conference on Machine Learning}.

\bibitem[Raileanu and Rockt{\"a}schel, 2021]{raileanu2020ride}
Raileanu, R. and Rockt{\"a}schel, T. (2021).
\newblock {RIDE: Rewarding impact-driven exploration for procedurally-generated environments}.
\newblock In {\em Proceedings of the International Conference on Learning Representations}.

\bibitem[Ramakrishnan et~al., 2020]{ramakrishnan2020occupancy}
Ramakrishnan, S.~K., Al-Halah, Z., and Grauman, K. (2020).
\newblock {Occupancy Anticipation for Efficient Exploration and Navigation}.
\newblock In {\em Proceedings of the European Conference on Computer Vision}.

\bibitem[Ramakrishnan et~al., 2022]{ramakrishnan2022poni}
Ramakrishnan, S.~K., Chaplot, D.~S., Al-Halah, Z., Malik, J., and Grauman, K. (2022).
\newblock {PONI: Potential Functions for ObjectGoal Navigation with Interaction-free Learning}.
\newblock In {\em Proceedings of the IEEE/CVF Conference on Computer Vision and Pattern Recognition}.

\bibitem[Ramakrishnan et~al., 2021a]{ramakrishnan2021hm3d}
Ramakrishnan, S.~K., Gokaslan, A., Wijmans, E., Maksymets, O., Clegg, A., Turner, J.~M., Undersander, E., Galuba, W., Westbury, A., Chang, A.~X., Savva, M., Zhao, Y., and Batra, D. (2021a).
\newblock Habitat-matterport 3d dataset ({HM}3d): 1000 large-scale 3d environments for embodied {AI}.
\newblock In {\em Advances in Neural Information Processing Systems}.

\bibitem[Ramakrishnan et~al., 2021b]{ramakrishnan2020exploration}
Ramakrishnan, S.~K., Jayaraman, D., and Grauman, K. (2021b).
\newblock {An Exploration of Embodied Visual Exploration}.
\newblock {\em International Journal of Computer Vision}.

\bibitem[Rawal et~al., 2023]{rawal2023aigen}
Rawal, N., Bigazzi, R., Baraldi, L., and Cucchiara, R. (2023).
\newblock {AIGeN: An Adversarial Approach for Instruction Generation in Vision-and-Language Navigation}.
\newblock {\em Under Review}.

\bibitem[Ren et~al., 2017]{ren2017faster}
Ren, S., He, K., Girshick, R., and Sun, J. (2017).
\newblock {Faster R-CNN: towards real-time object detection with region proposal networks}.
\newblock {\em IEEE Transactions on Pattern Analysis and Machine Intelligence}.

\bibitem[Rennie et~al., 2017]{rennie2017self}
Rennie, S.~J., Marcheret, E., Mroueh, Y., Ross, J., and Goel, V. (2017).
\newblock Self-critical sequence training for image captioning.
\newblock In {\em Proceedings of the IEEE/CVF Conference on Computer Vision and Pattern Recognition}.

\bibitem[Ronneberger et~al., 2015]{ronneberger2015u}
Ronneberger, O., Fischer, P., and Brox, T. (2015).
\newblock {U-Net: Convolutional Networks for Biomedical Image Segmentation}.
\newblock In {\em Proceedings of the International Conference on Medical Image Computing and Computer Assisted Intervention}.

\bibitem[Rosano et~al., 2020]{rosano2020embodied}
Rosano, M., Furnari, A., Gulino, L., and Farinella, G.~M. (2020).
\newblock {On Embodied Visual Navigation in Real Environments Through Habitat}.
\newblock In {\em Proceedings of the International Conference on Pattern Recognition}.

\bibitem[Saputra et~al., 2018]{saputra2018visual}
Saputra, M. R.~U., Markham, A., and Trigoni, N. (2018).
\newblock {Visual SLAM and structure from motion in dynamic environments: A survey}.
\newblock {\em ACM Computing Surveys}.

\bibitem[Savinov et~al., 2018]{savinov2018semi}
Savinov, N., Dosovitskiy, A., and Koltun, V. (2018).
\newblock Semi-parametric topological memory for navigation.
\newblock In {\em Proceedings of the International Conference on Learning Representations}.

\bibitem[Savva et~al., 2019]{savva2019habitat}
Savva, M., Kadian, A., Maksymets, O., Zhao, Y., Wijmans, E., Jain, B., Straub, J., Liu, J., Koltun, V., Malik, J., et~al. (2019).
\newblock {Habitat: A Platform for Embodied AI Research}.
\newblock In {\em Proceedings of the IEEE/CVF International Conference on Computer Vision}.

\bibitem[Sax et~al., 2019]{sax2018mid}
Sax, A., Emi, B., Zamir, A.~R., Guibas, L., Savarese, S., and Malik, J. (2019).
\newblock Mid-level visual representations improve generalization and sample efficiency for learning visuomotor policies.
\newblock In {\em Proceedings of the Conference on Robot Learning}.

\bibitem[Schmidhuber, 2010]{schmidhuber2010formal}
Schmidhuber, J. (2010).
\newblock {Formal Theory of Creativity, Fun, and Intrinsic Motivation}.
\newblock {\em IEEE Trans. on Autonomous Mental Development}.

\bibitem[Schulman et~al., 2017]{schulman2017proximal}
Schulman, J., Wolski, F., Dhariwal, P., Radford, A., and Klimov, O. (2017).
\newblock {Proximal Policy Optimization Algorithms}.
\newblock {\em arXiv preprint arXiv:1707.06347}.

\bibitem[Selin et~al., 2019]{selin2019efficient}
Selin, M., Tiger, M., Duberg, D., Heintz, F., and Jensfelt, P. (2019).
\newblock {Efficient Autonomous Exploration Planning of Large-Scale 3-D Environments}.
\newblock {\em IEEE Robotics and Automation Letters}.

\bibitem[Sharma et~al., 2018]{sharma2018conceptual}
Sharma, P., Ding, N., Goodman, S., and Soricut, R. (2018).
\newblock {Conceptual Captions: A Cleaned, Hypernymed, Image Alt-text Dataset For Automatic Image Captioning}.
\newblock In {\em Proceedings of the Annual Meeting of the Association for Computational Linguistics}.

\bibitem[Shridhar et~al., 2020]{ALFRED20}
Shridhar, M., Thomason, J., Gordon, D., Bisk, Y., Han, W., Mottaghi, R., Zettlemoyer, L., and Fox, D. (2020).
\newblock {ALFRED: A Benchmark for Interpreting Grounded Instructions for Everyday Tasks}.
\newblock In {\em Proceedings of the IEEE/CVF Conference on Computer Vision and Pattern Recognition}.

\bibitem[Singh et~al., 2022]{singh2022progprompt}
Singh, I., Blukis, V., Mousavian, A., Goyal, A., Xu, D., Tremblay, J., Fox, D., Thomason, J., and Garg, A. (2022).
\newblock Progprompt: Generating situated robot task plans using large language models.
\newblock {\em arXiv preprint arXiv:2209.11302}.

\bibitem[Sridharan and Mota, 2020]{sridharan2020commonsense}
Sridharan, M. and Mota, T. (2020).
\newblock {Commonsense Reasoning to Guide Deep Learning for Scene Understanding}.
\newblock In {\em Proceedings of the International Joint Conferences on Artificial Intelligence}.

\bibitem[Srinivasan et~al., 2021]{srinivasan2021wit}
Srinivasan, K., Raman, K., Chen, J., Bendersky, M., and Najork, M. (2021).
\newblock {WIT: Wikipedia-based Image Text Dataset for Multimodal Multilingual Machine Learning}.
\newblock In {\em Proceedings of the International ACM SIGIR Conference on Research and Development in Information Retrieval}.

\bibitem[Stachniss, 2009]{stachniss2009robotic}
Stachniss, C. (2009).
\newblock {\em {Robotic Mapping and Exploration}}.
\newblock Springer.

\bibitem[Straub et~al., 2019]{straub2019replica}
Straub, J., Whelan, T., Ma, L., Chen, Y., Wijmans, E., Green, S., Engel, J.~J., Mur-Artal, R., Ren, C., Verma, S., Clarkson, A., Yan, M., Budge, B., Yan, Y., Pan, X., Yon, J., Zou, Y., Leon, K., Carter, N., Briales, J., Gillingham, T., Mueggler, E., Pesqueira, L., Savva, M., Batra, D., Strasdat, H.~M., Nardi, R.~D., Goesele, M., Lovegrove, S., and Newcombe, R. (2019).
\newblock {The Replica Dataset: A Digital Replica of Indoor Spaces}.
\newblock {\em arXiv preprint arXiv:1906.05797}.

\bibitem[Sun et~al., 2011]{sun2011planning}
Sun, Y., Gomez, F., and Schmidhuber, J. (2011).
\newblock Planning to be surprised: Optimal bayesian exploration in dynamic environments.
\newblock In {\em AGI}.

\bibitem[Tang et~al., 2017]{tang2017exploration}
Tang, H., Houthooft, R., Foote, D., Stooke, A., Chen, O.~X., Duan, Y., Schulman, J., DeTurck, F., and Abbeel, P. (2017).
\newblock {\#Exploration: A study of count-based exploration for deep reinforcement learning}.
\newblock In {\em Advances in Neural Information Processing Systems}.

\bibitem[Telea, 2004]{telea2004image}
Telea, A. (2004).
\newblock An image inpainting technique based on the fast marching method.
\newblock {\em J. of Graphics Tools}.

\bibitem[Thomee et~al., 2016]{thomee2016yfcc100m}
Thomee, B., Shamma, D.~A., Friedland, G., Elizalde, B., Ni, K., Poland, D., Borth, D., and Li, L.-J. (2016).
\newblock {YFCC100M: The new data in multimedia research}.
\newblock {\em Communications of the ACM}.

\bibitem[Truong et~al., 2021]{truong2021bi}
Truong, J., Chernova, S., and Batra, D. (2021).
\newblock Bi-directional domain adaptation for sim2real transfer of embodied navigation agents.
\newblock {\em IEEE Robotics and Automation Letters}.

\bibitem[Truong et~al., 2022]{truong2022rethinking}
Truong, J., Rudolph, M., Yokoyama, N., Chernova, S., Batra, D., and Rai, A. (2022).
\newblock Rethinking sim2real: Lower fidelity simulation leads to higher sim2real transfer in navigation.
\newblock In {\em Proceedings of the Conference on Robot Learning}.

\bibitem[Vaswani et~al., 2017]{vaswani2017attention}
Vaswani, A., Shazeer, N., Parmar, N., Uszkoreit, J., Jones, L., Gomez, A.~N., Kaiser, {\L}., and Polosukhin, I. (2017).
\newblock Attention is all you need.
\newblock In {\em Advances in Neural Information Processing Systems}.

\bibitem[Vedantam et~al., 2015]{vedantam2015cider}
Vedantam, R., Lawrence~Zitnick, C., and Parikh, D. (2015).
\newblock {CIDEr: Consensus-based Image Description Evaluation}.
\newblock In {\em Proceedings of the IEEE/CVF Conference on Computer Vision and Pattern Recognition}.

\bibitem[Vinyals et~al., 2017]{vinyals2017show}
Vinyals, O., Toshev, A., Bengio, S., and Erhan, D. (2017).
\newblock {Show and Tell: Lessons Learned from the 2015 MSCOCO Image Captioning Challenge}.
\newblock {\em IEEE Transactions on Pattern Analysis and Machine Intelligence}.

\bibitem[Wijmans et~al., 2019]{wijmans2019dd}
Wijmans, E., Kadian, A., Morcos, A., Lee, S., Essa, I., Parikh, D., Savva, M., and Batra, D. (2019).
\newblock {DD-PPO: Learning Near-Perfect PointGoal Navigators from 2.5 Billion Frames}.
\newblock In {\em Proceedings of the International Conference on Learning Representations}.

\bibitem[Xia et~al., 2020]{xia2020interactive}
Xia, F., Shen, W.~B., Li, C., Kasimbeg, P., Tchapmi, M.~E., Toshev, A., Mart{\'\i}n-Mart{\'\i}n, R., and Savarese, S. (2020).
\newblock Interactive gibson benchmark: A benchmark for interactive navigation in cluttered environments.
\newblock {\em IEEE Robotics and Automation Letters}.

\bibitem[Xia et~al., 2018]{xia2018gibson}
Xia, F., Zamir, A.~R., He, Z., Sax, A., Malik, J., and Savarese, S. (2018).
\newblock {Gibson Env: Real-world perception for embodied agents}.
\newblock In {\em Proceedings of the IEEE/CVF Conference on Computer Vision and Pattern Recognition}.

\bibitem[Xu et~al., 2015]{xu2015show}
Xu, K., Ba, J., Kiros, R., Cho, K., Courville, A., Salakhutdinov, R., Zemel, R.~S., and Bengio, Y. (2015).
\newblock Show, attend and tell: Neural image caption generation with visual attention.
\newblock In {\em Proceedings of the International Conference on Machine Learning}.

\bibitem[Yadav et~al., 2022]{yadav2022habitat}
Yadav, K., Ramrakhya, R., Ramakrishnan, S.~K., Gervet, T., Turner, J., Gokaslan, A., Maestre, N., Chang, A.~X., Batra, D., Savva, M., et~al. (2022).
\newblock Habitat-matterport 3d semantics dataset.
\newblock {\em arXiv preprint arXiv:2210.05633}.

\bibitem[Yamauchi, 1997]{yamauchi1997frontier}
Yamauchi, B. (1997).
\newblock {A Frontier-Based Approach for Autonomous Exploration}.
\newblock In {\em Proceedings of the IEEE International Symposium on Computational Intelligence in Robotics and Automation}.

\bibitem[Ye et~al., 2021a]{ye2021auxiliaryON}
Ye, J., Batra, D., Das, A., and Wijmans, E. (2021a).
\newblock {Auxiliary tasks and exploration enable ObjectNav}.
\newblock In {\em Proceedings of the IEEE/CVF International Conference on Computer Vision}.

\bibitem[Ye et~al., 2021b]{ye2021auxiliaryPN}
Ye, J., Batra, D., Wijmans, E., and Das, A. (2021b).
\newblock {Auxiliary Tasks Speed Up Learning Point Goal Navigation}.
\newblock In {\em Proceedings of the Conference on Robot Learning}.

\bibitem[You et~al., 2019]{you2019large}
You, Y., Li, J., Reddi, S., Hseu, J., Kumar, S., Bhojanapalli, S., Song, X., Demmel, J., Keutzer, K., and Hsieh, C.-J. (2019).
\newblock {Large Batch Optimization for Deep Learning: Training BERT in 76 minutes}.
\newblock In {\em Proceedings of the International Conference on Learning Representations}.

\bibitem[Zamir et~al., 2018]{zamir2018taskonomy}
Zamir, A.~R., Sax, A., Shen, W., Guibas, L.~J., Malik, J., and Savarese, S. (2018).
\newblock Taskonomy: Disentangling task transfer learning.
\newblock In {\em Proceedings of the IEEE/CVF Conference on Computer Vision and Pattern Recognition}.

\bibitem[Zhang et~al., 2021a]{zhang2021vinvl}
Zhang, P., Li, X., Hu, X., Yang, J., Zhang, L., Wang, L., Choi, Y., and Gao, J. (2021a).
\newblock {VinVL: Revisiting visual representations in vision-language models}.
\newblock In {\em Proceedings of the IEEE/CVF Conference on Computer Vision and Pattern Recognition}.

\bibitem[Zhang et~al., 2020a]{zhang2020bebold}
Zhang, T., Xu, H., Wang, X., Wu, Y., Keutzer, K., Gonzalez, J.~E., and Tian, Y. (2020a).
\newblock Bebold: Exploration beyond the boundary of explored regions.
\newblock {\em arXiv preprint arXiv:2012.08621}.

\bibitem[Zhang et~al., 2021b]{zhang2021rstnet}
Zhang, X., Sun, X., Luo, Y., Ji, J., Zhou, Y., Wu, Y., Huang, F., and Ji, R. (2021b).
\newblock {RSTNet: Captioning with Adaptive Attention on Visual and Non-Visual Words}.
\newblock In {\em Proceedings of the IEEE/CVF Conference on Computer Vision and Pattern Recognition}.

\bibitem[Zhang et~al., 2020b]{zhang2020diagnosing}
Zhang, Y., Tan, H., and Bansal, M. (2020b).
\newblock {Diagnosing the Environment Bias in Vision-and-Language Navigation}.
\newblock {\em Proceedings of the International Joint Conferences on Artificial Intelligence}.

\bibitem[Zhao et~al., 2021]{zhao2021surprising}
Zhao, X., Agrawal, H., Batra, D., and Schwing, A.~G. (2021).
\newblock The surprising effectiveness of visual odometry techniques for embodied pointgoal navigation.
\newblock In {\em Proceedings of the IEEE/CVF International Conference on Computer Vision}.

\bibitem[Zhu et~al., 2018]{zhu2018deep}
Zhu, D., Li, T., Ho, D., Wang, C., and Meng, M. Q.-H. (2018).
\newblock {Deep Reinforcement Learning Supervised Autonomous Exploration in Office Environments}.
\newblock In {\em Proceedings of the IEEE International Conference on Robotics and Automation}.

\bibitem[Zhu et~al., 2021]{zhu2021soon}
Zhu, F., Liang, X., Zhu, Y., Yu, Q., Chang, X., and Liang, X. (2021).
\newblock Soon: Scenario oriented object navigation with graph-based exploration.
\newblock In {\em Proceedings of the IEEE/CVF Conference on Computer Vision and Pattern Recognition}.

\bibitem[Zhu et~al., 2017]{zhu2017target}
Zhu, Y., Mottaghi, R., Kolve, E., Lim, J.~J., Gupta, A., Fei-Fei, L., and Farhadi, A. (2017).
\newblock Target-driven visual navigation in indoor scenes using deep reinforcement learning.
\newblock In {\em Proceedings of the IEEE International Conference on Robotics and Automation}.

\end{thebibliography}
\end{spacing}

\cleardoublepage
\addcontentsline{toc}{chapter}{Glossary}
% \printglossaries

\cleardoublepage

\end{document}